\documentclass[nonacm, format=acmsmall, review=false, screen=true, prologue,table,xcdraw]{acmart}
\usepackage[utf8]{inputenc}

\usepackage{graphicx}
\usepackage{pdflscape}
\usepackage{afterpage}
\usepackage{multirow}

\newcolumntype{L}[1]{>{\raggedright\arraybackslash}p{#1}}
\newcolumntype{C}[1]{>{\centering\arraybackslash}p{#1}}
\newcolumntype{X}[1]{>{\centering\let\newline\\\arraybackslash\hspace{0pt}}m{#1}}

\usepackage{todonotes}

\usepackage[title]{appendix}

\usepackage{longtable}
\usepackage{ltablex}

\usepackage{enumitem}
\setlist{  
  listparindent=\parindent,
  parsep=0pt,
}

\begin{document}


\title{The Technological Emergence of AutoML}
\subtitle{A Survey of Performant Software and Applications in the Context of Industry}
\author{Alexander Scriven}
\affiliation{
  \institution{Complex Adaptive Systems Lab, Data Science Institute, University of Technology Sydney}
  \city{Sydney}
  \state{New South Wales}
  \postcode{2007}
  \country{Australia}
}
\email{alexander.scriven@uts.edu.au}

\author{David Jacob Kedziora}
\affiliation{%
  \institution{Complex Adaptive Systems Lab, Data Science Institute, University of Technology Sydney}
  \city{Sydney}
  \state{New South Wales}
  \postcode{2007}
  \country{Australia}
}
\email{david.kedziora@uts.edu.au}

\author{Katarzyna Musial}
\affiliation{%
  \institution{Complex Adaptive Systems Lab, Data Science Institute, University of Technology Sydney}
  \city{Sydney}
  \state{New South Wales}
  \postcode{2007}
  \country{Australia}
}
\email{katarzyna.musial-gabrys@uts.edu.au}

\author{Bogdan Gabrys}
\affiliation{%
  \institution{Complex Adaptive Systems Lab, Data Science Institute, University of Technology Sydney}
  \city{Sydney}
  \state{New South Wales}
  \postcode{2007}
  \country{Australia}
}
\email{bogdan.gabrys@uts.edu.au}

\renewcommand{\shortauthors}{Scriven et al.}

\begin{abstract}

With most technical fields, there exists a delay between fundamental academic research and practical industrial uptake.
Whilst some sciences have robust and well-established processes for commercialisation, such as the pharmaceutical practice of regimented drug trials, other fields face transitory periods in which fundamental academic advancements diffuse gradually into the space of commerce and industry.
For the still relatively young field of Automated/Autonomous Machine Learning (AutoML/AutonoML), that transitory period is under way, spurred on by a burgeoning interest from broader society.
Yet, to date, little research has been undertaken to assess the current state of this dissemination and its uptake.
Thus, this review makes two primary contributions to knowledge around this topic.
Firstly, it provides the most up-to-date and comprehensive survey of existing AutoML tools, both open-source and commercial.
Secondly, it motivates and outlines a framework for assessing whether an AutoML solution designed for real-world application is 'performant'; this framework extends beyond the limitations of typical academic criteria, considering a variety of stakeholder needs and the human-computer interactions required to service them.
Thus, additionally supported by an extensive assessment and comparison of academic and commercial case-studies, this review evaluates mainstream engagement with AutoML in the early 2020s, identifying obstacles and opportunities for accelerating future uptake.

\end{abstract}

\maketitle


\section{Introduction}
\label{Sec:intro}

Societal interest in machine learning (ML), especially the subtopic of deep learning (DL), has surged within recent years.
This is partially driven by the continuing success of these approaches in many application areas~\cite{grtr20, ozgu20, mibo21, piso21}, facilitated by both fundamental advances~\cite{koza06, chgu16, sith16, chgi17} and the increasing availability of computational resources.
Unsurprisingly, on the academic side, the field of artificial intelligence (AI) continues to dominate research outputs, as noted by the 2021 UNESCO Science Report~\cite{scle21}.
However, it is the current level of ML engagement in industry that is truly unprecedented.
For instance, the 2021 Global AI Adoption Index, commissioned by IBM, found that 80\% of 5501 global businesses are either using automation software or planning to within 12 months, and 74\% are exploring or deploying AI~\cite{ib21}.
The Gartner 2019 CIO Agenda survey, with 3000 respondents from across the globe, agrees with this trend, revealing that the proportion of firms deploying AI has increased from 10\% in 2015 to 37\% in 2019~\cite{ga19}.
Similar conclusions are echoed in the 2020 McKinsey `State of AI' report~\cite{mc20}.
Naturally, such a rate of mainstream permeation is also accompanied by intensifying discussions on how to use ML, and AI more broadly, in a socially responsible manner~\cite{ma17, leol18, joie19, mofl20, toai20}.

Nonetheless, despite the growing desire of industry to utilise ML, talent in data science remains scarce~\cite{pobu17, sr18}.
Both the Gartner and IBM studies agree that lack of expertise creates a barrier to AI adoption~\cite{ga19, ib21}, especially as, by and large, ML technology still requires specialist skills to implement and employ.
Worse yet, in practice, deploying ML solutions for real-world applications requires technical skills beyond the domain of data science.
Any shortfall in these broader talents will also adversely affect ML engagement in industry~\cite{au21, ga21a}.
So, faced with these realities, a business may ponder: does ML really have to rely so heavily on humans?
Enter `automated machine learning' (AutoML), a research endeavour that has become particularly popular over the last decade~\cite{elma19, huko19, lema19, es20, saha20, hezh21, zohu21}, striving to mechanise as many high-level ML operations as possible.
The appeal of this emergent field is multi-faceted, driven by many of the same motivations that inspire automation in general.
These include not just democratisation, enabling the broader public to leverage the power of ML approaches, but also efficiency boosts, redistributing the time and effort of existing talent to more valuable functions.

\begin{figure}[H]
  \centering
  \includegraphics[width=1\linewidth]{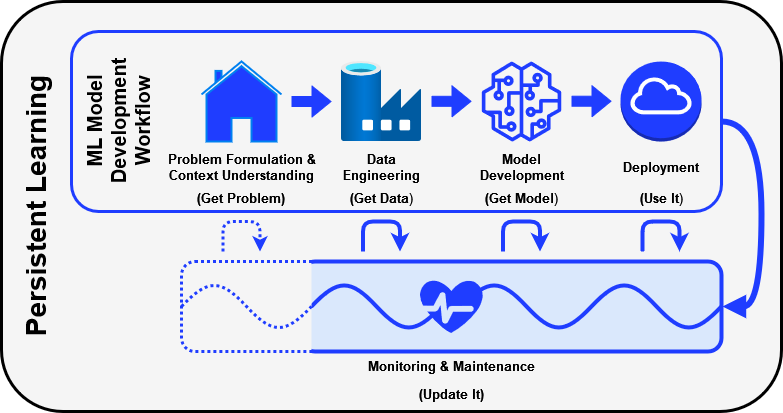}
  \caption{Schematic of a general machine learning (ML) workflow, which captures the phases involved in designing, constructing, deploying and maintaining an ML model for a real-world application.}
  \label{Fig:mlworkflow}
\end{figure}

Notably, within the modern era of AutoML, academia has already made much progress.
Admittedly, it can be challenging to contain this ever-widening field within a simple overview, and various works lean on taxonomies and categorical systems to aid this~\cite{kemu20, saha20, doke21, khke21}.
Consider then a conceptual representation of the processes that are involved in running a real-world ML application, i.e.~an ML workflow, as shown in Fig.~\ref{Fig:mlworkflow}.
With respect to this depiction, the bulk of AutoML research has traditionally focused on automating the model-development phase.
Advances in Bayesian optimisation, which continue to be employed~\cite{klfa17a, fr18}, are frequently credited for jump-starting this process, reducing human involvement in hyperparameter optimisation (HPO) and chipping away at the broader `combined algorithm selection and HPO' (CASH) problem~\cite{thhu13}.
Since then, this undertaking has evolved in many ways, such as by encapsulating neural architecture search (NAS)~\cite{wira19, rexi20, doli20, dota20}, which now forms the core of the AutoML-subfield known as `automated DL' (AutoDL).

However, as previously hinted, the scope of AutoML -- AutoDL included~\cite{doke21} -- has itself gradually expanded to encompass the rest of an ML workflow.
For instance, data engineering has received its own fair share of research attention.
Some works in this space focus on the initial stage of data preparation~\cite{khil15}, which may involve sampling and cleaning, while others contribute to the topic of automated feature engineering (AutoFE)~\cite{lath17}, covering both feature generation and selection.
Then there are phase-agnostic methodologies, such as meta-learning~\cite{albu15, lebu15, fiab17, va18, huko19, lega10}, that can theoretically be applied anywhere; these continue to free humans from micro-managing ML systems and supplying domain knowledge.
Of course, there is still much further to go.
Automating the phase of continuous monitoring and maintenance has recently been highlighted as a crucial prerequisite for truly autonomous machine learning (AutonoML)~\cite{kemu20}, where systems persist by adapting ML models to changes in data environment~\cite{kaga09, zlbi12}.
Progress in this space remains relatively nascent~\cite{luli19, chle21, ghwe21}.
Additionally, rigorous efforts to survey and benchmark state-of-the-art (SOTA) algorithms and approaches~\cite{es20, role20} are relatively sparse.
Nevertheless, the key takeaway from all of this is that, academically, the field of AutoML is rich with activity.

Unfortunately, the translation of pure theory to real-world practice is rarely smooth or one-to-one.
That is not to suggest that AutoML has been shunned by industry; to the absolute contrary, a prior dearth of tools to assist with developing ML models -- according to the IBM survey~\cite{ib21}, one of the top three obstacles for AI uptake -- has actually led to an explosion of commercial AutoML services.
Alongside numerous open-source packages, these offerings provide businesses plenty of options to choose from, as of the early 2020s, should they wish to apply ML approaches to problems of interest.
Yet a healthy scepticism remains warranted, especially where source code is confidential and promotional material is inherently biased.
It cannot be assumed that AutoML algorithms and architectures, developed in experimental environments that are well-controlled and sanitised, will deliver optimal outcomes once applied within messy real-world contexts.
Certainly, the academic case studies that exist~\cite{ormo17, payu19, tsko20}, evaluating one or more AutoML solutions within particular industrial domains, are too few in number to make broad claims.
So, it is worth asking the question: how are publicly available SOTA AutoML tools and services performing with respect to the demands of industry?

The notion of `performant' ML must be central to any pioneering survey that grapples with this question.
In most academic research, performance is usually gauged by purely technical metrics, such as model accuracy and training efficiency.
The focus is on how well a computer, in the absence of any human, can generate predictions/prescriptions via ML techniques.
On the other hand, industrial contexts are much more human-centric, where stakeholders may have a diversity of interests and obligations; the outcomes and impact of an ML application may only be very loosely correlated with technical performance.
Importantly, such matters cannot be ignored by academic AutoML researchers either, as stakeholder requirements can affect the very foundations of algorithms and architectures.
For instance, a need for interpretability may force ML model-selection pools to be constrained, a focus on fairness may require mechanisms for bias mitigation, and so on.

Simply put, the technological emergence of AutoML is driven by stakeholder need and the human-computer interaction (HCI) required to service it.
Correspondingly, it is impossible to gauge the current state of AutoML technology, especially in terms of whether it can support the needs of industry, without the careful development of an assessment framework anchored by a comprehensive set of HCI-weighted criteria for `performant ML'.
Certainly, the absence of such a systematic appraisal may not only obscure future directions for progress but, if deficiencies are not identified, may also have an eventual chilling effect on technological engagement, especially in the case of unmet expectations.

With all that stated, the primary goal of this review is to present a comprehensive snapshot of how AutoML has permeated into mainstream use within the early 2020s.
In contrast to two associated monographs that examined fundamental algorithms and approaches behind AutoML/AutoDL~\cite{kemu20, doke21}, this work surveys both their implementation and application in the context of industry.
It also defines what a `performant' AutoML system is -- HCI support is valued highly here -- and assesses how the current crop of available packages and services, as a whole, lives up to expectation.
To do so in a systematic manner, this review is structured as follows.
Section~\ref{Sec:mlwf} begins by elaborating on the notion of an ML workflow, conceptually framing AutoML in terms of the high-level operations required to develop, deploy and maintain an ML model.
Section~\ref{Sec:Motivation} uses this workflow to support the introduction of industry-related stakeholders and their interests/obligations.
These requirements are unified into a comprehensive set of criteria, supported by methods of assessment, that determine whether an AutoML system can be considered performant.
Section~\ref{Sec:Tools} then launches the survey in earnest, assessing the nature and capabilities of existing AutoML technology.
This begins with an examination of open-source AutoML packages; some of these are tools dedicated to a singular purpose, e.g.~HPO, while others are comprehensive systems that aim to automate a significant portion of an ML workflow.
The section additionally investigates AutoML systems that are designed for specific domains, as well as commercial products.
Subsequently, Section~\ref{Sec:Applications} assesses where AutoML technology has been used and how it has fared.
Academic work focusing on real-world applications is surveyed, as are vendor-based case studies.
All key findings and assessments are then synthesised in Section~\ref{Sec:discussion}, with commentary around how mature AutoML technology is, as well as whether there are obstacles and opportunities for future uptake.
Finally, Section~\ref{Sec:Conclusion} provides a concluding overview on the technological emergence of AutoML.

\section{The Machine Learning Workflow}
\label{Sec:mlwf}

Many academic works have presented diagrams that attempt to encapsulate the high-level operations of ML within one consolidated workflow~\cite{yawa18, elma19, drwe20, hezh21}, which we henceforth refer to as an MLWF.
One early forerunner in this endeavour, though not exclusive to ML, is the popular CRoss Industry Standard Process for Data Mining (CRISP-DM) model~\cite{sh00}, and several recent efforts have built upon this basis, e.g.~by additionally considering quality assurance~\cite{stbu21}.
In this section, we extend this model further, diverging where necessary, to align even closer with the modern practices of data science.
Such a summary will not be unfamiliar to academics and practitioners of data science, and many MLWFs found in AutoML literature are indeed similar, often only expanding/compressing one or more aspects.
Nonetheless, if this monograph is to grapple with the notion of performant ML, particularly within organisational settings that operate beyond pure experimental research, a robust characterisation of an MLWF is required.

Fundamentally, many papers that depict MLWFs agree that there are certain standard phases of ML operation, as captured by the `ML Model Development Workflow' component of Fig.~\ref{Fig:mlworkflow}.
Specifically, a typical ML application will flow from `Problem Formulation \& Context Understanding' through `Data Engineering' and `Model Development' to `Deployment'.
Some MLWFs also incorporate some form of `Monitoring \& Maintenance', although this is often presented almost as an afterthought.
An academic focus on one-and-done projects, as well as the computational expense of developing modern DL models, means that the challenge of dynamically changing data environments is often ignored, either negligently or deliberately.
However, there is a growing awareness within industry that persistent learning is essential, and we thus highlight `Monitoring \& Maintenance' as a unique phase within Fig.~\ref{Fig:mlworkflow}.
Indeed, while many MLWFs, such as a CRISP-DM representation~\cite{sh00}, provide double-headed arrows or other depictions of circularity between the first four phases, we associate that continuum of updates with the `Monitoring \& Maintenance' phase.
Granted, development during an ML project is frequently iterative, with previous phases of operation being revisited prior to deployment, but the primary intent of the `ML Model Development Workflow' is to move forward and bring an ML solution to production.
In contrast, it is the intent of `Monitoring \& Maintenance' to continually reassess and keep that ML solution relevant, even if -- the dashed lines in Fig.~\ref{Fig:mlworkflow} hint that this is rarely an academic concern -- an ML problem must be partially reformulated while keeping its present solution online.

Now, despite MLWF commonalities in the literature, it is essential to emphasise that perspectives are not universal, and academia often ignores matters relevant to business applications.
For instance, several core AutoML papers ignore the deployment phase outright~\cite{drwe20, es20, hezh21}.
Others do not dwell on this phase, associating it with the production of predictions~\cite{elma19, zohu21} or, via the display of prominent social media and tech company logos, suggesting that organisations are interested in this facet of ML~\cite{yawa18}.
Detail is scant.
In contrast, it is noteworthy that, when presenting MLWFs on websites, AutoML vendors frequently elaborate on aspects of deployment~\cite{19} and, also neglected by academia, monitoring and maintenance~\cite{ai, 20}.
As already discussed, this is a matter of focus; academia prioritises the development of high accuracy models, whereas industry cares equally, if not more, for sustainable operation.

\begin{figure}[H]
  \centering
  \includegraphics[width=1\linewidth]{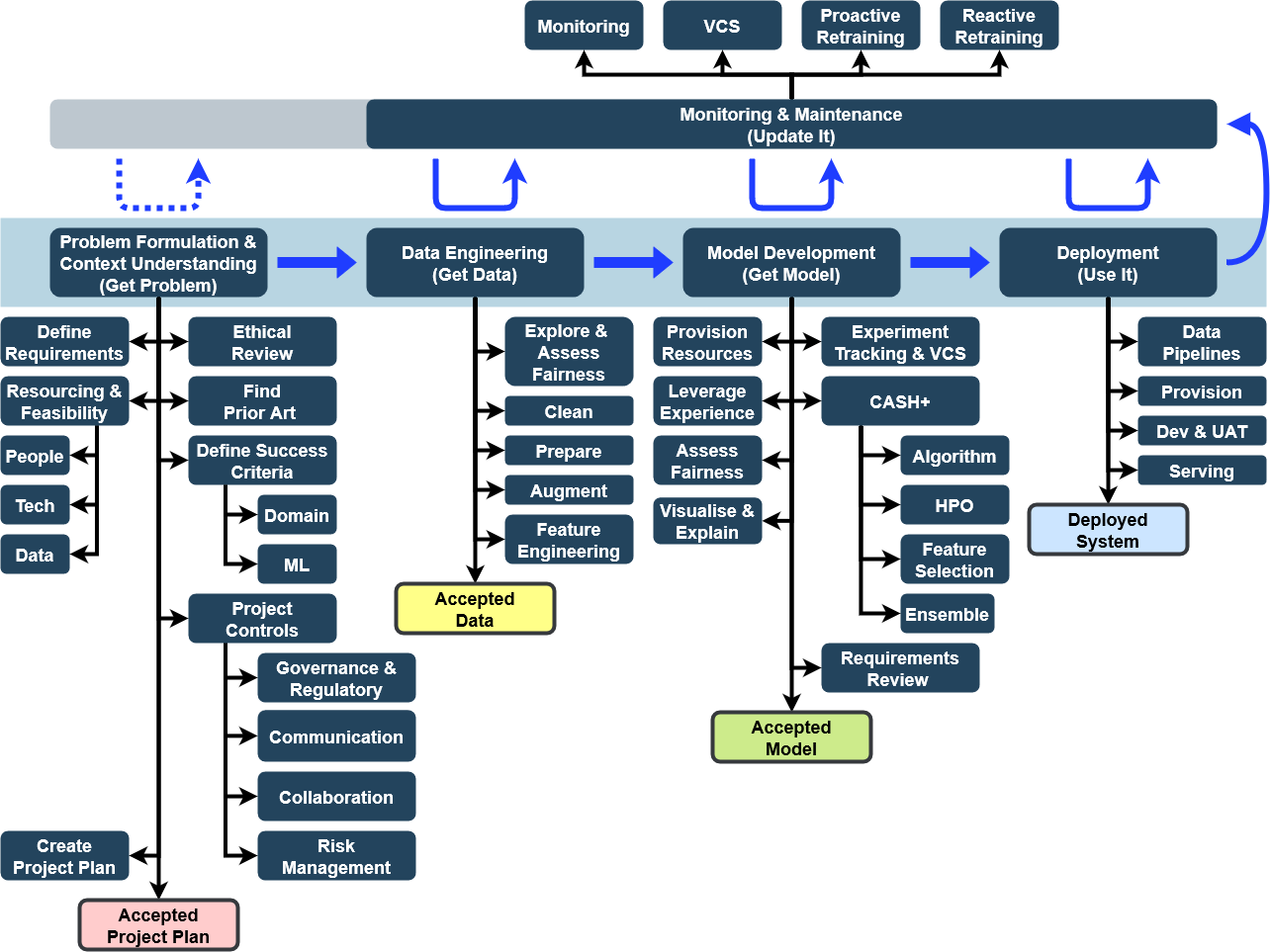}
  \caption{Key tasks within an overarching MLWF. The lighter-coloured boxes represent the outcome of an MLWF phase that is propagated onwards.}
  \label{Fig:mlworkflowsubtasks}
\end{figure}

Given this preface, we now present a deeper dive into the granular tasks that commonly constitute an MLWF within a business environment, as displayed in Fig.~\ref{Fig:mlworkflowsubtasks}.
Admittedly, these tasks are unlikely to be exhaustive for every ML application imaginable.
Individual tasks may also be unnecessary in various settings, even if good practice will likely still involve consideration ahead of rejection.
For instance, a financial project involving the approval/denial of credit via algorithmic means demands a greater contemplation of bias and ethics than is likely needed for the manufacturing-based prediction of machinery faults.
All the same, the breakdown in Fig.~\ref{Fig:mlworkflowsubtasks} is sufficiently informative to support a survey of ML tools and the extent to which they automate key tasks.

The first MLWF phase of \textbf{problem formulation \& context understanding} seeks to establish an agreeable plan of action, accepted by all relevant stakeholders, for undertaking the rest of an ML project.
Although this certainly includes academic elements of developing/acquiring expertise around a problem context and accumulating sufficient topical knowledge to support an ML effort, much of this stage involves largely organisational considerations that ensure the ML project is appropriately defined, scoped, and resourced.
First, an organisation must establish its project requirements, which are associated with why the ML application is being undertaken in the first place.
For example, a business may decide to leverage ML in predicting users at risk of churn so that its customer support teams can intervene and prevent this from occurring.
With these requirements formalised, this is then a good opportunity to begin considering `prior art'.
In modern times, this might include academic references, reputable blog posts, and similar work performed internally within the organisation.
Prior art can indicate how tractable the ML problem is, how best to approach solving it, and what kind of performance can be expected.
At this point, stakeholders can also determine whether a SOTA deep neural network is required or whether a simpler approximator, e.g.~a linear regressor, is sufficient for the established requirements.
In either case, whether appetite leans towards code reuse or pushing the frontiers of ML, resourcing and feasibility checks typically ensue.
Of course, the initial search for prior art already involves ensuring an ML project is conceptually sound, but this collection of sub-tasks covers other logistical matters.
These considerations include identifying/acquiring people to undertake the work, technological tools to assist them, and raw data sources to form the basis of modelling work.
As part of establishing a project plan, an organisation must also define how one can know that the project requirements have been satisfied, i.e.~its success criteria.
Determining this will generally attempt to marry organisational/domain factors with technical ML objectives and outputs.
For instance, a churn-concerned business may decide that, within acceptable timelines and on the balance of projected costs and rewards, a precision score of $85\%$ may be a satisfactory outcome.
Finally, an organisation must generally establish appropriate project-management controls to support a greenlit ML application.
This task includes considering governance and risk management, e.g.~delegating access permissions, task responsibilities, the authority to approve work, and so on.
Additionally, an organisation will typically use this phase to decide on communication channels and collaborative tools for key team members while also setting expectations and reporting processes.
Furthermore, the finalisation of a project plan will often include more detailed planning and project-management artefacts, such as Gantt charts or critical paths, but these nuances of organisational practices are too varied to generalise.

The second MLWF phase of \textbf{data engineering} involves the initial exploration, processing and enhancement of data for modelling purposes.
It is often professed that this stage takes up a significant portion of working time for any data scientist.
Indeed, a 2016 CrowdFlower report~\cite{cr16} claimed that the percentage was as high as $80\%$, summing `collecting data sets ($19\%$)' with `cleaning and organising data ($60\%$)'.
However, providing limited details on the number of respondents and methodology used, the report has been contested by other surveys, despite the widespread mainstream adoption of its claims.
For instance, the `2018 Kaggle Machine Learning \& Data Science Survey'~\cite{ka18}, with 23859 responses, yielded $11\%$ for gathering and $15\%$ for cleaning.
Of course, caution is still necessary when assessing outcomes from an open and uncontrolled survey based on volunteered self-reporting.
The Kaggle survey responders were diverse in occupation, e.g.~including students ($20\%$) and software engineers ($13\%$), were dominantly skewed towards an early career, i.e.~$60\%$ under 30 years of age, and heavily represented the United States ($19\%$) and India ($19\%$).
A 2020 version of the survey~\cite{ka21} could not reaffirm these claims as the question was absent; in 2020, the focus was on tools, techniques, and other questions regarding respondent skills and employment.
Elsewhere, an Anaconda survey, `The State of Data Science 2020'~\cite{an20a}, found worktime proportions of $19\%$ for `data loading' and $26\%$ for `data cleansing'.
It had a smaller sample size of 2360 respondents, but these appeared to be more of a professional data-science background.
Regardless, whether at $80\%$ or a more moderate value, the amount of time that goes into data engineering is not insubstantial.

Regarding the task breakdown in Fig.~\ref{Fig:mlworkflowsubtasks}, a common first step in data engineering is exploratory data analysis (EDA).
Data scientists will frequently assess the numerical properties of their datasets, visually inspecting graphical representations and identifying quality issues, e.g.~missing values and anomalies.
Different problem contexts will shape exactly what is undertaken in this section, and many helpful guides are available in many books and blogs~\cite{sh00, wigr16, ma21, sh21, va21a}.
Crucially, this is also an appropriate time to assess bias and fairness within the data inputs themselves; see Section~\ref{Sec:biasfairnesstools}.
Traditional EDA guides often ignore this facet, but recent years have seen a surge of attention and concern around ML and trustworthiness.
Correspondingly, matters of bias and fairness are some of the key criteria for performant ML outlined in Section~\ref{Sec:summarisedcriteria}.
Regardless, once EDA equips a data scientist with a sufficient understanding of a supplied data environment, they are then usually able to modify the data ahead of ML modelling.
Nomenclature and ontologies vary across the literature for the tasks involved in modifying data~\cite{va16, bogr19, re19, ga21b}, but here we settle on the sequence of cleaning, preparing, augmenting, and feature engineering.
Specifically, we define cleaning in relation to handling erroneous data, while preparation involves formatting input data so that an ML algorithm can access its information content.
Accordingly, the imputation of missing values is treated here as a cleaning task, while one-hot encoding categorical variables could be considered a preparatory step.
Data preparation also includes scaling, standardisation, and any preprocessing related to data type or structure, e.g.~handling timestamps or freeform text.
Then there is augmentation, where relevant ancillary data sources are joined to the current inputs.
Within this monograph, the novel data is considered entirely external, not engineered variations of existing data as some DL literature may define it~\cite{doke21}.
As an example of such augmentation, timestamped store-utilisation data used to predict foot traffic to a retail shop may become more informative when associated with public weather data.
Finally, feature engineering aims to transform existing variables in different ways, all in the hope that informative signals in the data may surface.
This task receives plenty of academic attention as it is arguably the least straightforward process.

Again, we stress that there are differing views in the data science community on how to categorise and arrange all these data-engineering tasks.
For instance, while we consider one-hot encoding as a form of data preparation here, it is valid to argue that new features are being engineered, i.e.~new columns are being added to tabular data.
Feature engineering is an even more complex notion to organise, especially in light of the standard filter/wrapper perspective discussed in an earlier AutonoML review~\cite{kemu20}.
For instance, if feature selection is done before considering an ML model, i.e.~in filter style, it would seem to belong in the data-engineering phase.
An example is filtering out features according to the outcomes of Pearson correlations or chi-squared tests.
However, if features are selected based on whether they improve the performance of a specific ML model, i.e.~in wrapper style, the algorithms that do so may be pipelined as part of the model-development phase.
An example is exclusion based on feature importance scores that the Random Forest ML algorithm provides.
Ultimately, we highlight these nuances but persist with the arrangement in Fig.~\ref{Fig:mlworkflowsubtasks}.
The tasks listed under data engineering are, in aggregate, holistically encompassing; minor variations do not significantly perturb the framework proposed by this monograph for assessing performant ML.

The third MLWF phase of \textbf{model development} is arguably the core of an ML application, proceeding once a cleaned, prepared and enhanced dataset is in hand.
As the task breakdown suggests, a large-scale ML application often involves several preparatory steps.
Beyond setting up requisite model-training infrastructure, an organisation may need to establish tools that track experimentation and model versioning while trials are being undertaken.
Eventually, though, there comes the actual process of fitting a mathematical model to a desirable function.
Even in the present time, AutoML literature predominantly focusses on selecting an ML algorithm and tuning hyperparameters, i.e.~solving the CASH problem~\cite{thhu13}, so we use the term CASH+ to be more encompassing of ML solutions.
Specifically, some ML applications and AutoML packages will bundle feature-selection or predictor-ensembling methods as part of an ML pipeline they pursue.
Also, it is worth noting that many researchers have investigated how to `leverage experience' when solving some subset of the CASH+ problem, e.g.~using opportunistically derived predictor rankings to constrain search spaces for ML algorithms~\cite{ngke21,keng22}.
The leveraging of experience thus covers the technical area of meta-learning~\cite{albu15, lebu15, va18, kemu20, lega10a, lega10, alga18}, but it also refers to leaning on domain experts for assistance in contextualising, understanding and making better decisions with preliminary model results~\cite{kaga09}.
We note that prior experience can inform any phase of the MLWF, but model development has most often been the focus of such research and development.

Notably, various metrics and visualisations may be produced throughout an ML application to understand the modelling activity.
However, once a preliminary final model has been produced, contextually driven tasks may be carried out to uncover what work was undertaken and what was ultimately produced.
Here, revisiting the notions of bias and fairness is crucial in producing models that meet trust-based requirements.
Again, Section~\ref{Sec:biasfairnesstools} elaborates on this topic, e.g.~on assessing data versus a model, but it is sufficient to note here that an ML model can be assessed and remedied if it proves problematically biased or unfair.
Another topic that is also deeply entwined with trust in ML is that of explainable AI (XAI).
Different organisational stakeholders will have different needs concerning XAI, as discussed in Section~\ref{Sec:keystakeholders}.
However, the model-development phase of an MLWF is an appropriate time to understand how an ML solution was generated and why it produces the outputs that it does.
This process often includes visualising performance metrics, global explainability artefacts such as feature importance, and the drivers of individual predictions/prescriptions.
Additionally, scenario planning tools may be made available here to understand the impact of potential interventions on these metrics and visualisations.
Finally, the generated ML model and all related artefacts, e.g.~XAI items, are assessed against the initial requirements for the project.
Some MLWFs consider this under an `evaluation' phase that often refers to simple technical metrics, e.g.~model accuracy and training time, but industrial applications often have many more requirements that an ML solution must satisfy before it can be approved for deployment.
The dearth of academic contemplation in this space is a primary motivation for this monograph and its comprehensive framework for performant ML.

The fourth MLWF phase of \textbf{deployment} and the fifth MLWF phase of \textbf{monitoring \& maintenance} incorporate all the tasks required to turn an experimental modelling project into a sustainable productionalised system embedded within some organisational process.
Broadly, they encompass much of what is nowadays referred to as `MLOps'.
Specifically, one of the first steps in deploying an ML solution is to convert all relevant data transformations into a robust pipeline.
This pipeline must connect raw data sources to the ML models running inference, and this transmission must be suitable for context, e.g.~batched, real-time, constrained for the internet of things, and so on.
Computational resources must also be provisioned to host the ML solution and support its inferences.
Now, one common practice in software engineering is undertaking user acceptance testing (UAT)~\cite{ci06, bl02}, ensuring that a system meets the expectations of end users.
Such practices are similarly relevant in organisational applications of ML where model results are intended for widespread consumption.

Eventually, once sufficiently tested and fully provisioned, an ML solution can be appropriately placed in production.
However, the performance of a static ML model can decay over time due to environmental dynamics, such as data drift, concept drift~\cite{zlga14, wehy16, luli19, maes19}, and other system disruptions.
Ideally, monitoring processes are established for all metrics of relevance, including technical performance, data properties of interest, and variables that assess deployed ML solution outcomes, e.g.~bias and fairness metrics.
An adaptation process can then be triggered automatically or after manual review if a monitored metric dips below a threshold.
The simplest form of adaptation is the full retraining of an ML model, but there are numerous redeployment techniques, e.g.~blue-green deployment, canary deployment~\cite{ahsy18}, and many more~\cite{ru20}.
Of course, as change occurs within a deployed solution, implementing a version control system (VCS) is advisable to mitigate the risks of unforeseen issues with updates; it is always helpful to roll back to prior safe versions.

In conclusion, we have schematised a general MLWF and, via Fig.~\ref{Fig:mlworkflowsubtasks}, elaborated on the typical tasks that an organisation may carry out in running an ML application.
As already mentioned, not everything here will be universally relevant.
Some ML efforts are solely angled towards uncovering data insights, and several vendors of automated tools operate within this market.
However, standard organisational use of ML involves taking a trained model from problem conceptualisation to production, generally relying on consistent delivery of business value via long-term consumer engagement.
Thus, the more tasks within the full MLWF that an AutoML tool automates, accounting for the standards of performant ML, the more appealing it is to industrial stakeholders.
In fact, if the extent and degree of automation sufficiently encompass the monitoring \& maintenance phase, such that an ML model learns persistently and autonomously, the era of AutonoML will have truly arrived~\cite{kemu20}.
In short, the MLWF framework helps anchor assessments of AutoML tools and their scope; this monograph will detail the broad spectrum of existing AutoML services.
However, to honestly assess the value of modern AutoML to industry, a simple question needs answering: who cares?

\section{Performant Machine Learning}
\label{Sec:Motivation}

Nuances aside, a common maxim in economics is that ``demand creates supply''.
Need and desire drive interest, investment, and innovation.
In complement to this rule, a product does not survive and thrive in an industrial setting without serving a purpose and generating a positive impact.
Indeed, while the clientele for ML technology may be extensive, it is also finite.
Over time, competition for stakeholder attention and engagement is an optimisation process, impelling tools and systems that support `performant ML' to bubble up into prominence.
Of course, as with biological evolution, this process is not perfect; odd and even detrimental `genotypes/phenotypes' could arise and become entrenched within a `population' of AutoML services.
Nevertheless, by and large, the quest for performant ML is the driving force behind AutoML technology.

So, what is performant ML?
Who decides?
Traditionally, academia has a very narrow scope when defining ML performance, as exemplified by typical textbooks on the topic~\cite{jawi13}.
Its focus is predominantly on metrics that judge how well an ML model approximates a desirable function, such as classification accuracy and various ratios involving a confusion matrix, e.g.~sensitivity, specificity, recall, area under the curve (AUC), and the $F_1$ score~\cite{fa06, po20}.
Occasionally, these metrics may also be paired with information on how long it took to optimise them, i.e.~the time costs of ML model training.
These considerations are particularly pertinent within hardware-conscious research beyond pure algorithmic advancement, where performance measures must be mindful of infrastructure~\cite{scmu16,dupa17}.
However, in industrial and applied contexts, it is reasonable to question whether these alone are the only metrics that matter.
One experimental investigation~\cite{zohu21} asserts that most CASH procedures perform reasonably similarly, at least in technical terms, concluding that the suitability of deploying AutoML frameworks for real-world use cases should consider factors beyond those of typical academic concern.
This review supports such a perspective.

Thus, having already detailed \textit{what} is typically involved in the practice of real-world ML, this section delves into \textit{who} would care for automating an MLWF and how they would judge the overall process as performant.
Specifically, in Section~\ref{Sec:keystakeholders}, we outline the key stakeholders involved in ML tasks within an industrial context, detailing their needs and the potential benefits they may realise from AutoML.
Then, in Section~\ref{Sec:summarisedcriteria}, we propose a comprehensive set of criteria by which the practical application of an MLWF can be evaluated.
Finally, in Section~\ref{Sec:automlrole}, we synthesise these considerations to assess the role that industry currently expects AutoML to play in supporting performant ML.

\subsection{Key Stakeholders and Requirements}
\label{Sec:keystakeholders}

Before establishing criteria for performant ML, one must first understand who would be involved in applying an MLWF within the context of industry.
Thus, Section~\ref{Sec:primary_stake} details the primary stakeholders that would be, and presently are, most engaged with the usage and outputs of AutoML.
Essentially, the discussion lists what these groups care about and how AutoML may factor into meeting these needs.
Secondary stakeholders are also briefly noted in Section~\ref{Sec:secondary_stake}, as their desires will likewise influence the continuing evolution of AutoML technology.

\subsubsection{Primary Stakeholders}
\label{Sec:primary_stake}

When discussing this collective, there are essentially two subcategories: data scientists and other technical users.
The latter term encompasses those who have the potential to use AutoML technologies directly but are not expert data scientists.
There naturally exists a spectrum on which such potential users can fit, from the trained professional to the technophobe with limited computer literacy.
This type of stakeholder, therefore, represents users who may lack technical skills but are valuable to an ML application due to their increased domain knowledge.
However, all primary stakeholders are still defined here by their close participation in ML analysis rather than involvement with any general infrastructure/architecture.
Thus, we exclude IT and generic data-engineering roles that would not typically interface with an AutoML system.
Also, it is understood that modern organisations are fluid, and roles may be transitional or hybrid, but the following categorisation should still be sufficient in encompassing the employment space related to ML applications.

\textbf{Data Scientists.}
This group of technicians represents the most prominent core stakeholder in AutoML technology.
Granted, democratisation is a central aim of the AutoML endeavour, but such a process is gradual and ongoing.
Realistically, data scientists remain the primary users interfacing with ML tools, meaning their needs dominate any discussion of the requirements that AutoML must satisfy.
Such considerations can be summarised as follows:
\begin{itemize}
	\item \textbf{Efficiency.}
	One of the key appeals of automation is that, ideally, it speeds up processes substantially.
	After all, machines are generally better than humans at formulaic tasks, maintaining high levels of consistency and endurance.
	The resulting procedural fluidity can be valuable to data scientists in two main forms.
	\begin{itemize}
		\item \textbf{Operational Efficiency.}
		This concept refers to saving time and effort expended by the staff of an organisation when managing a technical process.
		In context, data scientists will often form their own personal workflows for expediently tackling ML problems, but the manual application of these can still have high starting costs.
		For instance, template-driven approaches may need to be adjusted and tweaked per ML problem, while those who have not invested in such practices will likely need to code from scratch.
		Accordingly, there are many points along an MLWF where the automation of existing practices can speed up operations substantially.
		Crucially, none of these involve the science of ML; operational efficiency merely relates to the logistics that support an ML application.
		
		Of course, the desire to streamline work processes is common throughout industries focussed on maximising productivity and minimising cost.
		With the high salaries commanded by data science talent, there is an organisational impetus to mechanise operations that are high-volume and low-value, e.g.~via robotic process automation (RPA), so that data scientists are employed where their technical skills will have the greatest impact.
		Indeed, an IBM survey~\cite{ib21} found that, alongside saving costs ($58\%$) and freeing valuable time for employees (42\%), driving greater efficiencies ($58\%$) was a top reason for businesses using or considering automation tools.
		Nonetheless, interviews with data scientists~\cite{wawe19} indicate that employees also appreciate the prospective benefits of increased operational efficiency that AutoML may offer.
		
		Admittedly, because the interplay between automation and ergonomics is complicated, it can be challenging to draw bounds on the scope of AutoML under this requirement.
		For instance, the automation of project maintenance via Git, a VCS that a 2021 Stackoverflow developer survey~\cite{st21} found was used by over $93\%$ of 80000 respondents, will have had an undeniable impact on supporting streamlined ML.
		Automating collaboration between team members is another nuanced driver of efficient ML applications, almost ubiquitously ignored by academia, and data scientists have expressed a desire for tools that enhance communication and associated productivity~\cite{pa20b}.
		
        \item \textbf{Technical Efficiency.}
		This concept refers to a technical process running faster or with fewer resources.
		In context, this generally relates to the time and memory footprint involved in developing, deploying and maintaining an ML solution.
		At one implementational level, data scientists may appreciate efficiencies arising from reducing the time complexity of algorithmic processes, e.g.~by vectorising looped tasks.
		However, the development and release of theoretically novel algorithms can also significantly impact the speed of ML.
		In fact, the field of modern AutoML launched on the back of expedient ML model selection, as reviewed previously~\cite{kemu20}.
		Accordingly, while many data scientists will have some degree of reticence when adopting unfamiliar techniques, sufficient technical efficiency, exemplified by the history of convolutional neural networks, can overcome this barrier to uptake.
	\end{itemize}
    
    \item \textbf{Technical Performance.}
	While this review does not focus on metrics related to the standard `correctness' of ML models, this is primarily due to how heavily the concept has been discussed elsewhere.
	Certainly, it is inescapable that data scientists and dependent stakeholders seek ML solutions that are sufficiently representative of some desirable function, often a ground truth.
	However, while academia often seeks to push the limits of model validity, the costs and diminishing returns can be prohibitive within an industry setting.
	Research circles have noted such concerns, with some discussing how good is good enough \cite{foov18, laol20, thgr21}.
	
	That stated, AutoML is yet to shrug off an association with mixed technical performance.
	While developers of AutoML packages tend to promote the predictive power of their mechanisms and frameworks, independent benchmarks vary.
	Some suggest automated techniques achieve mediocre results compared to humans~\cite{zohu21}, some are more favourable~\cite{lu19}, and yet others sit in the middle, e.g.~stating that AutoML performs equal or better on 7 out of 12 tasks~\cite{habl20}.
	Although not an academic work, a recent poll on KDNuggets~\cite{kd20} likewise indicates this subdued outlook among data scientists more generally, asking the following question: ``How well do current AutoML solutions work, in your opinion?''
	With a scaling from 1 to 5, i.e.~`badly' to `super-human', the poll returned an average score of 2.4.
	However, there was a notable difference in average scores between those who tried AutoML (2.56) and those with only preconceptions (2.29).
	Moreover, no consideration was given to which tool was used.
	Ultimately, although the technical efficiency of AutoML does make it easier to reach improved technical performance, which is generally appealing to data scientists, it is not yet clear whether improved model validity should even be a primary selling point of AutoML.

    \item \textbf{Methodological Currency.}
    Essentially, other things being equal, data scientists prefer to operate as close to SOTA as possible.
	However, the modern AI field is progressing fast, and advances are constantly being made across the entire standard MLWF.
	This evolution is not a monolithic affair either.
	For instance, novel auto-augmentation data-engineering methods will likely have little reference to new deployment techniques for field-programmable gate arrays (FPGAs), even if both may be relevant to a DL application~\cite{doke21}.
	Accordingly, the so-called `unicorn' data scientist that is an expert in all the niche skills and topics across an MLWF is extremely rare, if not outright nonexistent~\cite{bako17, esga19}.
	Even keeping abreast of ML modelling alone can be challenging, noting that the industry-leading scikit-learn library -- it has Sklearn as an alias -- has, in version 0.24.2, 191 available estimators in the form of classifiers, regressors, clustering methods, and transformers~\cite{sklearnallestimators}.
	Given the already amorphous role of a data scientist at present~\cite{migi18}, a standard representative of this stakeholder group is likely to have varying degrees of expertise in different algorithms, HPO techniques, and other technical processes.
	Thus, AutoML can provide value to a data scientist by supporting access to unfamiliar techniques, whether brand new or renascent, subsequently improving operational and technical efficiencies.
    
    \item \textbf{Ease of Use.}
	Regardless of efficiency, a technical or operational process in an ML application loses its appeal if a data scientist cannot interface with it effectively.
	Of course, assessing ease of use for any computational tool is somewhat subjective, depending on who is using it and what they are using it for.
	Poor design or dependence on overly specialised skills can immediately hamper uptake.
	However, data scientists also vary in their personal preferences.
	Some are comfortable with code, some may seek a command-line interface (CLI) for its perceived simplicity and accessibility, and yet others will desire a graphical user interface (GUI) to interact with technical products~\cite{faol03}.
	This notion of a convenient user interface (UI) within the context of AutoML is reviewed deeper elsewhere~\cite{khke21}.
	
	Beyond personal preferences, technical tools score higher with data scientists if they are fit for purpose, integrating well into an existing MLWF and addressing specific use cases.
	For instance, when using ML to predict purchasing propensity in e-commerce, a technical stakeholder would likely appreciate any convenient method of accessing and manipulating data related to sales, customer demographics, website activity, etc.
	Granted, this lies within the purview of automated data engineering, but the emphasis here is on effectively configuring operational/technical processes for a specific use case.
	As another example, consider a data scientist working with a recommendation engine.
	Rather than composing a standard error measure over all samples, the stakeholder may prefer to work with a precision evaluation on some top-N recommendations~\cite{gush09}, an exponential decay that notes users are less likely to pick items down a ranked list~\cite{brhe98}, or some other non-standard metric~\cite{heko04}.
	As of the early 2020s, most AutoML packages strive to be as generally applicable as possible, but Section~\ref{Sec:specialisedtools} does provide limited examples of modern tools that conveniently specialise.
	
	As a final note, while an expansive set of programming approaches does exist, data scientists have clustered around certain popular open-source languages and frameworks for ML.
	Developing in these spaces automatically improves ease of use.
	Specifically, a 2019 KDnuggets Software Poll~\cite{kg19} highlighted Python and R as preferred languages from 2017 to 2019, although with a yearly decrease for R.
	It also listed Keras, scikit-learn and TensorFlow as popular ML libraries.
	These results were further corroborated by the Kaggle State of Machine Learning and Data Science report in 2020~\cite{ka21}, which was notably more focussed on practising data scientists.
	This report ranked scikit-learn, TensorFlow and Keras as the top ML frameworks, while also revealing that the top three languages regularly used by respondents -- multiple selections were allowed -- were Python (15530), SQL (7535), and R (4277).
	Additionally, at a ratio of 14241 to 1259, respondents recommended Python over R as the first language an aspiring data scientist should learn.
	Admittedly, it is unclear whether these programming languages will maintain a stranglehold on the mainstream in the long-term, with newer entrants like Julia acquiring small but growing fanbases~\cite{TiobeIndex}.
    
    \item \textbf{Explainability.}
	Understanding how an ML solution came to be and why it says what it says has surged in importance within academia over the last several years.
	However, this is not an unfamiliar requirement to data scientists who regularly interact with business stakeholders; part of the job is translating work and outputs into an understandable format.
	Unsurprisingly, an ability to communicate well is often cited as a core component of the skill profile for such a technician~\cite{cosa17, baba20a}.
	Indeed, a seminal 2012 Harvard Business Review article defined a data scientist as ``a hybrid of data hacker, analyst, communicator, and trusted adviser''~\cite{dapa12}.
	Likewise, an IBM survey~\cite{ib21} found that $91\%$ of businesses using AI say their ability to explain how a decision was arrived at is critical.
	Accordingly, many data scientists will appreciate computational tools that provide insight into what exactly they do.
	Satisfying this requirement boosts operational efficiency, but it also improves trust in an ML solution.
	This desire for transparency around data and ML modelling has been corroborated by recent interviews~\cite{drwe20}, albeit limited both in number and to students.
	Another set of interviews surveying 20 professional data scientists, limited by their association with the same organisation, likewise found a consensus need to surface what was done, e.g.~what algorithms or preprocessing techniques were used, and how it was done, e.g.~what hyperparameter values were chosen.
	Clearly, explainability and associated trust are essential issues to stakeholders, and a deeper dive into these topics is available elsewhere~\cite{khke21}.
    
\end{itemize}

\textbf{Analysts.}
This group of primary stakeholders is the first that can be considered to encompass `other' technical users.
Analysts typically have moderate exposure to techniques and technology involving data, possessing strong skills with popular business software, e.g.~Microsoft Excel, as well as reasonable fluency in SQL and some exposure to R and Python~\cite{da21, ma16}.
However, data visualisation will generally be enacted via popular software applications such as Tableau and PowerBI~\cite{ga21a}, rather than a technical coding library.
Of course, this is a generalisation as the spectrum of proficiency is broad.
Now, while the requirements of an analyst cover the same scope as a data scientist, priorities tend to differ.
Rather than technical efficiencies, given that analysts do not commonly practice ML and are thus unconcerned with optimising such processes, \textbf{ease of use} becomes particularly important.
Additionally, the core job function for an analyst requires proximity to business stakeholders, so one would also seek tools with a high degree of \textbf{explainability}.
Unlike data scientists, who lean towards understanding ML processes to instil confidence in the rigour and validity of an ML solution, analysts generally need explainability to bridge the gap between technical and non-technical stakeholders, as required by business intelligence/analytics (BI/BA) roles~\cite{wiar14, degr18, cejo16}.
Naturally, analysts working with AutoML tools are still likely to desire strong \textbf{technical performance} from an ML solution, but their standards will differ from that of a data scientist, who is far more likely to have benchmarked such metrics and is aware of what is currently SOTA.
Essentially, perspectives will be `anchored' differently depending on stakeholder experience with a technology thus far~\cite{fubo11}.

\textbf{Business Users.}
This group is yet another step removed from the technical expertise generally required for direct ML involvement.
Admittedly, many existing AutoML vendors market themselves as operable by ML novices; one could propose that any accountant, lawyer, line manager or other business stakeholder could participate in loading data, undertaking ML and deploying performant solutions as part of their decision-making workflow.
However, this is a lofty ideal even before considering the survey results in the rest of this review.
Several factors also complicate matters.
Firstly, business users are unlikely to have confidence using ML tools, even if ease of use is outstanding.
Indeed, despite visualisation tools and other methods for supporting explainability, AutoML-assisted technical operations, e.g.~the deployment of an ML solution, are likely to remain daunting.
Secondly, this type of stakeholder is unlikely to consider direct ML involvement within their remit.
The creation and management of analytical models typically fall to data scientists or analysts, and any organisational dearth of expertise here is probably better met by hiring talent to fill the gap.
In fact, it has been argued that enabling non-technical users to run an ML application may even be harmful~\cite{kr18, vega19, boko19}.
Nonetheless, business users remain critical stakeholders in an ML application, often acting as both the driving force and beneficiaries of its outputs.
Certainly, manager functions within business units are those that commission bodies of work to be completed and expect results from that expenditure of effort.
Thus, \textbf{technical performance} and \textbf{efficiency} are paramount, although more through a return-on-investment (ROI) lens that considers staffing time and organisational resources.
Conversely, any characteristic around interfacing with an AutoML system, e.g.~usability or explainability, is likely to be less of a direct concern, as organisations will usually rely on data scientists and analysts for reporting.
Indirectly, though, business users will still benefit from such facets, as they require confidence in an ML product and its alignment to business objectives.

\textbf{Deployment Technicians.}
This group encompasses those who move experimental ML solutions into production.
In smaller organisations, this role may blend with that of a data scientist, but larger businesses or those dealing with more mature technical functions often have a separate department dedicated to the policies, procedures and processes behind deploying technical products.
Normally, once an ML model has been created and is ready for consumption, it sits as an object within the same technical scope as other business applications, i.e.~ingesting data, writing data, and interacting with other systems.
How this solution is consumed will vary based on the intent behind an ML application, but modern business practices have established standardised roles focussed on deployment.
Traditionally called DevOps~\cite{aw00}, these functions, if specific to ML, are starting to be referred to as MLOps~\cite{ga20, da00, me20a}.
For those tasked with associated responsibilities, an ML pipeline is usually considered sacrosanct; its experimental accuracy is unquestioned, and matters such as explainability are irrelevant.
Instead, more technical considerations related to infrastructure are essential, and these have been the focus of various studies~\cite{crse18, dupa17, lovi20, scmu16, spta15, spve17}.
Like data scientists, deployment technicians care about \textbf{efficiency}, albeit in matters of inference and maintenance rather than model training, and they would also seek \textbf{methodological currency} from AutoML packages, given how quickly hardware and deployment techniques can evolve, e.g.~FPGAs and federated ML. 
Other considerations relating to an ML application include the ability to scale well~\cite{zohu21}, continously update~\cite{yash20}, handle dirty data~\cite{hanu20}, and adapt robustly to concept drift~\cite{maes19}.

\subsubsection{Secondary Stakeholders}
\label{Sec:secondary_stake}

This collective is not generally invested in a specific ML application like the primary stakeholders listed above.
However, the category remains important in discussing AutoML, as its constituents have roles and responsibilities that will both impact and be impacted by intensifying uptake of associated technologies.
Indeed, disregarding the requirements of the following organisational stakeholders would render an incomplete understanding of the dynamics that drive AutoML adoption in enterprise use cases.
Importantly, as before, we do not consider specific job titles here due to the fluidity of definitions within the modern workforce, instead focussing on organisational roles and responsibilities.

\textbf{Corporate Management.}
This group of secondary stakeholders encompasses the finance, human resources and other management units within an organisation, save for those included within a separate `risk and governance' subcategory below.
Crucially, for any organisation in the private sector or elsewhere, the allocation of finite resources is a strong motivator and constraint for business decisions and activities; corporate management is closely tied to those considerations.
Indeed, a recent Boston Consulting Group survey of senior executives at 1034 large organisations~\cite{bc21} found that the most significant driver for responsible AI use related to business benefits, as declared by $40\%$ of respondents.
This motivation was followed by customer expectations ($20\%$), risk mitigation ($16\%$), and regulatory compliance ($14\%$).
Of course, maintaining the health of a business organisation manifests in diverse requirements of a performant ML application, not all simple and direct.
For instance, an IBM survey, previously mentioned~\cite{ib21}, found that \textbf{explainability} is important to corporate management.
In fact, a CapGemini survey reveals that the proportion of executives interested in this area has increased from $32\%$ in 2019 to $78\%$ in 2020~\cite{ca20}.
However, such a stakeholder is typically not interested in understanding a specific ML model; they are frequently answerable to external entities, e.g.~customers and regulatory bodies, and thus adopt their interests.
Then there is \textbf{ease of use}, which corporate management is unlikely to ever avail itself of directly, but investing in accessible ML tools does benefit staff training and acquisition.
Granted, other requirements that AutoML could satisfy are more straightforwardly justified.
Strong \textbf{technical performance} of ML solutions can provide a competitive advantage within an industry and generate revenue.
Good \textbf{efficiency} can similarly save money, e.g.~operational efficiency frees time for existing staff and technical efficiency may save server costs.

\textbf{Risk and Governance Entities.}
This group is essentially dedicated to avoiding harm related to business practices, which, in context, refers specifically to the processes and outcomes of an ML application.
Given that conditions of uncertainty are inevitable within real-world settings, it is up to these entities, alongside management, to understand and mitigate associated risks.
Data governance, for instance, is increasingly of corporate interest, with the biannual Information Governance ANZ group survey finding in 2021 that $64\%$ of organisations had adopted a formal Information Governance (IG) framework, implementing associated policies and procedures, and 74\% had IG projects underway or planned for the following year~\cite{in21}.
Two years earlier, only $51\%$ were using a formal IG framework~\cite{in19}.
Evidently, there is a growing consensus that the implementation of an advanced technology should be subject to IG considerations and risk-based oversight.
Specifically, ML applications and the automation of their higher-level processes are expected to align with data governance and security practices via controllable access to data, models, and technical functionality.
Essentially, the running of an MLWF should be auditable.

Notably, there are many ways related to ML in which risk may arise, and some have nothing to do with its technical processes.
For instance, the field of data science is presently beset with significant variability in the skills and preferred approaches of its practitioners.
Admittedly, one could argue this is an issue in any industry that is heavily dependent on human expertise, e.g.~medicine or law.
However, the relative immaturity of industrial data science means that no educational or experiential thresholds are commonly agreed upon to signify that someone is a data scientist~\cite{deag17, hiir18, we18, krpe20}.
In fact, there is currently a proliferation of online courses and boot camps to assist people transitioning into the field~\cite{cava20}, many of debatable quality, and the absence of regulation means that it is not uncommon for prospective employees to simply change their job titles.
This inconsistency in skills and approaches can damage the quality of ML outputs, which is particularly dangerous in high-stakes problem contexts.
Moreover, even if every professional data scientist were a genuine expert, the lack of standardisation can still cause issues with reproducibility, which is a prominent concern across all sciences, including ML~\cite{pe15, bema20, haad20, rodr21}.
Thus, for stakeholders dedicated to risk and governance, AutoML has the appealing potential, in theory, to provide consistency and transparency in the application of ML, establishing a robust baseline in the practice of data science.
On the other hand, AutoML packages must themselves have appropriate safeguards for such an ideal to be realised, as ease of use erodes the accessibility barrier that prevents non-experts from inducing errors and possible harm via their ignorance.

Now, when discussing overlapping requirements with a data scientist, a naive expectation is that solid \textbf{technical performance} will stimulate trust in an ML solution simply by virtue of generating valid ML predictions/prescriptions.
However, the standard application of ML, even with a substantially accurate model, does not consider many nuanced drawbacks that can make an ML solution a poor fit for a real-world context.
These nuances can be very subtle, which is why any auditing bodies predominantly require \textbf{explainability} from the tools and processes used in an MLWF, if only to seek trust through transparency~\cite{toai20}.
Indeed, trust in AI has recently surged in importance within both academic and industrial circles of discussion~\cite{wawu19, drwe20, toai20, na21}.
Chief among the factors that can threaten trust are issues of bias and fairness, of which the public has become more aware and critical as data science progressively seeps into the lives of common people~\cite{nd21, sa21}.
The aforementioned IBM report~\cite{ib21} found $87\%$ of respondents professed that ``ensuring applications and services minimise bias'' is an essential aspect of AI.
However, the report also noted that skills shortages and a lack of assistive tools are the most considerable barriers to developing/managing trust in AI.
Naturally, this pressure for trustworthy ML has a financial motivation for many businesses, passed on from customers; a recent CapGemini survey of 800 organisations and 2900 consumers~\cite{ca20} found that $71\%$ of the latter want a clear explanation of results and $66\%$ expect AI to be fair and free of bias.
This expectation has increased awareness of AI discrimination among surveyed executives, from $35\%$ in 2019 to $65\%$ in 2020.
Sure enough, modelling in the literature posits that ignoring the societal requirement for debiasing can adversely impact business demand and associated profits~\cite{ukru21}.
Additionally, surveys~\cite{memo19}, taxonomies~\cite{mipo18} and instructional research~\cite{do17} are accumulating on this topic, sometimes concerning specific fields and applications, such as medicine~\cite{gita18} and hiring practices~\cite{dach19}, respectively.
Simply put, risk and governance entities are likely to desire increasing capabilities of assessing/managing bias and fairness within ML applications, and this will hold true of AutoML as well.

Finally, we note that this set of requirements, traditionally neglected by frontier ML theory and technology, cannot be ignored for long.
While extensive laws lag substantially behind the pace of ML progress, calls for regulation are gaining political traction worldwide~\cite{ah19}.
For instance, consideration of the issue has appeared in the Australian 2021-2022 Budget Fact sheets~\cite{of21}, while the US White House is launching a task force, i.e.~the National AI Research Resource, with a partial eye to such matters~\cite{th21}.
As some other examples, Standards Australia has recently developed a roadmap related to AI practices~\cite{au20} and the Office of the Australian Information Commissioner has published a `Guide to data analytics and the Australian Privacy Principles'~\cite{co18}.
Elsewhere, in April 2021, the European Union released a proposal for the regulation of AI among member states~\cite{un21}.
In essence, the societal context in which organisations employ ML is evolving, and business leaders are highlighting trust and explainability in ML as vital for meeting regulatory and compliance obligations~\cite{ib21}.
This review will not delve deeper into the international laws and policies being established for AI; the key takeaway is that an MLWF and its automation will be subject to increasing regulatory oversight in the coming years, especially within industries such as finance, law enforcement, and medicine.
Accordingly, organisations would likely appreciate transparency from associated tools and processes, as well as considerations of bias and fairness, should tension between business objectives and regulatory requirements ever arise.

\subsection{Unified Criteria}
\label{Sec:summarisedcriteria}

Having established the core requirements of primary/secondary stakeholders that engage with an ML application and associated tools, we can now distil and outline the key criteria by which AutoML in an industry setting can be assessed as supporting performant ML.
This proposed framework will anchor the subsequent review of open-source packages, both specialised in Section~\ref{Sec:ancillarytools} and holistic in Section~\ref{Sec:coretools}, as well as commercial offerings in Section~\ref{Sec:commercialtools}.
To best aid such an effort, each criterion below has been broken down further into several questions and associated scoring methods.
The questions have been designed to be answerable with publicly available information, if it exists, i.e.~source code and documentation for open-source tools or vendor websites for commercial products.
Additionally, these questions are slanted to recognise major challenges that face the ongoing uptake of AutoML technology, with one academic work~\cite{lema19} suggesting that obstacles fall into three main areas: search, technical speed/performance, and HCI.
In fact, given that grappling with these challenges is a continuous process, the responses to the proposed questionnaire are not always binary, e.g.~`no automation' or `full automation'.
Convenience functions and features that assist a user with an ML task, which suggest partial progress towards automation, warrant acknowledgement.
With that all stated, we now proceed to list the criteria.

\begin{itemize}

\item \textbf{Technical Performance.}
Any AutoML product that supports performant ML will always be judged by certain core metrics, i.e.~its potential to set/improve the `correctness' of an ML model.
This review aims to extend beyond such considerations, as necessary levels of solution validity differ dramatically between industries, use cases, risk profiles, and organisational agendas.
Some businesses can operate at the SOTA frontier, while others are dabbling in ML technologies for the first time.
Likewise, $25\%$ accuracy for a music recommendation service may be fantastic, while $75\%$ accuracy for a tumour classification system may be abysmal.
In short, predictive/prescriptive `correctness' is undoubtedly essential, but it is far from the be-all and end-all of ML requirements.
It is also the criterion that we do not delve into within this assessment framework for performant ML; experimental research is required to validate the technical performance of any AutoML system, and this is out of scope for this review.
Such attempts at benchmarking are also already numerous within the literature~\cite{gile19, trwa19b, baba20, depi20, zohu21}.

\item \textbf{Efficiency (22 Questions).}
As outlined in Section~\ref{Sec:primary_stake}, the pace and cost of running an MLWF are set on two fronts: operational and technical.
The former relates to processes that determine how effectively/productively the work of employees can be translated into advancing tasks within an MLWF.
The latter relates to the speed and resource consumption involved in developing, deploying and maintaining the technical ML solution itself.
Thus, several categories of questions have been established to evaluate how well an AutoML tool assists with overall efficiency.
First, there is an assessment of the effort required in tracking and managing experimentation during ML modelling.
Second, there is consideration around how easy it is to utilise prior art/work.
Packages rate highly on this sub-criterion if they are (1) capable of storing/managing a history of previous ML applications and (2) able to leverage that previous experience for future recommendation, e.g.~via meta-learning.
Efficient collaboration also aids in awareness of prior art, so evaluating its presence is included in this category.
Third, there is a determination of how much effort can be saved along work-intensive portions of an MLWF, i.e.~data exploration/preparation, feature engineering/selection, and actual modelling.
Because data preparation is a particular time-sink, it merits an extended set of spin-off questions at this point.
Finally, there is an appraisal of AutoML features, e.g.~configuration control, that may support technical efficiencies beyond those intrinsically linked to technical performance.
We now explicitly list the questions on efficiency.

\begin{longtable}{|C{1.5cm}|L{2cm}|L{2.5cm}|C{1cm}|C{1.5cm}|L{2.5cm}|}
\caption{Assessment Framework for Efficiency. Questions: E1-E3.}
\label{Table:Criteria_E1-3}
\\ \hline
\rowcolor[HTML]{B4C6E7} 
\multicolumn{1}{|c|}{\textbf{Criteria}} &
  \multicolumn{1}{c|}{\textbf{Sub-Criteria}} &
  \multicolumn{1}{c|}{\textbf{Question}} &
  \textbf{QCode} &
  \textbf{Scoring} &
  \multicolumn{1}{c|}{\textbf{Rubric}} \\ \hline\endhead
Efficiency & Effort in experiment management \& tracking & Does it provide   a model repository? & E1 & 0/1 & \begin{tabular}[t]{@{}L{\linewidth}}0: No\\      1: Yes\end{tabular} \\ \hline
Efficiency & Effort in experiment management \& tracking & Does   it provide model VCS?          & E2 & 0/1 & \begin{tabular}[t]{@{}L{\linewidth}}0: No\\      1: Yes\end{tabular} \\ \hline
Efficiency &
  Effort in experiment management \& tracking &
  Does   it provide experiment tracking features? &
  E3 &
  Scale 0:2 &
  \begin{tabular}[t]{@{}L{\linewidth}}0: No\\ 1: Yes for log storage/access, but with limited automation and/or visuals\\      2: Yes, with automatic log visualisation\end{tabular} \\ \hline
\end{longtable}

The sub-criteria in Table~\ref{Table:Criteria_E1-3} can be mapped to the MLWF in Fig.~\ref{Fig:mlworkflowsubtasks} as follows: E1 to \textit{Find Prior Art} within \textit{Problem Formulation \& Context Understanding} and E2/E3 to \textit{Experiment Tracking \& VCS} within \textit{Model Development} and \textit{VCS} within \textit{Monitoring \& Maintenance}.

\begin{longtable}{|C{1.5cm}|L{2cm}|L{2.5cm}|C{1cm}|C{1.5cm}|L{2.5cm}|}
\caption{Assessment Framework for Efficiency. Questions: E4-E6.}
\label{Table:Criteria_E4-6}
\\ \hline
\rowcolor[HTML]{B4C6E7} 
\multicolumn{1}{|c|}{\textbf{Criteria}} &
  \multicolumn{1}{c|}{\textbf{Sub-Criteria}} &
  \multicolumn{1}{c|}{\textbf{Question}} &
  \textbf{QCode} &
  \textbf{Scoring} &
  \multicolumn{1}{c|}{\textbf{Rubric}} \\ \hline\endhead
Efficiency &
  Effort in leveraging prior work \& collaboration &
  Does it offer a template/code repository? &
  E4 &
  Scale 0:2 &
  \begin{tabular}[t]{@{}L{\linewidth}}0: No\\ 1: Yes, templates/code can be generated by users \\ 2: Yes, templates/code can automatically kickstart projects\end{tabular} \\ \hline
Efficiency &
  Effort in leveraging prior work \& collaboration &
  Does   it suggest prior work? &
  E5 &
  0/1 &
  \begin{tabular}[t]{@{}L{\linewidth}}0: No\\ 1: Yes\end{tabular} \\ \hline
Efficiency &
  Effort in leveraging prior work \& collaboration &
  Does it facilitate project collaboration? &
  E6 &
  Scale 0:2 &
  \begin{tabular}[t]{@{}L{\linewidth}}0: No\\ 1: Yes, with basic features such as shared access to folders with project artefacts\\ 2: Yes, with advanced features\end{tabular} \\ \hline
\end{longtable}

The sub-criteria in Table~\ref{Table:Criteria_E4-6} can be mapped to the MLWF in Fig.~\ref{Fig:mlworkflowsubtasks} as follows: E4/E5 to \textit{Find Prior Art} within \textit{Problem Formulation \& Context Understanding} and E6 to \textit{Collaboration} within \textit{Problem Formulation \& Context Understanding}.

\begin{longtable}{|C{1.5cm}|L{2cm}|L{2.5cm}|C{1cm}|C{1.5cm}|L{2.5cm}|}
\caption{Assessment Framework for Efficiency. Questions: E7--E13.}
\label{Table:Criteria_E7-13}
\\ \hline
\rowcolor[HTML]{B4C6E7} 
\multicolumn{1}{|c|}{\textbf{Criteria}} &
  \multicolumn{1}{c|}{\textbf{Sub-Criteria}} &
  \multicolumn{1}{c|}{\textbf{Question}} &
  \textbf{QCode} &
  \textbf{Scoring} &
  \multicolumn{1}{c|}{\textbf{Rubric}} \\ \hline\endhead
Efficiency &
  MLWF Effort:   Data Exploration &
  Does it   automatically generate visualisations to assist in data exploration? &
  E7 &
  Scale 0:3 &
  \begin{tabular}[t]{@{}L{\linewidth}}0: No\\ 1: No, but convenience features are available\\ 2: Yes, to some degree\\ 3: Yes, and with automatic notification of issues or points of interest\end{tabular} \\ \hline
Efficiency &
  MLWF   Effort: Data Preparation &
  Does   it automatically prepare data for modelling? &
  E8 &
  Scale 0:2 &
  \begin{tabular}[t]{@{}L{\linewidth}}0: No\\ 1: No, but convenience features are available\\ 2: Yes, to some degree\end{tabular} \\ \hline
Efficiency &
  MLWF   Effort: Feature Engineering &
  Does   it automatically engineer features? &
  E9 &
  Scale 0:2 &
  \begin{tabular}[t]{@{}L{\linewidth}}0: No\\ 1: No, but convenience features are available\\ 2: Yes, to some degree\end{tabular} \\ \hline
Efficiency &
  MLWF   Effort: Feature Engineering &
  Does   it store features for later use by others? &
  E10 &
  0/1 &
  \begin{tabular}[t]{@{}L{\linewidth}}0: No\\ 1: Yes\end{tabular} \\ \hline
Efficiency &
  MLWF   Effort: Feature Selection &
  Does   it automatically select features? &
  E11 &
  Scale 0:2 &
  \begin{tabular}[t]{@{}L{\linewidth}}0: No\\ 1: No, but convenience features are available\\ 2: Yes, to some degree\end{tabular} \\ \hline
Efficiency &
  MLWF   Effort: Modelling &
  Does it specify HPO search spaces and algorithms by default? &
  E12 &
  0/1 &
  \begin{tabular}[t]{@{}L{\linewidth}}0: No\\ 1: Yes\end{tabular} \\ \hline
Efficiency &
  MLWF   Effort: Modelling &
  Does   it optimise an entire ML pipeline? &
  E13 &
  0/1 &
  \begin{tabular}[t]{@{}L{\linewidth}}0: No\\ 1: Yes\end{tabular} \\ \hline
\end{longtable}

The sub-criteria in Table~\ref{Table:Criteria_E7-13} can be mapped to the MLWF in Fig.~\ref{Fig:mlworkflowsubtasks} as follows: E7 to \textit{Explore \& Assess Fairness} within \textit{Data Engineering}, E8 to \textit{Clean} and \textit{Prepare} within \textit{Data Engineering}, E9 to \textit{Feature Engineering} within \textit{Data Engineering}, E10 to \textit{Find Prior Art} within \textit{Problem Formulation \& Context Understanding}, E11 to \textit{Feature Selection} within \textit{Model Development}, E12 to \textit{HPO} within \textit{Model Development}, and E13 to \textit{Data Engineering} and \textit{Model Development} generally.

\begin{longtable}{|C{1.5cm}|L{2cm}|L{2.5cm}|C{1cm}|C{1.5cm}|L{2.5cm}|}
\caption{Assessment Framework for Efficiency. Questions: E8A--E8F.}
\label{Table:Criteria_E8A-F}
\\ \hline
\rowcolor[HTML]{B4C6E7} 
\multicolumn{1}{|c|}{\textbf{Criteria}} &
  \multicolumn{1}{c|}{\textbf{Sub-Criteria}} &
  \multicolumn{1}{c|}{\textbf{Question}} &
  \textbf{QCode} &
  \textbf{Scoring} &
  \multicolumn{1}{c|}{\textbf{Rubric}} \\ \hline\endhead
Efficiency & MLWF Effort:   Data Preparation & Does it automate categorical   feature Processing?        & E8A & 0/1 &  \begin{tabular}[t]{@{}L{\linewidth}}0: No\\ 1: Yes\end{tabular} \\ \hline
Efficiency & MLWF   Effort: Data Preparation & Does it automate standardisation and normalisation?           & E8B & 0/1 &  \begin{tabular}[t]{@{}L{\linewidth}}0: No\\ 1: Yes\end{tabular} \\ \hline
Efficiency & MLWF   Effort: Data Preparation & Does it automate bucketing and binning?                  & E8C & 0/1 &  \begin{tabular}[t]{@{}L{\linewidth}}0: No\\ 1: Yes\end{tabular} \\ \hline
Efficiency & MLWF   Effort: Data Preparation & Does it automate text preprocessing?                      & E8D & 0/1 &  \begin{tabular}[t]{@{}L{\linewidth}}0: No\\ 1: Yes\end{tabular} \\ \hline
Efficiency & MLWF   Effort: Data Preparation & Does it automate time-period extraction?                & E8E & 0/1 &  \begin{tabular}[t]{@{}L{\linewidth}}0: No\\ 1: Yes\end{tabular} \\ \hline
Efficiency & MLWF   Effort: Data Preparation & Does it assist with class   imbalance via sampling techniques? & E8F & 0/1 &  \begin{tabular}[t]{@{}L{\linewidth}}0: No\\ 1: Yes\end{tabular} \\ \hline
\end{longtable}

The sub-criteria in Table~\ref{Table:Criteria_E8A-F} can be mapped to the MLWF in Fig.~\ref{Fig:mlworkflowsubtasks} as follows: E8A--E8F to \textit{Prepare} within \textit{Data Engineering}.

\begin{longtable}{|C{1.5cm}|L{2cm}|L{2.5cm}|C{1cm}|C{1.5cm}|L{2.5cm}|}
\caption{Assessment Framework for Efficiency. Questions: E14--E16.}
\label{Table:Criteria_E14-16}
\\ \hline
\rowcolor[HTML]{B4C6E7} 
\multicolumn{1}{|c|}{\textbf{Criteria}} &
  \multicolumn{1}{c|}{\textbf{Sub-Criteria}} &
  \multicolumn{1}{c|}{\textbf{Question}} &
  \textbf{QCode} &
  \textbf{Scoring} &
  \multicolumn{1}{c|}{\textbf{Rubric}} \\ \hline\endhead
Efficiency &
  Technical Efficiency &
  Does it undertake workload   optimisation? &
  E14 &
  Scale 0:2 &
  \begin{tabular}[t]{@{}L{\linewidth}}0: No\\ 1: No, but convenience features are available\\ 2: Yes, to some degree\end{tabular} \\ \hline
Efficiency & Technical   Efficiency & Does it allow   time limits for modelling?           & E15 & 0/1 & \begin{tabular}[t]{@{}L{\linewidth}}0: No\\ 1: Yes\end{tabular} \\ \hline
Efficiency & Technical   Efficiency & Does it allow   iteration/trial limits for modelling? & E16 & 0/1 & \begin{tabular}[t]{@{}L{\linewidth}}0: No\\ 1: Yes\end{tabular} \\ \hline
\end{longtable}

The sub-criteria in Table~\ref{Table:Criteria_E14-16} can be mapped to the MLWF in Fig.~\ref{Fig:mlworkflowsubtasks} as follows: E14 to \textit{Provision Resources} within \textit{Model Development} and E15/E16 to \textit{CASH+} within \textit{Model Development}.

\item \textbf{Dirty Data (5 Questions).}
This criterion specifically considers how robust an AutoML system is in the face of messy data, e.g.~format issues, missing values, outliers, etc.
It deserves its own category due to the considerable time and effort that tends to be invested into related tasks; see Section~\ref{Sec:datafeaturetools}.
We now explicitly list the questions on dirty data.

\begin{longtable}{|C{1.5cm}|L{2cm}|L{2.5cm}|C{1cm}|C{1.5cm}|L{2.5cm}|}
\caption{Assessment Framework for Dirty Data. Questions: DD1--DD5.}
\label{Table:Criteria_DD1-5}
\\ \hline
\rowcolor[HTML]{B4C6E7} 
\multicolumn{1}{|c|}{\textbf{Criteria}} &
  \multicolumn{1}{c|}{\textbf{Sub-Criteria}} &
  \multicolumn{1}{c|}{\textbf{Question}} &
  \textbf{QCode} &
  \textbf{Scoring} &
  \multicolumn{1}{c|}{\textbf{Rubric}} \\ \hline\endhead
Dirty   Data &
  Dirty Data &
  Does it automatically clean dirty data? &
  DD1 &
  Scale 0:2 &
  \begin{tabular}[t]{@{}L{\linewidth}}0: No\\ 1: No, but convenience features are available\\ 2: Yes, to some degree\end{tabular} \\ \hline
Dirty   Data &
  Dirty   Data &
  Does   it automatically infer data types? &
  DD2 &
  0/1 &
  \begin{tabular}[t]{@{}L{\linewidth}}0: No\\ 1: Yes\end{tabular} \\ \hline
Dirty   Data &
  Dirty   Data &
  Does   it automatically find and deal with missing values? &
  DD3 &
  Scale 0:2 &
  \begin{tabular}[t]{@{}L{\linewidth}}0: No\\ 1: Partially, as it finds missing values\\ 2: Yes\end{tabular} \\ \hline
Dirty   Data &
  Dirty   Data &
  Does   it automatically find and deal with outliers? &
  DD4 &
  Scale 0:2 &
  \begin{tabular}[t]{@{}L{\linewidth}}0: No\\ 1: Partially, as it finds outliers\\ 2: Yes\end{tabular} \\ \hline
Dirty   Data &
  Dirty   Data &
  Does   it undertake other domain-specific or advanced data cleaning operations? &
  DD5 &
  0/1 &
  \begin{tabular}[t]{@{}L{\linewidth}}0: No\\ 1: Yes\end{tabular} \\ \hline
\end{longtable}

The sub-criteria in Table~\ref{Table:Criteria_DD1-5} can be mapped to the MLWF in Fig.~\ref{Fig:mlworkflowsubtasks} as follows: DD1--DD5 to \textit{Clean} within \textit{Data Engineering}.

\item \textbf{Completeness \& Currency (13 Questions).}
This criterion considers completeness through the lens of technical domain coverage, i.e.~the types of ML problems that an AutoML package can handle.
For instance, those constrained to binary classification tasks will only ever be able to assist organisations with a subset of business problems.
Thus, the major subset of questions under this criterion determines the ML applications suitable for an AutoML system.
Admittedly, an associated low score is not a problem for specialist tools, but it reflects poorly on any general applicability claims.
The most `complete' AutoML systems should additionally be configurable for arbitrary business domains, i.e.~by enabling custom evaluation metrics for ML solutions.
Beyond such a focus, there is an appraisal concerning the integration of HPO methods and libraries, the latter being chosen for evaluation over individual ML algorithms to ensure a degree of abstraction.
After all, as mentioned earlier, an interface to scikit-learn 0.24.2 immediately provides access to 191 estimators.
Finally, this criterion also assesses methodological currency, ensuring technical domain coverage uses up-to-date techniques and approaches.
However, notably, this monograph surveys only open-source and commercial AutoML packages that are in popular use as of the early 2020s, meaning that a lack of currency typically applies only to `faded' tools; see Appendix~\ref{Sec:fadedpurehpo} and Appendix~\ref{Sec:fadedcoreautoml}.
We now explicitly list the questions on completeness \& currency.

\begin{longtable}{|C{2cm}|L{2cm}|L{2cm}|C{1cm}|C{1.5cm}|L{3cm}|}
\caption{Assessment Framework for Completeness \& Currency. Questions: CC1--CC13.}
\label{Table:Criteria_CC1-9}
\\ \hline
\rowcolor[HTML]{B4C6E7} 
\multicolumn{1}{|c|}{\textbf{Criteria}} &
  \multicolumn{1}{c|}{\textbf{Sub-Criteria}} &
  \multicolumn{1}{c|}{\textbf{Question}} &
  \textbf{QCode} &
  \textbf{Scoring} &
  \multicolumn{1}{c|}{\textbf{Rubric}} \\ \hline\endhead
Completeness   \& Currency &
  Technical   Domain Coverage &
  How does it handle unsupervised learning? &
  CC1 &
  Scale 0:2 &
  \begin{tabular}[t]{@{}L{\linewidth}}0: Not at all\\ 1: Via convenience features or platform extensions \\ 2: Full AutoML\end{tabular} \\ \hline
Completeness   \& Currency &
  Technical   Domain Coverage &
  How does it handle regression on tabular data? &
  CC2 &
  Scale 0:2 &
  \begin{tabular}[t]{@{}L{\linewidth}}0: Not at all\\ 1: Via convenience features or platform extensions \\ 2: Full AutoML\end{tabular} \\ \hline
Completeness   \& Currency &
  Technical   Domain Coverage &
  How does it handle classification on tabular data? &
  CC3 &
  Scale 0:2 &
  \begin{tabular}[t]{@{}L{\linewidth}}0: Not at all\\ 1: Via convenience features or platform extensions \\ 2: Full AutoML\end{tabular} \\ \hline
Completeness   \& Currency &
  Technical   Domain Coverage &
  How does it handle multi-class classification on tabular data? &
  CC4 &
  Scale 0:2 &
  \begin{tabular}[t]{@{}L{\linewidth}}0: Not at all\\ 1: Via convenience features or platform extensions \\ 2: Full AutoML\end{tabular} \\ \hline
Completeness   \& Currency &
  Technical   Domain Coverage &
  How does it handle time series and forecasting? &
  CC5 &
  Scale 0:2 &
  \begin{tabular}[t]{@{}L{\linewidth}}0: Not at all\\ 1: Via convenience features or platform extensions \\ 2: Full AutoML\end{tabular} \\ \hline
Completeness   \& Currency &
  Technical   Domain Coverage &
  How does it handle image-based problems? &
  CC6 &
  Scale 0:2 &
  \begin{tabular}[t]{@{}L{\linewidth}}0: Not at all\\ 1: Via convenience features or platform extensions \\ 2: Full AutoML\end{tabular} \\ \hline
Completeness   \& Currency &
  Technical   Domain Coverage &
  How does it handle text-based problems? &
  CC7 &
  Scale 0:2 &
  \begin{tabular}[t]{@{}L{\linewidth}}0: Not at all\\ 1: Via convenience features or platform extensions \\ 2: Full AutoML\end{tabular} \\ \hline
Completeness   \& Currency &
  Technical   Domain Coverage &
  Does it handle multi-modal tasks? &
  CC8 &
  0/1 &
 \begin{tabular}[t]{@{}L{\linewidth}}0: No\\ 1: Yes\end{tabular} \\ \hline
Completeness   \& Currency &
  Technical   Domain Coverage &
  How does it handle ensemble strategies? &
  CC9 &
  Scale 0:2 &
  \begin{tabular}[t]{@{}L{\linewidth}}0: Not at all\\ 1: Via convenience features or platform extensions \\ 2: Full AutoML\end{tabular} \\ \hline
  Completeness   \& Currency &
  Customisation &
  Does it allow custom evaluation metrics? &
  CC10 &
  0/1 &
 \begin{tabular}[t]{@{}L{\linewidth}}0: No\\ 1: Yes\end{tabular} \\ \hline
Completeness   \& Currency &
  HPO   Coverage &
  Which HPO techniques does it offer? &
  CC11 &
  Points 1:N &
  \begin{tabular}[t]{@{}L{\linewidth}}
Grid\\      Random\\      Bayesian\\     Multi-Armed Bandit\\ Genetic\\     Meta-learning
  \end{tabular} \\ \hline
Completeness   \& Currency &
  Library   Coverage &
  Which popular libraries does it interface with? &
  CC12 &
  Points 1:N &
  \begin{tabular}[t]{@{}L{\linewidth}}
  Sklearn\\      
  Keras\\
  TF\\   
  XGBoost\\      
  LightGBM\\     
  Catboost\\      
  Pytorch\\
  Ax\\
  R
  \end{tabular} \\ \hline
Completeness   \& Currency &
  Currency &
  Is it actively maintained? &
  CC13 &
  0/1 &
 \begin{tabular}[t]{@{}L{\linewidth}}0: No\\ 1: Yes\end{tabular} \\ \hline
\end{longtable}

The sub-criteria in Table~\ref{Table:Criteria_CC1-9} can be mapped to the MLWF in Fig.~\ref{Fig:mlworkflowsubtasks} as follows: CC1--CC10 to \textit{Data Engineering} and \textit{Model Development} generally and CC11--CC13 to \textit{CASH+} and \textit{Requirements Review} within \textit{Model Development}.

\item \textbf{Explainability (7 Questions).}
This criterion is core to any assessment of an AutoML system, even if the requirements manifest in different ways for different stakeholders~\cite{laad20}.
Indeed, for practising data scientists, this primarily encompasses explaining how an ML solution arises, how it arrives at an output, and why its performance level is what it is.
For other technical users, that insight into drivers for technical performance, e.g.~feature importance, remains essential.
As for corporate stakeholders, explainability must be present to ensure compliance with governance, regulatory and corporate social-responsibility requirements.
Of course, beyond the standard questions, scenario-building capabilities are also essential to note, as they expand the value of an ML solution beyond predictive power to insight generation.
Such a component would be especially desirable to semi-technical and business users that care more about understanding a problem context than any particular deployed model.
Finally, this criterion includes an evaluation of whether an AutoML package considers bias and fairness.
Specific tools are dedicated to this topic, so any appraisal must not only consider the identification of associated flaws but also the capacity for their remediation; see Section~\ref{Sec:biasfairnesstools}.
We now explicitly list the questions on explainability.

\begin{longtable}{|C{2cm}|L{2cm}|L{2cm}|C{1cm}|C{1.5cm}|L{3cm}|}
\hline
\endfoot    
\caption{Assessment Framework for Explainability. Questions: EX1--EX7.}
\label{Table:Criteria_EX1-5}
\\ \hline
\rowcolor[HTML]{B4C6E7} 
\multicolumn{1}{|c|}{\textbf{Criteria}} &
  \multicolumn{1}{c|}{\textbf{Sub-Criteria}} &
  \multicolumn{1}{c|}{\textbf{Question}} &
  \textbf{QCode} &
  \textbf{Scoring} &
  \multicolumn{1}{c|}{\textbf{Rubric}} \\ \hline\endhead
Explainability & Data Lineage          & Are data lineage   \& processing steps clear?        & EX1 & 0/1 &  \begin{tabular}[t]{@{}L{\linewidth}}0: No\\ 1: Yes\end{tabular} \\ \hline
Explainability & Model   Understanding & Is   it clear what modelling steps were undertaken? & EX2 & 0/1 &  \begin{tabular}[t]{@{}L{\linewidth}}0: No\\ 1: Yes\end{tabular} \\ \hline
Explainability &
  Model   Understanding &
  Does it automatically explain global model characteristics? &
  EX3 &
  Scale 0:2 &
  \begin{tabular}[t]{@{}L{\linewidth}}0: No\\ 1: No, but convenience features are available\\ 2: Yes, to some degree\end{tabular} \\ \hline
Explainability &
  Model   Understanding &
  Does it automatically explain local prediction-level artefacts? &
  EX4 &
  Scale 0:2 &
  \begin{tabular}[t]{@{}L{\linewidth}}0: No\\ 1: No, but convenience features are available\\ 2: Yes, to some degree\end{tabular} \\ \hline
Explainability & Scenario   Modelling  & Does it support scenario exploration?    & EX5 & 0/1 &  \begin{tabular}[t]{@{}L{\linewidth}}0: No\\ 1: Yes\end{tabular} \\ \hline
Bias   \& Fairness &
  Metrics &
  Does it automatically generate best-practice bias/fairness metrics for the model/data? &
  EX6 &
  Scale 0:2 &
  \begin{tabular}[t]{@{}L{\linewidth}}0: No\\ 1: No, but convenience features are available\\ 2: Yes, to some degree\end{tabular} \\ \hline
Bias \&   Fairness &
  Metrics &
  Does it automatically mitigate and/or remediate bias/fairness flaws in the model/data? &
  EX7 &
  Scale 0:2 &
  \begin{tabular}[t]{@{}L{\linewidth}}0: No\\ 1: No, but convenience features are available\\ 2: Yes, to some degree\end{tabular}
\end{longtable}

The sub-criteria in Table~\ref{Table:Criteria_EX1-5} can be mapped to the MLWF in Fig.~\ref{Fig:mlworkflowsubtasks} as follows: 
EX1--EX5 to \textit{Visualise \& Explain} and \textit{Requirements Review}  within  \textit{Model Development} and EX6--EX7 to \textit{Explore \& Assess Fairness} within \textit{Data Engineering} and \textit{Assess Fairness} within \textit{Model Development}.

\item \textbf{Ease of Use (5 Questions).}
As with explainability, this criterion also manifests differently for different stakeholders, primarily due to the variability in technical skills and operational requirements.
For instance, an AutoML tool only available via Python or R scripting immediately limits the userbase to technicians familiar with coding constructs, e.g.~variables.
Even amongst technicians, programming languages and data-science libraries that are less common will further restrict utility.
Thus, one of the pertinent questions to ask is whether a CLI is available, given that it is somewhat more accessible to general users.
Indeed, a CLI ideally requires nothing more than simple commands to be typed in and executed with a press of a return key.
Of course, future work may further evaluate the usability of individual UIs, but, for this monograph, it is sufficiently informative to delineate between AutoML technologies that have an accessible UI and those that do not.
As an aside, any particular quirks that assist with the translation of business problems to the system undertaking analytical work are also worth noting under this criterion, e.g.~natural language processing (NLP) engines and other exotic forms of HCI.
However, deeper discussions about HCI in AutoML are deferred to other reviews~\cite{khke21}.
We now explicitly list the questions on ease of use.

\begin{longtable}{|C{1.5cm}|L{2cm}|L{2.5cm}|C{1cm}|C{1.5cm}|L{2.5cm}|}
\caption{Assessment Framework for Ease of Use. Questions: EU1--EU5.}
\label{Table:Criteria_EU1-5}
\\ \hline
\rowcolor[HTML]{B4C6E7} 
\multicolumn{1}{|c|}{\textbf{Criteria}} &
  \multicolumn{1}{c|}{\textbf{Sub-Criteria}} &
  \multicolumn{1}{c|}{\textbf{Question}} &
  \textbf{QCode} &
  \textbf{Scoring} &
  \multicolumn{1}{c|}{\textbf{Rubric}} \\ \hline\endhead
Ease of   Use &
  Interface &
  Can it be interacted with via coding? &
  EU1 &
  0/1 &
 \begin{tabular}[t]{@{}L{\linewidth}}0: No\\ 1: Yes\end{tabular} \\ \hline
Ease of Use &
  Interface &
  Is   there a CLI with simple commands? &
  EU2 &
  0/1 &
 \begin{tabular}[t]{@{}L{\linewidth}}0: No\\ 1: Yes\end{tabular} \\ \hline
Ease of Use &
  Interface &
  Is   there a GUI? &
  EU3 &
  0/1 &
 \begin{tabular}[t]{@{}L{\linewidth}}0: No\\ 1: Yes\end{tabular} \\ \hline
Ease of Use &
  Interface &
  Is   it desktop-based or browser-based? &
  EU4 &
  Scale 0:2 &
 \begin{tabular}[t]{@{}L{\linewidth}}0:   Desktop only\\      1: Browser only\\      2: Both\end{tabular} \\ \hline
Ease of Use &
  Learning &
  Is there clear and extensive documentation and guidance available? &
  EU5 &
  Scale 0:2 &
 \begin{tabular}[t]{@{}L{\linewidth}}0:   No\\      1: Partially\\      2: Yes\end{tabular} \\ \hline
\end{longtable}

The sub-criteria in Table~\ref{Table:Criteria_EU1-5} are not technically mappable to individual tasks/phases of the MLWF in Fig.~\ref{Fig:mlworkflowsubtasks}.
They refer to the ways in which one can interact with an AutoML system, as well as how convenient these forms of HCI are.
Fundamentally, ease of use can impact every aspect of an MLWF, i.e.~wherever a user must interact with an AutoML system to complete a task.

\item \textbf{Deployment \& Management Effort (11 Questions).}
This criterion is perhaps the one that most distinguishes industrial concerns from academic considerations.
Indeed, once experimentation results in an ML solution, it takes significant technical effort to embed the object within a business decision-making process~\cite{scho15}.
Often, a production environment must continue to feed an ML model with transformed data, potentially in real-time and/or streaming fashion, then transfer generated predictions/prescriptions to an end user or other downstream systems.
Once deployed, an ML solution should also ideally be monitored for changes in defined metrics~\cite{brca17}, e.g.~those related to technical performance and fairness.
Whether reactively, in response to monitored information, or proactively, optimal modelling may additionally necessitate reapplying earlier MLWF processes as part of maintenance, such as model retraining.
Therefore, characteristics of an AutoML tool that assist or hinder this important criterion are crucial to appraise here.
We now explicitly list the questions on deployment \& management effort.

\begin{longtable}{|C{2cm}|L{2cm}|L{2cm}|C{1cm}|C{1.5cm}|L{3cm}|}
\caption{Assessment Framework for Deployment \& Management Effort. Questions: DM1--DM4.}
\label{Table:Criteria_DM1-DM4}
\\ \hline
\rowcolor[HTML]{B4C6E7} 
\multicolumn{1}{|c|}{\textbf{Criteria}} &
  \multicolumn{1}{c|}{\textbf{Sub-Criteria}} &
  \multicolumn{1}{c|}{\textbf{Question}} &
  \textbf{QCode} &
  \textbf{Scoring} &
  \multicolumn{1}{c|}{\textbf{Rubric}} \\ \hline\endhead
Deployment   \& Management Effort &
  Deployment &
  Does it use   model compression techniques? &
  DM1 &
  0/1 &
 \begin{tabular}[t]{@{}L{\linewidth}}0: No\\ 1: Yes\end{tabular} \\ \hline
Deployment \& Management Effort &
  Deployment &
  Can   it be deployed on-premise and/or in the cloud? &
  DM2 &
  Scale 0:2 &
 \begin{tabular}[t]{@{}L{\linewidth}}0:   No\\      1: Yes, only in the cloud\\      2: Yes, both\end{tabular} \\ \hline
Deployment \& Management Effort &
  Deployment &
  Does   it offer advanced deployment testing mechanisms, e.g.~A/B or champion-challenger? &
  DM3 &
  0/1 &
 \begin{tabular}[t]{@{}L{\linewidth}}0: No\\ 1: Yes\end{tabular} \\ \hline
Deployment \& Management Effort &
  Deployment &
  Does   it offer advanced deployment update mechanisms, e.g.~blue-green or canary? &
  DM4 &
  0/1 &
 \begin{tabular}[t]{@{}L{\linewidth}}0: No\\ 1: Yes\end{tabular} \\ \hline
\end{longtable}

The sub-criteria in Table~\ref{Table:Criteria_DM1-DM4} can be mapped to the MLWF in Fig.~\ref{Fig:mlworkflowsubtasks} as follows: DM1/DM2 to \textit{Provision} within \textit{Deployment}, DM3 to \textit{Serving} within \textit{Deployment} and \textit{Proactive Training} and \textit{Reactive Training} within \textit{Monitoring and Maintenance}, and DM4 to \textit{Serving} within \textit{Deployment}.

\begin{longtable}{|C{2cm}|L{2cm}|L{2cm}|C{1cm}|C{1.5cm}|L{3cm}|}
\caption{Assessment Framework for Deployment \& Management Effort. Questions: DM5--DM11.}
\label{Table:Criteria_DM5-DM11}
\\ \hline
\rowcolor[HTML]{B4C6E7} 
\multicolumn{1}{|c|}{\textbf{Criteria}} &
  \multicolumn{1}{c|}{\textbf{Sub-Criteria}} &
  \multicolumn{1}{c|}{\textbf{Question}} &
  \textbf{QCode} &
  \textbf{Scoring} &
  \multicolumn{1}{c|}{\textbf{Rubric}} \\ \hline\endhead
Deployment   \& Management Effort &
  Management &
  Does it   automatically set up monitoring? &
  DM5 &
  Scale 0:2 &
\begin{tabular}[t]{@{}L{\linewidth}}0: No, not   present at all\\      1: No, manual setup and/or configuration is required\\      2: Yes\end{tabular} \\ \hline
Deployment \& Management Effort & Management & Does   it monitor hardware usage and performance? & DM6  & 0/1 &  \begin{tabular}[t]{@{}L{\linewidth}}0: No\\ 1: Yes\end{tabular} \\ \hline
Deployment \& Management Effort & Management & Does   it monitor model performance metrics?              & DM7  & 0/1 &  \begin{tabular}[t]{@{}L{\linewidth}}0: No\\ 1: Yes\end{tabular} \\ \hline
Deployment \& Management Effort & Management & Does   it monitor data/concept drift?           & DM8  & 0/1 &  \begin{tabular}[t]{@{}L{\linewidth}}0: No\\ 1: Yes\end{tabular} \\ \hline
Deployment \& Management Effort & Management & Does   it monitor bias/fairness metrics?                & DM9  & 0/1 &  \begin{tabular}[t]{@{}L{\linewidth}}0: No\\ 1: Yes\end{tabular} \\ \hline
Deployment \& Management Effort &
  Management &
  Does   it reactively retrain based on monitoring triggers? &
  DM10 &
  Scale 0:3 &
\begin{tabular}[t]{@{}L{\linewidth}}0:   No\\      1: No, but convenience features can assist manual retraining\\      2: Yes, with triggers defined by user\\      3: Yes, with triggers provided by developers\end{tabular} \\ \hline
Deployment \& Management Effort & Management & Does   it proactively retrain?                                 & DM11 & 0/1 &  \begin{tabular}[t]{@{}L{\linewidth}}0: No\\ 1: Yes\end{tabular} \\ \hline
\end{longtable}

The sub-criteria in Table~\ref{Table:Criteria_DM5-DM11} can be mapped to the MLWF in Fig.~\ref{Fig:mlworkflowsubtasks} as follows: DM5--DM9 to \textit{Monitoring} within \textit{Monitoring and Maintenance}, DM10 to \textit{Reactive Training} within \textit{Monitoring and Maintenance}, and DM11 to \textit{Proactive Training} within \textit{Monitoring and Maintenance}.

\item \textbf{Governance (3 Questions).}
In an organisational context, an ML tool must align to existing data governance and security considerations to ensure regulatory and internal-policy compliance.
Therefore, this brief but important category assesses whether an AutoML tool aligns with relevant practices.
Questions include whether data access is appropriately managed~\cite{in17}, although, for ML, it is also essential to evaluate whether an organisation can control access to functionality.
Such system features are particularly pertinent during model deployment, as this phase bears significant risks around exposing organisational artefacts to external parties and inadvertently embedding immature projects into core business processes/environments.
Even with the best intentions to ensure security and controlled access, the ability to audit activity on internal systems is likewise desirable to confirm that appropriate activities are being undertaken by authorised entities~\cite{ryri14}.
We now explicitly list the questions on governance.

\begin{longtable}{|C{2cm}|L{2cm}|L{2cm}|C{1cm}|C{1.5cm}|L{3cm}|}
\caption{Assessment Framework for Governance. Questions: G1--G3.}
\label{Table:Criteria_G1-G3}
\\ \hline
\rowcolor[HTML]{B4C6E7} 
\multicolumn{1}{|c|}{\textbf{Criteria}} &
  \multicolumn{1}{c|}{\textbf{Sub-Criteria}} &
  \multicolumn{1}{c|}{\textbf{Question}} &
  \textbf{QCode} &
  \textbf{Scoring} &
  \multicolumn{1}{c|}{\textbf{Rubric}} \\ \hline\endhead
Governance \& Security & Governance & Does it offer   auditing of activity?                               & G1 & 0/1 & \begin{tabular}[t]{@{}L{\linewidth}}0: No\\ 1: Yes\end{tabular} \\ \hline
Governance \& Security & Security   & Does   it offer artefact access controls for the model/data?              & G2 & 0/1 & \begin{tabular}[t]{@{}L{\linewidth}}0: No\\ 1: Yes\end{tabular} \\ \hline
Governance \& Security & Security   & Does   it offer function access controls for training, deployment, etc.? & G3 & 0/1 & \begin{tabular}[t]{@{}L{\linewidth}}0: No\\ 1: Yes\end{tabular} \\ \hline
\end{longtable}

The sub-criteria in Table~\ref{Table:Criteria_G1-G3} are not technically mappable to individual tasks/phases of the MLWF in Fig.~\ref{Fig:mlworkflowsubtasks}.
They refer to how an organisation integrates with and accesses an AutoML system, as well as how secure these forms of HCI are.
Fundamentally, governance can impact every aspect of an MLWF, i.e.~wherever a user can interact with an AutoML system to influence a task.

\end{itemize}

\subsection{The Role of AutoML}
\label{Sec:automlrole}

The assessment framework proposed in this monograph reflects the desires of major stakeholder groups when engaging with ML in an industry setting.
Although the constituent questions and their measurement rubrics are diverse, there is a loose general trend: the higher an AutoML package scores for the proposed criteria, the more it is seen to support performant ML.
So, after such an amalgamation of varied requirements, it is worth asking whether the framework can be condensed into succinct insights about what drives AutoML uptake, now and in the future.
Basically, what does industry see as the role of AutoML, both present and prospective?

There appear to be several answers, as follows:
\begin{itemize}
    \item \textbf{Enhancing Data Science Practices.}
	The use of AutoML provides the potential to engage in ML processes that are, compared to manual efforts, more efficient, technically performant, robust, and explainable.
	Implementations with ongoing developer support will likely adopt the best available approaches for relevant MLWF tasks and stay up-to-date with the latest technologies.
	Moreover, beyond simply creating a model object, the ongoing AutoML endeavour to be genuinely end-to-end promises scalable deployment capabilities and an easier way to monitor/maintain performance subject to real-world data dynamics.
	Progress in these directions represents advancement towards true AutonoML and next-generation technical abilities~\cite{kemu20, khke21}.
	At the very least, from a purely financial perspective, efficiently generating and conveniently deploying ML solutions that potentially perform better than manual selections will offer both cost savings and revenue maximisation.
    \item \textbf{Democratising Data Science Practices.}
	Beyond pushing the limits of capabilities that ML practitioners are familiar with, AutoML offers a gateway to ML techniques and approaches for those who are not trained technicians.
	While not without risks and other implications, this inclusive `democratisation' stands to knock down existing skill barriers, with benefits flowing both ways.
	Specifically, organisations will likely leverage the power of data science with greater ease, while ML applications will profit from a more fluid influx of domain knowledge.
    \item \textbf{Standardising Data Science Practices.}
	Each AutoML system acts as a wrapper for a collection of methods that deal with targeted phases of an MLWF.
	There may be many services on offer as of the early 2020s, but there are still far fewer packages than individual techniques.
	Thus, centralising work efforts into operations framed by a common system provides many potential benefits, reproducibility among them.
	Standardisation of such practices also supports stronger security mechanisms and access controls alongside enhanced auditability and thus governance.
	Indeed, as society continues to expect increasingly more from its AI engagement, particularly on the ethical front, the existence of AutoML may make it easier to certify compliance.
	At the very least, the technology cannot exist at odds with data governance practices lest corporate decision makers lean towards caution, hindering industrial uptake and successful business integration.
\end{itemize}

\begin{table}[H]
\caption{\label{tab_criteria}Key criteria that an ML application should satisfy according to stakeholders. Considerations are marked F for fundamental and C for contextual.}
\resizebox{\textwidth}{!}{%
\begin{tabular}{|l|c|c|c|c|c|}
\hline
\rowcolor[HTML]{B4C6E7} 
\multicolumn{1}{|c|}{\cellcolor[HTML]{B4C6E7}\textbf{Criteria}} &
  \textbf{Data Scientist} &
  \textbf{Analyst} &
  \textbf{Deployment Technicians} &
  \textbf{Corporate} &
  \textbf{End Users} \\ \hline
Technical Performance           & F &   & F & C & F \\ \hline
Efficiency                      & F &  & F & C &   \\ \hline
Dirty   Data                      & F &   &   & C &   \\ \hline
Completeness \& Currency        & F &  &   &   &   \\ \hline
Explainability                  & C & C &   & C & F \\ \hline
Bias / Fairness                 & C & C &   & C & F \\ \hline
Ease of Use                     & F  & F  &  &   &   \\ \hline
Deployment \& Management Effort & F  &   & F & C &   \\ \hline
\end{tabular}%
}
\end{table}

In light of these overarching industrial expectations of AutoML, it is worth presenting one last condensed overview, specifically around which stakeholders care most about particular elements of the aggregated framework presented in Section~\ref{Sec:summarisedcriteria}.
Table~\ref{tab_criteria} does so, marking criteria by F if they are fundamental considerations to a stakeholder during engagement with an ML application.
Additionally, C denotes a contextual criterion, i.e.~one that is conditionally important to a stakeholder depending on organisational context.
Of course, as discussed earlier, different stakeholders have varying requirements and degrees thereof concerning each criterion; some will find certain concepts virtually irrelevant to their role.

Unsurprisingly, data scientists care about the greatest number of listed criteria, given the ongoing centrality of a technical role in an ML application.
After all, a core purpose of AutoML is to assist such practitioners closely with their technical tasks.
However, management is also broadly invested in AutoML technology supporting performant ML.
Beyond corporate stakeholders typically needing to sign off on software purchases, they also act as the bridge between the fundamental work of a data scientist or analyst and the business context in which an ML solution is to be deployed.
Thus, while managers usually do not operate AutoML technology, they still have wide-ranging conditional requirements that must be satisfied.

As for deployment technicians, these stakeholders become interested in AutoML once associated systems extend beyond the traditional core of ML model selection.
Specifically, such technicians will likely be keen to employ AutoML systems that incorporate and simplify MLOps processes.
Accordingly, their requirements are primarily infrastructural.
In a large enough analytics team, data scientists and deployment technicians will be distinct, so, while both care about technical performance and efficiency, the foci of both groups are different.
The former seeks to efficiently train up an ML pipeline with a high degree of validity, while the latter seeks to ensure a deployed solution services queries efficiently and maintains a baseline level of `correctness'.

Finally, end users may not interact directly with the development of an ML solution, but they remain affected by the entire process.
This statement holds whether the user is a target of low-stakes product recommendation or a more impactful application, e.g.~an AI-based hiring decision or loan determination.
Accordingly, trust -- this requires explainability -- is crucial for their engagement with an organisation that delivers the product.
If the end user is not satisfied that an ML solution is technically performant, unbiased and fair, they remove their financial support or, depending on the application, pursue recompense.
In effect, a happy client/customer indirectly validates using an AutoML service.

Notably, in the summation within Table~\ref{tab_criteria}, there is no explicit mention of the business user detailed in Section~\ref{Sec:primary_stake}.
Such a stakeholder without technical skill can be considered to sit within the corporate section, supplying domain knowledge and engaging with core technicians via an analyst as required.
In some framings of this structure, the role is considered a `domain expert', while others may refer to a `project sponsor'.
Regardless of terminology, this stakeholder does not traditionally have any direct involvement with an MLWF beyond perhaps the initial phase of problem formulation \& context understanding.
However, given the potential of AutoML to drive organisational value through democratisation, business users remain important; any AutoML system that facilitates HCI for these technical non-experts accelerates technological adoption.

The question then is as follows: if industrial uptake of AutoML is contingent on how well it enhances, democratises and standardises data science practices, how far along is the technology when meeting expectations?
In seeking to answer this question, incoming sections of the review examine how well modern offerings support performant ML, additionally attempting to glean whether AutoML has begun making a societal impact.

\section{AutoML Software}
\label{Sec:Tools}

The cataloguing of open-source and commercial AutoML services is splintered across various blogs, tutorials, social media posts, and GitHub-based collections.
Currently, there is no centralised index for this technology that is both reliable and up-to-date.
Thus, this review sifts through a compilation of sources to present one such encompassing snapshot of AutoML products as of the early 2020s.
Investigations supporting this survey, both broad and deep, have involved, for instance, popular search engines, blogs on data science and analytics~\cite{KDnuggets}, and curated repositories~\cite{awesomeproductionmachinelearning, awesomeAutoML}.

First, however, we detail the constraints around the scope of the review, i.e.~what software packages can be considered under the banner of AutoML.
Here, we define an AutoML `tool' as a computer program that, through some form of HCI, automates some element of the MLWF in Fig.~\ref{Fig:mlworkflowsubtasks}.
Crucially, the initial high-level process to receive concerted focus under the banner of AutoML was model selection, specifically HPO.
The field has gradually expanded its attention in the years since, even though the model-development phase remains a strong priority.
Given the variety of software packages currently accessible to the public, a further categorisation is thus possible.

If an AutoML tool allows a user to train an ML model on a given dataset as one of its internal functions, which is the core process of ML, we call it `comprehensive'.
Such a tool will also often be referred to here as an AutoML system, acknowledging the integrated nature of multiple mechanisms~\cite{kemu20}.
This definition does not mean that all comprehensive tools are identical in scope.
Some may or may not include ensemble mechanisms, data cleaning, etc.
However, they all act as wrappers around an ML solution object and, in some way, manage it.

In contrast, if an AutoML tool automates an aspect of an MLWF without directly training an ML model, we call it `dedicated'.
The term is used because this survey finds these ancillary tools predominantly focus on but a single element of the MLWF.
The reasons for this specialisation are numerous.
For instance, many HPO tools are actually built agnostically, such that they can be applied to arbitrary optimisation problems as formulaically as to selecting hyperparameter configurations for ML pipelines.
Hence, the dedication to the HPO task is, in some sense, a fruitful application of a more general theory.
Other times, a dedicated tool is simply designed to prioritise challenges beyond model development without becoming bogged down in unnecessarily comprehensive implementation.
Regardless, despite the diversity of dedicated tools in existence, this review finds that they are most concentrated in three areas: (1) data and feature engineering, (2) bias, fairness and explainability, and (3) HPO.

Naturally, whether dedicated or comprehensive, all AutoML tools were considered under exclusion criteria to ensure a meaningful analysis of industrial uptake.
The mere presence of a GitHub repository did not automatically qualify a tool for this review.
Exclusion criteria include the following:
\begin{itemize}
	\item A lack of open-source or commercial implementation ready for organisational use.
	This criterion immediately cuts out numerous academic proposals and experiments.
	Of course, this does not devalue the novel/interesting AutoML science that these projects engage in, e.g.~the reductivist evolutionary principles of AutoML-Zero~\cite{automlzero, reli20}.
	However, whether such theoretical concepts gain greater traction in mainstream usage is a matter for future assessment; it remains out of scope for this monograph.
	\item Association with a repository that possesses minimal commits and stars, often appearing as a quick burst within a short window of time and paired with the publication of an academic paper.
	Such a scenario suggests the repository serves more as an informational webpage or code storage facility than a means of hosting genuinely usable software for the global data-analytics community.
	\item Indication that a project is overly casual, i.e.~done for personal interest or publicity by an individual or small group.
	\item A lack of updates or maintenance activity for over 18 months, occasionally paired with an announcement of the tool being deprecated.
	However, some of these `faded' packages are still useful to mention as part of historical commentary; see Appendix~\ref{Sec:fadedpurehpo} and Appendix~\ref{Sec:fadedcoreautoml}.
\end{itemize}
In essence, these exclusion criteria all aim to reflect the subset of AutoML technology with which industry is seriously engaging at present.

Finally, we make one more relevant note; this review treats AutoDL as largely out of scope.
Here, we define an AutoDL tool as being both (1) focussed on NAS and (2) not primarily designed for tabular data.
Fortunately, this exclusion has not been an issue for the most part.
While AutoDL receives substantial interest as a subtopic of DL, the extreme computational resources required have stymied any significant dissemination of the technology throughout mainstream industry.
Granted, this state of affairs may change quickly at any time, and some blur in the discussion within this monograph is inescapable.
However, at this point in the early 2020s, AutoDL remains extremely focussed on the model-development phase of an MLWF and, overall, has not translated far beyond the academic proof-of-principle stage.
Deeper discussions are available elsewhere~\cite{doke21}.

We now proceed to discuss the surveyed AutoML tools from the perspective of performant ML, for which an assessment framework was introduced in Section~\ref{Sec:summarisedcriteria}.
Open-source software is explored first, with Section~\ref{Sec:ancillarytools} for dedicated tools and Section~\ref{Sec:coretools} for comprehensive systems.
Commercial AutoML products, which tend to be relatively opaque, are then analysed within Section~\ref{Sec:commercialtools}.

\subsection{Dedicated AutoML Tools}
\label{Sec:ancillarytools}

This section surveys and discusses open-source tools that, to some degree, automate a specific element of an MLWF.
Several of these programs are occasionally incorporated as mechanisms within other holistic software packages.
However, in complete isolation, they do not enable an ML application to be run; the core ML processes of model generation and management are typically absent.

Broad exclusion criteria have already been listed, but it is still worth providing examples of projects beyond the scope of this review.
Doing so raises awareness of how much research and development is ongoing within the field of AutoML.
For instance, conceptual tools lacking a substantive open-source implementation at the time of review include One Button Machine (OBM)~\cite{lath17}, Persistant~\cite{biab18}, and BigDansing, a system for big data cleansing~\cite{khil15}.
Then there are codebases that, while existing, have been given limited developmental/maintenance attention, often serving to supplement the publication of an academic paper.
Examples include ExploreKit~\cite{ExploreKit, kash16}, the Data Civilizer System~\cite{datacivilizersystem, defe17}, and NADEEF~\cite{NADEEF, daeb13}.
In fairness, NADEEF exemplifies software that appears to have been actively maintained at one point, but the lack of update in over five years implies it has since `faded'.
Finally, Cognito~\cite{cognito} is an example of exclusion based on appearing to be a small-scale personal/group project.
Such codebases tend to have lower levels of popularity, quality or maintenance compared to software entrenched within industry, although there is no such insinuation regarding Cognito specifically; some of these projects may eventually attain sufficient recognition to be more broadly adopted by organisations and their stakeholders.

Now, the incoming discussion around dedicated AutoML tools has been organised into three categories that encompass the majority of surveyed tools.
Section~\ref{Sec:datafeaturetools} covers data and feature engineering, Section~\ref{Sec:biasfairnesstools} relates to bias, fairness and explainability, and Section~\ref{Sec:purehpotools} addresses HPO.
Notably, data visualisation, arguably essential throughout an MLWF, is conspicuous in its absence.
This omission is not because there is a dearth of tools dedicated to such a form of HCI.
In contrast, there are simply too many packages in this space to consider for this survey.
Such consideration is also inappropriate for scope, as designing visual representations for data can be done entirely without ML in mind.
There are exceptions, however, e.g.~if a visualisation package is customised to optimally communicate specific ML metrics.
In practice, such products are still commonly dedicated to specific tasks, such as evaluating bias and fairness, thus slotting organically into one of the categories mentioned above.

\subsubsection{Data and Feature Engineering}
\label{Sec:datafeaturetools}

\begin{table}[h]
\caption{Dedicated AutoML Tools: Data and Feature Engineering.}
\label{table:datafeatureengtools}
\resizebox{\textwidth}{!}{%
\begin{tabular}{|l|l|l|l|}
\hline
\rowcolor[HTML]{B4C6E7} 
\multicolumn{1}{|c|}{\cellcolor[HTML]{B4C6E7}\textbf{Name}} &
  \multicolumn{1}{c|}{\cellcolor[HTML]{B4C6E7}\textbf{Categorisation}} &
  \multicolumn{1}{c|}{\cellcolor[HTML]{B4C6E7}\textbf{Github}} &
  \multicolumn{1}{c|}{\cellcolor[HTML]{B4C6E7}\textbf{Ref.}}
  
  \\ \hline Compose           & Data   Processing     & {\color[HTML]{0563C1} { https://github.com/FeatureLabs/compose}}             & \cite{compose, kagi16}
  
  \\ \hline Featuretools      & Feature   Engineering & {\color[HTML]{0563C1} { https://github.com/FeatureLabs/featuretools}}        & \cite{Featuretools, kave15}   
  
  \\ \hline Feature-engine    & Feature   Engineering & {\color[HTML]{0563C1} { https://github.com/solegalli/feature\_engine}}       & \cite{featureengine}
  
  \\ \hline Boruta            & Feature   Engineering & {\color[HTML]{0563C1} { https://github.com/scikit-learn-contrib/boruta\_py}} & \cite{compose, kuru10}       
  
  \\ \hline tsfresh           & Feature   Engineering & {\color[HTML]{0563C1} { https://github.com/blue-yonder/tsfresh}}             & \cite{Tsfresh, chbr18} \\ \hline    
    
\end{tabular}%
}
\end{table}

The first category of dedicated AutoML tools in this review encompasses those that focus solely on data preparation tasks.
The surveyed open-source packages classified as such are listed in Table~\ref{table:datafeatureengtools}.

Arguably, the most well-known within this set of codebases is Featuretools~\cite{Featuretools}, which can generate numerous features for subsequent ML model training.
As is typical for automated feature-engineering efforts, Featuretools provides a pool of `primitives', i.e.~basic feature transformations, such as `mean' and `sum'.
These primitives can then be stacked into even more complex transformations of data.
In contrast, Feature-engine~\cite{featureengine} avoids this form of deep feature synthesis, but it does provide an expansive variety of operations for data cleaning and preparation alongside feature generation.
The package, however, does not automatically apply every operation, so some manual involvement is still required in making appropriate choices.

In both cases, an ML model never has to enter the picture.
Mechanisms that prioritise feature generation are primarily sold on the basis of conveniently, and hopefully intelligently, providing many new perspectives on a dataset.
On the other hand, Boruta~\cite{Boruta} is solely dedicated to automated feature selection, which is often a more difficult task.
The challenge is deciding which features are actually helpful for generating an accurate ML solution.
In the case of Boruta, the package appears to act in wrapper style, as defined in Section~\ref{Sec:mlwf}.
Specifically, it relies on the technical performance of a model to guide the search for good features, although this dependence on an ML model object does not mean the mechanism is considered a comprehensive AutoML system under the definitions in this review.

The remaining two packages indicate the diversity of available dedicated tools operating at this phase of an MLWF.
The Compose program~\cite{compose} deserves its own data-processing subcategory in Table~\ref{table:datafeatureengtools}, operating upstream relative to the other tools.
It assists a stakeholder in constructing a training set from data, assuming one has not been previously prepared, and thus provides valuable labelling and extraction operations.
As for `tsfresh'~\cite{Tsfresh}, it is unique for being dedicated not only to a specific task, i.e.~feature engineering, but also to a specific technical domain.
The tool essentially generates features for problems and data that involve time series.
Such customisation to distinguish an AutoML service from the rest of the pack is a recurrent theme and is examined further in Section ~\ref{Sec:specialisedtools}.
However, despite the variety of dedicated tools on offer, it is still notable that only a handful can arguably be considered adequate for organisational use.
Certainly, there is a lot of research activity in automating data preparation, but, as an endeavour trailing HPO and automated model selection, perhaps it should not be surprising that technological translation here is nascent.

\subsubsection{Bias, Fairness and Explainability} 
\label{Sec:biasfairnesstools}

\begin{table}[h]
\caption{Dedicated AutoML Tools: Bias and Fairness.}
\label{table:biasfairnesstools}
\resizebox{\textwidth}{!}{%
\begin{tabular}{|l|l|l|l|}
\hline
\rowcolor[HTML]{B4C6E7} 
\multicolumn{1}{|c|}{\cellcolor[HTML]{B4C6E7}\textbf{Name}} &
  \multicolumn{1}{c|}{\cellcolor[HTML]{B4C6E7}\textbf{Categorisation}} &
  \multicolumn{1}{c|}{\cellcolor[HTML]{B4C6E7}\textbf{Github}} &
  \multicolumn{1}{c|}{\cellcolor[HTML]{B4C6E7}\textbf{Ref.}}
  
  \\ \hline AI Fairness   360 & Bias-Fairness         & {\color[HTML]{0563C1} { https://github.com/Trusted-AI/AIF360}}               & \cite{AIF360}

  \\ \hline FairLearn         & Bias-Fairness         & 
  {\color[HTML]{0563C1} { https://github.com/fairlearn/fairlearn}}               & \cite{fairlearn, bidu20}
  
  \\ \hline Audit AI         & Bias-Fairness         & 
  {\color[HTML]{0563C1} { https://github.com/pymetrics/audit-ai}}               & \cite{auditai}   
  
  \\ \hline aequitas         & Bias-Fairness         & 
  {\color[HTML]{0563C1} { https://github.com/dssg/aequitas}}               & \cite{aequitas, saku19}
    
  \\ \hline LIME         & Viz (Bias-Fairness)     &
  {\color[HTML]{0563C1} { https://github.com/marcotcr/lime}}               & \cite{lime, risi16}
  
  \\ \hline SHAP  & Viz (Bias-Fairness)     &
  {\color[HTML]{0563C1} { https://github.com/slundberg/shap}}               & \cite{shap, lule17}
  
    \\ \hline AIExplainability360  & Viz (Bias-Fairness)     &
  {\color[HTML]{0563C1} { https://github.com/Trusted-AI/AIX360}}               & \cite{AIExplainability360, arbe19}
  
  \\ \hline What If Tool  & Viz (Bias-Fairness)     &
  {\color[HTML]{0563C1} { https://github.com/PAIR-code/what-if-tool}}               & \cite{whatiftool}
  \\ \hline
    
\end{tabular}%
}
\end{table}

The second category of dedicated AutoML tools in this review encompasses those focussing solely on bias and fairness metrics, effectively grappling with explainability.
The surveyed open-source packages classified as such are listed in Table~\ref{table:biasfairnesstools}.

Given the recent surge in importance that the topic of ML trustworthiness has attained, this section warrants a more extensive discussion than the other two categories of dedicated AutoML tools.
First, it is essential to note that the bias of societal concern does not refer to the typical concept that technicians are familiar with, i.e.~the bias-variance trade-off that explains the high-bias underfitting and high-variance overfitting of ML models~\cite{gebi92}.
From an ethical standpoint, bias is instead defined as an inclination/prejudice for or against a person or group, and several academic works have attempted to further systematise this notion, e.g.~categorising six sources of bias~\cite{sugu20, memo19}.
Unsurprisingly, bias is intrinsically linked to fairness, defined elsewhere as ``the absence of any prejudice or favouritism toward an individual or a group based on their inherent or acquired characteristics''~\cite{chmi21}.
In turn, a lack of fairness in ML can have adverse impacts, including `allocation harms', where resources and opportunities are withheld from a particular group, and `quality-of-service harms', where predictive/prescriptive outcomes may be relatively poor for a particular group~\cite{fairlearn}.
Of course, the topic of bias and fairness is highly complex, and a more in-depth treatment of such issues in AutoML is available elsewhere~\cite{khke21}.
It is sufficient to note that bias can creep into an ML solution at many stages of an MLWF, e.g.~within data, ML algorithms, human decisions, and various processes.
Accordingly, the tools in this category support, to varying degrees, automatically diagnosing and mitigating bias-and-fairness issues.

The most popular of the dedicated tools in Table~\ref{table:biasfairnesstools} is arguably AI Fairness 360 (AIF360)~\cite{AIF360}, created by IBM.
At the time of survey, AIF360 was able to calculate three `individual fairness' metrics~\cite{zhne17, sphe18, fois19} and apply 13 bias mitigation algorithms based on external research~\cite{kaak12, kaca12, kaka12, zewu13, fefr15, hapr16, cawe17, plra17, agbe18a, kene18, zhle18, agdu19a, cehu20}.
The package also supports additional metrics evaluating `group fairness' and sample distortion.
However, while these complex calculations and procedures are automated, their selection is not.
Sample tutorial notebooks are provided by AIF360, but, within them, it is experts that decide on a metric and remedy.
Likewise, befitting an AutoML tool that is not considered comprehensive, integration into an ML application also requires manual involvement.
Additionally, this package has specific developer-environment requirements, which can be a barrier to entry for stakeholders with alternative operational practices.
However, this particular comment is more of a general reminder than a specific critique, given that few software applications approach genuine cross-platform universality.

Now, the way that HCI works for such a tool sparks a broader discussion: for an MLWF task dedicated to the careful and deliberate review/amendment of bias/fairness flaws, how much of the process should be automated?
One might argue that, at the very least, a bias-and-fairness tool should automatically calculate and visualise all relevant metrics, in much the same way as automated feature engineers in Section~\ref{Sec:datafeaturetools} systematically generate multiple perspectives of a dataset.
The counterargument is that an embarrassment of riches may still not make it any easier for a data scientist, let alone a non-technical user, to make the best operational decisions.
Metric and remedy selection is as important here as feature selection is to data engineering.
Thus, if full automation is deemed unwise in bias-and-fairness space, then some form of recommendation prompts would still likely provide added value to a user.
For now, the automation capability of most dedicated tools in Table~\ref{table:biasfairnesstools} does not extend far in this direction.
FairLearn~\cite{fairlearn}, developed and released by Microsoft, is akin to AIF360 in the level of development, documentation, and maintenance.
The package contains some overlap in terms of mitigation algorithms~\cite{hapr16, agbe18a, agdu19a}, but it also provides a dashboard with a more convenient overview of metrics and comparative analyses for both models and potentially discriminated groups.
This UI, along with very detailed documentation covering software usage and topical elaboration, probably marks the current limit of mechanisation for this aspect of the MLWF, at least among non-comprehensive tools.

Comparatively, Audit AI~\cite{auditai} is a less mature library focussed solely on bias-and-fairness diagnosis.
It supports the calculation of several metrics under techniques named `4/5th', `fisher', `z-test', `bayes factor', and `chi squared'.
Additionally, it provides `sim beta ratio' and `classifier posterior probabilities' tests for classification tasks, as well as analysis of variance (ANOVA) and a test of ``group proportions at different thresholds'' for regression tasks.
Overall, the software is light on documentation, which essentially comes as a readme file on GitHub.
However, it is notable for discussing regulatory needs in detail, e.g.~referring to a set of guidelines for employee selection procedures~\cite{UGESP} while motivating its 4/5ths rule.
Likewise, albeit without citation/justification, the package mentions implementing the Cochran-Mantel-Hanzel statistical test \cite{co54, maha59}, which is very common in regulatory practices.
In effect, Audit AI highlights the surging importance of governance to real-world ML, as stressed in Section~\ref{Sec:secondary_stake}.
Finally, `aequitas'~\cite{aequitas} also makes the cut for this survey, despite a relatively diminished amount of activity.
It enables calculating metrics derived from a confusion matrix, although applied to subgroups within data.
Moreover, it can be used as a code library or interacted with via a CLI, generating reports with simple graphs.
Thus, despite being relatively basic, it rounds out the tools dedicated directly to bias and fairness in this section.

Of course, there is an inherent dependence of ML trustworthiness on explainability; one cannot confidently assess bias and unfairness without understanding how an ML solution works.
This model-comprehension process is often best assisted by transforming technically arcane code into a more interpretable format.
Thus, although we keep the survey constrained, the handful of visualisation (viz) tools in Table~\ref{table:biasfairnesstools} is worthy of acknowledgement.
For instance, LIME~\cite{lime, risi16} is a package that pursues model-agnostic visualisations to assist in explaining black-box ML classifiers.
In turn, it is referenced by SHAP~\cite{shap, lule17}, which allows users to visualise and better understand numerous ML solutions, e.g.~whether structured as a linear model, tree, or a DL network.
Neither of these two tools explicitly provide instructions for bias-and-fairness application; it is up to a user to integrate these procedures into auditing operations.
Nonetheless, the packages are well maintained and documented, and the calculations/processes they automate simplify such tasks for ML stakeholders.
Indeed, AIExplainability360~\cite{AIExplainability360} is an example of a tool leveraging both LIME and SHAP alongside other software to promote explainability, despite not being updated recently and facing the common issue of users needing to gauge the utility/appropriateness of different functions themselves.
Finally, the What If Tool~\cite{whatiftool} by Google does return focus to fairness, allowing users to easily select and view five fairness metrics calculated for their ML model: `group unaware', demographic parity, equal opportunity, equal accuracy, and group thresholds.
There is an obvious crossover with AIF360 and FairLearn, but the library does retain a performance mindset alongside bias-and-fairness awareness, e.g.~when enabling users to analyse comparative statics for ML models. 

One overall conclusion is that there is a surprising level of activity under this category of dedicated AutoML tools.
The survey already narrowly excludes several packages, such as Parity Fairness~\cite{parityfairness}, which appears to be an incremental wrapper of AIF360 and FairLearn, FairML~\cite{fairml}, which seems to be a personal project for feature-importance graphs that faded in 2017, and Scikit-Fairness \cite{ScikitFairness}, which has been merged into FairLearn.
However, only AIF360 and FairLearn, both corporately sponsored, can be considered outliers in terms of maturity.
There is also much overlap amongst the packages.
For instance, AIF360 lists both FairLearn and LIME as dependencies.
Granted, the motivations that drive development in this space are likely to ensure continued progress; both AIF360 and FairLearn appear to be driven by socioethical considerations, while Audit AI presages the rise of regulatory obligations in industrial ML.

Nevertheless, all of the tools are presently constrained in terms of automation, requiring both upskilling and careful manual interaction for full engagement with an ML auditing task.
A thorough review of some of these tools reaches similar conclusions~\cite{lesi21}, noting that, despite stakeholder interest, it remains difficult to acquire the necessary skills/knowledge to make informed choices.
In essence, this topic is faced with the challenge of balancing accessibility with deep technicality.
The reviewing paper also found commonly held concerns around how to integrate these tools with existing MLWFs and grant them appropriate coverage.
Indeed, the challenge of integrating any dedicated mechanism into a larger automated system is nontrivial~\cite{kemu20}.
Regardless, given the increasing public awareness and scrutiny around ML trustworthiness, time will tell how this technological subspace evolves.

\subsubsection{Hyperparameter Optimisation}
\label{Sec:purehpotools}

\begin{longtable}{|L{2cm}|L{3cm}|C{1.5cm}|C{0.5cm}|L{3cm}|L{1.5cm}|}
\caption{Dedicated AutoML Tools: HPO.}
\label{table:hpotools}
\\ \hline
\rowcolor[HTML]{B4C6E7} 
\multicolumn{1}{|c|}{\textbf{Name}}
& \multicolumn{1}{c|}{\textbf{Requirements}}                                      & \textbf{GUI}
& \textbf{CLI} 
& \multicolumn{1}{c|}{\textbf{HPO Mechanisms}}                              
& \multicolumn{1}{c|}{\textbf{Ref.}}                            

\\ \hline\endhead
Auptimizer & Sklearn                    & Y & Y & Random, Grid, Multi-armed Bandit (MAB), Bayesian, Evolutionary      & \cite{Auptimizer, litr19} 

\\ \hline
Bayesian Optimisation & Sklearn                    & N    & N & Bayesian             & \cite{BayesianOptimization} 

\\ \hline
BayesOpt   & NONE            & N   & N & Bayesian           & \cite{Bayesopt, ma14}   

\\ \hline
BoTorch    & Torch           & N    & N & Bayesian             & \cite{BoTorch, baka19}    

\\ \hline
BTB        & Sklearn                    & N    & N & Bayesian              & \cite{BTB, smsa20}        

\\ \hline
DEAP       & NONE            & N    & N & Evolutionary,   Particle Swarm Optimisation (PSO)                  & \cite{DEAP, fora12}       

\\ \hline
Dragonfly  & NONE            & N           & Y & Bayesian          & \cite{DragonFly, kavy20}  

\\ \hline
GPflowOpt  & TF              & N           & N & Bayesian            & \cite{GPflowOpt, knva17}  

\\ \hline
Hyperopt   & Sklearn, Extras (LightGBM)          & N           & N & Bayesian          & \cite{Hyperopt, beya13a}   
\\ \hline
mlrMBO        & NONE                & N           & N & Bayesian         & \cite{biri18, mlrmbo}   

\\ \hline
Nevergrad  & Bayesian Optimisation, Torch, TF, Keras        & N           & Y & Evolutionary,   Random, PSO, Bayesian, Genetic, Mathematical      & \cite{Nevergrad}  

\\ \hline
Optuna     & Sklearn, Scikit-Optimize, MLflow, Extras (LightGBM, Torch, CatBoost, MXNet, XGBoost,   Keras, TF, Dask-ML, BoTorch, fastai) & N           & Y & Bayesian,   Evolutionary, Grid, Random, MAB         & \cite{Optuna, aksa19}     

\\ \hline
RBFOpt     & NONE            & N           & Y & Mathematical             & \cite{RbfOpt, cona18}     

\\ \hline
Scikit-Optimize       & Sklearn                    & N           & N & Bayesian         & \cite{scikitoptimize}       

\\ \hline
SMAC3      & Sklearn                    & N           & Y & Bayesian            & \cite{SMAC3, huho11}      

\\ \hline
Tune-sklearn          & Sklearn, Ray{[}Tune{]}                & N           & N & Random, Grid,   Bayesian         & \cite{tunesklearn}   

\\ \hline
\end{longtable}

The third category of dedicated AutoML tools in this review encompasses those that focus solely on HPO.
The surveyed open-source packages classified as such are listed in Table~\ref{table:hpotools}.

Given that the birthplace of modern AutoML is set in the topic of HPO~\cite{kemu20}, with plenty of prior discussion about optimisation methods and their implementations in the literature, an extensive commentary is not needed here.
Every tool in the table is considered dedicated, so they do not house an ML solution during its life cycle in an end-to-end MLWF.
Instead, the typical operation is for a user to provide a searchable hyperparameter space to the HPO package, then iteratively try out recommended configurations and update the package on how well those choices technically perform.
Even speed-up mechanisms such as successive halving are often presented to technicians as cues for when they need to train their ML models on larger subsets of data. It is noted here that whilst Early Stopping can be applied to a variety of HPO mechanisms, it is also included given it is offered as a method by which an end user aims to arrive at an optimal set of hyperparameters. As a definitional note, those which fall under numerical optimisation are grouped as 'Mathematical'. This includes, for example, Covariance matrix adaptation evolution strategy (CMA-ES).
Essentially, integrating these detached mechanisms into an automated model-development process requires some manual labour from stakeholders.

Tools that internalise ML models/algorithms, e.g.~ones wrapping scikit-learn or other packages, are naturally excluded from this section on account of leaning towards being comprehensive systems, i.e.~allowing a user to convert raw data into a trained ML solution.
Other exclusions are due to a lack of currency, often because meaningful activity has not been present in the repository within the last 18 months.
Sometimes, as with GPyOpt~\cite{gpyopt}, the software has been explicitly announced as deprecated.
Consequently, given that the first popularised systems of the modern AutoML era were developed around ten years ago, Appendix~\ref{Sec:fadedpurehpo} notes a relatively long history of faded HPO tools compared to other dedicated categories.

Several insights can be gleaned from Table~\ref{table:hpotools}, which notes package names and associated characteristics.
For instance, the table lists package dependencies, often pulled from a requirements.txt or setup.py file in a GitHub repository.
Notably, despite the efforts of this review to avoid delving into AutoDL, DL frameworks such as TensorFlow (TF), Keras and PyTorch (Torch) occasionally appear as such dependencies.
However, half of the packages operate around scikit-learn, even though a user must still manually code the training of an ML model.
A couple of subsequent columns in the table then assess how stakeholders interact with these packages.
In all cases beyond Auptimizer, the technical threshold for users is high; dedicated HPO tools are overwhelmingly designed to be paired with coding scripts.
That said, several packages do provide CLIs, which eases accessibility slightly.
Stakeholders avoid serious programming in such cases, simply editing a configuration file instead.

Finally, we note that the dedicated tools cover a range of HPO procedures in Table~\ref{table:hpotools}.
However, it is clear that Bayesian optimisation techniques have embedded themselves in widespread usage, even if this is partially a consequence of promotion by the research groups that initially researched this space, i.e.~a first-mover advantage.
For the same reason, bandit-based methods also have a strong representation, e.g.~via an associated Hyperband implementation offered by Auptimizer and Optuna.
In fact, recent years have seen research groups fuse the benefits of both strategies, such as with the Bayesian optimisation and Hyperband (BOHB) algorithm that Auptimizer and Optuna both provide.
Naturally, different implementations do attempt to improve upon existing methods, so there are often a variety of efficiency-based tweaks augmenting the fundamental techniques.
For instance, Tune-sklearn employs early stopping, while Auptimizer leverages successive halving, iteratively testing smaller sets of hyperparameter configurations on larger training-data samples.
Optuna does both.
Then, moving beyond these most popular mechanisms, evolutionary algorithms and particle swarm optimisation are also available, as well as other techniques termed `mathematical'.
These remnants include the use of radial basis functions in RBFOpt and sequential quadratic programming (SQP) in Nevergrad.
Ultimately, it is evident that the HPO category of dedicated AutoML tools is substantially mature.
Although the average non-technical stakeholder is unlikely to engage with such software, the technical user will probably appreciate the flexibility associated tools provide in their `detachable' form.

\subsection{Comprehensive AutoML Systems: Open-Source}
\label{Sec:coretools}

While dedicated tools certainly have their place in industrial use, the holy grail of AutoML is arguably to develop a framework that comprehensively automates \textit{every} relevant task from one end of an MLWF to the other~\cite{kemu20}.
The field is nowhere near that goal yet.
Nonetheless, within the last decade, many implementations have arisen that can automate the core facet of automated model development, i.e.~CASH, while managing the fundamental ML process of model training.
It is this internalised operation of converting data into an ML model that differentiates such packages from the dedicated HPO tools in Section~\ref{Sec:purehpotools}.
Some of these architectures have even extended their scope further out to neighbouring tasks and phases within an MLWF.


\begin{longtable}{|l|l|}
\caption{The list of surveyed comprehensive AutoML systems that are open-source and active.}
\label{Table:alltools}
\\
\hline
\rowcolor[HTML]{B4C6E7}
\textbf{Name}   & \textbf{Ref.}             \\ \hline
\endhead
Auto\_ViML    & \cite{AutoVimal}   \\ \hline
AutoGluon     & \cite{autogluon}   \\ \hline
AutoKeras     & \cite{autokeras}   \\ \hline
AutoML Alex   & \cite{AutoMLAlex}  \\ \hline
Auto-PyTorch  & \cite{AutoPyTorch}   \\ \hline
auto-sklearn  & \cite{autosklearn}   \\ \hline
carefree-learn & \cite{carefreelearn}  \\ \hline
FLAML         & \cite{flaml}   \\ \hline
GAMA          & \cite{gama}   \\ \hline
HyperGBM      & \cite{HyperGBM}   \\ \hline
Hyperopt-sklearn & \cite{hyperoptsklearn} \\ \hline
Igel           & \cite{igel}  \\ \hline
Lightwood      & \cite{lightwood}  \\ \hline
Ludwig         & \cite{ludwig}  \\ \hline
Mljar          & \cite{mljarsupervised}  \\ \hline
mlr3automl     & \cite{mlr3automl}  \\ \hline
OBOE           & \cite{oboe}  \\ \hline
PyCaret       & \cite{pycaret}   \\ \hline
TPOT          & \cite{tpot}   \\ \hline
\end{longtable}

In this section, we analyse 19 open-source tools, listed in Table~\ref{Table:alltools}, that the survey found to be comprehensive AutoML systems.
Each tool is assessed according to the criteria for performant ML introduced in Section~\ref{Sec:summarisedcriteria}, with information sourced from both available documentation and associated codebases.
Importantly, not all parts of the evaluation framework apply to every tool, e.g.~if the system is provided in the form of Python libraries, and these cases are highlighted by commentary when relevant.
Furthermore, this review finds a broad spectrum of capability, ranging from holistic implementations that assist stakeholders for several MLWF phases to those that barely meet the definition of comprehensive.
Additionally, we re-emphasise that the listed tools are considered current; historically notable faded projects are included in Appendix~\ref{Sec:fadedcoreautoml} for completeness.

With that context provided, the following findings are typically presented in tabular format.
They may note:
\begin{itemize}
	\item How each tool scores across a grouped set of criteria, thus providing a comparison between tools.
	Scores are also summed across all tools, thus assessing the general maturity of the AutoML `market' per sub-criterion.
	\item The number of tools that score at different levels for a particular sub-criterion.
	This aggregation is another insight into `market' maturity.
\end{itemize}
As Section~\ref{Sec:summarisedcriteria} details, all scores are usually binary, i.e.~1/0 for yes/no, or on an integer scale.
Occasionally, a sub-criterion involves listing tool features instead, and these are expanded into their own tables.

We now proceed to discuss the criteria, organised as follows: efficiency in Section~\ref{Sec:OpenEfficiency}, dirty data in Section~\ref{Sec:OpenDirtyData}, completeness \& currency in Section~\ref{Sec:OpenCompleteness}, explainability in Section~\ref{Sec:OpenExplainability}, ease of use in Section~\ref{Sec:OpenEase}, and the remaining elements of the assessment framework in Section~\ref{Sec:OpenRemnants}.

\subsubsection{Efficiency}
\label{Sec:OpenEfficiency}

\begin{table}[h]
\caption{Scores for open-source comprehensive AutoML systems (E1--E3). Evaluates the existence of a model repository (E1), a model VCS (E2), and experiment tracking (E3). See 
Table~\ref{Table:Criteria_E1-3} 
for rubric.}
\begin{tabular}{|l|c|c|c|c|}
\hline
\rowcolor[HTML]{B4C6E7} 
\textbf{Name}                                     & \textbf{E1} & \textbf{E2} & \textbf{E3} & \textbf{Sum (out of 4)} \\ \hline
AutoML Alex                                      & 0           & 0           & 2           & 2            \\ \hline
HyperGBM                                          & 0           & 0           & 2           & 2            \\ \hline
Auto\_ViML                                        & 0           & 0           & 0           & 0            \\ \hline
AutoGluon                                         & 0           & 0           & 0           & 0            \\ \hline
AutoKeras                                         & 0           & 0           & 0           & 0            \\ \hline
Auto-PyTorch                                      & 0           & 0           & 0           & 0            \\ \hline
auto-sklearn                                      & 0           & 0           & 0           & 0            \\ \hline
carefree-learn                                    & 0           & 0           & 0           & 0            \\ \hline
FLAML                                             & 0           & 0           & 0           & 0            \\ \hline
GAMA                                              & 0           & 0           & 0           & 0            \\ \hline
Hyperopt-sklearn                                  & 0           & 0           & 0           & 0            \\ \hline
Igel                                              & 0           & 0           & 0           & 0            \\ \hline
Lightwood                                         & 0           & 0           & 0           & 0            \\ \hline
Ludwig                                            & 0           & 0           & 0           & 0            \\ \hline
Mljar                                             & 0           & 0           & 0           & 0            \\ \hline
mlr3automl                                        & 0           & 0           & 0           & 0            \\ \hline
OBOE                                              & 0           & 0           & 0           & 0            \\ \hline
PyCaret                                           & 0           & 0           & 0           & 0            \\ \hline
TPOT                                              & 0           & 0           & 0           & 0            \\ \hline
\multicolumn{1}{|c|}{\cellcolor[HTML]{B4C6E7}Sum} & 0/19           & 0/19           & 4/38           &              \\ \hline
\end{tabular}
\label{Table:framework_E1_E3_tool}
\end{table}

\begin{table}[h]
\caption{Distribution of scores for open-source comprehensive AutoML systems (E3). Evaluates the existence of experiment tracking. Scores: 0 for none, 1 for storage/access with limited automation/visuals, 2 for storage/access with automatic log visualisation, and U for unclear.}
\label{Table:framework_E3_tool}
\begin{tabular}{|cc|}
\hline
\rowcolor[HTML]{B4C6E7} 
\multicolumn{2}{|c|}{\cellcolor[HTML]{B4C6E7}\textbf{E3}}                        \\ \hline
\rowcolor[HTML]{B4C6E7} 
\multicolumn{1}{|c|}{\cellcolor[HTML]{B4C6E7}\textbf{Score}} & \textbf{\# Tools} \\ \hline
\rowcolor[HTML]{FFFFFF} 
\multicolumn{1}{|c|}{\cellcolor[HTML]{FFFFFF}0}              & 17                \\ \hline
\rowcolor[HTML]{FFFFFF} 
\multicolumn{1}{|c|}{\cellcolor[HTML]{FFFFFF}1}              & 0                 \\ \hline
\rowcolor[HTML]{FFFFFF} 
\multicolumn{1}{|c|}{\cellcolor[HTML]{FFFFFF}2}              & 2                 \\ \hline
\rowcolor[HTML]{FFFFFF} 
\multicolumn{1}{|c|}{\cellcolor[HTML]{FFFFFF}U}              & 0                 \\ \hline
\end{tabular}
\end{table}

Most open-source AutoML tools arise from academic research.
So, concerning the assessment framework for performant ML, it is almost immediately evident that operational efficiency is not the highest of priorities.
Table~\ref{Table:framework_E1_E3_tool} shows that none of the surveyed systems provides a model repository (E1) or model VCS (E2), which would require a persistent data storage of some sort.
As for experiment tracking (E3), Table~\ref{Table:framework_E3_tool} affirms that only two tools have considered this notion, namely AutoML Alex and HyperGBM.
Both are automated to the point of providing convenient visuals of experimental logs.
However, AutoML Alex benefits here by plugging into the Optuna dashboard, a dedicated tool covered in Section~\ref{Sec:purehpotools}.

Now, because all surveyed open-source systems lack a centralised form of persistent data storage, they cannot provide several other features for operational efficiency.
For instance, we identified no repository of templates/code generated either by users or developers that could automatically kickstart an ML application (E4).
That is not to say that tutorials are unavailable; all packages were found to provide example code that a stakeholder can manually leverage.
They vary in substance and quality from minimal working examples to extensive end-to-end notebooks.
Next, the suggestion of prior work (E5), which essentially requires a recommendation mechanism on top of a template/code repository, is also absent.
Admittedly, a few surveyed systems claim to offer meta-learning capabilities, such as warm-starting, but such procedures do not typically kickstart the operational side of an ML application, nor do they have flexible VCS in mind.
Thus, meta-learning commentary is deferred until criterion CC11.
Finally, if there is no centralised framework for storing project artefacts, it is understandably hard to automate shared access and collaboration (E6).

\begin{table}[h]
\caption{Scores for open-source comprehensive AutoML systems (E7/E8). Evaluates the extent of automation for assisted data exploration (E7) and data preparation (E8). See Table~\ref{Table:Criteria_E7-13} for rubric.}
\label{Table:framework_E7_E8_tool}
\begin{tabular}{|l|c|c|c|}
\hline
\rowcolor[HTML]{B4C6E7} 
\textbf{Name}                                     & \textbf{E7} & \textbf{E8} & \textbf{Sum (out of 5)} \\ \hline
Auto\_ViML                                        & 0           & 2           & 2            \\ \hline
AutoGluon                                         & 0           & 2           & 2           \\ \hline
AutoKeras                                         & 0           & 2           & 2            \\ \hline
AutoML Alex                                      & 0           & 2           & 2            \\ \hline
Auto-PyTorch                                      & 0           & 2           & 2            \\ \hline
auto-sklearn                                      & 0           & 2           & 2            \\ \hline
carefree-learn                                    & 0           & 2           & 2            \\ \hline
FLAML                                             & 0           & 2           & 2            \\ \hline
GAMA                                              & 0           & 2           & 2            \\ \hline
HyperGBM                                          & 0           & 2           & 2            \\ \hline
Hyperopt-sklearn                                  & 0           & 2           & 2            \\ \hline
Igel                                              & 0           & 2           & 2            \\ \hline
Lightwood                                         & 0           & 2           & 2            \\ \hline
Ludwig                                            & 0           & 2           & 2            \\ \hline
Mljar                                             & 0           & 2           & 2            \\ \hline
mlr3automl                                        & 0           & 2           & 2            \\ \hline
OBOE                                              & 0           & 2           & 2            \\ \hline
PyCaret                                           & 0           & 2           & 2            \\ \hline
TPOT                                              & 0           & 2           & 2            \\ \hline
\multicolumn{1}{|c|}{\cellcolor[HTML]{B4C6E7}Sum} & 0/57           & 38/38          &              \\ \hline
\end{tabular}
\end{table}

Moving on to discussions of effort minimisation during phases of an MLWF, this survey could not identify any significant assistance with data exploration (E7) among open-source packages.
Such a result is likely due to their primary focus on model development, with general assumptions that input datasets are essentially predetermined.
Accordingly, in the open-source domain, the onus for exploring/understanding data remains on the ML stakeholder.
Supplementary tools explicitly dedicated to EDA must be sourced externally.
In contrast, Table~\ref{Table:framework_E7_E8_tool} indicates that all surveyed comprehensive AutoML systems do automate some form of data preprocessing (E8).

\begin{table}[h]
\caption{Coverage of data-preparation processes for open-source comprehensive AutoML systems (E8A--E8F). Evaluates categorical processing (E8A), standardisation/normalisation (E8B), bucketing/binning (E8C), text processing (E8D), time-period extraction (E8E), and management of class imbalance (E8F). Scores: 0 for absent and 1 for present.}
\label{Table:framework_E8A_E8F_tool}
\begin{tabular}{|l|c|c|c|c|c|c|c|}
\hline
\rowcolor[HTML]{B4C6E7} 
\textbf{Name}                                     & \textbf{E8-A} & \textbf{E8-B} & \textbf{E8-C} & \textbf{E8-D} & \textbf{E8-E} & \textbf{E8-F} & \textbf{Sum (out of 6)} \\ \hline
Auto\_ViML       & 1 & 1 & 1 & 1 & 1 & 1 & 6 \\ \hline
Ludwig           & 1 & 1 & 0 & 1 & 1 & 0 & 4 \\ \hline
Mljar            & 1 & 1 & 0 & 1 & 1 & 0 & 4 \\ \hline
PyCaret          & 1 & 1 & 1 & 0 & 0 & 1 & 4 \\ \hline
AutoGluon        & 1 & 0 & 0 & 1 & 1 & 0 & 3 \\ \hline
auto-sklearn     & 1 & 1 & 0 & 0 & 0 & 1 & 3 \\ \hline
FLAML            & 1 & 0 & 0 & 1 & 0 & 1 & 3 \\ \hline
HyperGBM         & 1 & 1 & 0 & 0 & 0 & 1 & 3 \\ \hline
Hyperopt-sklearn & 1 & 1 & 0 & 1 & 0 & 0 & 3 \\ \hline
Igel             & 1 & 1 & 0 & 0 & 1 & 0 & 3 \\ \hline
Lightwood        & 1 & 1 & 0 & 0 & 1 & 0 & 3 \\ \hline
AutoKeras        & 1 & 1 & 0 & 0 & 0 & 0 & 2 \\ \hline
AutoML Alex     & 1 & 1 & 0 & 0 & 0 & 0 & 2 \\ \hline
Auto-PyTorch     & 1 & 1 & 0 & 0 & 0 & 0 & 2 \\ \hline
carefree-learn   & 1 & 1 & 0 & 0 & 0 & 0 & 2 \\ \hline
mlr3automl       & 1 & 0 & 0 & 0 & 0 & 1 & 2 \\ \hline
OBOE             & 1 & 1 & 0 & 0 & 0 & 0 & 2 \\ \hline
TPOT             & 1 & 1 & 0 & 0 & 0 & 0 & 2 \\ \hline
GAMA             & 1 & 0 & 0 & 0 & 0 & 0 & 1 \\ \hline
\multicolumn{1}{|c|}{\cellcolor[HTML]{B4C6E7}Sum} & 19/19            & 15/19            & 2/19             & 6/19             & 6/19             & 6/19             &              \\ \hline
\end{tabular}
\end{table}

Of course, data preparation can involve many processes, so it is insufficiently informative to merely state that a comprehensive AutoML system assists or automates this MLWF phase.
Therefore, we also assess the coverage of each tool across several common forms of data preprocessing:
\begin{enumerate}[label=\Alph*.]
    \item Categorical processing, e.g.~dummy and one-hot encoding.
    \item Standardisation/normalisation, i.e.~feature rescaling.
    \item Bucketing/binning, i.e.~for continuous variables.
    \item Text preparation, e.g.~tokenising or embedding.
    \item Time-period extraction, e.g.~determining day of week from dates.
    \item Class-imbalance management, i.e.~sampling techniques.
\end{enumerate}

As Table~\ref{Table:framework_E8A_E8F_tool} reveals, there appears to be plenty of variability in the scope of automated data preparation.
The only consistency is that all comprehensive AutoML systems manipulate categorical features (E8A).
Most also simplify feature rescaling (E8B), although, as exemplified by the Ludwig package, this may sometimes need a flag to be set when employing AutoML techniques.
We did not consider toggle buttons as invalidating the presence of automation, provided that, once set, users do not need to configure the procedure manually, i.e.~the system takes care of it.

On the other end of the availability spectrum, only PyCaret and Auto\_ViML assist with binning continuous variables (E8C).
Granted, this process is not essential for many ML problems, but it does highlight the differentiation between tools.
As for text (E8D) and temporal data (E8E), both formats provide their own challenges.
It is clear from their correlated coverage in Table~\ref{Table:framework_E8A_E8F_tool} -- four out of six tools that can process one can also process the other -- that a small subset of comprehensive AutoML systems is prioritising problem extensibility, e.g.~supporting NLP and time-series forecasting.
Finally, there is the matter of class imbalance (E8F).
The statistics of observable data in real-world settings can be highly skewed, leading to flawed ML models and issues with fairness.
Hence, the dearth of open-source AutoML packages that assist with or automate sampling-based corrections is somewhat surprising, suggesting a mismatch between academic and industrial expectations of ML challenges.
Whatever the reason, class imbalance remains a risk for non-technical users, both in terms of detection and mitigation, that can stymie satisfactory outcomes and, consequently, AutoML democratisation.

\begin{table}[h]
\caption{Scores for open-source comprehensive AutoML systems (E9--E13). Evaluates the extent of automation for dataset feature generation (E9), reuse (E10), and selection (E11). Also evaluates whether HPO is completely specified by default (E12) and whether optimisation involves an entire ML pipeline (E13). See Table~\ref{Table:Criteria_E7-13} for rubric.}
\label{Table:framework_E9_E13_tool}
\begin{tabular}{|l|c|c|c|c|c|c|}
\hline
\rowcolor[HTML]{B4C6E7} 
\textbf{Name}
& \textbf{E9}
& \textbf{E10}
& \textbf{E11}
& \textbf{E12}
& \textbf{E13}
& \textbf{Sum (out of 7)} \\ \hline
TPOT                        & 2      & 0      & 2      & 1       & 1            & 6            \\ \hline
Auto\_ViML                  & 1      & 0      & 2      & 1       & 1            & 5            \\ \hline
AutoML Alex                & 2      & 0      & 2      & 1       & 0            & 5            \\ \hline
HyperGBM                    & 2      & 0      & 2      & 1       & 0            & 5            \\ \hline
auto-sklearn                & 0      & 0      & 2      & 1       & 1            & 4            \\ \hline
mlr3automl                  & 0      & 0      & 2      & 1       & 1            & 4            \\ \hline
PyCaret                     & 1      & 0      & 2      & 1       & 0            & 4            \\ \hline
Mljar                       & 0      & 0      & 2      & 1       & 0            & 3            \\ \hline
GAMA                        & 0      & 0      & 0      & 1       & 1            & 2            \\ \hline
Hyperopt-sklearn            & 0      & 0      & 0      & 1       & 1            & 2            \\ \hline
OBOE                        & 0      & 0      & 0      & 1       & 1            & 2            \\ \hline
AutoGluon                   & 0      & 0      & 0      & 1       & 0            & 1            \\ \hline
AutoKeras                   & 0      & 0      & 0      & 1       & 0            & 1            \\ \hline
Auto-PyTorch                & 0      & 0      & 0      & 1       & 0            & 1            \\ \hline
carefree-learn              & 0      & 0      & 0      & 1       & 0            & 1            \\ \hline
FLAML                       & 0      & 0      & 0      & 1       & 0            & 1            \\ \hline
Igel                        & 0      & 0      & 0      & 1       & 0            & 1            \\ \hline
Lightwood                   & 0      & 0      & 0      & 1       & 0            & 1            \\ \hline
Ludwig                      & 0      & 0      & 0      & 1       & 0            & 1            \\ \hline

\cellcolor[HTML]{B4C6E7}Sum & 8/38     & 0/19      & 16/38      & 19/19      & 7/19           &              \\ \hline
\end{tabular}
\end{table}

\begin{table}[h]
\caption{Distribution of scores for open-source comprehensive AutoML systems (E9). Evaluates the extent of automation for dataset feature generation. Scores: 0 for none, 1 for convenience features, 2 for substantial automation, and U for unclear.}
\label{Table:framework_E9_tool}
\begin{tabular}{|cc|}
\hline
\rowcolor[HTML]{B4C6E7} 
\multicolumn{2}{|c|}{\cellcolor[HTML]{B4C6E7}\textbf{E9}}                        \\ \hline
\rowcolor[HTML]{B4C6E7} 
\multicolumn{1}{|c|}{\cellcolor[HTML]{B4C6E7}\textbf{Score}} & \textbf{\# Tools} \\ \hline
\rowcolor[HTML]{FFFFFF} 
\multicolumn{1}{|c|}{\cellcolor[HTML]{FFFFFF}0}              & 14                \\ \hline
\rowcolor[HTML]{FFFFFF} 
\multicolumn{1}{|c|}{\cellcolor[HTML]{FFFFFF}1}              & 2                 \\ \hline
\rowcolor[HTML]{FFFFFF} 
\multicolumn{1}{|c|}{\cellcolor[HTML]{FFFFFF}2}              & 3                 \\ \hline
\rowcolor[HTML]{FFFFFF} 
\multicolumn{1}{|c|}{\cellcolor[HTML]{FFFFFF}U}              & 0                 \\ \hline
\end{tabular}
\end{table}

Admittedly, some of the above processes blur with the phase of feature engineering.
The nuance here is perhaps intent; data preparation aims to establish a baseline of information, while feature engineering seeks to improve the quality of this baseline via further transformation.
With that perspective, Table~\ref{Table:framework_E9_E13_tool} reveals which comprehensive open-source systems embrace the generative half of AutoFE.
It turns out, as Table~\ref{Table:framework_E9_tool} attests, there are not many.
Only three tools incorporate feature generation as part of their core automation: AutoML Alex, HyperGBM, and TPOT.
Another two, namely Auto\_ViML, and PyCaret, assist users in manually configuring/running feature generation.
Naturally, certain transformations may be relevant/applicable to more than one ML application, so, for the sake of consistency and effort minimisation, e.g.~ensuring that all departments in an organisation agree on a `sales' calculation, it would also be valuable to retain such knowledge for reuse (E10).
However, as already discussed for other criteria, no open-source system has implemented the level of persistence required to support such a mechanism.

\begin{table}[h]
\caption{Distribution of scores for open-source comprehensive AutoML systems (E11). Evaluates the extent of automation for dataset feature selection. Scores: 0 for none, 1 for convenience features, 2 for substantial automation, and U for unclear.}
\label{Table:framework_E11_tool}
\begin{tabular}{|cc|}
\hline
\rowcolor[HTML]{B4C6E7}
\multicolumn{2}{|c|}{\cellcolor[HTML]{B4C6E7}\textbf{E11}}                       \\ \hline
\rowcolor[HTML]{B4C6E7}
\multicolumn{1}{|c|}{\cellcolor[HTML]{B4C6E7}\textbf{Score}} & \textbf{\# Tools} \\ \hline
\rowcolor[HTML]{FFFFFF} 
\multicolumn{1}{|c|}{\cellcolor[HTML]{FFFFFF}0}              & 11                \\ \hline
\rowcolor[HTML]{FFFFFF} 
\multicolumn{1}{|c|}{\cellcolor[HTML]{FFFFFF}1}              & 0                 \\ \hline
\rowcolor[HTML]{FFFFFF} 
\multicolumn{1}{|c|}{\cellcolor[HTML]{FFFFFF}2}              & 8                 \\ \hline
\rowcolor[HTML]{FFFFFF} 
\multicolumn{1}{|c|}{\cellcolor[HTML]{FFFFFF}U}              & 0                 \\ \hline
\end{tabular}
\end{table}

Now, every system that touches feature generation in Table~\ref{Table:framework_E9_E13_tool} also has procedures for their sampling.
In fact, Table~\ref{Table:framework_E11_tool} shows that feature selection (E11) is sometimes available even when generation is not.
Moreover, all eight systems that support feature selection tend to do so at a substantially automated level.
This outcome means that an interested stakeholder can potentially mix and match the surveyed comprehensive AutoML systems with the dedicated tools in Section~\ref{Sec:datafeaturetools}.
We also stress that, just because a system offers a specific capability, this review does not necessarily judge the quality and granular extent of that capability.
So, for instance, TPOT can explore `polynomial' features, but requirements for deeper feature synthesis are better satisfied elsewhere.

\begin{table}[h]
\caption{Distribution of scores for open-source comprehensive AutoML systems (E13). Evaluates whether optimisation involves an entire ML pipeline. Scores: 0 for no, 1 for yes, and U for unclear.}
  \label{Table:framework_E13_tool}
\begin{tabular}{|cc|}
\hline
\rowcolor[HTML]{B4C6E7}
\multicolumn{2}{|c|}{\cellcolor[HTML]{B4C6E7}\textbf{E13}}                       \\ \hline
\rowcolor[HTML]{B4C6E7}
\multicolumn{1}{|c|}{\cellcolor[HTML]{B4C6E7}\textbf{Score}} & \textbf{\# Tools} \\ \hline
\rowcolor[HTML]{FFFFFF} 
\multicolumn{1}{|c|}{\cellcolor[HTML]{FFFFFF}0}              & 12                \\ \hline
\rowcolor[HTML]{FFFFFF} 
\multicolumn{1}{|c|}{\cellcolor[HTML]{FFFFFF}1}              & 7                 \\ \hline
\rowcolor[HTML]{FFFFFF} 
\multicolumn{1}{|c|}{\cellcolor[HTML]{FFFFFF}U}              & 0                 \\ \hline
\end{tabular}
\end{table}

Finally, although later criteria address the topic in greater depth, an operational-efficiency assessment also has to consider the model-development phase.
Indeed, it can be daunting for a user to specify HPO search spaces or algorithms, so ensuring that a comprehensive system has a default procedure for undertaking CASH (E12) is paramount.
Unsurprisingly, given that this review defines comprehensive systems to revolve around model development, exceptions that force a user to configure HPO are virtually nonexistent.
In contrast, optimising an extended ML pipeline (E13), complete with data preprocessing and model postprocessing components, is still relatively rare.
Table~\ref{Table:framework_E13_tool} shows only seven out of 19 open-source packages do so.
Such a state of the `market' makes sense, as CASH+ involves managing a much more expansive hyperparameter space; this requires implementing even more advanced theoretical techniques.

\begin{table}[h]
\caption{Scores for open-source comprehensive AutoML systems (E14--E16). Evaluates the availability of workload optimisation (E14), modelling time limits (E15), and modelling iteration limits (E16). See Table~\ref{Table:Criteria_E14-16} for rubric.}
\label{Table:framework_E14_E16_tool}
\begin{tabular}{|l|c|c|c|c|}
\hline
\rowcolor[HTML]{B4C6E7}
\textbf{Name}                                     & \textbf{E14} & \textbf{E15} & \textbf{E16} & \textbf{Sum (out of 4)} \\ \hline
FLAML                                             & 0            & 1            & 1            & 2            \\ \hline
HyperGBM                                          & 0            & 1            & 1            & 2            \\ \hline
Hyperopt-sklearn                                  & 0            & 1            & 1            & 2            \\ \hline
AutoKeras                                         & 0            & 0            & 1            & 1            \\ \hline
AutoML Alex                                      & 0            & 1            & 0            & 1            \\ \hline
Auto-PyTorch                                      & 0            & 1            & 0            & 1            \\ \hline
auto-sklearn                                      & 0            & 1            & 0            & 1            \\ \hline
GAMA                                              & 0            & 1            & 0            & 1            \\ \hline
Mljar                                             & 0            & 1            & 0            & 1            \\ \hline
mlr3automl                                        & 0            & 1            & 0            & 1            \\ \hline
OBOE                                              & 0            & 1            & 0            & 1            \\ \hline
TPOT                                              & 0            & 1            & 0            & 1            \\ \hline
Auto\_ViML                                        & 0            & 0            & 0            & 0            \\ \hline
AutoGluon                                         & 0            & 0            & 0            & 0            \\ \hline
carefree-learn                                    & 0            & 0            & 0            & 0            \\ \hline
Igel                                              & 0            & 0            & 0            & 0            \\ \hline
Lightwood                                         & 0            & 0            & 0            & 0            \\ \hline
Ludwig                                            & 0            & 0            & 0            & 0            \\ \hline
PyCaret                                           & 0            & 0            & 0            & 0            \\ \hline
\multicolumn{1}{|c|}{\cellcolor[HTML]{B4C6E7}Sum} & 0/38            & 11/19           & 4/19            &              \\ \hline
\end{tabular}
\end{table}

\begin{table}[h]
\caption{Distribution of scores for open-source comprehensive AutoML systems (E15). Evaluates the availability of modelling time limits. Scores: 0 for absent, 1 for present, and U for unclear.}
\label{Table:framework_E15_tool}
\begin{tabular}{|cc|}
\hline
\rowcolor[HTML]{B4C6E7}
\multicolumn{2}{|c|}{\cellcolor[HTML]{B4C6E7}\textbf{E15}}                       \\ \hline
\rowcolor[HTML]{B4C6E7}
\multicolumn{1}{|c|}{\cellcolor[HTML]{B4C6E7}\textbf{Score}} & \textbf{\# Tools} \\ \hline
\rowcolor[HTML]{FFFFFF} 
\multicolumn{1}{|c|}{\cellcolor[HTML]{FFFFFF}0}              & 8                 \\ \hline
\rowcolor[HTML]{FFFFFF} 
\multicolumn{1}{|c|}{\cellcolor[HTML]{FFFFFF}1}              & 11                \\ \hline
\rowcolor[HTML]{FFFFFF} 
\multicolumn{1}{|c|}{\cellcolor[HTML]{FFFFFF}U}              & 0                 \\ \hline
\end{tabular}
\end{table}

\begin{table}[h]
\caption{Distribution of scores for open-source comprehensive AutoML systems (E16). Evaluates the availability of modelling iteration limits. Scores: 0 for absent, 1 for present, and U for unclear.}
\label{Table:framework_E16_tool}
\begin{tabular}{|cc|}
\hline
\rowcolor[HTML]{B4C6E7}
\multicolumn{2}{|c|}{\cellcolor[HTML]{B4C6E7}\textbf{E16}}                       \\ \hline
\rowcolor[HTML]{B4C6E7}
\multicolumn{1}{|c|}{\cellcolor[HTML]{B4C6E7}\textbf{Score}} & \textbf{\# Tools} \\ \hline
\rowcolor[HTML]{FFFFFF} 
\multicolumn{1}{|c|}{\cellcolor[HTML]{FFFFFF}0}              & 15                \\ \hline
\rowcolor[HTML]{FFFFFF} 
\multicolumn{1}{|c|}{\cellcolor[HTML]{FFFFFF}1}              & 4                 \\ \hline
\rowcolor[HTML]{FFFFFF} 
\multicolumn{1}{|c|}{\cellcolor[HTML]{FFFFFF}U}              & 0                 \\ \hline
\end{tabular}
\end{table}

Switching from operational to technical efficiency, this monograph notes that the computational load of the model-development stage is mostly intrinsic to the algorithms implemented by an AutoML developer team.
It is beyond the scope of this review to benchmark which packages take less time and memory than others to tackle any specific ML problem.
However, we do note that none of the surveyed open-source comprehensive AutoML systems offers workload optimisation (E14).
Such an absence may be noticeable to large-scale enterprise deployments, which wrestle with many non-negligible costs, e.g.~those of cloud computing.
Likewise, it is arguable that workload optimisation could also benefit ML applications on local hardware, where an experimental/production environment may be highly constrained.
Managing costs and resources for ML work can be challenging even for a technician.
Fortunately, Table~\ref{Table:framework_E15_tool} shows that more than half of the surveyed tools allow users to limit model development by time.
Trial limits are also offered, although, as Table~\ref{Table:framework_E16_tool} shows, not as broadly.
This disparity is understandable, as iteration limits tend to be a step more technical than time limits, requiring a configurer to understand what is being iterated, i.e.~they need to know the algorithms involved in model development.

\subsubsection{Dirty Data}
\label{Sec:OpenDirtyData}

\begin{table}[h]
\caption{Scores for open-source comprehensive AutoML systems (DD1). Evaluates the extent of automation for cleaning dirty data. See Table~\ref{Table:Criteria_DD1-5} for rubric.}
\label{Table:framework_DD1_all_tool}
\begin{tabular}{|l|c|}
\hline
\rowcolor[HTML]{B4C6E7}
\textbf{Name}                                     & \textbf{DD1} \\ \hline
Auto\_ViML                                        & 2            \\ \hline
AutoGluon                                         & 2            \\ \hline
AutoKeras                                         & 2            \\ \hline
AutoML Alex                                      & 2            \\ \hline
Auto-PyTorch                                      & 2            \\ \hline
auto-sklearn                                      & 2            \\ \hline
carefree-learn                                    & 2            \\ \hline
FLAML                                             & 2            \\ \hline
GAMA                                              & 2            \\ \hline
HyperGBM                                          & 2            \\ \hline
Hyperopt-sklearn                                  & 2            \\ \hline
Igel                                              & 2            \\ \hline
Lightwood                                         & 2            \\ \hline
Ludwig                                            & 2            \\ \hline
Mljar                                             & 2            \\ \hline
mlr3automl                                        & 2            \\ \hline
OBOE                                              & 2            \\ \hline
PyCaret                                           & 2            \\ \hline
TPOT                                              & 2            \\ \hline
\multicolumn{1}{|c|}{\cellcolor[HTML]{B4C6E7}Sum} & 38/38           \\ \hline
\end{tabular}
\end{table}

At this point, open-source comprehensive AutoML systems have been assessed primarily on how they enhance operational efficiency.
Part of that analysis has examined the extent of automation during the preparation and subsequent engineering of dataset features.
However, this review defines such tasks under the assumption that incoming data is essentially clean, if perhaps sampled awkwardly, e.g.~with a class imbalance.
In contrast, real-world data has to contend with many gaps and formatting issues.
In fact, the `dirtiness' of these inputs is a significant and ubiquitous challenge in industrial applications, relatively avoidable for many academic research projects, which is why its management warrants a distinct criterion.
Given that context, it is perhaps surprising then that every surveyed tool in Table~\ref{Table:framework_DD1_all_tool} automates, at least to some extent, the cleaning of dirty data (DD1). 

\begin{table}[h]
\caption{Scores for open-source comprehensive AutoML systems (DD2--DD5). Evaluates the extent of automation for data-type inference (DD2), missing-value imputation (DD3), and outlier management (DD4). Also evaluates the existence of domain-specific/advanced cleaning operations (DD5). See Table~\ref{Table:Criteria_DD1-5} for rubric.}
\label{Table:framework_DD2_DD5_tool}
\begin{tabular}{|l|c|c|c|c|c|}
\hline
\rowcolor[HTML]{B4C6E7}
\textbf{Name}                                     & \textbf{DD2} & \textbf{DD3} & \textbf{DD4} & \textbf{DD5} & \textbf{Sum (out of 6)} \\ \hline
PyCaret                                           & 1            & 2            & 2            & 0            & 5            \\ \hline
AutoML Alex                                      & 1            & 1            & 2            & 0            & 4            \\ \hline
OBOE                                              & 1            & 1            & 2            & 0            & 4            \\ \hline
auto-sklearn                                      & 1            & 2            & 0            & 0            & 3            \\ \hline
Igel                                              & 1            & 2            & 0            & 0            & 3            \\ \hline
TPOT                                              & 1            & 2            & 0            & 0            & 3            \\ \hline
Auto\_ViML                                        & 1            & 1            & 0            & 0            & 2            \\ \hline
Auto-PyTorch                                      & 1            & 1            & 0            & 0            & 2            \\ \hline
carefree-learn                                    & 1            & 1            & 0            & 0            & 2            \\ \hline
HyperGBM                                          & 1            & 1            & 0            & 0            & 2            \\ \hline
Ludwig                                            & 0            & 2            & 0            & 0            & 2            \\ \hline
Mljar                                             & 1            & 1            & 0            & 0            & 2            \\ \hline
mlr3automl                                        & 1            & 1            & 0            & 0            & 2            \\ \hline
AutoGluon                                         & 1            & 0            & 0            & 0            & 1            \\ \hline
AutoKeras                                         & 1            & 0            & 0            & 0            & 1            \\ \hline
FLAML                                             & 1            & 0            & 0            & 0            & 1            \\ \hline
GAMA                                              & 1            & 0            & 0            & 0            & 1            \\ \hline
Hyperopt-sklearn                                  & 1            & 0            & 0            & 0            & 1            \\ \hline
Lightwood                                         & 1            & 0            & 0            & 0            & 1            \\ \hline
\multicolumn{1}{|c|}{\cellcolor[HTML]{B4C6E7}Sum} & 18/19           & 18/38           & 6/38            & 0/19            &              \\ \hline
\end{tabular}
\end{table}

\begin{table}[h]
\caption{Distribution of scores for open-source comprehensive AutoML systems (DD2). Evaluates the extent of automation for data-type inference. Scores: 0 for absent, 1 for present, and U for unclear.}
\label{Table:framework_DD2_tool}
\begin{tabular}{|cc|}
\hline
\rowcolor[HTML]{B4C6E7}
\multicolumn{2}{|c|}{\cellcolor[HTML]{B4C6E7}\textbf{DD2}}                       \\ \hline
\rowcolor[HTML]{B4C6E7}
\multicolumn{1}{|c|}{\cellcolor[HTML]{B4C6E7}\textbf{Score}} & \textbf{\# Tools} \\ \hline
\rowcolor[HTML]{FFFFFF} 
\multicolumn{1}{|c|}{\cellcolor[HTML]{FFFFFF}0}              & 1                 \\ \hline
\rowcolor[HTML]{FFFFFF} 
\multicolumn{1}{|c|}{\cellcolor[HTML]{FFFFFF}1}              & 18                \\ \hline
\rowcolor[HTML]{FFFFFF} 
\multicolumn{1}{|c|}{\cellcolor[HTML]{FFFFFF}U}              & 0                 \\ \hline
\end{tabular}
\end{table}

Naturally, the cleaning capabilities of the surveyed open-source systems are varied, as indicated by Table~\ref{Table:framework_DD2_DD5_tool}.
There is only one quasi-consistency that Table~\ref{Table:framework_DD2_tool} shows, in that almost all tools can infer the types of data passed into the system (DD2), potentially enabling specialised cleaning/processing.
The Ludwig package is a rare exception.
In fact, Ludwig appears to serve as a platform that is almost fully controlled via configuration files.
This reliance on manual specification means that the system barely scrapes into consideration as AutoML software.
Granted, required configurability is not an immediate disqualification, provided that the finer details of a process are sufficiently mechanised.

\begin{table}[h]
\caption{Distribution of scores for open-source comprehensive AutoML systems (DD3). Evaluates the extent of automation for missing-value imputation. Scores: 0 for none, 1 for automatic detection, 2 for automatic detection/resolution, and U for unclear.}
\label{Table:framework_DD3_tool}
\begin{tabular}{|cc|}
\hline
\rowcolor[HTML]{B4C6E7}
\multicolumn{2}{|c|}{\cellcolor[HTML]{B4C6E7}\textbf{DD3}}                       \\ \hline
\rowcolor[HTML]{B4C6E7}
\multicolumn{1}{|c|}{\cellcolor[HTML]{B4C6E7}\textbf{Score}} & \textbf{\# Tools} \\ \hline
\rowcolor[HTML]{FFFFFF} 
\multicolumn{1}{|c|}{\cellcolor[HTML]{FFFFFF}0}              & 6                 \\ \hline
\rowcolor[HTML]{FFFFFF} 
\multicolumn{1}{|c|}{\cellcolor[HTML]{FFFFFF}1}              & 8                 \\ \hline
\rowcolor[HTML]{FFFFFF} 
\multicolumn{1}{|c|}{\cellcolor[HTML]{FFFFFF}2}              & 5                 \\ \hline
\rowcolor[HTML]{FFFFFF} 
\multicolumn{1}{|c|}{\cellcolor[HTML]{FFFFFF}U}              & 0                 \\ \hline
\end{tabular}
\end{table}

In any case, the treatment of missing values (DD3) is an example of varied capability, with open-source packages scoring a summed 18 out of a maximal 38.
The distribution in Table~\ref{Table:framework_DD3_tool} reveals that this is simply a result of statistical spread, with AutoML developers almost uniformly choosing between ignorance, automated detection alone, and automated detection/resolution.
Where applicable, packages often differ in imputation strategy, regardless of whether they employ convenience features or outright automation.
For instance, TPOT, Mljar and AutoML Alex prefer taking a median of observed values, while Igel, PyCaret, Auto-PyTorch and mlr3automl replace missing values with a mean.
Both auto-sklearn and OBOE are a little more sophisticated, as imputation methods are considered part of an ML pipeline with optimisable hyperparameters.
As for the Ludwig package, it curiously defaults to replacement by zero values.
Additionally, this survey could not easily determine the technical details employed by HyperGBM, carefree-learn, and Auto\_ViML.

\begin{table}[h]
\caption{Distribution of scores for open-source comprehensive AutoML systems (DD4). Evaluates the extent of automation for outlier management. Scores: 0 for none, 1 for automatic detection, 2 for automatic detection/resolution, and U for unclear.}
\label{Table:framework_DD4_tool}
\begin{tabular}{|cc|}
\hline
\rowcolor[HTML]{B4C6E7}
\multicolumn{2}{|c|}{\cellcolor[HTML]{B4C6E7}\textbf{DD4}}                       \\ \hline
\rowcolor[HTML]{B4C6E7}
\multicolumn{1}{|c|}{\cellcolor[HTML]{B4C6E7}\textbf{Score}} & \textbf{\# Tools} \\ \hline
\rowcolor[HTML]{FFFFFF} 
\multicolumn{1}{|c|}{\cellcolor[HTML]{FFFFFF}0}              & 16                \\ \hline
\rowcolor[HTML]{FFFFFF} 
\multicolumn{1}{|c|}{\cellcolor[HTML]{FFFFFF}1}              & 0                 \\ \hline
\rowcolor[HTML]{FFFFFF} 
\multicolumn{1}{|c|}{\cellcolor[HTML]{FFFFFF}2}              & 3                 \\ \hline
\rowcolor[HTML]{FFFFFF} 
\multicolumn{1}{|c|}{\cellcolor[HTML]{FFFFFF}U}              & 0                 \\ \hline
\end{tabular}
\end{table}

Outlier management (DD4) is a far rarer capability among open-source systems, which makes sense when considering that a missing value is easier to identify than a statistical anomaly.
For the same reason, Table~\ref{Table:framework_DD4_tool} shows that if an AutoML developer bothers with detection, they will likely pursue full resolution.
Admittedly, the solution is typically a simple excision of anomalous data instances.
Regardless, minor notes for this sub-criterion include PyCaret requiring the user to set a flag when seeking to find/process anomalies, AutoML Alex employing the interquartile range (IQR) method, and OBOE exploring various techniques for outlier management.
Beyond these mechanisms, this survey found nothing else among the assessed tools that might qualify as advanced or domain-specific cleaning (DD5).

\subsubsection{Completeness and Currency}
\label{Sec:OpenCompleteness}


\begin{table}[h]
\small{
\caption{Scores for open-source comprehensive AutoML systems (CC1--CC9). Evaluates capabilities for unsupervised learning (CC1), regression on tabular data (CC2), standard classification on tabular data (CC3), multi-class classification on tabular data (CC4), time-series forecasting (CC5), image-based problem solving (CC6), text-based problem solving (CC7), multi-modal problem solving (CC8), and ensemble techniques (CC9). See Table~\ref{Table:Criteria_CC1-9} for rubric.}
\label{Table:framework_CC1_CC9_tool}
\begin{tabular}{|l|c|c|c|c|c|c|c|c|c|C{1.6cm}|}
\hline
\rowcolor[HTML]{B4C6E7}
\textbf{Name} &
  \textbf{CC1} &
  \textbf{CC2} &
  \textbf{CC3} &
  \textbf{CC4} &
  \textbf{CC5} &
  \textbf{CC6} &
  \textbf{CC7} &
  \textbf{CC8} &
  \textbf{CC9} &
  \textbf{Sum\newline(out of 17)} \\ \hline
TPOT             & 2 & 2 & 2 & 2 & 2 & 2 & 2 & 0 & 2 & 16 \\ \hline
AutoKeras        & 0 & 2 & 2 & 2 & 2 & 2 & 2 & 1 & 2 & 15 \\ \hline
auto-sklearn     & 1 & 2 & 2 & 2 & 2 & 1 & 1 & 0 & 2 & 13 \\ \hline
Ludwig           & 2 & 2 & 2 & 2 & 1 & 1 & 1 & 0 & 2 & 13 \\ \hline
AutoGluon        & 0 & 2 & 2 & 2 & 2 & 2 & 2 & 0 & 0 & 12 \\ \hline
Igel             & 1 & 2 & 2 & 2 & 1 & 1 & 2 & 0 & 1 & 12 \\ \hline
PyCaret          & 2 & 2 & 2 & 2 & 2 & 2 & 0 & 0 & 0 & 12 \\ \hline
FLAML            & 2 & 1 & 2 & 2 & 1 & 1 & 1 & 0 & 1 & 11 \\ \hline
Mljar            & 1 & 2 & 2 & 2 & 2 & 0 & 2 & 0 & 0 & 11 \\ \hline
Auto-PyTorch     & 1 & 2 & 2 & 2 & 1 & 1 & 1 & 0 & 0 & 10 \\ \hline
Auto\_ViML       & 0 & 2 & 2 & 2 & 2 & 0 & 0 & 0 & 0 & 8  \\ \hline
AutoML Alex     & 0 & 2 & 2 & 2 & 2 & 0 & 0 & 0 & 0 & 8  \\ \hline
carefree-learn   & 2 & 2 & 2 & 2 & 0 & 0 & 0 & 0 & 0 & 8  \\ \hline
GAMA             & 1 & 1 & 2 & 2 & 1 & 0 & 1 & 0 & 0 & 8  \\ \hline
HyperGBM         & 0 & 2 & 2 & 2 & 0 & 0 & 0 & 0 & 2 & 8  \\ \hline
Hyperopt-sklearn & 0 & 2 & 2 & 2 & 1 & 1 & 0 & 0 & 0 & 8  \\ \hline
Lightwood        & 0 & 2 & 2 & 2 & 1 & 0 & 1 & 0 & 0 & 8  \\ \hline
OBOE             & 1 & 1 & 1 & 1 & 1 & 1 & 1 & 0 & 1 & 8  \\ \hline
mlr3automl       & 1 & 1 & 1 & 1 & 1 & 1 & 1 & 0 & U & 7  \\ \hline
\multicolumn{1}{|c|}{\cellcolor[HTML]{B4C6E7}Sum} &
  17/38 &
  34/38 &
  36/38 &
  36/38 &
  25/38 &
  16/38 &
  18/38 &
  1/19 &
  13/38 &
   \\ \hline
\end{tabular}
}
\end{table}

While the ability to polish data and enhance its information content is crucial to ML applications, end-to-end AutoML will probably always revolve most tightly around the model-development phase of an MLWF.
Stakeholders deciding on which software to engage with will likely ask a fundamental question: what problems can these ML models solve?
The answer, at least for the surveyed open-source systems, is provided in Table~\ref{Table:framework_CC1_CC9_tool}.

\begin{table}[h]
\caption{Distribution of scores for open-source comprehensive AutoML systems (CC1). Evaluates capabilities for unsupervised learning. Scores: 0 for none, 1 for convenience features, 2 for substantial automation, and U for unclear.}
\label{Table:framework_CC1_tool}
\begin{tabular}{|cc|}
\hline
\rowcolor[HTML]{B4C6E7}
\multicolumn{2}{|c|}{\cellcolor[HTML]{B4C6E7}\textbf{CC1}}                       \\ \hline
\rowcolor[HTML]{B4C6E7}
\multicolumn{1}{|c|}{\cellcolor[HTML]{B4C6E7}\textbf{Score}} & \textbf{\# Tools} \\ \hline
\rowcolor[HTML]{FFFFFF} 
\multicolumn{1}{|c|}{\cellcolor[HTML]{FFFFFF}0}              & 7                 \\ \hline
\rowcolor[HTML]{FFFFFF} 
\multicolumn{1}{|c|}{\cellcolor[HTML]{FFFFFF}1}              & 7                 \\ \hline
\rowcolor[HTML]{FFFFFF} 
\multicolumn{1}{|c|}{\cellcolor[HTML]{FFFFFF}2}              & 5                 \\ \hline
\rowcolor[HTML]{FFFFFF} 
\multicolumn{1}{|c|}{\cellcolor[HTML]{FFFFFF}U}              & 0                 \\ \hline
\end{tabular}
\end{table}

Of all the assessed capabilities, unsupervised learning (CC1) would appear to be the most unusual.
While the process can still be expressed under the regular aims of finding a mathematical model that best approximates a desirable function, this desirable function has no prior supervisory hints as to what it may be.
Indeed, what typically happens is that principled techniques, such as clustering methods that minimise intra-cluster distances or inertia, are thrown at a problem in an exploratory fashion with the hope of teasing out patterns/insights that may satisfy stakeholders.
Accordingly, there is rarely any consistent way to apply CASH, as the objective function to maximise is human interest, with arcane dependencies on real-world context.
The only real exception is if the outcomes of an unsupervised-learning process help train subsequent ML predictors/prescriptors, the performances of which can be tested and validated.
In this case, automated unsupervised learning can be seen as a sophisticated form of feature generation.
Given all these nuances, Table~\ref{Table:framework_CC1_tool} is therefore notable for showing that unsupervised learning still has a healthy representation among surveyed tools, both in terms of convenience features and automation.

\begin{table}[h]
\caption{Distribution of scores for open-source comprehensive AutoML systems (CC2). Evaluates capabilities for regression on tabular data. Scores: 0 for none, 1 for convenience features, 2 for substantial automation, and U for unclear.}
\label{Table:framework_CC2_tool}
\begin{tabular}{|cc|}
\hline
\rowcolor[HTML]{B4C6E7}
\multicolumn{2}{|c|}{\cellcolor[HTML]{B4C6E7}\textbf{CC2}}                       \\ \hline
\rowcolor[HTML]{B4C6E7}
\multicolumn{1}{|c|}{\cellcolor[HTML]{B4C6E7}\textbf{Score}} & \textbf{\# Tools} \\ \hline
\rowcolor[HTML]{FFFFFF} 
\multicolumn{1}{|c|}{\cellcolor[HTML]{FFFFFF}0}              & 0                 \\ \hline
\rowcolor[HTML]{FFFFFF} 
\multicolumn{1}{|c|}{\cellcolor[HTML]{FFFFFF}1}              & 4                 \\ \hline
\rowcolor[HTML]{FFFFFF} 
\multicolumn{1}{|c|}{\cellcolor[HTML]{FFFFFF}2}              & 15                \\ \hline
\rowcolor[HTML]{FFFFFF} 
\multicolumn{1}{|c|}{\cellcolor[HTML]{FFFFFF}U}              & 0                 \\ \hline
\end{tabular}
\end{table}
\begin{table}[h]
\caption{Distribution of scores for open-source comprehensive AutoML systems (CC3). Evaluates capabilities for standard classification on tabular data. Scores: 0 for none, 1 for convenience features, 2 for substantial automation, and U for unclear.}
\label{Table:framework_CC3_tool}
\begin{tabular}{|cc|}
\hline
\rowcolor[HTML]{B4C6E7}
\multicolumn{2}{|c|}{\cellcolor[HTML]{B4C6E7}\textbf{CC3}}                       \\ \hline
\rowcolor[HTML]{B4C6E7}
\multicolumn{1}{|c|}{\cellcolor[HTML]{B4C6E7}\textbf{Score}} & \textbf{\# Tools} \\ \hline
\rowcolor[HTML]{FFFFFF} 
\multicolumn{1}{|c|}{\cellcolor[HTML]{FFFFFF}0}              & 0                 \\ \hline
\rowcolor[HTML]{FFFFFF} 
\multicolumn{1}{|c|}{\cellcolor[HTML]{FFFFFF}1}              & 2                 \\ \hline
\rowcolor[HTML]{FFFFFF} 
\multicolumn{1}{|c|}{\cellcolor[HTML]{FFFFFF}2}              & 17                \\ \hline
\rowcolor[HTML]{FFFFFF} 
\multicolumn{1}{|c|}{\cellcolor[HTML]{FFFFFF}U}              & 0                 \\ \hline
\end{tabular}
\end{table}
\begin{table}[h]
\caption{Distribution of scores for open-source comprehensive AutoML systems (CC4). Evaluates capabilities for multi-class classification on tabular data. Scores: 0 for none, 1 for convenience features, 2 for substantial automation, and U for unclear.}
\label{Table:framework_CC4_tool}
\begin{tabular}{|cc|}
\hline
\rowcolor[HTML]{B4C6E7}
\multicolumn{2}{|c|}{\cellcolor[HTML]{B4C6E7}\textbf{CC4}}                       \\ \hline
\rowcolor[HTML]{B4C6E7}
\multicolumn{1}{|c|}{\cellcolor[HTML]{B4C6E7}\textbf{Score}} & \textbf{\# Tools} \\ \hline
\rowcolor[HTML]{FFFFFF} 
\multicolumn{1}{|c|}{\cellcolor[HTML]{FFFFFF}0}              & 0                 \\ \hline
\rowcolor[HTML]{FFFFFF} 
\multicolumn{1}{|c|}{\cellcolor[HTML]{FFFFFF}1}              & 2                 \\ \hline
\rowcolor[HTML]{FFFFFF} 
\multicolumn{1}{|c|}{\cellcolor[HTML]{FFFFFF}2}              & 17                \\ \hline
\rowcolor[HTML]{FFFFFF} 
\multicolumn{1}{|c|}{\cellcolor[HTML]{FFFFFF}U}              & 0                 \\ \hline
\end{tabular}
\end{table}

Nonetheless, as expected, supervised learning is the predominant ML process that all comprehensive AutoML systems prioritise.
Tabular data tends to be the most simple structure to deal with, so regression (CC2) and classification (CC3) capabilities are well represented, as shown by Table~\ref{Table:framework_CC2_tool} and Table~\ref{Table:framework_CC3_tool}, respectively.
Only two packages constitute the difference, not quite automating regression while still doing so with classification.
As for multi-class classification (CC4), both the detailed Table~\ref{Table:framework_CC1_CC9_tool} and the aggregated Table~\ref{Table:framework_CC4_tool} indicate that it seems just as easy to offer as the standard binary version.

\begin{table}[h]
\caption{Distribution of scores for open-source comprehensive AutoML systems (CC5). Evaluates capabilities for time-series forecasting. Scores: 0 for none, 1 for convenience features, 2 for substantial automation, and U for unclear.}
\label{Table:framework_CC5_tool}
\begin{tabular}{|cc|}
\hline
\rowcolor[HTML]{B4C6E7}
\multicolumn{2}{|c|}{\cellcolor[HTML]{B4C6E7}\textbf{CC5}}                       \\ \hline
\rowcolor[HTML]{B4C6E7}
\multicolumn{1}{|c|}{\cellcolor[HTML]{B4C6E7}\textbf{Score}} & \textbf{\# Tools} \\ \hline
\rowcolor[HTML]{FFFFFF} 
\multicolumn{1}{|c|}{\cellcolor[HTML]{FFFFFF}0}              & 2                 \\ \hline
\rowcolor[HTML]{FFFFFF} 
\multicolumn{1}{|c|}{\cellcolor[HTML]{FFFFFF}1}              & 9                 \\ \hline
\rowcolor[HTML]{FFFFFF} 
\multicolumn{1}{|c|}{\cellcolor[HTML]{FFFFFF}2}              & 8                 \\ \hline
\rowcolor[HTML]{FFFFFF} 
\multicolumn{1}{|c|}{\cellcolor[HTML]{FFFFFF}U}              & 0                 \\ \hline
\end{tabular}
\end{table}
\begin{table}[h]
\caption{Distribution of scores for open-source comprehensive AutoML systems (CC6). Evaluates capabilities for image-based problem solving. Scores: 0 for none, 1 for convenience features, 2 for substantial automation, and U for unclear.}
\label{Table:framework_CC6_tool}
\begin{tabular}{|cc|}
\hline
\rowcolor[HTML]{B4C6E7}
\multicolumn{2}{|c|}{\cellcolor[HTML]{B4C6E7}\textbf{CC6}}                       \\ \hline
\rowcolor[HTML]{B4C6E7}
\multicolumn{1}{|c|}{\cellcolor[HTML]{B4C6E7}\textbf{Score}} & \textbf{\# Tools} \\ \hline
\rowcolor[HTML]{FFFFFF} 
\multicolumn{1}{|c|}{\cellcolor[HTML]{FFFFFF}0}              & 7                 \\ \hline
\rowcolor[HTML]{FFFFFF} 
\multicolumn{1}{|c|}{\cellcolor[HTML]{FFFFFF}1}              & 8                 \\ \hline
\rowcolor[HTML]{FFFFFF} 
\multicolumn{1}{|c|}{\cellcolor[HTML]{FFFFFF}2}              & 4                 \\ \hline
\rowcolor[HTML]{FFFFFF} 
\multicolumn{1}{|c|}{\cellcolor[HTML]{FFFFFF}U}              & 0                 \\ \hline
\end{tabular}
\end{table}
\begin{table}[h]
\caption{Distribution of scores for open-source comprehensive AutoML systems (CC7). Evaluates capabilities for text-based problem solving. Scores: 0 for none, 1 for convenience features, 2 for substantial automation, and U for unclear.}
\label{Table:framework_CC7_tool}
\begin{tabular}{|cc|}
\hline
\rowcolor[HTML]{B4C6E7}
\multicolumn{2}{|c|}{\cellcolor[HTML]{B4C6E7}\textbf{CC7}}                       \\ \hline
\rowcolor[HTML]{B4C6E7}
\multicolumn{1}{|c|}{\cellcolor[HTML]{B4C6E7}\textbf{Score}} & \textbf{\# Tools} \\ \hline
\rowcolor[HTML]{FFFFFF} 
\multicolumn{1}{|c|}{\cellcolor[HTML]{FFFFFF}0}              & 6                 \\ \hline
\rowcolor[HTML]{FFFFFF} 
\multicolumn{1}{|c|}{\cellcolor[HTML]{FFFFFF}1}              & 8                 \\ \hline
\rowcolor[HTML]{FFFFFF} 
\multicolumn{1}{|c|}{\cellcolor[HTML]{FFFFFF}2}              & 5                 \\ \hline
\rowcolor[HTML]{FFFFFF} 
\multicolumn{1}{|c|}{\cellcolor[HTML]{FFFFFF}U}              & 0                 \\ \hline
\end{tabular}
\end{table}
\begin{table}[h]
\caption{Distribution of scores for open-source comprehensive AutoML systems (CC8). Evaluates capabilities for multi-modal problem solving. Scores: 0 for absent, 1 for present, and U for unclear.}
\label{Table:framework_CC8_tool}
\begin{tabular}{|cc|}
\hline
\rowcolor[HTML]{B4C6E7}
\multicolumn{2}{|c|}{\cellcolor[HTML]{B4C6E7}\textbf{CC8}}                       \\ \hline
\rowcolor[HTML]{B4C6E7}
\multicolumn{1}{|c|}{\cellcolor[HTML]{B4C6E7}\textbf{Score}} & \textbf{\# Tools} \\ \hline
\rowcolor[HTML]{FFFFFF} 
\multicolumn{1}{|c|}{\cellcolor[HTML]{FFFFFF}0}              & 18                \\ \hline
\rowcolor[HTML]{FFFFFF} 
\multicolumn{1}{|c|}{\cellcolor[HTML]{FFFFFF}1}              & 1                 \\ \hline
\rowcolor[HTML]{FFFFFF} 
\multicolumn{1}{|c|}{\cellcolor[HTML]{FFFFFF}U}              & 0                 \\ \hline
\end{tabular}
\end{table}

Problems based on time-series forecasting (CC5) are not too exotic, and Table~\ref{Table:framework_CC5_tool} shows their coverage is moderate, with slightly fewer packages providing full automation over convenience features.
However, once stakeholder aims move to text-based problems (CC6) and image-based problems (CC7), there is a substantial drop-off in automated capability, as reflected in Table~\ref{Table:framework_CC6_tool} and Table~\ref{Table:framework_CC7_tool}, respectively.
Around a third of the surveyed systems do not support NLP tasks, image recognition, and the like.
Understandably, the data structures involved tend to be more complex, usually warranting DL techniques that are more commonly found in AutoDL software.
As for multi-modal problems (CC8), they are essentially a more challenging target of such capabilities, combining data sources in different formats, e.g.~tabular information, free-form text, and images.
It is not overly surprising that Table~\ref{Table:framework_CC8_tool} depicts only one package operating in this space, i.e.~AutoKeras.

\begin{table}[h]
\caption{Distribution of scores for open-source comprehensive AutoML systems (CC9). Evaluates capabilities for ensemble techniques. Scores: 0 for none, 1 for convenience features, 2 for substantial automation, and U for unclear.}
\label{Table:framework_CC9_tool}
\begin{tabular}{|cc|}
\hline
\rowcolor[HTML]{B4C6E7}
\multicolumn{2}{|c|}{\cellcolor[HTML]{B4C6E7}\textbf{CC9}}                       \\ \hline
\rowcolor[HTML]{B4C6E7}
\multicolumn{1}{|c|}{\cellcolor[HTML]{B4C6E7}\textbf{Score}} & \textbf{\# Tools} \\ \hline
\rowcolor[HTML]{FFFFFF} 
\multicolumn{1}{|c|}{\cellcolor[HTML]{FFFFFF}0}              & 10                \\ \hline
\rowcolor[HTML]{FFFFFF} 
\multicolumn{1}{|c|}{\cellcolor[HTML]{FFFFFF}1}              & 3                 \\ \hline
\rowcolor[HTML]{FFFFFF} 
\multicolumn{1}{|c|}{\cellcolor[HTML]{FFFFFF}2}              & 5                 \\ \hline
\rowcolor[HTML]{FFFFFF} 
\multicolumn{1}{|c|}{\cellcolor[HTML]{FFFFFF}U}              & 1                 \\ \hline
\end{tabular}
\end{table}

Finally, although ensemble techniques (CC9) represent not an ML problem but a way to solve them, their inclusion in an ML application can dramatically change the relationship between an ML solution and a constituent ML model~\cite{kemu20}.
In short, depending on context/implementation, leveraging ensembles may well feel like tackling a different type of ML problem.
Overall, Table~\ref{Table:framework_CC9_tool} suggests that most open-source comprehensive systems avoid these techniques, with only five offering significant automation.
Given that ensemble approaches are well known to be powerful in ML, e.g.~reducing predictive variance without increasing bias, one reason for this may be the computational complexity of robustly tuning solutions with multiple predictors.
How to do so effectively remains an open research question.

\begin{table}[h]
\caption{Coverage of model-selection processes for open-source comprehensive AutoML systems (CC10/CC11). Evaluates capabilities for custom evaluation metrics (CC10). Also evaluates the existence of HPO techniques (CC11), i.e.~grid search (GR), random search (RA), Bayesian optimisation (BA), multi-armed bandit strategies (MAB), genetic/evolutionary algorithms (GE), and meta-learning (M). For convenience, additionally classifies whether HPO mechanisms beyond grid/random search (GR+) are provided. Scores: 0 for absent and 1 for present.}
\label{Table:framework_CC10_CC13_tool}
\begin{tabular}{|l|c|c|c|c|c|c|c|c|c|}
\hline
\rowcolor[HTML]{B4C6E7}
\textbf{Name} &
  \textbf{CC10} &
  \textbf{GR} &
  \textbf{RA} &
  \textbf{BA} &
  \textbf{MAB} &
  \textbf{GE} &
  \textbf{M} &
  \textbf{SUM (out of 7)} &
  \textbf{GR+} \\ \hline
AutoML Alex     & 1 & 1 & 1 & 0 & 1 & 1 & 0 & 5 & 1 \\ \hline
carefree-learn   & 1 & 1 & 1 & 1 & 1 & 0 & 0 & 5 & 1 \\ \hline
PyCaret          & 0 & 1 & 1 & 1 & 1 & 0 & 0 & 4 & 1 \\ \hline
AutoGluon        & 1 & 0 & 1 & 1 & 1 & 0 & 0 & 4 & 1 \\ \hline
auto-sklearn     & 1 & 0 & 0 & 1 & 0 & 0 & 1 & 3 & 1 \\ \hline
FLAML            & 1 & 0 & 1 & 1 & 0 & 0 & 0 & 3 & 1 \\ \hline
HyperGBM         & 1 & 0 & 1 & 0 & 0 & 1 & 0 & 3 & 1 \\ \hline
mlr3automl       & 0 & 1 & 1 & 0 & 1 & 0 & 0 & 3 & 1 \\ \hline
Auto\_ViML       & 0 & 1 & 1 & 0 & 0 & 0 & 0 & 2 & 0 \\ \hline
AutoKeras        & 1 & 0 & 0 & 1 & 0 & 0 & 0 & 2 & 1 \\ \hline
GAMA             & 0 & 0 & 1 & 0 & 0 & 1 & 0 & 2 & 1 \\ \hline
Hyperopt-sklearn & 1 & 0 & 0 & 1 & 0 & 0 & 0 & 2 & 1 \\ \hline
Igel             & 0 & 1 & 1 & 0 & 0 & 0 & 0 & 2 & 0 \\ \hline
Lightwood        & 1 & 0 & 0 & 1 & 0 & 0 & 0 & 2 & 1 \\ \hline
Mljar            & 0 & 1 & 1 & 0 & 0 & 0 & 0 & 2 & 0 \\ \hline 
TPOT             & 1 & 0 & 0 & 0 & 0 & 1 & 0 & 2 & 1 \\ \hline
Auto-PyTorch     & 0 & 0 & 0 & 0 & 1 & 0 & 0 & 1 & 1 \\ \hline
Ludwig           & 0 & 0 & 0 & 1 & 0 & 0 & 0 & 1 & 1 \\ \hline
OBOE             & 0 & 0 & 0 & 0 & 0 & 0 & 1 & 1 & 1 \\ \hline
\cellcolor[HTML]{B4C6E7}\textbf{Sum} &
  10/19 &
  7/19 &
  11/19 &
  9/19 &
  6/19 &
  4/19 &
  2/19 &
 &
  16/19 \\ \hline
\end{tabular}
\end{table}
\begin{table}[h]
\caption{Distribution of scores for open-source comprehensive AutoML systems (CC10). Evaluates capabilities for custom evaluation metrics. Scores: 0 for absent, 1 for present, and U for unclear.}
\label{Table:framework_CC10_tool}
\begin{tabular}{|cc|}
\hline
\rowcolor[HTML]{B4C6E7}
\multicolumn{2}{|c|}{\cellcolor[HTML]{B4C6E7}\textbf{CC10}}                      \\ \hline
\rowcolor[HTML]{B4C6E7}
\multicolumn{1}{|c|}{\cellcolor[HTML]{B4C6E7}\textbf{Score}} & \textbf{\# Tools} \\ \hline
\rowcolor[HTML]{FFFFFF} 
\multicolumn{1}{|c|}{\cellcolor[HTML]{FFFFFF}0}              & 9                 \\ \hline
\rowcolor[HTML]{FFFFFF} 
\multicolumn{1}{|c|}{\cellcolor[HTML]{FFFFFF}1}              & 10                \\ \hline
\rowcolor[HTML]{FFFFFF} 
\multicolumn{1}{|c|}{\cellcolor[HTML]{FFFFFF}U}              & 0                 \\ \hline
\end{tabular}
\end{table}

Now, regardless of ML application, a core focus of comprehensive AutoML is finding the best ML solution for an ML problem; this involves optimising hyperparameters, which may or may not include the type of ML model itself.
Optimisation, both at the low level of parameters -- this is known as model training -- and the high level of hyperparameters, is driven by objective functions.
Customising such evaluation metrics (CC10) can prove very attractive to industry stakeholders, allowing them to define success for specific contexts and domains.
For instance, recommendation engines may prefer to optimise accuracy for their top N suggestions rather than deal with a thorough aggregate, even if the recommendation quality rapidly falls outside this top-N selection.
After all, end-users are unlikely to scroll very far through a list of suggestions~\cite{gush09}.
Accordingly, acknowledging the benefits, around half of the surveyed open-source systems offer metric customisability, as detailed by Table~\ref{Table:framework_CC10_CC13_tool} and summed within Table~\ref{Table:framework_CC10_tool}.

As for HPO strategies (CC11), Table~\ref{Table:framework_CC10_CC13_tool} elaborates on which types have currently found purchase amongst open-source software.
The most basic form, grid search (GR), is provided by approximately a third of the surveyed tools.
Of course, grid search is not renowned for its performance, partially because a regularly defined grid is somewhat arbitrary.
A comical package named HungaBunga~\cite{HungaBunga} exemplifies this, offering users the `power' of grid-search HPO at extreme granularity.
Random search (RA) is then an alternative, although it is often lumped in with grid search as another unsophisticated approach.
Such a perspective is why this review finds that 16 of 19 open-source systems go beyond grid/random search (GR+).
In fairness, though, random search has proven to be remarkably competitive, especially for its simplicity and within highly dimensional spaces.
Unsurprisingly, the technique is the most widely covered in this space, offered by ten packages.

Out of the more sophisticated methods, Bayesian approaches (BA) prove to be the most popular, offered by nine open-source systems.
This result mimics their prevalence for dedicated HPO tools, discussed in Section~\ref{Sec:purehpotools}.
It remains challenging to determine whether Bayesian techniques are truly theoretically and implementationally optimal for the modern era of hardware or whether they simply profit from a first-mover advantage.
Most likely, both factors play a part.
The same goes for the popularity of multi-armed bandit strategies (MAB), represented by the Hyperband implementation, which has been embraced by the Bayesian optimisation community in the form of BOHB and associated methods.
Finally, genetic algorithms and other evolutionary approaches (GE) have minor representation in four systems, although this relative sparsity may arguably be due to modern hardware still failing to leverage their relative robustness.
Naturally, there are also other HPO techniques on offer that do not fit in the categories above, but they are typically confined to one AutoML framework each, e.g.~fine-tuning based on hill-climbing for Mljar and BlendSearch for FLAML.

Optimisation strategies for AutoML are, of course, constantly evolving.
Meta-learning (M), the ability to accelerate model development based on prior experience, is a noticeable gap in this space.
Only two tools claim to provide such a capability, i.e.~auto-sklearn and OBOE.
Admittedly, effective meta-learning requires a challenging infrastructure to be in place.
Local long-term storage is unlikely to be sufficiently informative, as a single user may not generate enough history to benefit their future ML applications.
Instead, meta-learning is ideally fuelled by a hive repository that services many users, and developing/managing this may be out of reach for most open-source developers.
Granted, the theory behind the concept is also continuously being explored and refined~~\cite{lebu15,albu15,ngke21,keng22}.
Nonetheless, other innovations have surfaced around HPO implementations, primarily relating to configurability.
For instance, AutoGluon allows users to toggle the `quality' of an AutoML run, limiting fit/inference time.
Likewise, FLAML has a `low\_cost\_partial\_config' parameter that enables the quick generation of low-cost models.
In the meantime, Mljar provides multiple modes of operation that tune the balance of expected technical performance and explainability from the desired ML solution, i.e.~`explain', `perform', `compete', and `Optuna'.
This review finds that such customisability is more frequent among vendor-based AutoML products, as seen in Section~\ref{Sec:commercialtools}, so the provision of such services among open-source systems suggests the acknowledgement that model validity is not everything; some users and scenarios prioritise speed or explainability.

\begin{table}[h]
\caption{Coverage of popular libraries for open-source comprehensive AutoML systems (CC12).}
\label{Table:framework_CC12_tool}
\centering
~\footnotesize{
\begin{tabular}{|p{0.15\linewidth}|c|c|c|c|c|c|c|c|c|}
\hline
\rowcolor[HTML]{B4C6E7}
\textbf{Name} &
  \textbf{Sklearn} &
  \textbf{Keras} &
  \textbf{TF} &
  \textbf{XGBoost} &
  \textbf{LightGBM} &
  \textbf{CatBoost} &
  \textbf{PyTorch} &
  \textbf{Ax} &
  \textbf{R} \\ \hline
Auto\_ViML       & 1 &   &   & 1 &   & 1                      &   &   &   \\ \hline
AutoGluon        & 1 &   &   &   & 1 & 1                      &   &   &   \\ \hline
AutoKeras        &   & 1 &   &   &   &                        &   &   &   \\ \hline
AutoML Alex     & 1 &   &   & 1 & 1 & 1 &   &   &   \\ \hline
Auto-PyTorch     &   &   &   &   &   &                        & 1 &   &   \\ \hline
auto-sklearn     & 1 &   &   &   &   &                        &   &   &   \\ \hline
carefree-learn   &   &   &   &   &   &                        & 1 &   &   \\ \hline
FLAML            & 1 &   &   & 1 & 1 &                        &   &   &   \\ \hline
GAMA             & 1 &   &   &   &   &                        &   &   &   \\ \hline
HyperGBM         &   &   &   & 1 & 1 & 1                      &   &   &   \\ \hline
Hyperopt-sklearn & 1 &   &   &   &   &                        &   &   &   \\ \hline
Igel             & 1 &   &   &   &   &                        &   &   &   \\ \hline
Lightwood        &   &   &   &   &   &                        &   & 1 &   \\ \hline
Ludwig           &   &   & 1 &   &   &                        &   &   &   \\ \hline
Mljar            & 1 &   &   & 1 & 1 & 1 &   &   &   \\ \hline
mlr3automl       &   &   &   &   &   &                        &   &   & 1 \\ \hline
OBOE             & 1 &   &   &   &   &                        &   &   &   \\ \hline
PyCaret          & 1 &   &   & 1 & 1 & 1 &   &   &   \\ \hline
TPOT             & 1 &   &   &   &   &                        &   &   &   \\ \hline
\multicolumn{1}{|c|}{\cellcolor[HTML]{B4C6E7}Sum} &
  12/19 &
  1/19 &
  1/19 &
  6/19 &
  \multicolumn{1}{c|}{6/19} &
  \multicolumn{1}{c|}{6/19} &
  \multicolumn{1}{c|}{2/19} &
  \multicolumn{1}{c|}{1/19} &
  \multicolumn{1}{c|}{1/19} \\ \hline
\end{tabular}
}
\end{table}

The final major sub-criterion under completeness \& currency notes which major algorithmic libraries each comprehensive system interfaces with (CC12).
As Table~\ref{Table:framework_CC12_tool} shows, Sklearn is dominant, which accords with the widespread modern uptake of Python for scientific programming.
Such a concentration can be risky, as problems in an oversubscribed dependency will propagate to numerous AutoML packages.
However, in practice, the community is large enough that any issues should be rapidly fixed.
There is also a smaller but still notable presence of boosting libraries referenced within the open-source codebases, such as XGBoost, LightGBM, and CatBoost.
Less well represented are the DL packages, i.e.~Keras, TF, and Torch.
Only one or two systems interface with each one.
Of course, such a result would likely change if this review intended to dive deeper into NAS software.
Finally, as outliers, one open-source system operates with the Facebook Ax library, while mlr3automl is linked with the R ecosystem.
We use the programming language R to refer to this latter category, as, unlike with the Pythonic concentration of ML algorithms into a package like Sklearn, the dependencies of mlr3automl are splintered and hard to group.

As a side note, there is a sub-criterion dedicated to whether a piece of AutoML software is being actively maintained (CC13).
However, as mentioned when discussing exclusions at the beginning of Section~\ref{Sec:Tools}, every package in the main body of this review has seen significant activity in recent times.
Faded systems of interest are listed in Appendix~\ref{Sec:fadedcoreautoml}.

\subsubsection{Explainability}
\label{Sec:OpenExplainability}

\begin{table}[h]
\caption{Scores for open-source comprehensive AutoML systems (EX3--EX7). Evaluates capabilities for enhancing global interpretability (EX3), enhancing local interpretability (EX4), scenario analysis (EX5), bias/fairness assessment (EX6), and bias/fairness management (EX7). See Table~\ref{Table:Criteria_EX1-5} for rubric.}
\label{Table:framework_EX4_EX7_tool}
\begin{tabular}{|l|c|c|c|c|c|c|}
\hline
\rowcolor[HTML]{B4C6E7}
\textbf{Name}               & \textbf{EX3} & \textbf{EX4} & \textbf{EX5} & \textbf{EX6} & \textbf{EX7} & \textbf{Sum (out of 9)} \\ \hline
Auto\_ViML       & 2 & 0 & 0 & 0 & 0 & 2 \\ \hline
AutoGluon        & 2 & 0 & 0 & 0 & 0 & 2 \\ \hline
GAMA             & 2 & 0 & 0 & 0 & 0 & 2 \\ \hline
Lightwood        & 2 & 0 & 0 & 0 & 0 & 2 \\ \hline
Ludwig           & 2 & 0 & 0 & 0 & 0 & 2 \\ \hline
Mljar            & 2 & 0 & 0 & 0 & 0 & 2 \\ \hline
PyCaret          & 2 & 0 & 0 & 0 & 0 & 2 \\ \hline
AutoKeras        & 0 & 0 & 0 & 0 & 0 & 0 \\ \hline
AutoML Alex     & 0 & 0 & 0 & 0 & 0 & 0 \\ \hline
Auto-PyTorch     & 0 & 0 & 0 & 0 & 0 & 0 \\ \hline
auto-sklearn     & 0 & 0 & 0 & 0 & 0 & 0 \\ \hline
carefree-learn   & 0 & 0 & 0 & 0 & 0 & 0 \\ \hline
FLAML            & 0 & 0 & 0 & 0 & 0 & 0 \\ \hline
HyperGBM         & 0 & 0 & 0 & 0 & 0 & 0 \\ \hline
Hyperopt-sklearn & 0 & 0 & 0 & 0 & 0 & 0 \\ \hline
Igel             & 0 & 0 & 0 & 0 & 0 & 0 \\ \hline
mlr3automl       & 0 & 0 & 0 & 0 & 0 & 0 \\ \hline
OBOE             & 0 & 0 & 0 & 0 & 0 & 0 \\ \hline
TPOT             & 0 & 0 & 0 & 0 & 0 & 0 \\ \hline
\cellcolor[HTML]{B4C6E7}Sum & 
14/38           & 
0/38            &
0/19            & 
0/38            & 
0/38            &              \\ \hline
\end{tabular}
\end{table}

\begin{table}[h]
\caption{Distribution of scores for open-source comprehensive AutoML systems (EX3). Evaluates capabilities for enhancing global interpretability. Scores: 0 for none, 1 for convenience features, 2 for substantial automation, and U for unclear.}
  \label{Table:framework_EX3_tool}
\begin{tabular}{|cc|}
\hline
\rowcolor[HTML]{B4C6E7}
\multicolumn{2}{|c|}{\cellcolor[HTML]{B4C6E7}\textbf{EX3}}                       \\ \hline
\rowcolor[HTML]{B4C6E7}
\multicolumn{1}{|c|}{\cellcolor[HTML]{B4C6E7}\textbf{Score}} & \textbf{\# Tools} \\ \hline
\rowcolor[HTML]{FFFFFF} 
\multicolumn{1}{|c|}{\cellcolor[HTML]{FFFFFF}0}              & 12                \\ \hline
\rowcolor[HTML]{FFFFFF} 
\multicolumn{1}{|c|}{\cellcolor[HTML]{FFFFFF}1}              & 0                 \\ \hline
\rowcolor[HTML]{FFFFFF} 
\multicolumn{1}{|c|}{\cellcolor[HTML]{FFFFFF}2}              & 7                 \\ \hline
\rowcolor[HTML]{FFFFFF} 
\multicolumn{1}{|c|}{\cellcolor[HTML]{FFFFFF}U}              & 0                 \\ \hline
\end{tabular}
\end{table}

Despite research and debate stretching back decades, trust in AI has only really become of critical mainstream interest in recent years.
This surge is partially driven by the accelerating uptake of ML technologies within broad sectors of industry and society, as well as the increasing friction between model outcomes and real-world nuances.
Thus, with many open-source implementations of AutoML hailing from academic roots, it is not surprising that developers have been slow to prioritise the explainability requirements of non-technical users.
Indeed, we found little evidence of facilities among the surveyed tools that help clarify data lineage (EX1) and modelling steps (EX2), although TPOT does produce pipeline visualisations and HyperGBM does provide experimental progress graphs via a dashboard.
Overall, as indicated by Table~\ref{Table:framework_EX4_EX7_tool}, the only mechanisms related to explainability that have a significant presence among open-source systems are those related to global model interpretability (EX3).
The provision of such services varies in quality and extensiveness, but Table~\ref{Table:framework_EX3_tool} asserts that seven tools do so with sufficient automation.
At one end of the spectrum, AutoGluon provides a table of global feature importance, while, at the other, PyCaret supplies a dashboard and extensive graphs.

For now, no tool appears to enhance local interpretability (EX4), i.e.~by allowing specific predictions to be analysed for the forces that drive them.
Capabilities for scenario analysis (EX5) are similarly missing.
This absence is likely to be keenly felt by stakeholders wishing to employ ML for prescriptive analytics, where the expected impact of an action is best evaluated against a counterfactual prediction, i.e.~the outcome that occurs when the action is not taken.
As for bias and fairness issues, the survey found no notable presence of mechanisms for either identification (EX6) or mitigation (EX7).
Accordingly, the current open-source `market' of AutoML limits itself in terms of appeal to industry stakeholders, at least from an explainability perspective.
However, technically savvy users do have the option of integrating the dedicated tools listed in Section~\ref{Sec:biasfairnesstools}, once their own limitations have been acknowledged.
Such a workaround may patch any relevant gaps within an MLWF.
Furthermore, the dearth of explainability services in the open-source space needs to be seen in the context of ongoing trends; the next generation of AutoML software will likely have had more time to react to the current societal focus on trust in AI.

\subsubsection{Ease of Use}
\label{Sec:OpenEase}

\begin{table}[h]
\caption{Scores for open-source comprehensive AutoML systems (EU1--EU5). Evaluates the availability of interactions via coding (EU1), CLI (EU2), and GUI (EU3). Also evaluates software client type (EU4) and level of documentation (EU5). See Table~\ref{Table:Criteria_EU1-5} for rubric.}
\label{Table:framework_EU1_EU5_tool}
\begin{tabular}{|l|c|c|c|c|c|c|}
\hline
\rowcolor[HTML]{B4C6E7}
\textbf{Name}                        & \textbf{EU1} & \textbf{EU2} & \textbf{EU3} & \textbf{EU4} & \textbf{EU5} & \textbf{Sum (out of 7)} \\ \hline
GAMA             & 1 & 1 & 1 & 2 & 2 & 7 \\ \hline
HyperGBM         & 1 & 1 & 1 & 2 & 1 & 6 \\ \hline
Igel             & 1 & 1 & 1 & 2 & 1 & 6 \\ \hline
Ludwig           & 1 & 1 & 0 & 0 & 2 & 4 \\ \hline
TPOT             & 1 & 1 & 0 & 0 & 2 & 4 \\ \hline
AutoGluon        & 1 & 0 & 0 & 0 & 2 & 3 \\ \hline
AutoKeras        & 1 & 0 & 0 & 0 & 2 & 3 \\ \hline
AutoML Alex     & 1 & 0 & 0 & 0 & 2 & 3 \\ \hline
auto-sklearn     & 1 & 0 & 0 & 0 & 2 & 3 \\ \hline
carefree-learn   & 1 & 0 & 0 & 0 & 2 & 3 \\ \hline
Mljar            & 1 & 0 & 0 & 0 & 2 & 3 \\ \hline
PyCaret          & 1 & 0 & 0 & 0 & 2 & 3 \\ \hline
Auto\_ViML       & 1 & 0 & 0 & 0 & 1 & 2 \\ \hline
Auto-PyTorch     & 1 & 0 & 0 & 0 & 1 & 2 \\ \hline
FLAML            & 1 & 0 & 0 & 0 & 1 & 2 \\ \hline
Hyperopt-sklearn & 1 & 0 & 0 & 0 & 1 & 2 \\ \hline
Lightwood        & 1 & 0 & 0 & 0 & 1 & 2 \\ \hline
mlr3automl       & 1 & 0 & 0 & 0 & 1 & 2 \\ \hline
OBOE             & 1 & 0 & 0 & 0 & 1 & 2 \\ \hline
\cellcolor[HTML]{B4C6E7}\textbf{Sum} & 
19/19           & 
5/19            & 
3/19            & 
6/38            & 
29/38           &              \\ \hline
\end{tabular}
\end{table}

\begin{table}[h]
\caption{Distribution of scores for open-source comprehensive AutoML systems (EU2). Evaluates the availability of interactions via CLI. Scores: 0 for absent, 1 for present, and U for unclear.}
\label{Table:framework_EU2_tool}
\begin{tabular}{|cc|}
\hline
\rowcolor[HTML]{B4C6E7}
\multicolumn{2}{|c|}{\cellcolor[HTML]{B4C6E7}\textbf{EU2}}                       \\ \hline
\rowcolor[HTML]{B4C6E7}
\multicolumn{1}{|c|}{\cellcolor[HTML]{B4C6E7}\textbf{Score}} & \textbf{\# Tools} \\ \hline
\rowcolor[HTML]{FFFFFF} 
\multicolumn{1}{|c|}{\cellcolor[HTML]{FFFFFF}0}              & 14                \\ \hline
\rowcolor[HTML]{FFFFFF} 
\multicolumn{1}{|c|}{\cellcolor[HTML]{FFFFFF}1}              & 5                 \\ \hline
\rowcolor[HTML]{FFFFFF} 
\multicolumn{1}{|c|}{\cellcolor[HTML]{FFFFFF}U}              & 0                 \\ \hline
\end{tabular}
\end{table}

\begin{table}[h]
\caption{Distribution of scores for open-source comprehensive AutoML systems (EU3). Evaluates the availability of interactions via GUI. Scores: 0 for absent, 1 for present, and U for unclear.}
\label{Table:framework_EU3_tool}
\begin{tabular}{|cc|}
\hline
\rowcolor[HTML]{B4C6E7}
\multicolumn{2}{|c|}{\cellcolor[HTML]{B4C6E7}\textbf{EU3}}                       \\ \hline
\rowcolor[HTML]{B4C6E7}
\multicolumn{1}{|c|}{\cellcolor[HTML]{B4C6E7}\textbf{Score}} & \textbf{\# Tools} \\ \hline
\rowcolor[HTML]{FFFFFF} 
\multicolumn{1}{|c|}{\cellcolor[HTML]{FFFFFF}0}              & 16                \\ \hline
\rowcolor[HTML]{FFFFFF} 
\multicolumn{1}{|c|}{\cellcolor[HTML]{FFFFFF}1}              & 3                 \\ \hline
\rowcolor[HTML]{FFFFFF} 
\multicolumn{1}{|c|}{\cellcolor[HTML]{FFFFFF}U}              & 0                 \\ \hline
\end{tabular}
\end{table}

Assessing open-source comprehensive AutoML systems for their problem-solving capabilities is all fine and well, but even the most mechanised frameworks will be avoided if stakeholders struggle to interface with them.
Thus, the ease-of-use criterion aims to gauge accessibility to some degree, although deeper discussions of HCI are better reserved for another review~\cite{khke21}.
Of course, the codebases of all open-source tools are, by definition, open to the public, so Table~\ref{Table:framework_EU1_EU5_tool} starts off by highlighting that all surveyed systems can be used via scripting.
Most of the packages are written in Python, so the technical barrier is arguably lower than it could be.
Nonetheless, for non-technical users, a simple CLI (EU2) is the next step in accessibility.
Immediately, support for this kind of HCI drops from 19 packages to five, as Table~\ref{Table:framework_EU2_tool} shows.
Furthermore, only three of these five, as aggregated by Table~\ref{Table:framework_EU3_tool}, bother to develop a GUI (EU3).

\begin{table}[h]
\caption{Distribution of scores for open-source comprehensive AutoML systems (EU4). Evaluates software client type. Scores: 0 for desktop only, 1 for browser only, 2 for desktop or browser, and U for unclear.}
\label{Table:framework_EU4_tool}
\begin{tabular}{|cc|}
\hline
\rowcolor[HTML]{B4C6E7}
\multicolumn{2}{|c|}{\cellcolor[HTML]{B4C6E7}\textbf{EU4}}                       \\ \hline
\rowcolor[HTML]{B4C6E7}
\multicolumn{1}{|c|}{\cellcolor[HTML]{B4C6E7}\textbf{Score}} & \textbf{\# Tools} \\ \hline
\rowcolor[HTML]{FFFFFF} 
\multicolumn{1}{|c|}{\cellcolor[HTML]{FFFFFF}0}              & 16                \\ \hline
\rowcolor[HTML]{FFFFFF} 
\multicolumn{1}{|c|}{\cellcolor[HTML]{FFFFFF}1}              & 0                 \\ \hline
\rowcolor[HTML]{FFFFFF} 
\multicolumn{1}{|c|}{\cellcolor[HTML]{FFFFFF}2}              & 3                 \\ \hline
\rowcolor[HTML]{FFFFFF} 
\multicolumn{1}{|c|}{\cellcolor[HTML]{FFFFFF}U}              & 0                 \\ \hline
\end{tabular}
\end{table}
\begin{table}[h]
\caption{Distribution of scores for open-source comprehensive AutoML systems (EU5). Evaluates level of documentation. Scores: 0 for none, 1 for partial, 2 for extensive, and U for unclear.}
\label{Table:framework_EU5_tool}
\begin{tabular}{|cc|}
\hline
\rowcolor[HTML]{B4C6E7}
\multicolumn{2}{|c|}{\cellcolor[HTML]{B4C6E7}\textbf{EU5}}                       \\ \hline
\rowcolor[HTML]{B4C6E7}
\multicolumn{1}{|c|}{\cellcolor[HTML]{B4C6E7}\textbf{Score}} & \textbf{\# Tools} \\ \hline
\rowcolor[HTML]{FFFFFF} 
\multicolumn{1}{|c|}{\cellcolor[HTML]{FFFFFF}0}              & 0                 \\ \hline
\rowcolor[HTML]{FFFFFF} 
\multicolumn{1}{|c|}{\cellcolor[HTML]{FFFFFF}1}              & 9                 \\ \hline
\rowcolor[HTML]{FFFFFF} 
\multicolumn{1}{|c|}{\cellcolor[HTML]{FFFFFF}2}              & 10                 \\ \hline
\rowcolor[HTML]{FFFFFF} 
\multicolumn{1}{|c|}{\cellcolor[HTML]{FFFFFF}U}              & 0                 \\ \hline
\end{tabular}
\end{table}

Given these results, it will become evident in Section~\ref{Sec:commercialtools} that commercial AutoML products are much more concerned with accessibility.
Here, this contrast immediately suggests that the notion of democratisation remains a buzzword in the academically rooted open-source sphere, at least compared to the goal of `data-science enhancement' specified in Section~\ref{Sec:automlrole}.
This reasoning also explains why few non-commercial packages are designed for accessible clients (EU4), e.g.~as a browser-based application rather than desktop software.
Only the rare GUI providers do so, i.e.~GAMA, HyperGBM, and Igel.
To their credit, as noted in Table~\ref{Table:framework_EU4_tool}, they support both browser \textit{and} desktop access.
In any case, Table~\ref{Table:framework_EU5_tool} indicates that all surveyed systems are reasonably well documented on average, although the breadth and detail vary considerably.
For instance, at the time of review, mlr3automl provided only a readme file on its GitHub page, admittedly including a vignette, whereas PyCaret was supported by comprehensive and sectioned documentation hosted on a separate website.

At this point, we highlight two open-source packages that did not make the cut for the core survey but still exemplify the diversity of developmental efforts in AutoML.
In particular, they implement unique approaches to HCI.
First is Libra~\cite{libra}, which contains an NLP engine intending to translate naturally stated queries into machine-understandable instructions.
Such a framework marks a rare case of an AutoML tool tackling automation at the earliest stages of an MLWF, i.e.~problem formulation \& context understanding.
Significant advances of this type would make it easier for stakeholders beyond data scientists to translate a business objective into an ML problem, e.g.~by establishing whether an intended task can be reframed as unsupervised clustering or supervised classification/regression.
The other noteworthy package is Otto~\cite{otto}, which operates as a chatbot.
A user essentially answers a series of questions and is returned a scikit-learn script that can be run to generate desired ML models.
While not technically a `comprehensive' system, as an ML model is not explicitly created and managed, Otto thus demonstrates an alternative way to view ML applications.
Ultimately, time will tell whether Libra and Otto are simply anomalies or whether they are early forerunners to a future wave of HCI innovation in the technological space of AutoML.

\subsubsection{Remaining Criteria}
\label{Sec:OpenRemnants}

By and large, this review found that open-source AutoML systems do not address the last criteria related to performant ML.
Specifically, none of the surveyed tools has invested in the automation of any deployment \& management effort (DM1--DM11), while mechanisms for governance \& security (G1--G3) are likewise absent.
This result may be partially due to all packages operating primarily offline.
That stated, some of the software does offer assistance for deployment, which is worth noting.
For instance, AutoGluon and PyCaret both provide tutorials on how to productionise their generated ML models. Additionally, research run by the AutoGluon team has also investigated model compression \cite{famu20}.
Given its close links to Amazon Web Services (AWS), the former unsurprisingly promotes SageMaker for deployment, while the latter suggests various options.
As for Ludwig and Igel, both wrap FastAPI within assistive functions to build an API endpoint for a trained model.
Then there is the carefree-learn package, which seemingly has a sister `deployment' repository \cite{carefreelearndeploy}, albeit with little detail and documentation.
Ultimately though, while governance \& security can presently be excused as luxury features, the gap in automated coverage along the latter stages of an MLWF is glaring, at least for organisations that rely on running real-world ML applications regularly.
Perhaps managing deployment and ongoing maintenance, i.e.~MLOps, is too costly and out of scope for most open-source developers.
However, as will be seen, once money gets involved, the status quo for AutoML services shifts.

\subsection{Comprehensive AutoML Systems: Commercial}
\label{Sec:commercialtools}

As reviewed at length, interested stakeholders have numerous open-source tools available that can automate parts of an ML application.
A decent amount is `comprehensive' in that the packages can manage many core tasks around generating a good ML model without any human involvement.
However, there are apparent gaps in their coverage and priorities relating to performant ML.
For instance, there is very little in the way of deploying and maintaining an ML solution, i.e.~the latter phases of an MLWF.
Additionally, open-source UIs are not typically designed with general accessibility in mind.
Thus, one may argue that, with its academic roots, open-source AutoML software is almost intended as an optional addendum to an ML application; its focus is to enhance the productivity of already technically capable data scientists.
However, as Section~\ref{Sec:keystakeholders} detailed, there are many non-technical stakeholders who are keen to leverage ML but are unwilling to spend time and money on data-science training.
At the broader scale, it is not even feasible to satiate the surging demand for robust data-driven decision-making by educating more data scientists alone.
Ideally, autonomously operating ML systems will be what meets these industry-wide needs.
Nevertheless, the point here is that lay users require a far greater degree of hand-holding through the processes of an ML application.
Accordingly, there is a profit to be made for any vendors willing to provide services in this market, but, for that coin to be earned, some product differentiation must necessarily exist.
In light of this, we review the commercial comprehensive AutoML systems that have arisen to meet the broader demand of industries and organisations.

Now, to some extent, this survey collated a list of commercial packages in the same manner as for open-source software, i.e.~via popular search engines, blogs, and other relevant websites.
Examples of common information sources include AnalyticsVidhya~\cite{AnalyticsVidhya}, KDNuggets~\cite{KDnuggets}, Gartner~\cite{GartnerCompetitors}, and Wikipedia~\cite{Wikipedia}.
However, the commercial sphere does introduce novel nuances.
For instance, with vendors competing for clientele, some summary software details could be inspected on comparison sites, such as G2~\cite{G2com} and AlternativeTo~\cite{AlternativeTo}.

Another consequence of competition is that developers in this space have seemingly sought to find and occupy niches with greater intensity than in the open-source market.
Sometimes, this differentiation appears at the fringes of the provided AutoML services, given that most vendors still aim to supply `comprehensive' coverage of an MLWF.
However, there are `dedicated' commercial tools as well, even if they are a mix of too few and too opaque to warrant the extended discussion their open-source counterparts received in Section~\ref{Sec:ancillarytools}.
As examples of such dedicated commercial tools, Hazy~\cite{Hazy} and MostlyAI~\cite{MOSTLYAI} both focus on data preparation, fleshing out model-training inputs with statistically similar synthetic data.
There is an added business benefit: using fake data maintains privacy around the real instances.
Elsewhere, provided that a user does all the preliminary coding and modelling, there are platforms for logging, sorting, viewing and assessing experiments.
These include Neptune~\cite{Neptuneai}, Comet~\cite{Comet}, and DeterminedAI~\cite{DeterminedAI}.
Then there is MLOps, another area of AutoML tool dedication.
The associated packages that \textit{only} focus on this facet are limited and do not qualify under a review of comprehensive systems, but they highlight that the deployment and maintenance phases of an MLWF are of surging interest.


\begin{longtable}{|l|l|}
\caption{The list of surveyed comprehensive AutoML systems that are commercial and active.}
\label{Table:allvendors}
\\
\hline
\rowcolor[HTML]{B4C6E7}
\textbf{Name}   & \textbf{Ref.}             \\ \hline
Alteryx         & \cite{Alteryx} \\ \hline
Auger        & \cite{Auger} \\ \hline
SageMaker (AWS)   & \cite{AmazonSageMakerAutopilot} \\ \hline
B2Metric        & \cite{B2Metric} \\ \hline
Big Squid       & \cite{BigSquid} \\ \hline
BigML           & \cite{BigML} \\ \hline
cnvrg.io           & \cite{cnvrgio} \\ \hline
Compellon       & \cite{Compellon} \\ \hline
D2iQ            & \cite{D2iQKaptain} \\ \hline
Databricks      & \cite{DataBricks} \\ \hline
Dataiku         & \cite{Dataiku} \\ \hline
DataRobot       & \cite{DataRobot} \\ \hline
Deep Cognition   & \cite{Deepcognition} \\ \hline
Einblick        & \cite{Einblick} \\ \hline
Cloud AutoML (Google)          & \cite{GoogleloudAutoML} \\ \hline
H2O          & \cite{h2o3} \\ \hline
Watson Studio (IBM)            & \cite{IBMWatsonStudio} \\ \hline
Knime           & \cite{Knime} \\ \hline
Azure AutoML (Microsoft)       & \cite{MicrosoftAzureAutoml} \\ \hline
MyDataModels    & \cite{MyDataModels} \\ \hline
Number Theory & \cite{NumberTheory} \\ \hline
RapidMiner      & \cite{Rapidminer} \\ \hline
Viya (SAS)             & \cite{SASVIYA} \\ \hline
Spell           & \cite{SPELL} \\ \hline
TIMi            & \cite{Timi} \\ \hline
\end{longtable}

Of course, to keep the survey feasible and maximally relevant to the aims of this review, a number of packages were further excluded.
Some products appear to wrap around other AutoML software and thus do not contribute to unique commentary, e.g.~Iguazio~\cite{Iguazio}, Domino~\cite{Domino}, and Qubole~\cite{Qubole}.
However, we also cannot overstate the challenge -- completely expected -- of analysing closed-source software, i.e.~its opacity.
Specifically, while a shortlist of 37 commercial products was compiled, only the 25 comprehensive AutoML systems in Table~\ref{Table:allvendors} supplied enough details for meaningful assessment.
For completeness, the remaining 12 are listed in Appendix~\ref{Sec:opaquevendor}.
Even then, information on the selected 25 packages is replete with gaps, at least to the public eye, and many more `unclear' scores are to be expected from their evaluation.
Incidentally, Table~\ref{Table:allvendors} notes, in places, both the AutoML product and the overseeing organisation.
We will often refer to the vendor names in the incoming analysis rather than a product name, as some software effectively operates within a grander ecosystem, implemented with deep connectivity to features drawn from a suite of other applications.

Before proceeding, it is worth making a few prefacing comments.
First, it is interesting to consider the incoming assessment in the context of a 2020 Kaggle State of Machine Learning and Data Science report~\cite{ka21}, which asked about the use of enterprise ML tools.
The question sampled data scientists who use the dominant AWS, Google Cloud and Microsoft Azure platforms, so it is no surprise that the flagships SageMaker ($16.5\%$), Google Cloud ML ($14.8\%$) and Azure ML Studio ($12.9\%$) received strong responses.
However, $55.2\%$ of the respondents claimed to use no ML tool on the cloud.
The report also found that $33\%$ of data scientists, presumably unbound by the platform-related subsampling, do not use AutoML tools.
The remaining usage statistics were listed as $13.9\%$ for Google Cloud AutoML, $9.5\%$ for H2O Driverless AI, $8.4\%$ for DataRobot AutoML, and $6.5\%$ for Databricks AutoML.
The takeaway is that, while AutoML has made definite inroads, the technology is still far from indispensable to ML practitioners.
At the same time, its potential, alongside data science as a whole, is becoming increasingly recognised by industry every year.
There has been a noteworthy spate of acquisitions in recent times, e.g.~Sigopt by Intel in October 2020~\cite{20}, BigSquid and Compellon by Qlik and Clearsense, respectively, in September 2021~\cite{ql21, cl21}, Ople.AI by Aktana in October 2021~\cite{ak21}, and Spell by Reddit in 2022~\cite{ha22}.
In essence, large BI and data-processing companies appear to be chasing the competitive advantage that such capabilities can provide, which aligns very well with the democratising goal of AutoML.

With those perspectives in mind, the following findings are presented in a virtually identical fashion to the open-source software analysis in Section~\ref{Sec:coretools}.
Accordingly, we now proceed to discuss the criteria, organised as follows: efficiency in Section~\ref{Sec:CommercialEfficiency}, dirty data in Section~\ref{Sec:CommercialDirtyData}, completeness \& currency in Section~\ref{Sec:CommercialCompleteness}, explainability in Section~\ref{Sec:CommercialExplainability}, ease of use in Section~\ref{Sec:CommercialEase}, deployment \& management effort in Section~\ref{Sec:CommercialDeployment}, and governance \& security in Section~\ref{Sec:CommercialGovernance}.
A final overarching commentary on commercial comprehensive AutoML systems is provided in Section~\ref{Sec:CommercialDiscussion}.

\subsubsection{Efficiency}
\label{Sec:CommercialEfficiency}

\begin{table}[h]
\caption{Scores for commercial comprehensive AutoML systems (E1--E3). Evaluates the existence of a model repository (E1), a model VCS (E2), and experiment tracking (E3). See Table~\ref{Table:Criteria_E1-3} for rubric.}
\label{Table:framework_E1_E3}
\begin{tabular}{|l|c|c|c|c|}
\hline
\rowcolor[HTML]{B4C6E7}
\textbf{Name} & \textbf{E1} & \textbf{E2} & \textbf{E3} & \textbf{Sum  (out of 4)} \\ \hline
cnvrg.io           & 1 & 1 & 2 & 4 \\ \hline
Databricks      & 1 & 1 & 2 & 4 \\ \hline
Dataiku         & 1 & 1 & 2 & 4 \\ \hline
DataRobot       & 1 & 1 & 2 & 4 \\ \hline
Microsoft       & 1 & 1 & 2 & 4 \\ \hline
SageMaker   & 1 & 1 & 2 & 4 \\ \hline
SAS             & 1 & 1 & 2 & 4 \\ \hline
Spell           & 1 & 1 & 2 & 4 \\ \hline
Alteryx         & 1 & 0 & 2 & 3 \\ \hline
Auger        & 1 & 0 & 2 & 3 \\ \hline
Big Squid       & 1 & 0 & 2 & 3 \\ \hline
BigML           & 1 & 1 & 1 & 3 \\ \hline
Deep Cognition   & 0 & 1 & 2 & 3 \\ \hline
Google          & 1 & 0 & 2 & 3 \\ \hline
H2O          & 1 & 0 & 2 & 3 \\ \hline
MyDataModels    & 1 & 0 & 2 & 3 \\ \hline
RapidMiner      & 1 & 1 & 1 & 3 \\ \hline
D2iQ            & 0 & 0 & 2 & 2 \\ \hline
Einblick        & U & U & 2 & 2 \\ \hline
IBM             & 1 & 0 & 1 & 2 \\ \hline
Number Theory & 1 & 0 & 1 & 2 \\ \hline
KNIME           & 0 & 0 & 1 & 1 \\ \hline
B2Metric        & U & U & U & 0 \\ \hline
Compellon       & 0 & 0 & 0 & 0 \\ \hline
TIMi           & 0 & 0 & 0 & 0 \\ \hline
\multicolumn{1}{|c|}{\cellcolor[HTML]{B4C6E7}Sum}           & 18/25          & 11/25          & 39/50          &              \\ \hline
\end{tabular}
\end{table}

\begin{table}[h]
\caption{Distribution of scores for commercial comprehensive AutoML systems (E1). Evaluates the existence of a model repository. Scores: 0 for absent, 1 for present, and U for unclear.}
\label{Table:framework_E1}
\begin{tabular}{|cc|}
\hline
\rowcolor[HTML]{B4C6E7}
\multicolumn{2}{|c|}{\cellcolor[HTML]{B4C6E7}\textbf{E1}}                          \\ \hline
\rowcolor[HTML]{B4C6E7}
\multicolumn{1}{|c|}{\cellcolor[HTML]{B4C6E7}\textbf{Score}} & \textbf{\# Vendors} \\ \hline
\rowcolor[HTML]{FFFFFF} 
\multicolumn{1}{|c|}{\cellcolor[HTML]{FFFFFF}0}              & 5                   \\ \hline
\rowcolor[HTML]{FFFFFF} 
\multicolumn{1}{|c|}{\cellcolor[HTML]{FFFFFF}1}              & 18                  \\ \hline
\rowcolor[HTML]{FFFFFF} 
\multicolumn{1}{|c|}{\cellcolor[HTML]{FFFFFF}U}              & 2                   \\ \hline
\end{tabular}
\end{table}

\begin{table}[h]
\caption{Distribution of scores for commercial comprehensive AutoML systems (E2). Evaluates the existence of a model VCS. Scores: 0 for absent, 1 for present, and U for unclear.}
\label{Table:framework_E2}
\begin{tabular}{|cc|}
\hline
\rowcolor[HTML]{B4C6E7}
\multicolumn{2}{|c|}{\cellcolor[HTML]{B4C6E7}\textbf{E2}}                          \\ \hline
\rowcolor[HTML]{B4C6E7}
\multicolumn{1}{|c|}{\cellcolor[HTML]{B4C6E7}\textbf{Score}} & \textbf{\# Vendors} \\ \hline
\rowcolor[HTML]{FFFFFF} 
\multicolumn{1}{|c|}{\cellcolor[HTML]{FFFFFF}0}              & 12                  \\ \hline
\rowcolor[HTML]{FFFFFF} 
\multicolumn{1}{|c|}{\cellcolor[HTML]{FFFFFF}1}              & 11                  \\ \hline
\rowcolor[HTML]{FFFFFF} 
\multicolumn{1}{|c|}{\cellcolor[HTML]{FFFFFF}U}              & 2                   \\ \hline
\end{tabular}
\end{table}

\begin{table}[h]
\caption{Distribution of scores for commercial comprehensive AutoML systems (E3). Evaluates the existence of experiment tracking. Scores: 0 for none, 1 for storage/access with limited automation/visuals, 2 for storage/access with automatic log visualisation, and U for unclear.}
\label{Table:framework_E3}
\begin{tabular}{|cc|}
\hline
\rowcolor[HTML]{B4C6E7}
\multicolumn{2}{|c|}{\cellcolor[HTML]{B4C6E7}\textbf{E3}}                          \\ \hline
\rowcolor[HTML]{B4C6E7}
\multicolumn{1}{|c|}{\cellcolor[HTML]{B4C6E7}\textbf{Score}} & \textbf{\# Vendors} \\ \hline
\rowcolor[HTML]{FFFFFF} 
\multicolumn{1}{|c|}{\cellcolor[HTML]{FFFFFF}0}              & 2                   \\ \hline
\rowcolor[HTML]{FFFFFF} 
\multicolumn{1}{|c|}{\cellcolor[HTML]{FFFFFF}1}              & 5                   \\ \hline
\rowcolor[HTML]{FFFFFF} 
\multicolumn{1}{|c|}{\cellcolor[HTML]{FFFFFF}2}              & 17                  \\ \hline
\rowcolor[HTML]{FFFFFF} 
\multicolumn{1}{|c|}{\cellcolor[HTML]{FFFFFF}U}              & 1                   \\ \hline
\end{tabular}
\end{table}

Differentiation between open-source and commercial AutoML products is immediately evident from the moment effort in tracking and management during experimentation is assessed.
The details are provided in Table~\ref{Table:framework_E1_E3}.
Indeed, vendors are geared for real-world applications, where the generation and refinement of ML models, before and after deployment, are expected to be iterative and exploratory.
Thus, as summed by Table~\ref{Table:framework_E1}, 18 out of 25 commercial development teams have identifiably elected to incorporate some form of persistence via an ML model repository (E1).
Such a mechanism typically serves as a basis for a model VCS (E2); sure enough, this direct upgrade is the case for ten packages.
Overall, Table~\ref{Table:framework_E2} highlights that 11 out of 25 surveyed systems provide the capability for stakeholders to roll back model iterations automatically.
Although far from ubiquitous, this integration of robust software-development practices shows growing market maturity.
At the very least, almost all AutoML products provide some form of experiment tracking (E3), and, as Table~\ref{Table:framework_E3} shows, 17 out of these 22 do so at sophisticated levels, enabling users to inspect the history of an ML application conveniently.

Naturally, the 25 surveyed commercial systems have already attained a spectrum of scores at this point.
Those ranked highest tend to have more of an MLOps focus, e.g.~Databricks and cnvrg.io.
A couple positioned lower in the rankings are trickier to assess, such as Einblink, previously Northstar, which has a unique modelling approach.
Nonetheless, on average, the commercial AutoML sector cares more about this space than open-source developers.
One possible interpretation is that academically sourced implementations are focussed on pushing the limits of full black-box automation, while vendors, maybe sensitive to business/legal ramifications, are much keener on allowing stakeholders to be as involved in an ML application, within reason, as they want.

\begin{table}[h]
\caption{Scores for commercial comprehensive AutoML systems (E4--E6). Evaluates the existence of a template/code repository (E4). Also evaluates capabilities for prior-work recommendation (E5) and project collaboration (E6). See Table~\ref{Table:Criteria_E4-6} for rubric.}
\label{Table:framework_E4_E6}
\begin{tabular}{|l|c|c|c|c|}
\hline
\rowcolor[HTML]{B4C6E7}
\textbf{Name}                                     & \textbf{E4} & \textbf{E5} & \textbf{E6} & \textbf{Sum (out of 5)} \\ \hline
Databricks                                        & 2           & 0           & 2           & 4            \\ \hline
Dataiku                                           & 2           & 0           & 2           & 4            \\ \hline
cnvrg.io                                             & 2           & 0           & 1           & 3            \\ \hline
DataRobot                                         & 2           & 0           & 1           & 3            \\ \hline
H2O                                            & 2           & 0           & 1           & 3            \\ \hline
Microsoft                                         & 1           & 0           & 2           & 3            \\ \hline
SageMaker                                     & 2           & 0           & 1           & 3            \\ \hline
SAS                                               & 2           & 0           & 1           & 3            \\ \hline
Deep Cognition                                     & 2           & 0           & 0           & 2            \\ \hline
IBM                                               & 1           & 0           & 1           & 2            \\ \hline
KNIME                                             & 2           & 0           & 0           & 2            \\ \hline
Number Theory                                   & 1           & 0           & 1           & 2            \\ \hline
Alteryx                                           & 0           & 0           & 1           & 1            \\ \hline
Einblick                                          & U           & 0           & 1           & 1            \\ \hline
Google                                            & 1           & 0           & 0           & 1            \\ \hline
RapidMiner                                        & 1           & 0           & 0           & 1            \\ \hline
Spell                                             & 0           & 0           & 1           & 1            \\ \hline
Auger                                          & U           & U           & U           & 0            \\ \hline
B2Metric                                          & U           & 0           & U           & 0            \\ \hline
Big Squid                                         & 0           & 0           & 0           & 0            \\ \hline
BigML                                             & 0           & 0           & 0           & 0            \\ \hline
Compellon                                         & 0           & 0           & 0           & 0            \\ \hline
D2iQ                                              & 0           & 0           & 0           & 0            \\ \hline
MyDataModels                                      & 0           & 0           & 0           & 0            \\ \hline
TIMi                                             & 0           & 0           & 0           & 0            \\ \hline
\multicolumn{1}{|c|}{\cellcolor[HTML]{B4C6E7}Sum} & 23/50          & 0/25           & 16/50          &              \\ \hline
\end{tabular}
\end{table}

\begin{table}[h]
\caption{Distribution of scores for commercial comprehensive AutoML systems (E4). Evaluates the existence of a template/code repository. Scores: 0 for none, 1 for manual generation, 2 for automatic leveraging, and U for unclear.}
\label{Table:framework_E4}
\begin{tabular}{|cc|}
\hline
\rowcolor[HTML]{B4C6E7}
\multicolumn{2}{|c|}{\cellcolor[HTML]{B4C6E7}\textbf{E4}}                          \\ \hline
\rowcolor[HTML]{B4C6E7}
\multicolumn{1}{|c|}{\cellcolor[HTML]{B4C6E7}\textbf{Score}} & \textbf{\# Vendors} \\ \hline
\rowcolor[HTML]{FFFFFF} 
\multicolumn{1}{|c|}{\cellcolor[HTML]{FFFFFF}0}              & 7                   \\ \hline
\rowcolor[HTML]{FFFFFF} 
\multicolumn{1}{|c|}{\cellcolor[HTML]{FFFFFF}1}              & 5                   \\ \hline
\rowcolor[HTML]{FFFFFF} 
\multicolumn{1}{|c|}{\cellcolor[HTML]{FFFFFF}2}              & 9                   \\ \hline
\rowcolor[HTML]{FFFFFF} 
\multicolumn{1}{|c|}{\cellcolor[HTML]{FFFFFF}U}              & 4                   \\ \hline
\end{tabular}
\end{table}

The next sub-criteria of assessment revolve around the acceleration of an ML application via the leveraging of prior work and collaboration.
In contrast to open-source systems, Table~\ref{Table:framework_E4_E6} indicates that commercial products have some decent coverage here.
Perhaps appreciating that lay users do not have a background in ML to lean on, more than half of the surveyed systems provide repositories for templates or coded scripts (E4), as aggregated by Table~\ref{Table:framework_E4}.
In many of these cases, the frameworks are set up to support users in generating and sharing these documents.
However, some packages, exemplified by SageMaker and its Jumpstart mechanism~\cite{aw21}, allow projects to be automatically kickstarted by templates for common use cases.
That stated, none of the 25 surveyed vendors seems to have a more sophisticated means of suggesting prior art (E5).
This result is interesting to note, given the long-running academic focus on meta-learning and the era of intelligent discovery assistants (IDAs) that preceded the current wave of AutoML~\cite{kemu20,albu15,lebu15}.
Admittedly, a deeper examination of other survey-excluded systems may be warranted ahead of any strident conclusions.
For instance, TAZI~\cite{Tazi}, Iguazio~\cite{Iguazio} and Kortical~\cite{Kortical} all claim to support off-the-shelf use cases, and Domino~\cite{Domino} does so similarly via its Domino Knowledge Center.
However, within the survey sample, only Auger hints at possible meta-learning; a blog post mentions that its optimiser is selected appropriately for a given dataset, perhaps leveraging a history of similar ML problems~\cite{au21a}.

\begin{table}[h]
\caption{Distribution of scores for commercial comprehensive AutoML systems (E6). Evaluates capabilities for project collaboration. Scores: 0 for none, 1 for basic features, 2 for advanced features, and U for unclear.}
\label{Table:framework_E6}
\begin{tabular}{|cc|}
\hline
\rowcolor[HTML]{B4C6E7}
\multicolumn{2}{|c|}{\cellcolor[HTML]{B4C6E7}\textbf{E6}}                          \\ \hline
\rowcolor[HTML]{B4C6E7}
\multicolumn{1}{|c|}{\cellcolor[HTML]{B4C6E7}\textbf{Score}} & \textbf{\# Vendors} \\ \hline
\rowcolor[HTML]{FFFFFF} 
\multicolumn{1}{|c|}{\cellcolor[HTML]{FFFFFF}0}              & 10                  \\ \hline
\rowcolor[HTML]{FFFFFF} 
\multicolumn{1}{|c|}{\cellcolor[HTML]{FFFFFF}1}              & 10                  \\ \hline
\rowcolor[HTML]{FFFFFF} 
\multicolumn{1}{|c|}{\cellcolor[HTML]{FFFFFF}2}              & 3                   \\ \hline
\rowcolor[HTML]{FFFFFF} 
\multicolumn{1}{|c|}{\cellcolor[HTML]{FFFFFF}U}              & 2                   \\ \hline
\end{tabular}
\end{table}

Mechanisms to support collaboration on a project (E6) are present in around half of the commercial AutoML systems.
Most have been assessed by Table~\ref{Table:framework_E6} to be fairly rudimentary, e.g.~simply allowing project folders to be shared.
For some instances, e.g.~SageMaker~\cite{aw21a} and IBM~\cite{IBMWatsonStudio}, a score of 1 was granted because, while notebooks themselves could be shared, each user had to work with a unique copy of the code, thus risking desynchronisation issues.
In contrast, Microsoft and Databricks~\cite{da21a} exemplify a score of 2 being granted for the presence of live co-editing.
That stated, there is a discussion to be had around the utility of such a feature, as it is unlikely to become a common software practice when such a robust industry-standard VCS such as Git exists.
In the meantime, Einblick, the commercialisation of the academic product Northstar~\cite{kr18}, supports a unique collaboration model via a whiteboard-style `visual analytics' interface upon which users drag and drop connectable nodes.
This form of non-standard HCI has obvious benefits but also several drawbacks.
For one thing, the unusual approach means it is more difficult to assess the system for certain functions, e.g.~around deployment and management.
The jury is also out on whether such a hands-on UI, advocated in promotional material, will actually live up to its aims of enhancing productivity for data-science teams at scale.
Granted, this uncertainty is to be expected for a nascent technology; the long-term success of radical innovations in AutoML will probably not be evident for many years.

\begin{table}[h]
\caption{Scores for commercial comprehensive AutoML systems (E7/E8). Evaluates the extent of automation for assisted data exploration (E7) and data preparation (E8). See Table~\ref{Table:Criteria_E7-13} for rubric.}
\label{Table:framework_E7_E8}
\begin{tabular}{|l|c|c|c|}
\hline
\rowcolor[HTML]{B4C6E7}
\textbf{Name}                                     & \textbf{E7} & \textbf{E8} & \textbf{Sum (out of 5)} \\ \hline
DataRobot                                         & 3           & 2           & 5            \\ \hline
Auger                                          & 2           & 2           & 4            \\ \hline
Big Squid                                         & 2           & 2           & 4            \\ \hline
Dataiku                                           & 2           & 2           & 4            \\ \hline
H2O                                            & 2           & 2           & 4            \\ \hline
Microsoft                                         & 2           & 2           & 4            \\ \hline
Compellon                                         & 3           & 0           & 3            \\ \hline
IBM                                               & 2           & 1           & 3            \\ \hline
Number Theory                                   & 2           & 1           & 3            \\ \hline
Alteryx                                           & 1           & 1           & 2            \\ \hline
cnvrg.io                                             & 1           & 1           & 2            \\ \hline
D2iQ                                              & 2           & 0           & 2            \\ \hline
Databricks                                        & 1           & 1           & 2            \\ \hline
Einblick                                          & 1           & 1           & 2            \\ \hline
Google                                            & 0           & 2           & 2            \\ \hline
KNIME                                             & 1           & 1           & 2            \\ \hline
MyDataModels                                      & 2           & U           & 2            \\ \hline
RapidMiner                                        & 1           & 1           & 2            \\ \hline
SageMaker                                     & 1           & 1           & 2            \\ \hline
SAS                                               & 1           & 1           & 2            \\ \hline
B2Metric                                          & 0           & 1           & 1            \\ \hline
BigML                                             & 0           & 1           & 1            \\ \hline
Deep Cognition                                     & 0           & 1           & 1            \\ \hline
TIMi                                             & 1           & 0           & 1            \\ \hline
Spell                                             & 0           & 0           & 0            \\ \hline
\multicolumn{1}{|c|}{\cellcolor[HTML]{B4C6E7}Sum} & 33/75          & 27/50          &              \\ \hline
\end{tabular}
\end{table}

\begin{table}[h]
\caption{Distribution of scores for commercial comprehensive AutoML systems (E7). Evaluates the extent of automation for assisted data exploration. Scores: 0 for none, 1 for convenience features, 2 for substantial automation, 3 for substantial automation with useful alerts, and U for unclear.}
  \label{Table:framework_E7}
\begin{tabular}{|cc|}
\hline
\rowcolor[HTML]{B4C6E7}
\multicolumn{2}{|c|}{\cellcolor[HTML]{B4C6E7}\textbf{E7}}                          \\ \hline
\rowcolor[HTML]{B4C6E7}
\multicolumn{1}{|c|}{\cellcolor[HTML]{B4C6E7}\textbf{Score}} & \textbf{\# Vendors} \\ \hline
\rowcolor[HTML]{FFFFFF} 
\multicolumn{1}{|c|}{\cellcolor[HTML]{FFFFFF}0}              & 5                   \\ \hline
\rowcolor[HTML]{FFFFFF} 
\multicolumn{1}{|c|}{\cellcolor[HTML]{FFFFFF}1}              & 9                   \\ \hline
\rowcolor[HTML]{FFFFFF} 
\multicolumn{1}{|c|}{\cellcolor[HTML]{FFFFFF}2}              & 9                   \\ \hline
\rowcolor[HTML]{FFFFFF} 
\multicolumn{1}{|c|}{\cellcolor[HTML]{FFFFFF}3}              & 2                   \\ \hline
\rowcolor[HTML]{FFFFFF} 
\multicolumn{1}{|c|}{\cellcolor[HTML]{FFFFFF}U}              & 0                   \\ \hline
\end{tabular}
\end{table}

Moving on to matters of effort reduction across an MLWF, we begin by assessing how a dataset ready for ML arises.
Immediately evident, there is a stark contrast between Table~\ref{Table:framework_E7_E8} and the previous analysis for open-source tools: EDA (E7).
Again, most vendors are geared for real-world applications run by relatively non-academic stakeholders, meaning that data sources are likely to be messy and, initially, poorly understood.
Thus, most surveyed commercial systems facilitate some form of exploration, allowing issues, insights, structures and patterns in data to be sought out.
However, Table~\ref{Table:framework_E7} indicates varying levels of support, with nine packages still requiring users to click about with convenience features.
Another nine go a step further into actual automation, generating visualisations to guide user interpretations, e.g.~histograms, charts, and graphs.
Only two, DataRobot and Compellon, advise a user how best to sift through these visuals, thus earning the highest score of 3.
Indeed, it is a reoccurring theme in this review that automation is not binary, and the mechanised generation of information alone is not ideal for effort reduction without being accompanied by some form of mechanised selection/emphasis.
Nonetheless, given that contextual insights are usually the domain of BI, with ML more focussed on the actual models, it is noteworthy that EDA has already been integrated so substantially into commercial AutoML.

\begin{table}[h]
\caption{Distribution of scores for commercial comprehensive AutoML systems (E8). Evaluates the extent of automation for data preparation. Scores: 0 for none, 1 for convenience features, 2 for substantial automation, and U for unclear.}
\label{Table:framework_E8}
\begin{tabular}{|cc|}
\hline
\rowcolor[HTML]{B4C6E7}
\multicolumn{2}{|c|}{\cellcolor[HTML]{B4C6E7}\textbf{E8}}                          \\ \hline
\rowcolor[HTML]{B4C6E7}
\multicolumn{1}{|c|}{\cellcolor[HTML]{B4C6E7}\textbf{Score}} & \textbf{\# Vendors} \\ \hline
\rowcolor[HTML]{FFFFFF} 
\multicolumn{1}{|c|}{\cellcolor[HTML]{FFFFFF}0}              & 4                   \\ \hline
\rowcolor[HTML]{FFFFFF} 
\multicolumn{1}{|c|}{\cellcolor[HTML]{FFFFFF}1}              & 13                  \\ \hline
\rowcolor[HTML]{FFFFFF} 
\multicolumn{1}{|c|}{\cellcolor[HTML]{FFFFFF}2}              & 7                   \\ \hline
\rowcolor[HTML]{FFFFFF} 
\multicolumn{1}{|c|}{\cellcolor[HTML]{FFFFFF}U}              & 1                   \\ \hline
\end{tabular}
\end{table}

In contrast, there is less excuse for a comprehensive AutoML system to neglect the automation of data preparation (E8), given the substantial burden it places on a data scientist.
Thus, the fact that Table~\ref{Table:framework_E8} indicates more than half of the surveyed services provide only convenience features, with four vendors appearing to ignore the topic outright, suggests an oversight in the market.
Stakeholders essentially need to be mindful of this hidden `upstream' cost that an otherwise appealing product may neglect.
That stated, while this review has systematised an MLWF that suggests an organic evolution of AutonoML technology, i.e.~steadily outwards from the model-development phase, it is worth remembering that vendors pursuing business benefits need not subscribe to an academic view of AutoML.
Indeed, Compellon and TIMi exemplify frameworks leaning towards exploratory analytics over the deployment of ML models, seemingly targeting business users with their messaging.
Ignoring data preparation may be intentional on their part, and Section~\ref{Sec:CommercialDiscussion} provides some extended commentary on this fragmentation of the `commercial AutoML agenda'.
Of course, it is also possible that the tools do undertake data-preparation work but simply do not promote it, possibly because it is not a priority.

\begin{table}[h]
\caption{Coverage of data-preparation processes for commercial comprehensive AutoML systems (E8A--E8F). Evaluates categorical processing (E8A), standardisation/normalisation (E8B), bucketing/binning (E8C), text processing (E8D), time-period extraction (E8E), and management of class imbalance (E8F). Scores: 0 for absent and 1 for present.}
\label{Table:framework_E8A_E8F}
\begin{tabular}{|l|c|c|c|c|c|c|c|}
\hline
\rowcolor[HTML]{B4C6E7}
\textbf{Name}                                     & \textbf{E8-A} & \textbf{E8-B} & \textbf{E8-C} & \textbf{E8-D} & \textbf{E8-E} & \textbf{E8-F} & \textbf{Sum (out of 6)} \\ \hline
Dataiku   & 1 & 1 & 1 & 1 & 1 & 1 & 6 \\ \hline
DataRobot & 1 & 1 & 1 & 1 & 1 & 1 & 6 \\ \hline
Google    & 1 & 1 & 1 & 1 & 1 & 0 & 5 \\ \hline
Auger  & 1 & 1 & 0 & 0 & 1 & 1 & 4 \\ \hline
H2O    & 1 & 0 & 1 & 1 & 1 & 0 & 4 \\ \hline
Microsoft & 1 & 1 & 0 & 1 & 1 & 0 & 4 \\ \hline
Big Squid & 1 & 1 & 0 & 0 & 0 & 0 & 2 \\ \hline
\multicolumn{1}{|c|}{\cellcolor[HTML]{B4C6E7}Sum} & 7/7             & 6/7             & 4/7             & 5/7             & 6/7             & 3/7             &              \\ \hline
\end{tabular}
\end{table}

This closed-source opacity becomes murkier as analysis becomes steadily more granular, and the Table~\ref{Table:framework_E8A_E8F} breakdown of what is included under data preparation (E8A--E8F) can only be applied to a handful of packages.
It is also no surprise that the table only lists the seven commercial systems that boast an automated form of this capability; the details effectively become part of their promotion.
In any case, it is notable that industry-heavyweight SageMaker Autopilot does not provide any automatic assistance with data preparation.
Another minor comment is that the data-preparation mechanism employed by Google does work with weighted arrays, which no other vendors consider.

\begin{table}[h]
\caption{Scores for commercial comprehensive AutoML systems (E9--E11). Evaluates the extent of automation for dataset feature generation (E9), reuse (E10), and selection (E11). See Table~\ref{Table:Criteria_E7-13} for rubric.}
\label{Table:framework_E9_E13}
\begin{tabular}{|l|c|c|c|c|}
\hline
\rowcolor[HTML]{B4C6E7}
\textbf{Name}               & \textbf{E9} & \textbf{E10} & \textbf{E11} & \textbf{Sum (out of 5)} \\ \hline
DataRobot                   & 2           & 1            & 2            & 5            \\ \hline
Dataiku                     & 1.5           & 1            & 2            & 4.5            \\ \hline
B2Metric                    & 2           & 0            & 2            & 4            \\ \hline
H2O                      & 2           & 0            & 2            & 4            \\ \hline
RapidMiner                  & 2           & 0            & 2            & 4            \\ \hline
IBM                         & 1           & 0            & 2            & 3            \\ \hline
Microsoft                   & 2           & 0            & 1            & 3            \\ \hline
Alteryx                     & 1.5         & 0            & 1            & 2.5          \\ \hline
Compellon                   & 0           & 0            & 2            & 2            \\ \hline
Databricks                  & 1           & 1            & 0            & 2            \\ \hline
Einblick                    & 1           & 0            & 1            & 2            \\ \hline
KNIME                       & 1           & 0            & 1            & 2            \\ \hline
Number Theory             & 1           & 0            & 1            & 2            \\ \hline
SageMaker               & 1           & 1            & 0            & 2            \\ \hline
Auger                    & 0           & 0            & 1            & 1            \\ \hline
Big Squid                   & 0           & 0            & 1            & 1            \\ \hline
BigML                       & 0           & 1            & 0            & 1            \\ \hline
Deep Cognition               & 0           & 0            & 1            & 1            \\ \hline
Google                      & 0           & 0            & 1            & 1            \\ \hline
MyDataModels                & 0           & 0            & 1            & 1            \\ \hline
SAS                         & 0           & 0            & 1            & 1            \\ \hline
cnvrg.io                       & 0           & 0            & U            & 0            \\ \hline
D2iQ                        & 0           & 0            & 0            & 0            \\ \hline
Spell                       & 0           & 0            & 0            & 0            \\ \hline
TIMi                       & 0           & 0            & 0            & 0            \\ \hline
\cellcolor[HTML]{B4C6E7}Sum & 19/50          & 4/25            & 25/50           &              \\ \hline
\end{tabular}
\end{table}

\begin{table}[h]
\caption{Distribution of scores for commercial comprehensive AutoML systems (E9). Evaluates the extent of automation for dataset feature generation. Scores: 0 for none, 1 for convenience features, 2 for substantial automation, and U for unclear. The 1.5 denotes existence of convenience plugins enabling automation.}
\label{Table:framework_E9}
\begin{tabular}{|cc|}
\hline
\rowcolor[HTML]{B4C6E7}
\multicolumn{2}{|c|}{\cellcolor[HTML]{B4C6E7}\textbf{E9}}                          \\ \hline
\rowcolor[HTML]{B4C6E7}
\multicolumn{1}{|c|}{\cellcolor[HTML]{B4C6E7}\textbf{Score}} & \textbf{\# Vendors} \\ \hline
\rowcolor[HTML]{FFFFFF} 
\multicolumn{1}{|c|}{\cellcolor[HTML]{FFFFFF}0}              & 12                  \\ \hline
\rowcolor[HTML]{FFFFFF} 
\multicolumn{1}{|c|}{\cellcolor[HTML]{FFFFFF}1}              & 6                   \\ \hline
\rowcolor[HTML]{FFFFFF} 
\multicolumn{1}{|c|}{\cellcolor[HTML]{FFFFFF}1.5}            & 2                   \\ \hline
\rowcolor[HTML]{FFFFFF} 
\multicolumn{1}{|c|}{\cellcolor[HTML]{FFFFFF}2}              & 5                   \\ \hline
\rowcolor[HTML]{FFFFFF} 
\multicolumn{1}{|c|}{\cellcolor[HTML]{FFFFFF}U}              & 0                   \\ \hline
\end{tabular}
\end{table}

Turning to AutoFE, Table~\ref{Table:framework_E9_E13} shows that the surveyed vendors have a moderate coverage of this space.
In fact, commercial systems are granted a higher average score for dataset feature generation (E9) than open-source systems in Section~\ref{Sec:OpenEfficiency}, i.e.~19/25 versus 8/19.
However, Table~\ref{Table:framework_E9} indicates that half of the surveyed systems still do not bother with this aspect, and half again of those that do choose to stick with convenience features, e.g.~manually triggered functions that sum/average dataset columns.
Notably, two AutoML services, Alteryx and Dataiku, are unique in that they support the automated generation of dataset features via an optional plugin.
For this reason, they are marked with scores of 1.5, reflecting the blur between a convenience feature and an automated mechanism.
As for the operator `nodes' of KNIME, they are considered feature transformations that require a degree of user manipulation, so the associated system scores a 1.

\begin{table}[h]
\caption{Distribution of scores for commercial comprehensive AutoML systems (E10). Evaluates the extent of automation for dataset feature reuse. Scores: 0 for absent, 1 for present, and U for unclear.}
\label{Table:framework_E10}
\begin{tabular}{|cc|}
\hline
\rowcolor[HTML]{B4C6E7}
\multicolumn{2}{|c|}{\cellcolor[HTML]{B4C6E7}\textbf{E10}}                         \\ \hline
\rowcolor[HTML]{B4C6E7}
\multicolumn{1}{|c|}{\cellcolor[HTML]{B4C6E7}\textbf{Score}} & \textbf{\# Vendors} \\ \hline
\rowcolor[HTML]{FFFFFF} 
\multicolumn{1}{|c|}{\cellcolor[HTML]{FFFFFF}0}              & 20                  \\ \hline
\rowcolor[HTML]{FFFFFF} 
\multicolumn{1}{|c|}{\cellcolor[HTML]{FFFFFF}1}              & 5                   \\ \hline
\rowcolor[HTML]{FFFFFF} 
\multicolumn{1}{|c|}{\cellcolor[HTML]{FFFFFF}U}              & 0                   \\ \hline
\end{tabular}
\end{table}

Now, generating useful features can be a complex affair, so it would make sense to store them for future reuse (E10).
This reasoning is especially the case where an organisation may use the same datasets for numerous ML projects run by different teams.
Such a notion is similar to constructing/managing data marts in a data warehouse with corporately controlled metric registers, thus ensuring consistency and accelerating preparatory processes.
Admittedly, one could argue that this is a luxury compared to other priorities, likely only to become more valuable as the AutoML market further matures.
Nonetheless, Table~\ref{Table:framework_E10} shows that four reviewed vendors have already considered the innovation.

\begin{table}[h]
\caption{Distribution of scores for commercial comprehensive AutoML systems (E11). Evaluates the extent of automation for dataset feature selection. Scores: 0 for none, 1 for convenience features, 2 for substantial automation, and U for unclear.}
\label{Table:framework_E11}
\begin{tabular}{|cc|}
\hline
\rowcolor[HTML]{B4C6E7}
\multicolumn{2}{|c|}{\cellcolor[HTML]{B4C6E7}\textbf{E11}}                         \\ \hline
\rowcolor[HTML]{B4C6E7}
\multicolumn{1}{|c|}{\cellcolor[HTML]{B4C6E7}\textbf{Score}} & \textbf{\# Vendors} \\ \hline
\rowcolor[HTML]{FFFFFF} 
\multicolumn{1}{|c|}{\cellcolor[HTML]{FFFFFF}0}              & 6                   \\ \hline
\rowcolor[HTML]{FFFFFF} 
\multicolumn{1}{|c|}{\cellcolor[HTML]{FFFFFF}1}              & 11                  \\ \hline
\rowcolor[HTML]{FFFFFF} 
\multicolumn{1}{|c|}{\cellcolor[HTML]{FFFFFF}2}              & 7                   \\ \hline
\rowcolor[HTML]{FFFFFF} 
\multicolumn{1}{|c|}{\cellcolor[HTML]{FFFFFF}U}              & 1                   \\ \hline
\end{tabular}
\end{table}

Of course, whether they are generated or recalled, dataset features must be chosen carefully from the pool of available options when used in modelling.
Overall, commercial systems assist strongly with dataset feature selection (E11), much more than generation, which mimics the trend seen for open-source tools.
However, this varies substantially between packages, and Table~\ref{Table:framework_E11} reveals that a majority of 11 offer convenience features, e.g.~the ability to manually drop features via a checkbox/widget, sometimes assisted by an informed view of what to select.
Another seven packages then go further and provide substantial automation.
Interestingly, BigML, Auger, Microsoft and RapidMiner exemplify services that approach this process from a perspective of exclusion rather than inclusion, advising which dataset features are `bad'.
For instance, Microsoft avoids high cardinality, while RapidMiner targets those of `low quality' and redundancies revealed by high correlations.
Crucially, most of these selection processes, whether automated or not, are applied prior to the existence of any ML model, i.e.~they are filter-style, not wrapper-style.

As a side note, this review could not make determinations for two sub-criteria regarding commercial AutoML, namely whether HPO search spaces and algorithms are already specified by default (E12) and whether HPO operates on an ML pipeline of components (E13).
It is very likely that, for the former sub-criterion, all packages should be marked `yes' with a 1, as model selection processes are generally obscured from users in a comprehensive system; they should not lock up in the absence of manual specification.
Such a result would subsequently mean that the sub-criterion does not differentiate between any of the open-source and closed-source software presented in this review.
However, there are open-source tools excluded from this survey that do force users to make choices prior to HPO, so the question on specification remains valid to ask.
As for the latter sub-criterion, it is difficult to guess the structure of ML solutions under the hood of a commercial system.
However, optimising extended pipelines is challenging from a theoretical/implementational standpoint.
It is already rare among the more technically driven open-source offerings, even being elevated as a core innovation within a package named AutoWEKA4MCPS \cite{sabu16,sabu19,ngmu21, AutoWeka4MCPS}.
Thus, most probably, few commercial systems, if any, will presently lump preprocessors and postprocessors into an HPO space.

\begin{table}[h]
\caption{Scores for commercial comprehensive AutoML systems (E14--E16). Evaluates the availability of workload optimisation (E14), modelling time limits (E15), and modelling iteration limits (E16). See Table~\ref{Table:Criteria_E14-16} for rubric.}
\label{Table:framework_E14_E16}
\begin{tabular}{|l|c|c|c|c|}
\hline
\rowcolor[HTML]{B4C6E7}
\textbf{Name}                                     & \textbf{E14} & \textbf{E15} & \textbf{E16} & \textbf{Sum (out of 4)} \\ \hline
DataRobot                                         & 2            & 1            & 0            & 3            \\ \hline
Auger                                          & 0            & 1            & 1            & 2            \\ \hline
BigML                                             & 0            & 1            & 1            & 2            \\ \hline
Databricks                                        & 0            & 1            & 1            & 2            \\ \hline
Dataiku                                           & 0            & 1            & 1            & 2            \\ \hline
Alteryx                                           & 0            & 0            & 1            & 1            \\ \hline
cnvrg.io                                             & 1            & 0            & 0            & 1            \\ \hline
D2iQ                                              & 1            & 0            & 0            & 1            \\ \hline
H2O                                            & 0            & 1            & 0            & 1            \\ \hline
IBM                                               & 0            & 0            & 1            & 1            \\ \hline
Microsoft                                         & 0            & 1            & 0            & 1            \\ \hline
SageMaker                                     & 1            & 0            & 0            & 1            \\ \hline
Spell                                             & 1            & 0            & 0            & 1            \\ \hline
B2Metric                                          & 0            & 0            & 0            & 0            \\ \hline
Big Squid                                         & 0            & 0            & 0            & 0            \\ \hline
Compellon                                         & 0            & U            & U            & 0            \\ \hline
Deep Cognition                                     & 0            & 0            & 0            & 0            \\ \hline
Einblick                                          & 0            & 0            & 0            & 0            \\ \hline
Google                                            & 0            & 0            & 0            & 0            \\ \hline
KNIME                                             & 0            & 0            & 0            & 0            \\ \hline
MyDataModels                                      & 0            & U            & U            & 0            \\ \hline
Number Theory                                   & 0            & 0            & 0            & 0            \\ \hline
RapidMiner                                        & 0            & 0            & 0            & 0            \\ \hline
SAS                                               & 0            & 0            & 0            & 0            \\ \hline
TIMi                                             & 0            & 0            & 0            & 0            \\ \hline
\multicolumn{1}{|c|}{\cellcolor[HTML]{B4C6E7}Sum} & 6/50            & 7/25            & 6/25            &              \\ \hline
\end{tabular}
\end{table}

\begin{table}[h]
\caption{Distribution of scores for commercial comprehensive AutoML systems (E14). Evaluates the availability of workload optimisation. Scores: 0 for none, 1 for convenience features, 2 for substantial automation, and U for unclear.}
\label{Table:framework_E14}
\begin{tabular}{|cc|}
\hline
\rowcolor[HTML]{B4C6E7}
\multicolumn{2}{|c|}{\cellcolor[HTML]{B4C6E7}\textbf{E14}}                         \\ \hline
\rowcolor[HTML]{B4C6E7}
\multicolumn{1}{|c|}{\cellcolor[HTML]{B4C6E7}\textbf{Score}} & \textbf{\# Vendors} \\ \hline
\rowcolor[HTML]{FFFFFF} 
\multicolumn{1}{|c|}{\cellcolor[HTML]{FFFFFF}0}              & 20                  \\ \hline
\rowcolor[HTML]{FFFFFF} 
\multicolumn{1}{|c|}{\cellcolor[HTML]{FFFFFF}1}              & 4                   \\ \hline
\rowcolor[HTML]{FFFFFF} 
\multicolumn{1}{|c|}{\cellcolor[HTML]{FFFFFF}2}              & 1                   \\ \hline
\rowcolor[HTML]{FFFFFF} 
\multicolumn{1}{|c|}{\cellcolor[HTML]{FFFFFF}U}              & 0                   \\ \hline
\end{tabular}
\end{table}

\begin{table}[h]
\caption{Distribution of scores for commercial comprehensive AutoML systems (E15). Evaluates the availability of modelling time limits. Scores: 0 for absent, 1 for present, and U for unclear.}
\label{Table:framework_E15}
\begin{tabular}{|cc|}
\hline
\rowcolor[HTML]{B4C6E7}
\multicolumn{2}{|c|}{\cellcolor[HTML]{B4C6E7}\textbf{E15}}                         \\ \hline
\rowcolor[HTML]{B4C6E7}
\multicolumn{1}{|c|}{\cellcolor[HTML]{B4C6E7}\textbf{Score}} & \textbf{\# Vendors} \\ \hline
\rowcolor[HTML]{FFFFFF} 
\multicolumn{1}{|c|}{\cellcolor[HTML]{FFFFFF}0}              & 16                  \\ \hline
\rowcolor[HTML]{FFFFFF} 
\multicolumn{1}{|c|}{\cellcolor[HTML]{FFFFFF}1}              & 7                   \\ \hline
\rowcolor[HTML]{FFFFFF} 
\multicolumn{1}{|c|}{\cellcolor[HTML]{FFFFFF}U}              & 2                   \\ \hline
\end{tabular}
\end{table}

\begin{table}[h]
\caption{Distribution of scores for commercial comprehensive AutoML systems (E16). Evaluates the availability of modelling iteration limits. Scores: 0 for absent, 1 for present, and U for unclear.}
\label{Table:framework_E16}
\begin{tabular}{|cc|}
\hline
\rowcolor[HTML]{B4C6E7}
\multicolumn{2}{|c|}{\cellcolor[HTML]{B4C6E7}\textbf{E16}}                         \\ \hline
\rowcolor[HTML]{B4C6E7}
\multicolumn{1}{|c|}{\cellcolor[HTML]{B4C6E7}\textbf{Score}} & \textbf{\# Vendors} \\ \hline
\rowcolor[HTML]{FFFFFF} 
\multicolumn{1}{|c|}{\cellcolor[HTML]{FFFFFF}0}              & 17                  \\ \hline
\rowcolor[HTML]{FFFFFF} 
\multicolumn{1}{|c|}{\cellcolor[HTML]{FFFFFF}1}              & 6                   \\ \hline
\rowcolor[HTML]{FFFFFF} 
\multicolumn{1}{|c|}{\cellcolor[HTML]{FFFFFF}U}              & 2                   \\ \hline
\end{tabular}
\end{table}

Finally, on matters of technical efficiency, Table~\ref{Table:framework_E14_E16} shows that the commercial space of workload management is not too dissimilar from that of open-source AutoML systems.
However, as aggregated by Table~\ref{Table:framework_E14}, there are uniquely a few packages that assist with workload optimisation (E14).
Most of these provide only convenience features and notably have a promotional focus on MLOps, e.g.~Spell, cnvrg.io and D2iQ.
Granted, via the AWS ecosystem, SageMaker also has the option of plugging into Auto Scaling to manage relevant EC2 server instances.
A similar capability may be possible for Google and Microsoft as well, but it has not been explicitly promoted in their cases.

Naturally, time limits (E15) and trial limits (E16) are available among AutoML systems, highlighted by Table~\ref{Table:framework_E15} and Table~\ref{Table:framework_E16}, respectively.
However, support for such modelling control is surprisingly rare, and providers of one are often those offering the other.
It is also notable that, barring DataRobot, any packages supporting workload management facilitate either infrastructural optimisation or algorithmic constraints, but never both.
This hints that there is a divide in the approach to resource management between vendors focussed on MLOps and those leaning into user-driven model development.
As for DataRobot, it is the only commercial system that appears to substantially automate workload optimisation, iteratively increasing data size and computational resources as it reduces the number of candidate models during a search.
This process, of course, is very similar to successive halving, so other products may do so, too, without explicitly promoting it.
Whatever the case, the takeaway is that resource management has not yet received significant attention from either open-source or closed-source AutoML tools.
Perhaps fine-tuning the technical efficiency of an ML application is not yet a priority compared with ensuring that training and deploying ML models are robust functions.
Indeed, given reasonable cloud-computing prices and supplied operational efficiencies elsewhere, the sub-criteria above may not become vital differentiators until the market matures further.

\subsubsection{Dirty Data}
\label{Sec:CommercialDirtyData}

\begin{table}[h]
\caption{Scores for commercial comprehensive AutoML systems (DD1). Evaluates the extent of automation for cleaning dirty data. See Table~\ref{Table:Criteria_DD1-5} for rubric.}
\label{Table:framework_DD1_all}
\begin{tabular}{|l|c|}
\hline
\rowcolor[HTML]{B4C6E7}
\textbf{Name}                                     & \textbf{DD1} \\ \hline
Auger                                          & 2            \\ \hline
Big Squid                                         & 2            \\ \hline
Dataiku                                           & 2            \\ \hline
DataRobot                                         & 2            \\ \hline
Google                                            & 2            \\ \hline
H2O                                            & 2            \\ \hline
Microsoft                                         & 2            \\ \hline
MyDataModels                                      & 2            \\ \hline
Alteryx                                           & 1            \\ \hline
B2Metric                                          & 1            \\ \hline
BigML                                             & 1            \\ \hline
cnvrg.io                                             & 1            \\ \hline
Databricks                                        & 1            \\ \hline
Deep Cognition                                     & 1            \\ \hline
Einblick                                          & 1            \\ \hline
IBM                                               & 1            \\ \hline
KNIME                                             & 1            \\ \hline
Number Theory                                   & 1            \\ \hline
RapidMiner                                        & 1            \\ \hline
SageMaker                                     & 1            \\ \hline
SAS                                               & 1            \\ \hline
Compellon                                         & 0            \\ \hline
D2iQ                                              & 0            \\ \hline
Spell                                             & 0            \\ \hline
TIMi                                             & 0            \\ \hline
\multicolumn{1}{|c|}{\cellcolor[HTML]{B4C6E7}Sum} & 29/50           \\ \hline
\end{tabular}
\end{table}

\begin{table}[h]
\caption{Distribution of scores for commercial comprehensive AutoML systems (DD1). Evaluates the extent of automation for cleaning dirty data. Scores: 0 for none, 1 for convenience features, 2 for substantial automation, and U for unclear.}
\label{Table:framework_DD1}
\begin{tabular}{|cc|}
\hline
\rowcolor[HTML]{B4C6E7}
\multicolumn{2}{|c|}{\cellcolor[HTML]{B4C6E7}\textbf{DD1}}                         \\ \hline
\rowcolor[HTML]{B4C6E7}
\multicolumn{1}{|c|}{\cellcolor[HTML]{B4C6E7}\textbf{Score}} & \textbf{\# Vendors} \\ \hline
\rowcolor[HTML]{FFFFFF} 
\multicolumn{1}{|c|}{\cellcolor[HTML]{FFFFFF}0}              & 4                   \\ \hline
\rowcolor[HTML]{FFFFFF} 
\multicolumn{1}{|c|}{\cellcolor[HTML]{FFFFFF}1}              & 13                  \\ \hline
\rowcolor[HTML]{FFFFFF} 
\multicolumn{1}{|c|}{\cellcolor[HTML]{FFFFFF}2}              & 8                   \\ \hline
\rowcolor[HTML]{FFFFFF} 
\multicolumn{1}{|c|}{\cellcolor[HTML]{FFFFFF}U}              & 0                   \\ \hline
\end{tabular}
\end{table}

As already assessed for open-source tools, this criterion examines whether commercial AutoML systems can aid stakeholders with the initial cleaning of data (DD1) prior even to its preparation for ingestion by an ML model.
Accordingly, the evaluations are displayed in Table~\ref{Table:framework_DD1_all}.
Further elaborated in Table~\ref{Table:framework_DD1}, we find that most vendors only supply convenience features for this task; users still need to make decisions and take explicit action.
However, eight packages do provide substantial automation, and seven of these were the same that also scored highest for the automation of data preparation in Table~\ref{Table:framework_E7_E8}.
This outcome is not surprising.
If development effort is to be expended on the data-engineering phase of an MLWF, it makes sense to support cleaning and preparation simultaneously.

\begin{table}[h]
\caption{Scores for commercial comprehensive AutoML systems (DD2--DD5). Evaluates the extent of automation for data-type inference (DD2), missing-value imputation (DD3), and outlier management (DD4). Also evaluates the existence of domain-specific/advanced cleaning operations (DD5). See Table~\ref{Table:Criteria_DD1-5} for rubric.}
\label{Table:framework_DD2_DD5}
\begin{tabular}{|l|c|c|c|c|c|}
\hline
\rowcolor[HTML]{B4C6E7}
\textbf{Name}                                     & \textbf{DD2} & \textbf{DD3} & \textbf{DD4} & \textbf{DD5} & \textbf{Sum (out of 6)} \\ \hline
DataRobot                                         & 1            & 2            & 2            & 1            & 6            \\ \hline
H2O                                            & 1            & 2            & 1            & 0            & 4            \\ \hline
Auger                                          & 1            & 2            & 0            & 0            & 3            \\ \hline
Big Squid                                         & 1            & 2            & 0            & 0            & 3            \\ \hline
Dataiku                                           & 1            & 2            & 0            & 0            & 3            \\ \hline
Google                                            & 1            & 2            & 0            & 0            & 3            \\ \hline
Microsoft                                         & 1            & 2            & 0            & 0            & 3            \\ \hline
MyDataModels                                      & U            & 1            & 1            & 0            & 2            \\ \hline
\multicolumn{1}{|c|}{\cellcolor[HTML]{B4C6E7}Sum} & 7/8            & 15/16           & 4/16            & 1/8            &              \\ \hline
\end{tabular}
\end{table}

\begin{table}[h]
\caption{Distribution of scores for commercial comprehensive AutoML systems (DD2). Evaluates the extent of automation for data-type inference. Scores: 0 for absent, 1 for present, and U for unclear.}
\label{Table:framework_DD2}
\begin{tabular}{|cc|}
\hline
\rowcolor[HTML]{B4C6E7}
\multicolumn{2}{|c|}{\cellcolor[HTML]{B4C6E7}\textbf{DD2}}                         \\ \hline
\rowcolor[HTML]{B4C6E7}
\multicolumn{1}{|c|}{\cellcolor[HTML]{B4C6E7}\textbf{Score}} & \textbf{\# Vendors} \\ \hline
\rowcolor[HTML]{FFFFFF} 
\multicolumn{1}{|c|}{\cellcolor[HTML]{FFFFFF}0}              & 0                   \\ \hline
\rowcolor[HTML]{FFFFFF} 
\multicolumn{1}{|c|}{\cellcolor[HTML]{FFFFFF}1}              & 7                   \\ \hline
\rowcolor[HTML]{FFFFFF} 
\multicolumn{1}{|c|}{\cellcolor[HTML]{FFFFFF}U}              & 1                   \\ \hline
\end{tabular}
\end{table}

\begin{table}[h]
\caption{Distribution of scores for commercial comprehensive AutoML systems (DD3). Evaluates the extent of automation for missing-value imputation. Scores: 0 for none, 1 for automatic detection, 2 for automatic detection/resolution, and U for unclear.}
\label{Table:framework_DD3}
\begin{tabular}{|cc|}
\hline
\rowcolor[HTML]{B4C6E7}
\multicolumn{2}{|c|}{\cellcolor[HTML]{B4C6E7}\textbf{DD3}}                         \\ \hline
\rowcolor[HTML]{B4C6E7}
\multicolumn{1}{|c|}{\cellcolor[HTML]{B4C6E7}\textbf{Score}} & \textbf{\# Vendors} \\ \hline
\rowcolor[HTML]{FFFFFF} 
\multicolumn{1}{|c|}{\cellcolor[HTML]{FFFFFF}0}              & 0                   \\ \hline
\rowcolor[HTML]{FFFFFF} 
\multicolumn{1}{|c|}{\cellcolor[HTML]{FFFFFF}1}              & 1                   \\ \hline
\rowcolor[HTML]{FFFFFF} 
\multicolumn{1}{|c|}{\cellcolor[HTML]{FFFFFF}2}              & 7                   \\ \hline
\rowcolor[HTML]{FFFFFF} 
\multicolumn{1}{|c|}{\cellcolor[HTML]{FFFFFF}U}              & 0                   \\ \hline
\end{tabular}
\end{table}

Of course, opacity remains an issue here as it did for data preparation, so Table~\ref{Table:framework_DD2_DD5} only breaks down capabilities for commercial systems that claim significant automation.
As summarised by Table~\ref{Table:framework_DD2}, almost all packages can identifiably infer data types (DD2).
Likewise, Table~\ref{Table:framework_DD3} indicates that almost all packages can detect and resolve missing values (DD3), even though the preferred processes differ.
For instance, Microsoft, DataRobot, Big Squid and Auger all appear solely to impute.
Dataiku and H2O additionally provide an option to drop missing values.
Embedding approaches are also viable, being supported by both Dataiku and Google.
The latter vendor even seems capable of assessing the severity of a null value, which may suggest leaving it alone.
Essentially, handling missing values appears to be a standard offering for automated data cleaning.

\begin{table}[h]
\caption{Distribution of scores for commercial comprehensive AutoML systems (DD4). Evaluates the extent of automation for outlier management. Scores: 0 for none, 1 for automatic detection, 2 for automatic detection/resolution, and U for unclear.}
\label{Table:framework_DD4}
\begin{tabular}{|cc|}
\hline
\rowcolor[HTML]{B4C6E7}
\multicolumn{2}{|c|}{\cellcolor[HTML]{B4C6E7}\textbf{DD4}}                         \\ \hline
\rowcolor[HTML]{B4C6E7}
\multicolumn{1}{|c|}{\cellcolor[HTML]{B4C6E7}\textbf{Score}} & \textbf{\# Vendors} \\ \hline
\rowcolor[HTML]{FFFFFF} 
\multicolumn{1}{|c|}{\cellcolor[HTML]{FFFFFF}0}              & 5                   \\ \hline
\rowcolor[HTML]{FFFFFF} 
\multicolumn{1}{|c|}{\cellcolor[HTML]{FFFFFF}1}              & 2                   \\ \hline
\rowcolor[HTML]{FFFFFF} 
\multicolumn{1}{|c|}{\cellcolor[HTML]{FFFFFF}2}              & 1                   \\ \hline
\rowcolor[HTML]{FFFFFF} 
\multicolumn{1}{|c|}{\cellcolor[HTML]{FFFFFF}U}              & 0                   \\ \hline
\end{tabular}
\end{table}

\begin{table}[h]
\caption{Distribution of scores for commercial comprehensive AutoML systems (DD5). Evaluates the existence of domain-specific/advanced cleaning operations. Scores: 0 for absent, 1 for present, and U for unclear.}
\label{Table:framework_DD5}
\begin{tabular}{|cc|}
\hline
\rowcolor[HTML]{B4C6E7}
\multicolumn{2}{|c|}{\cellcolor[HTML]{B4C6E7}\textbf{DD5}}                         \\ \hline
\rowcolor[HTML]{B4C6E7}
\multicolumn{1}{|c|}{\cellcolor[HTML]{B4C6E7}\textbf{Score}} & \textbf{\# Vendors} \\ \hline
\rowcolor[HTML]{FFFFFF} 
\multicolumn{1}{|c|}{\cellcolor[HTML]{FFFFFF}0}              & 7                   \\ \hline
\rowcolor[HTML]{FFFFFF} 
\multicolumn{1}{|c|}{\cellcolor[HTML]{FFFFFF}1}              & 1                   \\ \hline
\rowcolor[HTML]{FFFFFF} 
\multicolumn{1}{|c|}{\cellcolor[HTML]{FFFFFF}U}              & 0                   \\ \hline
\end{tabular}
\end{table}

In contrast, outlier management (DD4) is a rarity.
The reasons are the same as discussed in Section~\ref{Sec:OpenDirtyData}, and, sure enough, Table~\ref{Table:framework_DD4} finds only three services handling automatic detection of anomalies.
Notably, the one that claims to automatically resolve outliers, i.e.~DataRobot, is also the only comprehensive system within this review that supplies advanced cleaning operations (DD5).
Scoring 1 in Table~\ref{Table:framework_DD5}, the software is seemingly able to identify `inliers', i.e.~erroneous data points within the interior of a statistical distribution.
Naturally, such an analysis is more complex than simply highlighting distant outliers.

\subsubsection{Completeness and Currency}
\label{Sec:CommercialCompleteness}

\begin{table}[h]
\small{
\caption{Scores for commercial comprehensive AutoML systems (CC1--CC9). Evaluates capabilities for unsupervised learning (CC1), regression on tabular data (CC2), standard classification on tabular data (CC3), multi-class classification on tabular data (CC4), time-series forecasting (CC5), image-based problem solving (CC6), text-based problem solving (CC7), multi-modal problem solving (CC8), and ensemble techniques (CC9). See Table~\ref{Table:Criteria_CC1-9} for rubric.}
\label{Table:framework_CC1_CC9}
\begin{tabular}{|l|c|c|c|c|c|c|c|c|c|C{1.5cm}|}
\hline
\rowcolor[HTML]{B4C6E7}
\textbf{Name} &
  \textbf{CC1} &
  \textbf{CC2} &
  \textbf{CC3} &
  \textbf{CC4} &
  \textbf{CC5} &
  \textbf{CC6} &
  \textbf{CC7} &
  \textbf{CC8} &
  \textbf{CC9} &
  \textbf{Sum\newline(out of 17)} \\ \hline
DataRobot       & 2 & 2 & 2 & 2 & 2 & 2 & 2 & 1 & 2 & 17 \\ \hline
H2O          & 2 & 2 & 2 & 2 & 2 & 2 & 2 & 0 & 2 & 16 \\ \hline
Dataiku         & 2 & 2 & 2 & 2 & 2 & 2 & 1 & 0 & 2 & 15 \\ \hline
Microsoft       & 1 & 2 & 2 & 2 & 2 & 1 & 1 & 0 & 2 & 13 \\ \hline
BigML           & 2 & 2 & 2 & 2 & 2 & 2 & 0 & 0 & 0 & 12 \\ \hline
Google          & 0 & 2 & 2 & 2 & 2 & 2 & 2 & 0 & 0 & 12 \\ \hline
Number Theory & 1 & 2 & 2 & 2 & 1 & 1 & 2 & 0 & 1 & 12 \\ \hline
Einblick        & 1 & 2 & 2 & 2 & 2 & 0 & 2 & 0 & 0 & 11 \\ \hline
RapidMiner      & 2 & 1 & 2 & 2 & 1 & 1 & 1 & 0 & 1 & 11 \\ \hline
SageMaker   & 1 & 2 & 2 & 2 & 1 & 1 & 1 & 0 & 0 & 10 \\ \hline
Alteryx         & 0 & 2 & 2 & 2 & 1 & 1 & 0 & 0 & 0 & 8  \\ \hline
Auger        & 0 & 2 & 2 & 2 & 0 & 0 & 0 & 0 & 2 & 8  \\ \hline
B2Metric        & 2 & 2 & 2 & 2 & 0 & 0 & 0 & 0 & 0 & 8  \\ \hline
Big Squid       & 0 & 2 & 2 & 2 & 2 & 0 & 0 & 0 & 0 & 8  \\ \hline
Databricks      & 0 & 2 & 2 & 2 & 2 & 0 & 0 & 0 & 0 & 8  \\ \hline
IBM             & 0 & 2 & 2 & 2 & 1 & 0 & 1 & 0 & 0 & 8  \\ \hline
KNIME           & 1 & 1 & 1 & 1 & 1 & 1 & 1 & 0 & 1 & 8  \\ \hline
SAS             & 1 & 1 & 2 & 2 & 1 & 0 & 1 & 0 & 0 & 8  \\ \hline
cnvrg.io           & 1 & 1 & 1 & 1 & 1 & 1 & 1 & 0 & U & 7  \\ \hline
Deep Cognition   & 0 & 0 & 2 & 2 & 0 & 2 & 0 & 0 & U & 6  \\ \hline
MyDataModels    & 0 & 2 & 2 & 2 & 0 & 0 & 0 & 0 & U & 6  \\ \hline
Spell           & 0 & 1 & 1 & 1 & 1 & 1 & 1 & 0 & 0 & 6  \\ \hline
TIMi           & 0 & 2 & 2 & 0 & 0 & 0 & 0 & 0 & U & 4  \\ \hline
Compellon       & U & U & 2 & U & U & U & U & U & U & 2  \\ \hline
D2iQ            & U & U & U & U & U & U & U & U & 0 & 0  \\ \hline
\multicolumn{1}{|c|}{\cellcolor[HTML]{B4C6E7}Sum} &
  19/50 &
  39/50 &
  45/50 &
  41/50 &
  27/50 &
  20/50 &
  19/50 &
  1/25 &
  13/50 &
   \\ \hline
\end{tabular}
}
\end{table}

When assessing the types of ML problems that commercial comprehensive AutoML systems are designed for, Table~\ref{Table:framework_CC1_CC9} reveals a spectrum of coverage.
The survey results are not entirely dissimilar to the evaluations in Section~\ref{Sec:OpenCompleteness}, although, for the summed scores, commercial software does have a less generalist/automated tail that goes below the open-source range of 7 to 16.
Admittedly, opacity makes it harder to judge a couple of these outlying vendor products.

\begin{table}[h]
\caption{Distribution of scores for commercial comprehensive AutoML systems (CC1). Evaluates capabilities for unsupervised learning. Scores: 0 for none, 1 for convenience features, 2 for substantial automation, and U for unclear.}
\label{Table:framework_CC1}
\begin{tabular}{|cc|}
\hline
\rowcolor[HTML]{B4C6E7}
\multicolumn{2}{|c|}{\cellcolor[HTML]{B4C6E7}\textbf{CC1}}                         \\ \hline
\rowcolor[HTML]{B4C6E7}
\multicolumn{1}{|c|}{\cellcolor[HTML]{B4C6E7}\textbf{Score}} & \textbf{\# Vendors} \\ \hline
\rowcolor[HTML]{FFFFFF} 
\multicolumn{1}{|c|}{\cellcolor[HTML]{FFFFFF}0}              & 10                  \\ \hline
\rowcolor[HTML]{FFFFFF} 
\multicolumn{1}{|c|}{\cellcolor[HTML]{FFFFFF}1}              & 7                   \\ \hline
\rowcolor[HTML]{FFFFFF} 
\multicolumn{1}{|c|}{\cellcolor[HTML]{FFFFFF}2}              & 6                   \\ \hline
\rowcolor[HTML]{FFFFFF} 
\multicolumn{1}{|c|}{\cellcolor[HTML]{FFFFFF}U}              & 2                   \\ \hline
\end{tabular}
\end{table}
\begin{table}[h]
\caption{Distribution of scores for commercial comprehensive AutoML systems (CC2). Evaluates capabilities for regression on tabular data. Scores: 0 for none, 1 for convenience features, 2 for substantial automation, and U for unclear.}
\label{Table:framework_CC2}
\begin{tabular}{|cc|}
\hline
\rowcolor[HTML]{B4C6E7}
\multicolumn{2}{|c|}{\cellcolor[HTML]{B4C6E7}\textbf{CC2}}                         \\ \hline
\rowcolor[HTML]{B4C6E7}
\multicolumn{1}{|c|}{\cellcolor[HTML]{B4C6E7}\textbf{Score}} & \textbf{\# Vendors} \\ \hline
\rowcolor[HTML]{FFFFFF} 
\multicolumn{1}{|c|}{\cellcolor[HTML]{FFFFFF}0}              & 1                   \\ \hline
\rowcolor[HTML]{FFFFFF} 
\multicolumn{1}{|c|}{\cellcolor[HTML]{FFFFFF}1}              & 5                   \\ \hline
\rowcolor[HTML]{FFFFFF} 
\multicolumn{1}{|c|}{\cellcolor[HTML]{FFFFFF}2}              & 17                  \\ \hline
\rowcolor[HTML]{FFFFFF} 
\multicolumn{1}{|c|}{\cellcolor[HTML]{FFFFFF}U}              & 2                   \\ \hline
\end{tabular}
\end{table}
\begin{table}[h]
\caption{Distribution of scores for commercial comprehensive AutoML systems (CC3). Evaluates capabilities for standard classification on tabular data. Scores: 0 for none, 1 for convenience features, 2 for substantial automation, and U for unclear.}
\label{Table:framework_CC3}
\begin{tabular}{|cc|}
\hline
\rowcolor[HTML]{B4C6E7}
\multicolumn{2}{|c|}{\cellcolor[HTML]{B4C6E7}\textbf{CC3}}                         \\ \hline
\rowcolor[HTML]{B4C6E7}
\multicolumn{1}{|c|}{\cellcolor[HTML]{B4C6E7}\textbf{Score}} & \textbf{\# Vendors} \\ \hline
\rowcolor[HTML]{FFFFFF} 
\multicolumn{1}{|c|}{\cellcolor[HTML]{FFFFFF}0}              & 0                   \\ \hline
\rowcolor[HTML]{FFFFFF} 
\multicolumn{1}{|c|}{\cellcolor[HTML]{FFFFFF}1}              & 3                   \\ \hline
\rowcolor[HTML]{FFFFFF} 
\multicolumn{1}{|c|}{\cellcolor[HTML]{FFFFFF}2}              & 21                  \\ \hline
\rowcolor[HTML]{FFFFFF} 
\multicolumn{1}{|c|}{\cellcolor[HTML]{FFFFFF}U}              & 1                   \\ \hline
\end{tabular}
\end{table}
\begin{table}[h]
\caption{Distribution of scores for commercial comprehensive AutoML systems (CC4). Evaluates capabilities for multi-class classification on tabular data. Scores: 0 for none, 1 for convenience features, 2 for substantial automation, and U for unclear.}
\label{Table:framework_CC4}
\begin{tabular}{|cc|}
\hline
\rowcolor[HTML]{B4C6E7}
\multicolumn{2}{|c|}{\cellcolor[HTML]{B4C6E7}\textbf{CC4}}                         \\ \hline
\rowcolor[HTML]{B4C6E7}
\multicolumn{1}{|c|}{\cellcolor[HTML]{B4C6E7}\textbf{Score}} & \textbf{\# Vendors} \\ \hline
\rowcolor[HTML]{FFFFFF} 
\multicolumn{1}{|c|}{\cellcolor[HTML]{FFFFFF}0}              & 1                   \\ \hline
\rowcolor[HTML]{FFFFFF} 
\multicolumn{1}{|c|}{\cellcolor[HTML]{FFFFFF}1}              & 3                   \\ \hline
\rowcolor[HTML]{FFFFFF} 
\multicolumn{1}{|c|}{\cellcolor[HTML]{FFFFFF}2}              & 19                  \\ \hline
\rowcolor[HTML]{FFFFFF} 
\multicolumn{1}{|c|}{\cellcolor[HTML]{FFFFFF}U}              & 2                   \\ \hline
\end{tabular}
\end{table}

In any case, Table~\ref{Table:framework_CC1} suggests that it is almost a coin flip whether a package will assist/automate unsupervised learning (CC1) or ignore it entirely.
Understandably, a lack of a clearly defined target can make it harder to provide case studies or business pitches that objectively highlight superior ML model validity.
On the other hand, unsupervised learning meshes well with EDA for the systems that lean more in that direction.
So, its provision ultimately depends on the agenda of a vendor.
Supervised learning, however, is almost compulsory to offer.
Noted by Table~\ref{Table:framework_CC2}, which considers regression on tabular data (CC2), the vast majority of packages provide significant automation in this space.
Given that supervised learning is the fundamental task that innovations in ML model selection originally targeted, comprehensive systems struggle to be labelled as AutoML if they do not even automate this core process.
Scores for classification capabilities on tabular data, both standard (CC3) and multi-class (CC4), affirm this argument.
They are summarised in Table~\ref{Table:framework_CC3} and Table~\ref{Table:framework_CC4}, respectively.
Arguably, classification is slightly easier than regression, so it has even better coverage in terms of automation. 

\begin{table}[h]
\caption{Distribution of scores for commercial comprehensive AutoML systems (CC5). Evaluates capabilities for time-series forecasting. Scores: 0 for none, 1 for convenience features, 2 for substantial automation, and U for unclear.}
\label{Table:framework_CC5}
\begin{tabular}{|cc|}
\hline
\rowcolor[HTML]{B4C6E7}
\multicolumn{2}{|c|}{\cellcolor[HTML]{B4C6E7}\textbf{CC5}}                         \\ \hline
\rowcolor[HTML]{B4C6E7}
\multicolumn{1}{|c|}{\cellcolor[HTML]{B4C6E7}\textbf{Score}} & \textbf{\# Vendors} \\ \hline
\rowcolor[HTML]{FFFFFF} 
\multicolumn{1}{|c|}{\cellcolor[HTML]{FFFFFF}0}              & 5                   \\ \hline
\rowcolor[HTML]{FFFFFF} 
\multicolumn{1}{|c|}{\cellcolor[HTML]{FFFFFF}1}              & 9                   \\ \hline
\rowcolor[HTML]{FFFFFF} 
\multicolumn{1}{|c|}{\cellcolor[HTML]{FFFFFF}2}              & 9                   \\ \hline
\rowcolor[HTML]{FFFFFF} 
\multicolumn{1}{|c|}{\cellcolor[HTML]{FFFFFF}U}              & 2                   \\ \hline
\end{tabular}
\end{table}
\begin{table}[h]
\caption{Distribution of scores for commercial comprehensive AutoML systems (CC6). Evaluates capabilities for image-based problem solving. Scores: 0 for none, 1 for convenience features, 2 for substantial automation, and U for unclear.}
\label{Table:framework_CC6}
\begin{tabular}{|cc|}
\hline
\rowcolor[HTML]{B4C6E7}
\multicolumn{2}{|c|}{\cellcolor[HTML]{B4C6E7}\textbf{CC6}}                         \\ \hline
\rowcolor[HTML]{B4C6E7}
\multicolumn{1}{|c|}{\cellcolor[HTML]{B4C6E7}\textbf{Score}} & \textbf{\# Vendors} \\ \hline
\rowcolor[HTML]{FFFFFF} 
\multicolumn{1}{|c|}{\cellcolor[HTML]{FFFFFF}0}              & 9                   \\ \hline
\rowcolor[HTML]{FFFFFF} 
\multicolumn{1}{|c|}{\cellcolor[HTML]{FFFFFF}1}              & 8                   \\ \hline
\rowcolor[HTML]{FFFFFF} 
\multicolumn{1}{|c|}{\cellcolor[HTML]{FFFFFF}2}              & 6                   \\ \hline
\rowcolor[HTML]{FFFFFF} 
\multicolumn{1}{|c|}{\cellcolor[HTML]{FFFFFF}U}              & 2                   \\ \hline
\end{tabular}
\end{table}
\begin{table}[h]
\caption{Distribution of scores for commercial comprehensive AutoML systems (CC7). Evaluates capabilities for text-based problem solving. Scores: 0 for none, 1 for convenience features, 2 for substantial automation, and U for unclear.}
\label{Table:framework_CC7}
\begin{tabular}{|cc|}
\hline
\rowcolor[HTML]{B4C6E7}
\multicolumn{2}{|c|}{\cellcolor[HTML]{B4C6E7}\textbf{CC7}}                         \\ \hline
\rowcolor[HTML]{B4C6E7}
\multicolumn{1}{|c|}{\cellcolor[HTML]{B4C6E7}\textbf{Score}} & \textbf{\# Vendors} \\ \hline
\rowcolor[HTML]{FFFFFF} 
\multicolumn{1}{|c|}{\cellcolor[HTML]{FFFFFF}0}              & 9                   \\ \hline
\rowcolor[HTML]{FFFFFF} 
\multicolumn{1}{|c|}{\cellcolor[HTML]{FFFFFF}1}              & 9                   \\ \hline
\rowcolor[HTML]{FFFFFF} 
\multicolumn{1}{|c|}{\cellcolor[HTML]{FFFFFF}2}              & 5                   \\ \hline
\rowcolor[HTML]{FFFFFF} 
\multicolumn{1}{|c|}{\cellcolor[HTML]{FFFFFF}U}              & 2                   \\ \hline
\end{tabular}
\end{table}
\begin{table}[h]
\caption{Distribution of scores for commercial comprehensive AutoML systems (CC8). Evaluates capabilities for multi-modal problem solving. Scores: 0 for absent, 1 for present, and U for unclear.}
\label{Table:framework_CC8}
\begin{tabular}{|cc|}
\hline
\rowcolor[HTML]{B4C6E7}
\multicolumn{2}{|c|}{\cellcolor[HTML]{B4C6E7}\textbf{CC8}}                         \\ \hline
\rowcolor[HTML]{B4C6E7}
\multicolumn{1}{|c|}{\cellcolor[HTML]{B4C6E7}\textbf{Score}} & \textbf{\# Vendors} \\ \hline
\rowcolor[HTML]{FFFFFF} 
\multicolumn{1}{|c|}{\cellcolor[HTML]{FFFFFF}0}              & 22                  \\ \hline
\rowcolor[HTML]{FFFFFF} 
\multicolumn{1}{|c|}{\cellcolor[HTML]{FFFFFF}1}              & 1                   \\ \hline
\rowcolor[HTML]{FFFFFF} 
\multicolumn{1}{|c|}{\cellcolor[HTML]{FFFFFF}U}              & 2                   \\ \hline
\end{tabular}
\end{table}
\begin{table}[h]
\caption{Distribution of scores for commercial comprehensive AutoML systems (CC9). Evaluates capabilities for ensemble techniques. Scores: 0 for none, 1 for convenience features, 2 for substantial automation, and U for unclear.}
\label{Table:framework_CC9}
\begin{tabular}{|cc|}
\hline
\rowcolor[HTML]{B4C6E7}
\multicolumn{2}{|c|}{\cellcolor[HTML]{B4C6E7}\textbf{CC9}}                         \\ \hline
\rowcolor[HTML]{B4C6E7}
\multicolumn{1}{|c|}{\cellcolor[HTML]{B4C6E7}\textbf{Score}} & \textbf{\# Vendors} \\ \hline
\rowcolor[HTML]{FFFFFF} 
\multicolumn{1}{|c|}{\cellcolor[HTML]{FFFFFF}0}              & 12                  \\ \hline
\rowcolor[HTML]{FFFFFF} 
\multicolumn{1}{|c|}{\cellcolor[HTML]{FFFFFF}1}              & 3                   \\ \hline
\rowcolor[HTML]{FFFFFF} 
\multicolumn{1}{|c|}{\cellcolor[HTML]{FFFFFF}2}              & 5                   \\ \hline
\rowcolor[HTML]{FFFFFF} 
\multicolumn{1}{|c|}{\cellcolor[HTML]{FFFFFF}U}              & 5                   \\ \hline
\end{tabular}
\end{table}

Again, as with open-source tools, Table~\ref{Table:framework_CC5} reveals a drop in coverage for time-series forecasting (CC5).
Similarly, by the time discussion moves to image-based tasks (CC6) and NLP (CC7), Table~\ref{Table:framework_CC6} and Table~\ref{Table:framework_CC7} suggest, respectively, that these capabilities are now relatively niche.
This result is not surprising, as the more complex data inputs are typically reserved for AutoDL approaches, which, beyond being out of scope for this review, are currently nowhere near the technological maturity of standard AutoML.
Consequently, if handling a single source of non-tabular data is a rare capability, the result shown in Table~\ref{Table:framework_CC8} is also not unexpected; only one commercial system claims to handle multi-modal tasks (CC8), i.e.~DataRobot.
Granted, AutoML frameworks that optimise ML pipelines could theoretically merge multiple preprocessing components dedicated to different data modalities, but it is unclear whether any existing implementation can genuinely support such a process.
Finally, Table~\ref{Table:framework_CC9} shows that only eight vendors identifiably confront the nuances of tackling ML problems with ensemble techniques (CC9), although the quantity of U scores is uniquely high, as methodological details tend to be obfuscated for closed-source software.
It is possible that, in the early days of ML adoption by the business community, basic models are sufficient to provide substantial benefits already.
The power of ensembles may not yet be warranted, especially given their typical complexity and associated adverse impact on explainability.

\begin{table}[h]
\caption{Coverage of model-selection processes for commercial comprehensive AutoML systems (CC10/CC11). Evaluates capabilities for custom evaluation metrics (CC10). Also evaluates the existence of HPO techniques (CC11), i.e.~grid search (GR), random search (RA), Bayesian optimisation (BA), multi-armed bandit strategies (MAB), genetic/evolutionary algorithms (GE), and meta-learning (M). For convenience, additionally classifies whether HPO mechanisms beyond grid/random search (GR+) are provided. Scores: 0 for absent, 1 for present, and U for unclear. The 0.5 denotes the HPO technique can be manually included as an operator `node'.}
\label{Table:framework_CC10_CC13}
\begin{tabular}{|l|c|c|c|c|c|c|c|c|c|}
\hline
\rowcolor[HTML]{B4C6E7}
\textbf{Name} &
  \textbf{CC10} &
  \textbf{GR} &
  \textbf{RA} &
  \textbf{BA} &
  \textbf{MAB} &
  \textbf{GE} &
   \textbf{M} &
  \textbf{SUM (out of 7)} &
  \textbf{GR+} \\ \hline
D2iQ            & 1 & 1   & 1   & 1   & 1   & 0& 0   & 5   & 1 \\ \hline
Dataiku         & 1 & 1   & 1   & 1   & 0   & 0& 0   & 4   & 1 \\ \hline
H2O          & 1 & 1   & 1   & 0   & 0   & 1& 0   & 4   & 1 \\ \hline
SAS             & U & 1   & 1   & 1   & 0   & 1& 0   & 4   & 1 \\ \hline
Spell           & 1 & 1   & 1   & 1   & 0   & 0& 0   & 4   & 1 \\ \hline
B2Metric        & 0 & 1   & 1   & 1   & 0   & 0& 0   & 3   & 1 \\ \hline
Databricks      & 1 & 0   & 0   & 1   & 0   & 1& 0   & 3   & 1 \\ \hline
KNIME           & 1 & 0.5 & 0.5 & 0.5 & 0.5 & 0& 0   & 3   & 1 \\ \hline
SageMaker   & 1 & 0   & 1   & 1   & 0   & 0   & 0 & 3  & 1 \\ \hline
RapidMiner      & U & 0.5 & 0.5 & 0.5 & 0.5 & 0.5& 0 & 2.5 & 1 \\ \hline
BigML           & 1 & 0   & 0   & 1   & 0   & 0& 0   & 2   & 1 \\ \hline
cnvrg.io           & 1 & 1   & 0   & 0   & 0   & 0& 0   & 2   & 0 \\ \hline
Microsoft       & 0 & 1   & 1   & 0   & 0   & 0 & 0  & 2   & 0 \\ \hline
Auger        & 0 & 0   & 0   & 1   & 0   & 0& 0   & 1   & 1 \\ \hline
Einblick        & 0 & 0   & 0   & 1   & 0   & 0& 0   & 1   & 1 \\ \hline
Google          & 1 & U   & U   & U   & U   & U& 0   & 1   & U \\ \hline
Number Theory & 1 & U   & U   & U   & U   & U& 0   & 1   & U \\ \hline
Alteryx         & 0 & U   & U   & U   & U   & U& 0   & 0   & U \\ \hline
Big Squid       & 0 & U   & U   & U   & U   & U& 0   & 0   & U \\ \hline
Compellon       & U & U   & U   & U   & U   & U& 0   & 0   & U \\ \hline
DataRobot       & 0 & U   & U   & U   & U   & U& 0   & 0   & U \\ \hline
Deep Cognition   & 0 & U   & U   & U   & U   & U& 0   & 0   & U \\ \hline
IBM             & 0 & 0   & 0   & 0   & 0   & 0& 0   & 0*   & 0 \\ \hline
MyDataModels    & 0 & U   & U   & U   & U   & U& 0   & 0   & U \\ \hline
TIMi           & 0 & U   & U   & U   & U   & U& 0   & 0   & U \\ \hline
\cellcolor[HTML]{B4C6E7}\textbf{Sum} &
  11/25 &
  9/25 &
  9/25 &
  11/25 &
  2/25 &
  3.5/25 &
  0/25 &
   &
  13/25 \\ \hline
\end{tabular}
\end{table}
\begin{table}[h]
\caption{Distribution of scores for commercial comprehensive AutoML systems (CC10). Evaluates capabilities for custom evaluation metrics. Scores: 0 for absent, 1 for present, and U for unclear.}
\label{Table:framework_CC10}
\begin{tabular}{|cc|}
\hline
\rowcolor[HTML]{B4C6E7}
\multicolumn{2}{|c|}{\cellcolor[HTML]{B4C6E7}\textbf{CC10}}                        \\ \hline
\rowcolor[HTML]{B4C6E7}
\multicolumn{1}{|c|}{\cellcolor[HTML]{B4C6E7}\textbf{Score}} & \textbf{\# Vendors} \\ \hline
\rowcolor[HTML]{FFFFFF} 
\multicolumn{1}{|c|}{\cellcolor[HTML]{FFFFFF}0}              & 11                  \\ \hline
\rowcolor[HTML]{FFFFFF} 
\multicolumn{1}{|c|}{\cellcolor[HTML]{FFFFFF}1}              & 11                  \\ \hline
\rowcolor[HTML]{FFFFFF} 
\multicolumn{1}{|c|}{\cellcolor[HTML]{FFFFFF}U}              & 3                   \\ \hline
\end{tabular}
\end{table}

Now, as previously emphasised, an increasingly granular review of commercial AutoML confronts decreasing transparency around system details.
Thus, while it is reasonably clear which ML problems can be addressed by the ML solutions provided by a package, it is far less obvious how these solutions are derived.
Accordingly, over a third of the systems surveyed in Table~\ref{Table:framework_CC10_CC13} have information gaps regarding their mechanical processes around CASH.
However, what is clear from Table~\ref{Table:framework_CC10}, is that vendors are evenly split around whether they support customisable metrics (CC10).
Recalling Section~\ref{Sec:CommercialEfficiency}, an interpretation was suggested that many commercial AutoML features appear to favour the primacy of user control when compared to open-source tools, possibly to minimise potential liability, but the result here implies that there is still a broad range of how much freedom is genuinely supplied.
For instance, in this particular space, DataRobot and Microsoft exemplify vendors making the decisions, while Google and SageMaker support user-driven customisation.
Of course, a cynical perspective might be that it benefits commercial systems to sell the \textit{impression} of user control while still making the choices where they actually matter.
However, in fairness, there is a diversity of agendas and intended audiences within the AutoML marketplace, as discussed in Section~\ref{Sec:CommercialDiscussion}; these can determine how much technical control can be safely granted to a stakeholder.

Turning to the HPO methods employed (CC11), a couple of commercial systems involve users pipelining ML processes by connecting operator `nodes', e.g.~RapidMiner and KNIME.
It is difficult to determine whether the manual installation of associated HPO nodes aligns with the spirit of full AutoML, so they have been represented here with values of 0.5.
Regardless, wherever details can be extracted, most AutoML products supply the basic algorithms of grid search (GR) and random search (RA).
In fact, beyond these methods (GR+), heavyweight Microsoft goes no further.
However, the remaining packages that can be commented about almost ubiquitously support Bayesian optimisation (BA).

That stated, the sparsity of multi-armed bandit techniques (MAB) -- 3/16 identifiable commercial providers versus 6/19 open-source offerings -- hints that the technical depth of the commercial market may be somewhat limited.
This outcome may be due to a lag in how quickly theoretical advances disseminate into the commercial sphere.
Alternatively, if this is a genuine indicator of the marketplace, vendors may simply find no need to implement more sophisticated methods if strong operational performance is available elsewhere, e.g.~via experiment tracking, a decent UI, convenience features, and so on.
The H2O package, for instance, is one of the earlier entrants into the AutoML sphere that eschewed popular Bayesian techniques for more straightforward grid/random search methods; evidently, the implementation was sufficiently performant to maintain its existence.
Notably, it has since become one of the few commercial adopters of genetic/evolutionary algorithms (GE).
Finally, other than Auger perhaps hinting at meta-learning, there is no evidence that any vendor explores this functionality, so its mention is excluded from Table~\ref{Table:framework_CC10_CC13}.
Additionally, as a side note, while IBM scores zero on the standard forms of HPO, it appears unique in employing derivative-free optimisation via the RBFOpt package.

Moving on to the last sub-criteria, this review finds that closed-source opacity prevents any clear overview of which popular ML libraries are interfaced with (CC12).
Proprietary algorithms may have been developed in-house, which further limits their exposure.
We do note though that Google uses TF, D2iQ employs TF and Torch, and Dataiku interfaces with Sklearn, XGBoost, and H2O.
As for the sub-criterion on currency (CC13), we re-emphasise that all the surveyed tools in this analysis were found to be active at the time of survey.

\subsubsection{Explainability}
\label{Sec:CommercialExplainability}

As has been reiterated many times, evaluating the detailed mechanics of a commercial AutoML system is challenging.
For one thing, unlike open-source software, there are financial costs to fully exploring the routine processes of a tool; a product trial is unlikely to be sufficiently informative.
Moreover, even with full access, the detailed code underlying these services remains obscured.
Thus, this review, primarily a collation of published material about current AutoML software, cannot assess every sub-criterion.
For instance, we could not adequately gauge whether vendors provide facilities to clarify data lineage (EX1) and modelling steps (EX2).
Naturally, the value of such an assessment for judging performant ML does remain, and deeper investigations are recommended in the future.

\begin{table}[h]
\caption{Scores for commercial comprehensive AutoML systems (EX3--EX7). Evaluates capabilities for enhancing global interpretability (EX3), enhancing local interpretability (EX4), scenario analysis (EX5), bias/fairness assessment (EX6), and bias/fairness management (EX7). See Table~\ref{Table:Criteria_EX1-5} for rubric.}
\label{Table:framework_EX3_EX7}
\begin{tabular}{|l|c|c|c|c|c|c|}
\hline
\rowcolor[HTML]{B4C6E7}
\textbf{Name}               & \textbf{EX3} & \textbf{EX4} & \textbf{EX5} & \textbf{EX6} & \textbf{EX7} & \textbf{Sum (out of 9)} \\ \hline
Dataiku         & 2 & 2 & 1 & 2 & 0 & 7 \\ \hline
IBM             & 2 & 0 & 1 & 2 & 2 & 7 \\ \hline
SageMaker   & 2 & 2 & 0 & 2 & 1 & 7 \\ \hline
DataRobot       & 2 & 2 & 0 & 2 & 0 & 6 \\ \hline
Google          & 2 & 2 & 1 & 1 & 0 & 6 \\ \hline
Big Squid       & 2 & 2 & 1 & 0 & 0 & 5 \\ \hline
H2O          & 2 & 2 & 0 & 1 & 0 & 5 \\ \hline
Microsoft       & 2 & 2 & 0 & 1 & 0 & 5 \\ \hline
RapidMiner      & 2 & 0 & 1 & 1 & 0 & 4 \\ \hline
SAS             & 2 & 2 & 0 & 0 & 0 & 4 \\ \hline
MyDataModels    & 2 & 0 & 1 & 0 & 0 & 3 \\ \hline
BigML           & 2 & 0 & 0 & 0 & 0 & 2 \\ \hline
Compellon       & 1 & 0 & 1 & 0 & 0 & 2 \\ \hline
Databricks      & 1 & 1 & 0 & 0 & 0 & 2 \\ \hline
Deep Cognition   & 2 & 0 & 0 & 0 & 0 & 2 \\ \hline
Einblick        & 1 & 1 & 0 & 0 & 0 & 2 \\ \hline
Number Theory & 2 & 0 & 0 & 0 & 0 & 2 \\ \hline
TIMi           & 2 & 0 & 0 & 0 & 0 & 2 \\ \hline
cnvrg.io           & 1 & 0 & 0 & 0 & 0 & 1 \\ \hline
D2iQ            & 1 & 0 & 0 & 0 & 0 & 1 \\ \hline
Spell           & 1 & 0 & 0 & 0 & 0 & 1 \\ \hline
Alteryx         & 0 & 0 & 0 & 0 & 0 & 0 \\ \hline
Auger        & U & U & U & 0 & 0 & 0 \\ \hline
B2Metric        & U & U & U & 0 & 0 & 0 \\ \hline
KNIME           & 0 & 0 & 0 & 0 & 0 & 0 \\ \hline
\cellcolor[HTML]{B4C6E7}Sum & 
36/50           & 
18/50           & 
7/25            & 
12/50           & 
3/50            &              \\ \hline
\end{tabular}
\end{table}

\begin{table}[h]
\caption{Distribution of scores for commercial comprehensive AutoML systems (EX3). Evaluates capabilities for enhancing global interpretability. Scores: 0 for none, 1 for convenience features, 2 for substantial automation, and U for unclear.}
\label{Table:framework_EX3}
\begin{tabular}{|cc|}
\hline
\rowcolor[HTML]{B4C6E7}
\multicolumn{2}{|c|}{\cellcolor[HTML]{B4C6E7}\textbf{EX3}}                         \\ \hline
\rowcolor[HTML]{B4C6E7}
\multicolumn{1}{|c|}{\cellcolor[HTML]{B4C6E7}\textbf{Score}} & \textbf{\# Vendors} \\ \hline
\rowcolor[HTML]{FFFFFF} 
\multicolumn{1}{|c|}{\cellcolor[HTML]{FFFFFF}0}              & 2                   \\ \hline
\rowcolor[HTML]{FFFFFF} 
\multicolumn{1}{|c|}{\cellcolor[HTML]{FFFFFF}1}              & 6                   \\ \hline
\rowcolor[HTML]{FFFFFF} 
\multicolumn{1}{|c|}{\cellcolor[HTML]{FFFFFF}2}              & 15                  \\ \hline
\rowcolor[HTML]{FFFFFF} 
\multicolumn{1}{|c|}{\cellcolor[HTML]{FFFFFF}U}              & 2                   \\ \hline
\end{tabular}
\end{table}
\begin{table}[h]
\caption{Distribution of scores for commercial comprehensive AutoML systems (EX4). Evaluates capabilities for enhancing local interpretability. Scores: 0 for none, 1 for convenience features, 2 for substantial automation, and U for unclear.}
\label{Table:framework_EX4}
\begin{tabular}{|cc|}
\hline
\rowcolor[HTML]{B4C6E7}
\multicolumn{2}{|c|}{\cellcolor[HTML]{B4C6E7}\textbf{EX4}}                         \\ \hline
\rowcolor[HTML]{B4C6E7}
\multicolumn{1}{|c|}{\cellcolor[HTML]{B4C6E7}\textbf{Score}} & \textbf{\# Vendors} \\ \hline
\rowcolor[HTML]{FFFFFF} 
\multicolumn{1}{|c|}{\cellcolor[HTML]{FFFFFF}0}              & 13                  \\ \hline
\rowcolor[HTML]{FFFFFF} 
\multicolumn{1}{|c|}{\cellcolor[HTML]{FFFFFF}1}              & 2                   \\ \hline
\rowcolor[HTML]{FFFFFF} 
\multicolumn{1}{|c|}{\cellcolor[HTML]{FFFFFF}2}              & 8                   \\ \hline
\rowcolor[HTML]{FFFFFF} 
\multicolumn{1}{|c|}{\cellcolor[HTML]{FFFFFF}U}              & 2                   \\ \hline
\end{tabular}
\end{table}

Nonetheless, once considerations of explainability move from system mechanics to system inputs/outputs, evaluations are easier to make, and these are detailed in Table~\ref{Table:framework_EX3_EX7}.
Indeed, if existent, the interpretability features discussed here are often well integrated into a UI and strongly promoted.
Accordingly, Table~\ref{Table:framework_EX3} reveals that it is almost standard among vendors to explain global aspects of a model (EX3) automatically.
Most of the time, this is identifying the importance of each input dataset feature to the potentially complex input-output mapping that a user receives.
In contrast, Table~\ref{Table:framework_EX4} shows that local interpretability at the level of individual predictions (EX4) is mostly overlooked, though its enhancement is often well automated when addressed.
Given that individual predictions/prescriptions are often associated with unique entities, e.g.~choosing whether a loan is approved, businesses typically cannot afford to make complacent decisions even at this granular level.
Thus, a lack of improvement in this area could stifle industrial uptake.

\begin{table}[h]
\caption{Distribution of scores for commercial comprehensive AutoML systems (EX5). Evaluates capabilities for scenario analysis. Scores: 0 for absent, 1 for present, and U for unclear.}
\label{Table:framework_EX5}
\begin{tabular}{|cc|}
\hline
\rowcolor[HTML]{B4C6E7}
\multicolumn{2}{|c|}{\cellcolor[HTML]{B4C6E7}\textbf{EX5}}                         \\ \hline
\rowcolor[HTML]{B4C6E7}
\multicolumn{1}{|c|}{\cellcolor[HTML]{B4C6E7}\textbf{Score}} & \textbf{\# Vendors} \\ \hline
\rowcolor[HTML]{FFFFFF} 
\multicolumn{1}{|c|}{\cellcolor[HTML]{FFFFFF}0}              & 16                  \\ \hline
\rowcolor[HTML]{FFFFFF} 
\multicolumn{1}{|c|}{\cellcolor[HTML]{FFFFFF}1}              & 7                   \\ \hline
\rowcolor[HTML]{FFFFFF} 
\multicolumn{1}{|c|}{\cellcolor[HTML]{FFFFFF}U}              & 2                   \\ \hline
\end{tabular}
\end{table}

The distribution of scores only continues to worsen for each subsequent sub-criterion.
Table~\ref{Table:framework_EX5} indicates that less than a third of surveyed vendors identifiably support scenario capabilities (EX5), which would allow a user to simulate the impact of changing variables on model outputs at a global/local level.
Granted, similar to sensitivity analysis in other fields, this functionality leans more towards leveraging ML for exploration and insights rather than model production.
Its provision may well depend on how a vendor markets its AutoML product.
For a churn-related example of how scenario analysis may function, BigSquid claims its simulation tool allows stakeholders to see (1) the probability of a person churning, (2) the features that locally drive this churn, and (3) how this probability changes when certain variables are tuned.
Compellon has a similar offering, but it goes a step further in allowing a user to specify a desired reduction in churn, which the tool responds to by highlighting both impactful features to target and the expected drop in churn associated with their variation.
Such a comparison suggests a distinction between manual exploration and automatic optimisation, but judging the effectiveness of these scenario-based offerings is out of scope for this review.
Elsewhere, while not included within the 25 comprehensive systems here, Ople.AI notably promotes a scenario tool with a unique `ROI' tab that converts model metrics into financial values, aiming to translate technical outputs into a business understanding efficiently.

\begin{table}[h]
\caption{Distribution of scores for commercial comprehensive AutoML systems (EX6). Evaluates capabilities for bias/fairness assessment. Scores: 0 for none, 1 for convenience features, 2 for substantial automation, and U for unclear.}
\label{Table:framework_EX6}
\begin{tabular}{|cc|}
\hline
\rowcolor[HTML]{B4C6E7}
\multicolumn{2}{|c|}{\cellcolor[HTML]{B4C6E7}\textbf{EX6}}                         \\ \hline
\rowcolor[HTML]{B4C6E7}
\multicolumn{1}{|c|}{\cellcolor[HTML]{B4C6E7}\textbf{Score}} & \textbf{\# Vendors} \\ \hline
\rowcolor[HTML]{FFFFFF} 
\multicolumn{1}{|c|}{\cellcolor[HTML]{FFFFFF}0}              & 17                  \\ \hline
\rowcolor[HTML]{FFFFFF} 
\multicolumn{1}{|c|}{\cellcolor[HTML]{FFFFFF}1}              & 4                   \\ \hline
\rowcolor[HTML]{FFFFFF} 
\multicolumn{1}{|c|}{\cellcolor[HTML]{FFFFFF}2}              & 4                   \\ \hline
\rowcolor[HTML]{FFFFFF} 
\multicolumn{1}{|c|}{\cellcolor[HTML]{FFFFFF}U}              & 0                   \\ \hline
\end{tabular}
\end{table}
\begin{table}[h]
\caption{Distribution of scores for commercial comprehensive AutoML systems (EX7). Evaluates capabilities for bias/fairness management. Scores: 0 for none, 1 for convenience features, 2 for substantial automation, and U for unclear.}
\label{Table:framework_EX7}
\begin{tabular}{|cc|}
\hline
\rowcolor[HTML]{B4C6E7}
\multicolumn{2}{|c|}{\cellcolor[HTML]{B4C6E7}\textbf{EX7}}                         \\ \hline
\rowcolor[HTML]{B4C6E7}
\multicolumn{1}{|c|}{\cellcolor[HTML]{B4C6E7}\textbf{Score}} & \textbf{\# Vendors} \\ \hline
\rowcolor[HTML]{FFFFFF} 
\multicolumn{1}{|c|}{\cellcolor[HTML]{FFFFFF}0}              & 23                  \\ \hline
\rowcolor[HTML]{FFFFFF} 
\multicolumn{1}{|c|}{\cellcolor[HTML]{FFFFFF}1}              & 1                   \\ \hline
\rowcolor[HTML]{FFFFFF} 
\multicolumn{1}{|c|}{\cellcolor[HTML]{FFFFFF}2}              & 1                   \\ \hline
\rowcolor[HTML]{FFFFFF} 
\multicolumn{1}{|c|}{\cellcolor[HTML]{FFFFFF}U}              & 0                   \\ \hline
\end{tabular}
\end{table}

Regarding bias and fairness issues, their identification (EX6) and management (EX7) have not yet convincingly found their way into AutoML implementations within the commercial sphere.
This state of affairs starkly contrasts the volume of discussion held within the media, business community, and academic literature.
Indeed, only a third of surveyed vendors, half of these via convenience features, are shown by Table~\ref{Table:framework_EX6} to surface associated metrics.
Mitigation and remediation, either assisted or automated, are shown by Table~\ref{Table:framework_EX7} to be virtually nonexistent.
This absence of functionality provides an ongoing risk for current AutoML providers, especially as regulations and policies for compliance continue to build, all aiming to safeguard trust in AI.
However, even if the liability for unfair outcomes rests solely with the client, this feature gap is not ideal.
For most options, stakeholders presently need to undertake their own bias/fairness analysis, yet understanding how and why relevant metrics are calculated is an advanced skill in its own right~\cite{lesi21}.
Only IBM thus far seems to assess and handle such issues automatically, and the processes involved are not well detailed.
That all stated, some lag is always to be expected between the time a concern rises into mainstream consciousness and the time it is pragmatically addressed.
There is no current indication that the AutoML marketplace will not adapt to meet these evolving requirements held by industrial organisations and broader society.

\subsubsection{Ease of Use}
\label{Sec:CommercialEase}

\begin{table}[h]
\caption{Scores for commercial comprehensive AutoML systems (EU1--EU5). Evaluates the availability of interactions via coding (EU1), CLI (EU2), and GUI (EU3). Also evaluates software client type (EU4) and level of documentation (EU5). See Table~\ref{Table:Criteria_EU1-5} for rubric.}
\label{Table:framework_EU1_EU5}
\begin{tabular}{|l|c|c|c|c|c|c|}
\hline
\rowcolor[HTML]{B4C6E7}
\textbf{Name}                        & \textbf{EU1} & \textbf{EU2} & \textbf{EU3} & \textbf{EU4} & \textbf{EU5} & \textbf{Sum (out of 7)} \\ \hline
BigML           & 1 & 0 & 1 & 2 & 2 & 6 \\ \hline
cnvrg.io        & 1 & 1 & 1 & 1 & 2 & 6 \\ \hline
Dataiku         & 1 & 1 & 1 & 1 & 2 & 6 \\ \hline
Google          & 1 & 1 & 1 & 1 & 2 & 6 \\ \hline
H2O             & 1 & 0 & 1 & 2 & 2 & 6 \\ \hline
KNIME           & 1 & 0 & 1 & 2 & 2 & 6 \\ \hline
Microsoft       & 1 & 1 & 1 & 1 & 2 & 6 \\ \hline
RapidMiner      & 1 & 0 & 1 & 2 & 2 & 6 \\ \hline
SageMaker       & 1 & 1 & 1 & 1 & 2 & 6 \\ \hline
Alteryx         & 0 & 0 & 1 & 2 & 2 & 5 \\ \hline
Auger           & 1 & 1 & 1 & 1 & 1 & 5 \\ \hline
D2iQ            & 1 & 1 & 0 & 1 & 2 & 5 \\ \hline
DataRobot       & 1 & 0 & 1 & 1 & 2 & 5 \\ \hline
Deep Cognition  & 1 & 0 & 1 & 1 & 2 & 5 \\ \hline
IBM             & 1 & 0 & 1 & 1 & 2 & 5 \\ \hline
SAS             & 1 & 1 & 1 & 1 & 1 & 5 \\ \hline
Spell           & 1 & 1 & 0 & 1 & 2 & 5 \\ \hline
Big Squid       & 0 & 0 & 1 & 1 & 2 & 4 \\ \hline
Databricks      & 1 & 0 & 0 & 1 & 2 & 4 \\ \hline
Einblick        & 0 & 0 & 1 & 1 & 2 & 4 \\ \hline
MyDataModels    & 0 & 0 & 1 & 1 & 1 & 3 \\ \hline
B2Metric        & 0 & 0 & 1 & 1 & 0 & 2 \\ \hline
Compellon       & 0 & 0 & 1 & 1 & 0 & 2 \\ \hline
Number Theory   & 0 & 0 & 1 & 1 & 0 & 2 \\ \hline
TIMi            & 0 & 0 & 1 & 0 & 0 & 1 \\ \hline
\cellcolor[HTML]{B4C6E7}\textbf{Sum} & 
17/25           & 
9/25           & 
22/25           & 
29/50           & 
39/50           &              \\ \hline
\end{tabular}
\end{table}
\begin{table}[h]
\caption{Distribution of scores for commercial comprehensive AutoML systems (EU1). Evaluates the availability of interactions via coding. Scores: 0 for absent, 1 for present, and U for unclear.}
\label{Table:framework_EU1}
\begin{tabular}{|cc|}
\hline
\rowcolor[HTML]{B4C6E7}
\multicolumn{2}{|c|}{\cellcolor[HTML]{B4C6E7}\textbf{EU1}}                         \\ \hline
\rowcolor[HTML]{B4C6E7}
\multicolumn{1}{|c|}{\cellcolor[HTML]{B4C6E7}\textbf{Score}} & \textbf{\# Vendors} \\ \hline
\rowcolor[HTML]{FFFFFF} 
\multicolumn{1}{|c|}{\cellcolor[HTML]{FFFFFF}0}              & 8                   \\ \hline
\rowcolor[HTML]{FFFFFF} 
\multicolumn{1}{|c|}{\cellcolor[HTML]{FFFFFF}1}              & 17                  \\ \hline
\rowcolor[HTML]{FFFFFF} 
\multicolumn{1}{|c|}{\cellcolor[HTML]{FFFFFF}U}              & 0                   \\ \hline
\end{tabular}
\end{table}

This survey does not assess the effectiveness of HCI options employed by comprehensive AutoML systems, leaving broader commentary to another review~\cite{khke21}.
However, it can highlight how users are expected to interface with these tools, which correlates with intended audiences and accessibility.
Unsurprisingly, reflecting the diverging slants of free and commercial software towards technicians and lay users, respectively, the open-source evaluations in Section~\ref{Sec:OpenEase} end up markedly different from those in Table~\ref{Table:framework_EU1_EU5}.
This assertion about divergence is, of course, a generalisation, as all AutoML implementations strive to serve as many stakeholders as possible, but the trends are nonetheless apparent.
For instance, Table~\ref{Table:framework_EU1} shows that eight surveyed vendors do not even support text-based coding (EU1).
This outcome contrasts with the overwhelming expectation that users, presumably data scientists, are meant to script with the functions provided in open-source software.
After all, few free tools even provide alternative forms of HCI.

\begin{table}[h]
\caption{Distribution of scores for commercial comprehensive AutoML systems (EU2). Evaluates the availability of interactions via CLI. Scores: 0 for absent, 1 for present, and U for unclear.}
\label{Table:framework_EU2}
\begin{tabular}{|cc|}
\hline
\rowcolor[HTML]{B4C6E7}
\multicolumn{2}{|c|}{\cellcolor[HTML]{B4C6E7}\textbf{EU2}}                         \\ \hline
\rowcolor[HTML]{B4C6E7}
\multicolumn{1}{|c|}{\cellcolor[HTML]{B4C6E7}\textbf{Score}} & \textbf{\# Vendors} \\ \hline
\rowcolor[HTML]{FFFFFF} 
\multicolumn{1}{|c|}{\cellcolor[HTML]{FFFFFF}0}              & 16                  \\ \hline
\rowcolor[HTML]{FFFFFF} 
\multicolumn{1}{|c|}{\cellcolor[HTML]{FFFFFF}1}              & 9                   \\ \hline
\rowcolor[HTML]{FFFFFF} 
\multicolumn{1}{|c|}{\cellcolor[HTML]{FFFFFF}U}              & 0                   \\ \hline
\end{tabular}
\end{table}
\begin{table}[h]
\caption{Distribution of scores for commercial comprehensive AutoML systems (EU3). Evaluates the availability of interactions via GUI. Scores: 0 for absent, 1 for present, and U for unclear.}
\label{Table:framework_EU3}
\begin{tabular}{|cc|}
\hline
\rowcolor[HTML]{B4C6E7}
\multicolumn{2}{|c|}{\cellcolor[HTML]{B4C6E7}\textbf{EU3}}                         \\ \hline
\rowcolor[HTML]{B4C6E7}
\multicolumn{1}{|c|}{\cellcolor[HTML]{B4C6E7}\textbf{Score}} & \textbf{\# Vendors} \\ \hline
\rowcolor[HTML]{FFFFFF} 
\multicolumn{1}{|c|}{\cellcolor[HTML]{FFFFFF}0}              & 3                   \\ \hline
\rowcolor[HTML]{FFFFFF} 
\multicolumn{1}{|c|}{\cellcolor[HTML]{FFFFFF}1}              & 22                  \\ \hline
\rowcolor[HTML]{FFFFFF} 
\multicolumn{1}{|c|}{\cellcolor[HTML]{FFFFFF}U}              & 0                   \\ \hline
\end{tabular}
\end{table}

Now, while some vendors provide a CLI (EU2), as indicated by Table~\ref{Table:framework_EU2}, such a form of UI does not seem fashionable for either open-source or closed-source software.
Instead, with three exceptions, Table~\ref{Table:framework_EU3} confirms that GUIs (EU3) are the go-to within the commercial sphere of AutoML.
Of course, designing an effective graphical interface takes extra effort beyond what is needed for the core AutoML engine, so developing such a UI is more likely to be spurred on when financial incentives are involved.
However, the resulting benefits are clear: convenient controls, aesthetic appeal, and broader accessibility.
The downside is that GUIs form a layer of abstraction, constraining user freedom to whatever functions have been encapsulated by graphical widgets.
Broad accessibility and deep access are rarely both supported by the same UI, meaning that data scientists, in the absence of low-level coding, may struggle to fully/flexibly leverage the technical power a closed-source framework could provide.

\begin{table}[h]
\caption{Distribution of scores for commercial comprehensive AutoML systems (EU4). Evaluates software client type. Scores: 0 for desktop only, 1 for browser only, 2 for desktop or browser, and U for unclear.}
\label{Table:framework_EU4}
\begin{tabular}{|cc|}
\hline
\rowcolor[HTML]{B4C6E7}
\multicolumn{2}{|c|}{\cellcolor[HTML]{B4C6E7}\textbf{EU4}}                         \\ \hline
\rowcolor[HTML]{B4C6E7}
\multicolumn{1}{|c|}{\cellcolor[HTML]{B4C6E7}\textbf{Score}} & \textbf{\# Vendors} \\ \hline
\rowcolor[HTML]{FFFFFF} 
\multicolumn{1}{|c|}{\cellcolor[HTML]{FFFFFF}0}              & 1                   \\ \hline
\rowcolor[HTML]{FFFFFF} 
\multicolumn{1}{|c|}{\cellcolor[HTML]{FFFFFF}1}              & 19                  \\ \hline
\rowcolor[HTML]{FFFFFF} 
\multicolumn{1}{|c|}{\cellcolor[HTML]{FFFFFF}2}              & 5                   \\ \hline
\rowcolor[HTML]{FFFFFF} 
\multicolumn{1}{|c|}{\cellcolor[HTML]{FFFFFF}U}              & 0                   \\ \hline
\end{tabular}
\end{table}
\begin{table}[h]
\caption{Distribution of scores for commercial comprehensive AutoML systems (EU5). Evaluates level of documentation. Scores: 0 for none, 1 for partial, 2 for extensive, and U for unclear.}
\label{Table:framework_EU5}
\begin{tabular}{|cc|}
\hline
\rowcolor[HTML]{B4C6E7}
\multicolumn{2}{|c|}{\cellcolor[HTML]{B4C6E7}\textbf{EU5}}                         \\ \hline
\rowcolor[HTML]{B4C6E7}
\multicolumn{1}{|c|}{\cellcolor[HTML]{B4C6E7}\textbf{Score}} & \textbf{\# Vendors} \\ \hline
\rowcolor[HTML]{FFFFFF} 
\multicolumn{1}{|c|}{\cellcolor[HTML]{FFFFFF}0}              & 4                   \\ \hline
\rowcolor[HTML]{FFFFFF} 
\multicolumn{1}{|c|}{\cellcolor[HTML]{FFFFFF}1}              & 3                   \\ \hline
\rowcolor[HTML]{FFFFFF} 
\multicolumn{1}{|c|}{\cellcolor[HTML]{FFFFFF}2}              & 18                  \\ \hline
\rowcolor[HTML]{FFFFFF} 
\multicolumn{1}{|c|}{\cellcolor[HTML]{FFFFFF}U}              & 0                   \\ \hline
\end{tabular}
\end{table}

Another indicator that commercial AutoML and open-source AutoML diverge in their intended audience is the client type of its software (EU4).
Specifically, while all open-source comprehensive systems tend to be hosted on a desktop, potentially requiring technical know-how around their installation, Table~\ref{Table:framework_EU4} shows vendors prefer browser-based access.
This trend aligns with the software-as-a-service (SaaS) movement that has recently become popular.
In fact, commercial packages that support both browsers and desktops are often older entrants to the market, e.g.~KNIME, RapidMiner, and Alteryx.
As for the level of documentation (EU5) available, Table~\ref{Table:framework_EU5} suggests that vendors are more diligent on the whole than open-source developers, potentially needing to robustly support users with minimal ML expertise.
That stated, there are a few services with low scores here, although the associated vendors may simply reserve shielded documentation for paid customers.

Whatever the case, ease of use is rarely considered when evaluating AutoML technology, at least compared with technical performance metrics.
This criterion is currently associated with few research papers, and each only examines a small sample size of stakeholders~\cite{wawe19, drwe20}.
Accordingly, this review welcomes future systematic treatments of this topic, as they may shed further light on how much usability influences AutoML adoption by industrial organisations.

\subsubsection{Deployment and Management Effort}
\label{Sec:CommercialDeployment}

Open-source comprehensive AutoML systems appear to focus heavily on developing the best ML models possible.
Although they are still intended for general use, including by enterprises, such software is often closely linked to academic endeavours and thus prioritises exploiting technical innovations in ML.
Commercial software, on the other hand, appears to care less about enhancing an ML solution and more about how to make it pragmatically useful to the paying client.
The degree of this differentiation is debatable, but the concrete outcome is clear: commercial comprehensive AutoML systems distinguish themselves from free tools by investing heavily in MLOps capabilities.

However, as with all the preceding stages of the MLWF displayed in Fig.~\ref{Fig:mlworkflowsubtasks}, the deployment phase and subsequent monitoring and maintenance are clusterings of various possible tasks.
For instance, one process of interest in the academic literature is model compression (DM1), where a predictor with a complicated structure is simplified into a much smaller ML model that faithfully approximates the original function.
This procedure ideally trades a negligible amount of solution validity for massive savings in computational space and inference time.
Nonetheless, despite occasional associated research \cite{famu20}, we could not find this feature promoted by any AutoML implementation.
Seemingly, general concerns about model size remain negligible.
For one thing, businesses may still be leveraging enough value from classic ML models to leave compression techniques unwarranted.
The cheapening costs of cloud usage may also be limiting the pressure of computational constraints on all but the highest-volume vendors.
That all said, as businesses begin to leverage more complicated ML solutions for their possible added value, perhaps even with AutoDL maturing as a technology, model compression may become particularly appealing.

\begin{table}[h]
\caption{Scores for commercial comprehensive AutoML systems (DM2--DM4). Evaluates capabilities for deployment in specific environments (DM2). Also evaluates the existence of advanced mechanisms for deployment tests (DM3) and deployment updates (DM4). See Table~\ref{Table:Criteria_DM1-DM4} for rubric.}
\label{Table:framework_DM2_DM4}
\begin{tabular}{|l|c|c|c|c|}
\hline
\rowcolor[HTML]{B4C6E7}
\textbf{Name}  & \textbf{DM2} & \textbf{DM3} & \textbf{DM4} & \textbf{Sum (out of 4)} \\ \hline
cnvrg.io       & 2 & 1 & 1 & 4 \\ \hline
Dataiku        & 2 & 1 & 0 & 3 \\ \hline
DataRobot      & 2 & 1 & 0 & 3 \\ \hline
H2O            & 2 & 1 & 0 & 3 \\ \hline
Number Theory  & 2 & 1 & 0 & 3 \\ \hline
RapidMiner     & 2 & 1 & 0 & 3 \\ \hline
SAS            & 2 & 1 & 0 & 3 \\ \hline
Alteryx        & 2 & U & 0 & 2 \\ \hline
Big Squid      & 2 & 0 & 0 & 2 \\ \hline
BigML          & 2 & 0 & 0 & 2 \\ \hline
D2iQ           & 2 & 0 & 0 & 2 \\ \hline
Deep Cognition & 2 & 0 & 0 & 2 \\ \hline
IBM            & 2 & 0 & 0 & 2 \\ \hline
KNIME          & 2 & 0 & 0 & 2 \\ \hline
Microsoft      & 2 & 0 & 0 & 2 \\ \hline
Spell          & 2 & 0 & 0 & 2 \\ \hline
B2Metric       & U & 1 & 0 & 1 \\ \hline
Databricks     & 1 & U & 0 & 1 \\ \hline
Google         & 1 & 0 & 0 & 1 \\ \hline
SageMaker      & 1 & 0 & 0 & 1 \\ \hline
Auger          & U & 0 & 0 & 0 \\ \hline
Compellon      & 0 & 0 & 0 & 0 \\ \hline
Einblick       & 0 & 0 & 0 & 0 \\ \hline
MyDataModels   & 0 & 0 & 0 & 0 \\ \hline
TIMi           & 0 & 0 & 0 & 0 \\ \hline
\cellcolor[HTML]{B4C6E7}\textbf{Sum} & 
35/50           & 
8/25 & 
1/25 &   \\ \hline
\end{tabular}
\end{table}
\begin{table}[h]
\caption{Distribution of scores for commercial comprehensive AutoML systems (DM2). Evaluates capabilities for deployment in specific environments. Scores: 0 for none, 1 for cloud only, 2 for cloud or on-premise, and U for unclear.}
\label{Table:framework_DM2}
\begin{tabular}{|cc|}
\hline
\rowcolor[HTML]{B4C6E7}
\multicolumn{2}{|c|}{\cellcolor[HTML]{B4C6E7}\textbf{DM2}}                         \\ \hline
\rowcolor[HTML]{B4C6E7}
\multicolumn{1}{|c|}{\cellcolor[HTML]{B4C6E7}\textbf{Score}} & \textbf{\# Vendors} \\ \hline
\rowcolor[HTML]{FFFFFF} 
\multicolumn{1}{|c|}{\cellcolor[HTML]{FFFFFF}0}              & 4                   \\ \hline
\rowcolor[HTML]{FFFFFF} 
\multicolumn{1}{|c|}{\cellcolor[HTML]{FFFFFF}1}              & 3                   \\ \hline
\rowcolor[HTML]{FFFFFF} 
\multicolumn{1}{|c|}{\cellcolor[HTML]{FFFFFF}2}              & 16                  \\ \hline
\rowcolor[HTML]{FFFFFF} 
\multicolumn{1}{|c|}{\cellcolor[HTML]{FFFFFF}U}              & 2                   \\ \hline
\end{tabular}
\end{table}

For now, some standard questions can be asked of commercial AutoML systems, especially around deployment.
They and their evaluated answers are detailed in Table~\ref{Table:framework_DM2_DM4}.
First, a commercial service must be assessed for whether it even supports the productionisation of an ML solution.
If yes, it is essential to know where that product is deployed (DM2).
It turns out, as Table~\ref{Table:framework_DM2} shows, that the vast majority of vendors are flexible with their production environments; only four surveyed packages identifiably ignore this element of an MLWF.
However, stakeholders should still be mindful that a few commercial heavyweights are restricted solely to the cloud, e.g.~Google and SageMaker.
This constraint may be too limiting for organisations that, for security reasons, desire on-premise deployments for their ML applications.

\begin{table}[h]
\caption{Distribution of scores for commercial comprehensive AutoML systems (DM3). Evaluates the existence of advanced mechanisms for deployment tests. Scores: 0 for absent, 1 for present, and U for unclear.}
\label{Table:framework_DM3}
\begin{tabular}{|cc|}
\hline
\rowcolor[HTML]{B4C6E7}
\multicolumn{2}{|c|}{\cellcolor[HTML]{B4C6E7}\textbf{DM3}}                         \\ \hline
\rowcolor[HTML]{B4C6E7}
\multicolumn{1}{|c|}{\cellcolor[HTML]{B4C6E7}\textbf{Score}} & \textbf{\# Vendors} \\ \hline
\rowcolor[HTML]{FFFFFF} 
\multicolumn{1}{|c|}{\cellcolor[HTML]{FFFFFF}0}              & 15                  \\ \hline
\rowcolor[HTML]{FFFFFF} 
\multicolumn{1}{|c|}{\cellcolor[HTML]{FFFFFF}1}              & 8                   \\ \hline
\rowcolor[HTML]{FFFFFF} 
\multicolumn{1}{|c|}{\cellcolor[HTML]{FFFFFF}U}              & 2                   \\ \hline
\end{tabular}
\end{table}

\begin{table}[h]
\caption{Distribution of scores for commercial comprehensive AutoML systems (DM4). Evaluates the existence of advanced mechanisms for deployment updates. Scores: 0 for absent, 1 for present, and U for unclear.}
\label{Table:framework_DM4}
\begin{tabular}{|cc|}
\hline
\rowcolor[HTML]{B4C6E7}
\multicolumn{2}{|c|}{\cellcolor[HTML]{B4C6E7}\textbf{DM4}}                         \\ \hline
\rowcolor[HTML]{B4C6E7}
\multicolumn{1}{|c|}{\cellcolor[HTML]{B4C6E7}\textbf{Score}} & \textbf{\# Vendors} \\ \hline
\rowcolor[HTML]{FFFFFF} 
\multicolumn{1}{|c|}{\cellcolor[HTML]{FFFFFF}0}              & 24                  \\ \hline
\rowcolor[HTML]{FFFFFF} 
\multicolumn{1}{|c|}{\cellcolor[HTML]{FFFFFF}1}              & 1                   \\ \hline
\rowcolor[HTML]{FFFFFF} 
\multicolumn{1}{|c|}{\cellcolor[HTML]{FFFFFF}U}              & 0                   \\ \hline
\end{tabular}
\end{table}

Next up, once automated deployment support is identified, the most pertinent question is how this deployment works in practice.
For a real-world ML application, an organisation may iterate through numerous ML solutions, so Table~\ref{Table:framework_DM3} summarises whether commercial systems have advanced methods of testing deployments (DM3), particularly in relation to models that are presently online.
Such procedures include A/B testing and champion-challenger setups; they also tend to form the basis of monitoring approaches, which will be discussed shortly.
The takeaway here is that, while vendors do overwhelmingly assist in deploying ML solutions, only a small number automatically check how a prospective deployment compares against a previous one.
As for sophisticated means of shifting to that new deployment (DM4), which can be critical if end-users are actively employing predictions/prescriptions from an existing ML model, Table~\ref{Table:framework_DM4} notes that only one service explicitly provides.
Specifically, cnvrg.io supports a canary deployment strategy, where a new ML solution can be rolled out incrementally to subsets of users.
Evidently, while enthusiastic, the embrace of MLOps by the commercial AutoML marketplace is still very nascent.

\begin{table}[h]
\caption{Scores for commercial comprehensive AutoML systems (DM5--DM11). Evaluates the setup procedure for solution monitoring (DM5). Also evaluates capabilities for monitoring hardware performance (DM6), model performance (DM7), data/concept drift (DM8), and bias/fairness metrics (DM9). Additionally evaluates the existence of reactive retraining (DM10) and proactive retraining (DM11). See Table~\ref{Table:Criteria_DM5-DM11} for rubric.}
\label{Table:framework_DM5_DM11}
\begin{tabular}{|l|c|c|c|c|c|c|c|c|}
\hline
\rowcolor[HTML]{B4C6E7}
\textbf{Name}               & \textbf{DM5} & \textbf{DM6} & \textbf{DM7} & \textbf{DM8} & \textbf{DM9} & \textbf{DM10} & \textbf{DM11} & \textbf{Sum (out of 10)} \\ \hline
DataRobot       & 2 & 1 & 1 & 1 & 1 & 2 & 0 & 8 \\ \hline
Microsoft       & 1 & 1 & 1 & 1 & 1 & 2 & 0 & 7 \\ \hline
SageMaker   & 2 & 1 & 1 & 1 & 1 & 1 & 0 & 7 \\ \hline
cnvrg.io           & 2 & 1 & 1 & 0 & 0 & 2 & 0 & 6 \\ \hline
Dataiku         & 2 & 0 & 1 & 1 & 0 & 2 & 0 & 6 \\ \hline
H2O          & 2 & 0 & 1 & 1 & 0 & 2 & 0 & 6 \\ \hline
IBM             & 2 & 0 & 1 & 1 & 1 & 1 & 0 & 6 \\ \hline
Auger        & 2 & 0 & 1 & 0 & 0 & 2 & 0 & 5 \\ \hline
B2Metric        & 2 & U & U & U & U & 2 & 1 & 5 \\ \hline
Number Theory & 2 & 0 & 1 & 0 & 0 & 2 & 0 & 5 \\ \hline
Alteryx         & 2 & 1 & 1 & 0 & 0 & U & 0 & 4 \\ \hline
Google          & 1 & 1 & 1 & 1 & 0 & 0 & 0 & 4 \\ \hline
KNIME           & 2 & 1 & 0 & 0 & 0 & 1 & 0 & 4 \\ \hline
RapidMiner      & 1 & 0 & 1 & 1 & 0 & 1 & 0 & 4 \\ \hline
SAS             & 1 & 0 & 1 & 0 & 0 & 2 & 0 & 4 \\ \hline
D2iQ            & 2 & 1 & 0 & 0 & 0 & 0 & 0 & 3 \\ \hline
Spell           & 1 & 1 & 1 & 0 & 0 & 0 & 0 & 3 \\ \hline
Databricks      & 1 & 1 & 0 & 0 & 0 & U & 0 & 2 \\ \hline
BigML           & 1 & U & U & U & U & 0 & 0 & 1 \\ \hline
Big Squid       & U & U & U & U & U & 0 & 0 & 0 \\ \hline
Compellon       & 0 & 0 & 0 & 0 & 0 & 0 & 0 & 0 \\ \hline
Deep Cognition   & 0 & 0 & 0 & 0 & 0 & 0 & 0 & 0 \\ \hline
Einblick        & 0 & 0 & 0 & 0 & 0 & 0 & 0 & 0 \\ \hline
MyDataModels    & 0 & 0 & 0 & 0 & 0 & 0 & 0 & 0 \\ \hline
TIMi           & 0 & 0 & 0 & 0 & 0 & 0 & 0 & 0 \\ \hline
\cellcolor[HTML]{B4C6E7}Sum & 
31/50           & 
10/25           & 
14/25           & 
8/25            & 
4/25            & 
22/75            & 
1/25             &              \\ \hline
\end{tabular}
\end{table}

Deployment capabilities, of course, are broadly applicable, i.e.~equally relevant to one-and-done ML projects.
Iterations can also be warranted simply for implementational reasons, e.g.~fine-tuning end-user access to ML solution outputs.
Thus, the notion of deployment alone does not capture the sense of continuous learning, which is a crucial prerequisite to AutonoML and next-generation frameworks~\cite{kemu20}.
This concept is not just a topic of intellectual curiosity either; many businesses during the COVID pandemic have wised up to the need for adaptation.
Hence, the remaining sub-criteria here assess commercial comprehensive systems for the other important facets of MLOps: monitoring and maintenance.
The evaluations are detailed in Table~\ref{Table:framework_DM5_DM11}.

\begin{table}[h]
\caption{Distribution of scores for commercial comprehensive AutoML systems (DM5). Evaluates the setup procedure for solution monitoring. Scores: 0 for none, 1 for convenience features, 2 for substantial automation, and U for unclear.}
\label{Table:framework_DM5}
\begin{tabular}{|cc|}
\hline
\rowcolor[HTML]{B4C6E7}
\multicolumn{2}{|c|}{\cellcolor[HTML]{B4C6E7}\textbf{DM5}}                         \\ \hline
\rowcolor[HTML]{B4C6E7}
\multicolumn{1}{|c|}{\cellcolor[HTML]{B4C6E7}\textbf{Score}} & \textbf{\# Vendors} \\ \hline
\rowcolor[HTML]{FFFFFF} 
\multicolumn{1}{|c|}{\cellcolor[HTML]{FFFFFF}0}              & 5                   \\ \hline
\rowcolor[HTML]{FFFFFF} 
\multicolumn{1}{|c|}{\cellcolor[HTML]{FFFFFF}1}              & 7                   \\ \hline
\rowcolor[HTML]{FFFFFF} 
\multicolumn{1}{|c|}{\cellcolor[HTML]{FFFFFF}2}              & 12                  \\ \hline
\rowcolor[HTML]{FFFFFF} 
\multicolumn{1}{|c|}{\cellcolor[HTML]{FFFFFF}U}              & 1                   \\ \hline
\end{tabular}
\end{table}
\begin{table}[h]
\caption{Distribution of scores for commercial comprehensive AutoML systems (DM6). Evaluates capabilities for monitoring hardware performance. Scores: 0 for absent, 1 for present, and U for unclear.}
\label{Table:framework_DM6}
\begin{tabular}{|cc|}
\hline
\rowcolor[HTML]{B4C6E7}
\multicolumn{2}{|c|}{\cellcolor[HTML]{B4C6E7}\textbf{DM6}}                         \\ \hline
\rowcolor[HTML]{B4C6E7}
\multicolumn{1}{|c|}{\cellcolor[HTML]{B4C6E7}\textbf{Score}} & \textbf{\# Vendors} \\ \hline
\rowcolor[HTML]{FFFFFF} 
\multicolumn{1}{|c|}{\cellcolor[HTML]{FFFFFF}0}              & 12                  \\ \hline
\rowcolor[HTML]{FFFFFF} 
\multicolumn{1}{|c|}{\cellcolor[HTML]{FFFFFF}1}              & 10                  \\ \hline
\rowcolor[HTML]{FFFFFF} 
\multicolumn{1}{|c|}{\cellcolor[HTML]{FFFFFF}U}              & 3                   \\ \hline
\end{tabular}
\end{table}
\begin{table}[h]
\caption{Distribution of scores for commercial comprehensive AutoML systems (DM7). Evaluates capabilities for monitoring model performance. Scores: 0 for absent, 1 for present, and U for unclear.}
\label{Table:framework_DM7}
\begin{tabular}{|cc|}
\hline
\rowcolor[HTML]{B4C6E7}
\multicolumn{2}{|c|}{\cellcolor[HTML]{B4C6E7}\textbf{DM7}}                         \\ \hline
\rowcolor[HTML]{B4C6E7}
\multicolumn{1}{|c|}{\cellcolor[HTML]{B4C6E7}\textbf{Score}} & \textbf{\# Vendors} \\ \hline
\rowcolor[HTML]{FFFFFF} 
\multicolumn{1}{|c|}{\cellcolor[HTML]{FFFFFF}0}              & 8                   \\ \hline
\rowcolor[HTML]{FFFFFF} 
\multicolumn{1}{|c|}{\cellcolor[HTML]{FFFFFF}1}              & 14                  \\ \hline
\rowcolor[HTML]{FFFFFF} 
\multicolumn{1}{|c|}{\cellcolor[HTML]{FFFFFF}U}              & 3                   \\ \hline
\end{tabular}
\end{table}
\begin{table}[h]
\caption{Distribution of scores for commercial comprehensive AutoML systems (DM8). Evaluates capabilities for monitoring data/concept drift. Scores: 0 for absent, 1 for present, and U for unclear.}
\label{Table:framework_DM8}
\begin{tabular}{|cc|}
\hline
\rowcolor[HTML]{B4C6E7}
\multicolumn{2}{|c|}{\cellcolor[HTML]{B4C6E7}\textbf{DM8}}                         \\ \hline
\rowcolor[HTML]{B4C6E7}
\multicolumn{1}{|c|}{\cellcolor[HTML]{B4C6E7}\textbf{Score}} & \textbf{\# Vendors} \\ \hline
\rowcolor[HTML]{FFFFFF} 
\multicolumn{1}{|c|}{\cellcolor[HTML]{FFFFFF}0}              & 14                  \\ \hline
\rowcolor[HTML]{FFFFFF} 
\multicolumn{1}{|c|}{\cellcolor[HTML]{FFFFFF}1}              & 8                   \\ \hline
\rowcolor[HTML]{FFFFFF} 
\multicolumn{1}{|c|}{\cellcolor[HTML]{FFFFFF}U}              & 3                   \\ \hline
\end{tabular}
\end{table}
\begin{table}[h]
\caption{Distribution of scores for commercial comprehensive AutoML systems (DM9). Evaluates capabilities for monitoring bias/fairness metrics. Scores: 0 for absent, 1 for present, and U for unclear.}
\label{Table:framework_DM9}
\begin{tabular}{|cc|}
\hline
\rowcolor[HTML]{B4C6E7}
\multicolumn{2}{|c|}{\cellcolor[HTML]{B4C6E7}\textbf{DM9}}                         \\ \hline
\rowcolor[HTML]{B4C6E7}
\multicolumn{1}{|c|}{\cellcolor[HTML]{B4C6E7}\textbf{Score}} & \textbf{\# Vendors} \\ \hline
\rowcolor[HTML]{FFFFFF} 
\multicolumn{1}{|c|}{\cellcolor[HTML]{FFFFFF}0}              & 18                  \\ \hline
\rowcolor[HTML]{FFFFFF} 
\multicolumn{1}{|c|}{\cellcolor[HTML]{FFFFFF}1}              & 4                   \\ \hline
\rowcolor[HTML]{FFFFFF} 
\multicolumn{1}{|c|}{\cellcolor[HTML]{FFFFFF}U}              & 3                   \\ \hline
\end{tabular}
\end{table}

First, this review finds that a moderate number of vendors do support some form of monitoring.
As summarised by Table~\ref{Table:framework_DM5}, half of the commercial tools automate its setup (DM5), with many others supplying convenience features.
Curiously, although the field of MLOps tends to care about how best to manage infrastructure around an ML model object, Table~\ref{Table:framework_DM6} reveals that less than half of the sampled systems use this monitoring to keep a close eye on hardware usage and performance (DM6).
In contrast, it is somewhat more common, as shown by Table~\ref{Table:framework_DM7}, to track technical metrics for model performance (DM7).
However, even if 14 out of 25 packages do watch for model deterioration, such a practice is relatively basic.
Indeed, indicative that few commercial systems genuinely exploit more advanced theoretical innovations, Table~\ref{Table:framework_DM8} notes that less than a third identifiably bother to inspect the data itself for deleterious dynamics.
Preemptively catching data/concept drift (DM8) is understandably more complicated than simply noticing that the validity of an ML model is decreasing over time, but it is also much more informative.
As for monitoring bias/fairness metrics (DM9), Table~\ref{Table:framework_DM9} supports previous discussion in Section~\ref{Sec:CommercialExplainability}.
Specifically, if so few AutoML tools even calculate these variables to begin with, it is no surprise that examining their time dependence is very rare.

\begin{table}[h]
\caption{Distribution of scores for commercial comprehensive AutoML systems (DM10). Evaluates the existence of reactive retraining. Scores: 0 for none, 1 for manual with convenience features, 2 for automatic with user-defined triggers, 3 for automatic with developer-defined triggers, and U for unclear.}
\label{Table:framework_DM10}
\begin{tabular}{|cc|}
\hline
\rowcolor[HTML]{B4C6E7}
\multicolumn{2}{|c|}{\cellcolor[HTML]{B4C6E7}\textbf{DM10}}                        \\ \hline
\rowcolor[HTML]{B4C6E7}
\multicolumn{1}{|c|}{\cellcolor[HTML]{B4C6E7}\textbf{Score}} & \textbf{\# Vendors} \\ \hline
\rowcolor[HTML]{FFFFFF} 
\multicolumn{1}{|c|}{\cellcolor[HTML]{FFFFFF}0}              & 10                  \\ \hline
\rowcolor[HTML]{FFFFFF} 
\multicolumn{1}{|c|}{\cellcolor[HTML]{FFFFFF}1}              & 4                   \\ \hline
\rowcolor[HTML]{FFFFFF} 
\multicolumn{1}{|c|}{\cellcolor[HTML]{FFFFFF}2}              & 9                   \\ \hline
\rowcolor[HTML]{FFFFFF} 
\multicolumn{1}{|c|}{\cellcolor[HTML]{FFFFFF}3}              & 0                   \\ \hline
\rowcolor[HTML]{FFFFFF} 
\multicolumn{1}{|c|}{\cellcolor[HTML]{FFFFFF}U}              & 2                   \\ \hline
\end{tabular}
\end{table}
\begin{table}[h]
\caption{Distribution of scores for commercial comprehensive AutoML systems (DM11). Evaluates the existence of proactive retraining. Scores: 0 for absent, 1 for present, and U for unclear.}
\label{Table:framework_DM11}
\begin{tabular}{|cc|}
\hline
\rowcolor[HTML]{B4C6E7}
\multicolumn{2}{|c|}{\cellcolor[HTML]{B4C6E7}\textbf{DM11}}                        \\ \hline
\rowcolor[HTML]{B4C6E7}
\multicolumn{1}{|c|}{\cellcolor[HTML]{B4C6E7}\textbf{Score}} & \textbf{\# Vendors} \\ \hline
\rowcolor[HTML]{FFFFFF} 
\multicolumn{1}{|c|}{\cellcolor[HTML]{FFFFFF}0}              & 24                  \\ \hline
\rowcolor[HTML]{FFFFFF} 
\multicolumn{1}{|c|}{\cellcolor[HTML]{FFFFFF}1}              & 1                   \\ \hline
\rowcolor[HTML]{FFFFFF} 
\multicolumn{1}{|c|}{\cellcolor[HTML]{FFFFFF}U}              & 0                   \\ \hline
\end{tabular}
\end{table}

Finally, maintenance is the follow-up to monitoring, a process during which AutoML systems ideally attend to the accumulated flaws they have identified.
In theory, any framework that employs mechanisms to compare/update deployed ML solutions is already geared to be adaptive.
Thus, it is no surprise that Table~\ref{Table:framework_DM10} finds 13 surveyed commercial services identifiably support stakeholders with reactive model retraining (DM10).
Sometimes the user is only provided with a convenient pathway to enact this procedure, but most of these AutoML systems take care of updates automatically based on user-specified triggers.
As for the proactive form of retraining (DM11), where an ML solution is constantly being improved after deployment rather than simply being preserved against detrimental data dynamics, it is shown by Table~\ref{Table:framework_DM11} to be much rarer.
Admittedly, there are AutoML tools not included within this analysis that claim to support such capabilities.
For instance, TAZI \cite{Tazi} states it has a mechanism, i.e.~`TAZI Hunt', that is continuously looking for better models.
Elsewhere, Pecan notes on its website that ``our platform continuously monitors and optimizes your models''.
This review cannot verify the effectiveness of either due to a lack of publicly available detail.

As a concluding comment, while reactive retraining is reasonably common in the commercial AutoML sphere, the technical sophistication of these mechanisms should not be presumed.
Indeed, not one surveyed comprehensive system counted in Table~\ref{Table:framework_DM10} takes full responsibility for retraining, i.e.~via developer-provided triggers, although, granted, this may again be a consequence of vendors limiting their liability for poor outcomes.
Furthermore, the standard champion-challenger setup may be interpreted as comparing various ML `experts', but it does not ensemble them in any meaningful way to preserve local information.
Indeed, model retraining tends to be all-or-nothing, facing risks such as catastrophic interference/forgetting.
There are definitely more potent algorithms and frameworks to be found in the academic literature, and these may be required for next-generation AutonoML systems~\cite{kemu20, khke21, kaga09, ruga11, tsga12, bafa21}.
However, it is also an open question regarding when adaptation should be triggered, and cost-benefit analyses are required~\cite{zlbu15}.
Server prices in cloud infrastructures are usage-based, so model updates have a direct and quantifiable monetary cost; proactive retraining may not be worthwhile.
Ultimately, the best practices for persistent AutoML and adaptation to data dynamics will continue to be hashed out on both the academic and technological fronts.

\subsubsection{Governance and Security}
\label{Sec:CommercialGovernance}

\begin{table}[h]
\caption{Scores for commercial comprehensive AutoML systems (G1--G3). Evaluates capabilities for auditing (G1). Also evaluates the existence of access controls for model/data artefacts (G2) and ML application processes (G3). See Table~\ref{Table:Criteria_G1-G3} for rubric.}
\label{Table:framework_G1_G3}
\begin{tabular}{|l|c|c|c|c|}
\hline
\rowcolor[HTML]{B4C6E7}
\textbf{Name}               & \textbf{G1} & \textbf{G2} & \textbf{G3} & \textbf{Sum (out of 3)} \\ \hline
BigML                       & 1           & 1           & 1           & 3            \\ \hline
Databricks                  & 1           & 1           & 1           & 3            \\ \hline
Dataiku                     & 1           & 1           & 1           & 3            \\ \hline
DataRobot                   & 1           & 1           & 1           & 3            \\ \hline
Google                      & 1           & 1           & 1           & 3            \\ \hline
IBM                         & 1           & 1           & 1           & 3            \\ \hline
KNIME                       & 1           & 1           & 1           & 3            \\ \hline
Microsoft                   & 1           & 1           & 1           & 3            \\ \hline
RapidMiner                  & 1           & 1           & 1           & 3            \\ \hline
SageMaker               & 1           & 1           & 1           & 3            \\ \hline
SAS                         & 1           & 1           & 1           & 3            \\ \hline
Alteryx                     & 1           & 1           & 0           & 2            \\ \hline
B2Metric                    & 1           & 1           & U           & 2            \\ \hline
D2iQ                        & 0           & 1           & 1           & 2            \\ \hline
Einblick                    & 0           & 1           & 1           & 2            \\ \hline
Spell                       & 0           & 1           & 1           & 2            \\ \hline
cnvrg.io                       & 0           & 1           & 0           & 1            \\ \hline
Auger                    & U           & U           & U           & 0            \\ \hline
Big Squid                   & 0           & 0           & 0           & 0            \\ \hline
Compellon                   & U           & U           & U           & 0            \\ \hline
Deep Cognition               & 0           & 0           & 0           & 0            \\ \hline
H2O                      & 0           & 0           & 0           & 0            \\ \hline
MyDataModels                & U           & U           & U           & 0            \\ \hline
Number Theory             & U           & U           & U           & 0            \\ \hline
TIMi                       & 0           & 0           & 0           & 0            \\ \hline
\cellcolor[HTML]{B4C6E7}Sum & 
13/25          & 
17/25          & 
14/25          &              \\ \hline
\end{tabular}
\end{table}
\begin{table}[h]
\caption{Distribution of scores for commercial comprehensive AutoML systems (G1).  Evaluates capabilities for auditing. Scores: 0 for absent, 1 for present, and U for unclear.}
\label{Table:framework_G1}
\begin{tabular}{|cc|}
\hline
\rowcolor[HTML]{B4C6E7}
\multicolumn{2}{|c|}{\cellcolor[HTML]{B4C6E7}\textbf{G1}}                          \\ \hline
\rowcolor[HTML]{B4C6E7}
\multicolumn{1}{|c|}{\cellcolor[HTML]{B4C6E7}\textbf{Score}} & \textbf{\# Vendors} \\ \hline
\rowcolor[HTML]{FFFFFF} 
\multicolumn{1}{|c|}{\cellcolor[HTML]{FFFFFF}0}              & 8                   \\ \hline
\rowcolor[HTML]{FFFFFF} 
\multicolumn{1}{|c|}{\cellcolor[HTML]{FFFFFF}1}              & 13                  \\ \hline
\rowcolor[HTML]{FFFFFF} 
\multicolumn{1}{|c|}{\cellcolor[HTML]{FFFFFF}U}              & 4                   \\ \hline
\end{tabular}
\end{table}

Enterprise application of ML exists within a different environment of concerns to that of a hobbyist seeking low-stakes ML models for personal use.
Businesses caring about regulatory compliance and privacy matters may therefore be reluctant to employ free AutoML tools, which do not seem to support such safeguards.
Accordingly, as for the case of MLOps, Table~\ref{Table:framework_G1_G3} shows that this criterion is a significant differentiator between the commercial and open-source markets.
There is, however, a range of aggregate scores among the vendors.
Large corporations are already very familiar with the operational requirements of businesses, so the AutoML services of certain heavyweights score well in this area, e.g.~Google, Microsoft, and SageMaker.
Furthermore, much infrastructure supporting performant ML with governance and security does not need to be overly specialised, so vendors with existing mature frameworks in place for other applications can benefit here.
For instance, when considering auditability (G1), corporations that supply general cloud access already tend to have suitable governance mechanisms to leverage.
In combination with other packages that specifically emphasise suitability for regulatory requirements, such as BigML, it turns out that over half of the surveyed commercial systems identifiably support auditability.
This result is shown in Table~\ref{Table:framework_G1}.

\begin{table}[h]
\caption{Distribution of scores for commercial comprehensive AutoML systems (G2).  Evaluates the existence of access controls for model/data artefacts. Scores: 0 for absent, 1 for present, and U for unclear.}
\label{Table:framework_G2}
\begin{tabular}{|cc|}
\hline
\rowcolor[HTML]{B4C6E7}
\multicolumn{2}{|c|}{\cellcolor[HTML]{B4C6E7}\textbf{G2}}                          \\ \hline
\rowcolor[HTML]{B4C6E7}
\multicolumn{1}{|c|}{\cellcolor[HTML]{B4C6E7}\textbf{Score}} & \textbf{\# Vendors} \\ \hline
\rowcolor[HTML]{FFFFFF} 
\multicolumn{1}{|c|}{\cellcolor[HTML]{FFFFFF}0}              & 4                   \\ \hline
\rowcolor[HTML]{FFFFFF} 
\multicolumn{1}{|c|}{\cellcolor[HTML]{FFFFFF}1}              & 17                  \\ \hline
\rowcolor[HTML]{FFFFFF} 
\multicolumn{1}{|c|}{\cellcolor[HTML]{FFFFFF}U}              & 4                   \\ \hline
\end{tabular}
\end{table}
\begin{table}[h]
\caption{Distribution of scores for commercial comprehensive AutoML systems (G3).  Evaluates the existence of access controls for ML application processes. Scores: 0 for absent, 1 for present, and U for unclear.}
\label{Table:framework_G3}
\begin{tabular}{|cc|}
\hline
\rowcolor[HTML]{B4C6E7}
\multicolumn{2}{|c|}{\cellcolor[HTML]{B4C6E7}\textbf{G3}}                          \\ \hline
\rowcolor[HTML]{B4C6E7}
\multicolumn{1}{|c|}{\cellcolor[HTML]{B4C6E7}\textbf{Score}} & \textbf{\# Vendors} \\ \hline
\rowcolor[HTML]{FFFFFF} 
\multicolumn{1}{|c|}{\cellcolor[HTML]{FFFFFF}0}              & 6                   \\ \hline
\rowcolor[HTML]{FFFFFF} 
\multicolumn{1}{|c|}{\cellcolor[HTML]{FFFFFF}1}              & 14                  \\ \hline
\rowcolor[HTML]{FFFFFF} 
\multicolumn{1}{|c|}{\cellcolor[HTML]{FFFFFF}U}              & 5                   \\ \hline
\end{tabular}
\end{table}

As for security involving the data and model (G2), Table~\ref{Table:framework_G2} highlights that at least 17 vendors have installed some artefact access controls in their AutoML services.
Furthermore, if there are protection mechanisms for objects, then functions are also typically safeguarded (G3), e.g.~to prevent unauthorised users from overwriting existing ML deployments.
Only a few exceptions make up the differing distribution of scores in Table~\ref{Table:framework_G3}, e.g.~Alteryx and cnvrg.io.
Of course, as policies around responsible AI continue to evolve internationally, this criterion may warrant expansion into other detailed questions in the future.
However, for now, these evaluations should help emphasise that there are many facets to the smooth operation of a performant ML application; not all of them are technical.

\subsubsection{Discussion}
\label{Sec:CommercialDiscussion}

In hindsight~\cite{kemu20} and with foresight~\cite{khke21}, many aspects regarding the progression of AutoML may appear expected.
It can thus be tempting to ascribe a singular pathway to the field along which developments tread, especially given an academic community that is largely collaborative.
This sense of a convergent roadmap is further prompted by the modern realities of scientific funding and popular fervour; it is usually not a good academic career move to be \textit{too} radical.
However, when it comes to the competitive commercial sphere, this review finds that, by and large, all bets are off.
The only guaranteed driving force shared across all AutoML vendors is money, which requires engagement from stakeholder clients.
Accordingly, in this unusual swarm optimisation of performant ML, it is contrastingly desirable for vendor entities to diverge and differentiate, seeking their own profitable niches.
They, of course, remain grouped by the survey requirement that they supply a `comprehensive' AutoML system.
Nonetheless, such an environment supports the fragmentation of agendas, each with its own power to influence the ongoing evolution of AutoML technology.
Therefore, we highlight specific trends and clusters that have surfaced as part of this survey.

\textbf{A licence to explore.}
Relevant systems:
\begin{itemize}
    \item Einblick
    \item Compellon
    \item MyDataModels
    \item TIMi
\end{itemize}

Some AutoML systems are developed and promoted to dive deep into data.
They tend to score lower for deployment-related criteria while solidly supporting EDA and related auxiliary functions.
So, for example, they will likely not productionise an ML model that automatically identifies users at risk of churn.
Instead, an intended use case may be a marketing manager (1) uploading a spreadsheet of customers, (2) seeking to examine the driving features of churn, and (3) experimenting with various scenarios.
Moreover, as the listed tools focus on harvesting and understanding insights from data, they tend to prioritise clear visualisations and explainability.

\textbf{The primacy of MLOps.}
Relevant systems:
\begin{itemize}
    \item Google
    \item SageMaker
    \item cnvrg.io
    \item D2iQ
    \item Databricks
    \item SPELL
\end{itemize}

On the other end of the MLWF to EDA, it is often apparent which AutoML services prioritise MLOps.
These commercial systems score well on deployment and management metrics, but they often support other core functions of an ML application in a more rudimentary manner, if at all.
Amidst this list, Google and SageMaker are particularly interesting.
The associated corporations already had a business in providing cloud services and DevOps before touching AutoML, so their entrance into the marketplace may be a logical extension of supplying MLOps support to developers.
Essentially, their services have grown from the opposite direction to teams that started automating model selection and now focus on productionisation.
As a side note, heavyweight Microsoft has not been included here, given that the AutoML framework seems more suited to less-technical users, often providing a GUI rather than relying on users to code around convenience features.
These factors suggest a `from-scratch' approach to AutoML instead of a gradual expansion originating in DevOps.

\textbf{Some assembly required.}
Relevant systems:
\begin{itemize}
    \item KNIME (open marketplace)
    \item RapidMiner (open marketplace)
    \item cnvrg.io (semi-open marketplace)
    \item Alteryx (closed marketplace)
    \item Deep Cognition (closed marketplace)
    \item Einblick (closed marketplace)
    \item Number Theory (closed marketplace)
\end{itemize}

Several commercial AutoML frameworks are designed under a `building block' paradigm, where users typically fashion a network of nodes to represent and process an ML application.
Depending on the implementation, these nodes may be ML pipeline components, i.e.~specific data transformations, or higher-level ML operations.
A mix of the two may even be possible.
Whatever the case, it has occasionally been challenging to assess associated systems for their automated capabilities.
Indeed, the flexibility of free assembly has a drawback that, without appropriate defaults/constraints, users need to manually find and vet these building blocks before integrating them into modelling work.
Additionally, under this paradigm, the supply of possible nodes can vary.
Closed marketplaces, noted in the list of relevant systems, tend to have restricted offerings with assurances of functionality by the developer.
In contrast, the open marketplaces of KNIME and RapidMiner, where community members submit their own building blocks, can stimulate high scores for completeness and currency.
This boost, however, comes with the cost of further complicating trust in ML.
Both approaches have pros and cons.

\textbf{The legacy experience.}
Relevant systems:
\begin{itemize}
    \item BigML
    \item MyDataModels
    \item TIMi
\end{itemize}

The topics of design and aesthetics are somewhat subjective; individual user responses to various developer decisions cannot easily be predicted.
However, some trends drive the software industry, and many vendors have embraced the paradigm of SaaS.
Present-day products are often easily accessible to general stakeholders via a browser, exhibiting UI artefacts generated by Bootstrap and other modern front-end web development frameworks.
The tools listed here are thus notable for not existing within the same design sphere as many others.
Nonetheless, it remains an open question whether a legacy user experience (UX) influences their uptake and, if so, by how much.

\textbf{Degrees of transparency.} 
Relevant systems:
\begin{itemize}
    \item Dataiku (highly performant + highly transparent)
    \item DataRobot (highly performant + moderately transparent)
    \item H2O (moderately performant + highly transparent)
    \item IBM (moderately performant + moderately transparent)
    \item Auger (moderately performant + minimally transparent)
    \item B2Metric (moderately performant + minimally transparent)
    \item BigSquid (moderately performant + minimally transparent)
\end{itemize}

All commercial AutoML tools analysed within this section have been `comprehensive', automating key model-development processes central to traditional AutoML while also, in many cases, supporting other tasks along an MLWF.
Under this definition, the tools listed here are those within the analysis that have scored reasonably well across a breadth of criteria for performant ML.
They are not necessarily `better' than their competitors, but, at the same time, it does not hurt to be so thorough.
Indeed, some sub-criteria can be `make or break' for potential clients, including whether on-premise deployment is supported, whether a system is auditable, whether business-appropriate data can be easily prepared, and so on.

That all said, transparency also matters when stakeholders consider their options.
Such an assertion needs to be emphasised as, zero-scoring unknowns aside, the assessment framework for performant ML did not penalise commercial systems for every obscured detail.
In fact, several tools at the bottom of the list above were often given the benefit of the doubt regarding promoted claims.
Admittedly, this lack of transparency is common among smaller organisations.
A cynical perspective may be that a commercial fog allows gaps in services to be veiled. 
More charitably, one could argue that smaller vendors must be especially defensive of their competitive advantages within a business environment.
Whatever the case, it is likely that stakeholders will increasingly lean towards more transparent offerings as the commercial sphere becomes progressively more populated.
Ultimately, the future will tell whether the AutoML marketplace will remain as diverse and fragmented as it currently is or whether an effective oligopoly driven by consensus consumer desires will emerge.

\section{AutoML Applications}
\label{Sec:Applications}

At this point, it is worth recalling that a core purpose of this review is to comment on the translation of AutoML theory/design into AutoML implementation/technology as of the early 2020s.
Crucially, Section~\ref{Sec:CommercialDiscussion} acknowledged that both academia and the commercial sphere have varying levels of freedom to explore/generate new innovations in the field.
On the academic side, public funding -- we refer to fundamental-research grants -- necessarily hews to some degree of conservatism, so that taxpayer money is not `wasted'.
However, once endowed, these resources can often allow researchers to pursue threads of interest without immediate applicability.
In contrast, a privately funded startup can be as radical as it wants, subject to the approval of its investors, but its survival depends on the ability to maintain a healthy income, i.e.~those radical choices need to pay off.
Thus, it is the consumers that primarily determine whether a provider stays in business and continues to influence the marketplace.
The implication here is that, while innovation and experimentation will always influence AutoML products, the assertion in Section~\ref{Sec:Motivation} continues to hold merit within industry settings and broader society: ``demand creates supply''.

Unfortunately, the problem with evaluating a nascent technology is that it is often unclear what the demand profile for its use is, let alone how it will eventually look once it stabilises.
For instance, recent years have witnessed debates around internet access becoming enshrined as a human right, whereas, a century ago, virtually no one could predict that this non-existent concept would become a need.
So, to get a handle on the demand that drives AutoML technology, present and expected, this review has pursued a combination of several approaches.
In Section~\ref{Sec:keystakeholders}, we reasoned at a granular level what target stakeholders want from an ML application to claim that it is `performant'.
In Section~\ref{Sec:Tools}, we surveyed extant AutoML software, following the logic that its supply is an oblique reflection of its demand.

Here, we examine the final piece of the puzzle: ML applications that have been processed via AutoML practices and systems.
After all, stakeholder desires and the tools developed to satiate them are both symbols of hypothesis; they \textit{suggest} AutoML will be helpful to industry without ever \textit{proving} this to be true.
Of course, such forecasting is still valuable in these early years of the technology, especially while the evidence of industrial use is still building up.
Nonetheless, a survey of applications finds that there is already enough content to, in a preliminary fashion, directly analyse the `fulfilled' demand for AutoML.
Accordingly, with appropriate constraints, Section~\ref{Sec:AcademicApplications} investigates what academic literature reveals about the real-world use of this technology.
Then, acknowledging that some AutoML software has been preemptively designed for specific applications, Section~\ref{Sec:specialisedtools} discusses the phenomenon of domain specialisation.
Finally, although the associated narrative must be approached carefully, Section~\ref{Sec:IndustrialApplications} examines commentary on applications according to the vendors of AutoML.

\subsection{Academic Reports}
\label{Sec:AcademicApplications}

The most reliable evidence of successfully completed AutoML applications is likely to be found within the academic literature, thanks to an enforced degree of peer review by independent experts.
Naturally, the downside is that this perspective is also very limited, likely overlooking the more mundane usage of AutoML by commercial clients.
Academic publications need to be of sufficient scientific importance.
A further possible complication is that AutoML is meant to be an instrument for stakeholders to achieve other aims.
Ideally, it should not be the promotional point of any research, making it harder to identify relevant applications.
Nonetheless, this is not presently an obstacle.
Given that much AutoML literature within the past decade is on theory and design, its practical use is relatively novel and expected to be highlighted.
After all, the field is still looking to build an authoritative portfolio of archetypal examples to demonstrate rigorous proof of principle.

So, having acknowledged the nuances of an academic perspective, this section presents results from a literature survey.
However, because this monograph is focussed on the broader use of AutoML technology, stringent constraints on the scope were necessary.
First of all, the use of AutoML software by its own development team, without any significant industrial collaboration, did not make the cut any more than a self-citation would be considered a sign of broader research impact.
Also, general benchmark papers, of which there are many~\cite{chis21, lipa22a}, were likewise excluded.
For instance, testing experiments on UCI datasets~\cite{lipa22} is not necessarily representative of real-world uptake.
Publications deemed suitable for this survey need to target a specific industrial problem and associated dataset, proposing a solution obtained via AutoML.
High-level reviews were thus similarly ignored, even though one such article is notable for considering adaptivity in real-world contexts~\cite{sply18}; this is both rare and increasingly recognised as important.

Naturally, such a tightly constrained survey of academic literature is not trivial to undertake.
Citations of a publication that marks the release of an AutoML package cannot be blindly collated.
For instance, the seminal Auto-WEKA \cite{thhu13} has been referenced numerous times to justify employing Bayesian optimisation~\cite{czpo15, wija19} without any concurrent use of the implementation.
Similarly, perusing the literature with keywords related to ML and automation throws up many irrelevant results.
As an example, in doing so, one may encounter a manuscript titled ``automated staff assignment for building maintenance using natural language processing''~\cite{mozh20}.
This publication and others~\cite{hoth20} employ ML to mechanise a manual operation; they do not automate the higher-level processes of ML.
Of course, conversely to these false positives, such keywords can also miss valid applications that are not explicitly cognisant of the AutoML aspect but still use associated software.
As previously mentioned, these examples are presently rare, but they do exist.

Another challenge with an application survey is deciding how to treat the history of AutoML.
Particularly in earlier years, several proposed methods and advancements were not given names.
Others that were named cannot presently be linked to an open-source repository or extant software product, e.g.~SABLE~\cite{baga15}, FLASH~\cite{zhba16}, Learn-O-Matic~\cite{sefe12}, Predict-ML~\cite{lu16}, Net-Net~\cite{bamu18}, and HyperSPACE~\cite{hakr16}.
While this means exclusion from the earlier analyses of tools, many such AutoML approaches have a legitimate presence in an industry use case.
They may not be employed again with ease, but they still prototypically demonstrate how organisations may leverage AutoML in pursuing real-world business objectives.
So, in general, applications based on faded techniques and frameworks have been included in this survey.
At the same time, to provide a current lens for this review, only a ten-year window for academic reports has been considered, ending on the 3rd of August 2022.
This restriction should not be seen as diminishing the importance of earlier works.
For instance, a series of publications involving the chemical process industry is noteworthy for its prescient and sophisticated focus on automated adaptation~\cite{kaga09, kaga09a, kaga09b, kaga09c, kaga10, kaga10a, kagr11}, i.e.~a crucial element of monitoring and maintenance within an MLWF.

\begin{table}[h]
  \footnotesize{
  \caption{References to surveyed academic publications detailing AutoML applications. The references are grouped by major industries, subfields, and sub-subfields.}
  \label{table:academic_applications_all_papers}
  \begin{tabular}{|C{1.5cm}cc|L{4.2cm}|}
  \hline
  \rowcolor[HTML]{B4C6E7}
  \multicolumn{1}{|c|}{\cellcolor[HTML]{B4C6E7}\textbf{Industry}} &
    \multicolumn{1}{c|}{\cellcolor[HTML]{B4C6E7}\textbf{Subfield}} &
    \textbf{Sub-Subfield} &
    \multicolumn{1}{c|}{\cellcolor[HTML]{B4C6E7}\textbf{Ref.}} \\ \hline
  \multicolumn{1}{|c|}{\multirow{10}{*}{Health \& Biomedical (63)}} &
    \multicolumn{1}{c|}{\multirow{5}{*}{Diagnosis (32)}} &
    Breast Cancer (6) &
    \cite{maso19,mima22,raka21,refa21,sisa16,sise20} \\ \cline{3-4} 
  \multicolumn{1}{|c|}{} & \multicolumn{1}{c|}{} & Mental Health (6) &
    \cite{dapi19,dapi20,homa19,lefu20,ryvi22,valu21} \\ \cline{3-4} 
  \multicolumn{1}{|c|}{} & \multicolumn{1}{c|}{} &  \multicolumn{1}{c|}{\multirow{3}{*}{Other (20)}} &
    \cite{alsc18,ba20,bali22,bero21,cafe22,dire21,famu21,imra17,inco22,jakh22,megr17,payu19,riwo19,rora18,scsc22,sipa21,srk.21,tari21,zear22,zhba16} \\ \cline{2-4} 
  \multicolumn{1}{|c|}{} & \multicolumn{2}{c|}{\multirow{3}{*}{Condition Management  (19)}} &
    \cite{bash21,brst18,chli22,damu20,hago21,hole22,huyu21,koko19,lahi22,lich21a,lilu21,lust21,maar22,pe21,rilu22,sech21,sioz21,tile19,wapo21} \\ \cline{2-4} 
  \multicolumn{1}{|c|}{} & \multicolumn{2}{c|}{Genetics \& Biomarker Research (8)} &
    \cite{bape21,cobe17,levi19,male21,maro21a,mumi21,ormo17,tsch22} \\ \cline{2-4} 
  \multicolumn{1}{|c|}{} & \multicolumn{2}{c|}{Processing Medical Data (4)} &
    \cite{hakr16,lu16,lust17,olur16} \\ \hline
  \multicolumn{1}{|c|}{\multirow{4}{*}{Transport   \& Logistics (15)}} &
    \multicolumn{2}{c|}{Transport Demand (6)} &
    \cite{anma20,antr18,camc21,f.v.19,piar22,tsfa21} \\ \cline{2-4} 
  \multicolumn{1}{|c|}{} &
    \multicolumn{2}{c|}{Transport Infrastructure Management (5)} &
    \cite{hefi21,koga19,leli21,qixu21,qomo21} \\ \cline{2-4} 
  \multicolumn{1}{|c|}{} &
    \multicolumn{2}{c|}{Optimising   Logistics (2)} &
    \cite{deme21,hosh14} \\ \cline{2-4} 
  \multicolumn{1}{|c|}{} &
    \multicolumn{2}{c|}{Road   Accidents (2)} &
    \cite{anma21,shbi21} \\ \hline
  \multicolumn{1}{|c|}{\multirow{4}{*}{Chemical   \& Material Science (15)}} &
    \multicolumn{2}{c|}{Chemical Process (7)} &
    \cite{baga15,baga17,baga18,sa17,sabu16a,sabu19,zlga14} \\ \cline{2-4} 
  \multicolumn{1}{|c|}{} &
    \multicolumn{2}{c|}{Electrical   and Photovoltaic Compounds (3)} &
    \cite{luzh22,ngsa18,pa20a} \\ \cline{2-4} 
  \multicolumn{1}{|c|}{} &
    \multicolumn{2}{c|}{Other   - Material Science (3)} &
    \cite{desc21,mepl21,path18} \\ \cline{2-4} 
  \multicolumn{1}{|c|}{} &
    \multicolumn{2}{c|}{Other   - Chemistry (2)} &
    \cite{hash19,viha19} \\ \hline
  \multicolumn{1}{|c|}{\multirow{4}{*}{Information   Technology (13)}} &
    \multicolumn{2}{c|}{\multirow{2}{*}{Cybersecurity (10)}} &
    \cite{chbr21,hagu19,iswa21,lama21,lano21,lasa20,pupa21,sa21a,shxi21,vame21} \\ \cline{2-4} 
  \multicolumn{1}{|c|}{} &
    \multicolumn{2}{c|}{Software   Development (2)} &
    \cite{gala17,mopa22} \\ \cline{2-4} 
  \multicolumn{1}{|c|}{} &
    \multicolumn{2}{c|}{System   Robustness (1)} &
    \cite{zhgu16} \\ \hline
  \multicolumn{1}{|c|}{\multirow{7}{*}{Energy   \& Utilities (12)}} &
    \multicolumn{2}{c|}{Water (5)} &
    \cite{khba21,mugo22,taxi22,vese21,zhye20} \\ \cline{2-4} 
  \multicolumn{1}{|c|}{} &
    \multicolumn{1}{c|}{\multirow{4}{*}{Power Generation (4)}} &
    Solar (1) &
    \cite{dawa20} \\ \cline{3-4} 
  \multicolumn{1}{|c|}{} &
    \multicolumn{1}{c|}{} &
    Hydrocarbons (1) &
    \cite{yama22} \\ \cline{3-4} 
  \multicolumn{1}{|c|}{} &
    \multicolumn{1}{c|}{} &
    Wind (1) &
    \cite{prra22} \\ \cline{3-4} 
  \multicolumn{1}{|c|}{} &
    \multicolumn{1}{c|}{} &
    Other (1) &
    \cite{sadh22} \\ \cline{2-4} 
  \multicolumn{1}{|c|}{} &
    \multicolumn{2}{c|}{Power   Grid Optimisation (2)} &
    \cite{maes19,waba19} \\ \cline{2-4} 
  \multicolumn{1}{|c|}{} &
    \multicolumn{2}{c|}{Waste   Management (1)} &
    \cite{luhe20} \\ \hline
  \multicolumn{3}{|c|}{Agriculture (8)} &
    \cite{anan21,kali21,kiod22,kooi22,lisa22,nymv20,pail22,zhsu21} \\ \hline
  \multicolumn{3}{|c|}{Education (7)} &
    \cite{edim22,gama22,kadr21,kopa21,nopi22,same21,tsko20} \\ \hline
  \multicolumn{3}{|c|}{Meteorological (5)} &
    \cite{mamo22,me21,sefe12,susc21,zher20} \\ \hline
  \multicolumn{3}{|c|}{Radiography \&   Physics (5)} &
    \cite{lili22,maom20,paba21,poal21,zhge18} \\ \hline
  \multicolumn{3}{|c|}{Finance (5)} &
    \cite{adka21,adka21a,huhu17,juch21,mamo19} \\ \hline
  \multicolumn{3}{|c|}{Manufacturing \&   Machinery (5)} &
    \cite{buts19,gezi22,ladu21,mazh20,meku21} \\ \hline
  \multicolumn{3}{|c|}{Robotics (3)} &
    \cite{alga17,caal19,ebbe17} \\ \hline
  \multicolumn{3}{|c|}{Aviation (3)} &
    \cite{aiso21,cofr21,leri12} \\ \hline
  \multicolumn{3}{|c|}{Telecom (2)} &
    \cite{fepi21,zhli22} \\ \hline
  \multicolumn{3}{|c|}{Ecology (2)} &
    \cite{bamu18,flga21} \\ \hline
  \multicolumn{3}{|c|}{Retail (2)} &
    \cite{chtr13,jadh20} \\ \hline
  \multicolumn{3}{|c|}{Advertising (1)} &
    \cite{chya20} \\ \hline
  \multicolumn{3}{|c|}{Professional   Services (1)} &
    \cite{slsa21} \\ \hline
  \multicolumn{3}{|c|}{Sport (1)} &
    \cite{game21} \\ \hline
  \multicolumn{3}{|c|}{Media (1)} &
    \cite{abta22} \\ \hline
  \end{tabular}
  }
  \end{table}

In total, 169 papers have been included in this survey.
They are cited within Table~\ref{table:academic_applications_all_papers}; discussion around the listed groupings is deferred to later.
For each report, the following elements were extracted:
\begin{itemize}
    \item The tools used
    \item The application domain
    \item The metrics used in the experimental work
    \item The justifications given for the use of AutoML technology
    \item The stages of an MLWF involved in the study
    \item The year of publication
\end{itemize}

\begin{figure}[h]
  \centering
  \includegraphics[width=1\linewidth]{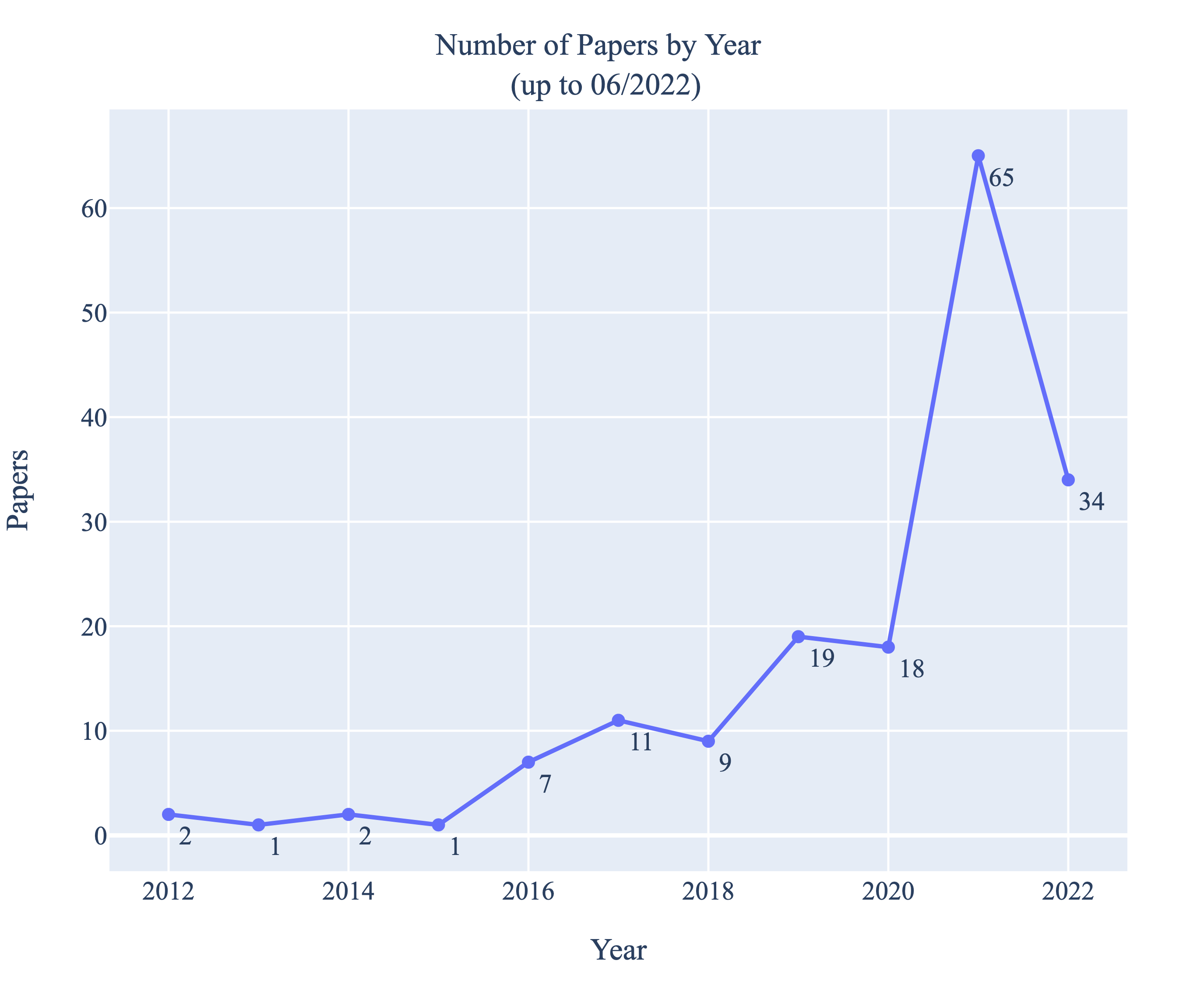}
  \caption{The yearly numbers of academic publications reporting on AutoML applications.}
  \label{Fig:papers_by_year}
\end{figure}

To begin the analysis based on this curated information, Fig.~\ref{Fig:papers_by_year} charts how many publications have reported on AutoML applications each year.
Given the early August survey cut-off, the projection for 2022 should at least be on par with 2021.
So, it is evident that the practical use of AutoML is surging, even in the eyes of academia.
Granted, the overall volume of work in this space is increasing, but this result also highlights that the AutoML field is no longer purely in the fundamental research stage.
Efforts to leverage its technological benefits within various domains are accumulating.

\begin{figure}[h]
  \centering
  \includegraphics[width=1\linewidth]{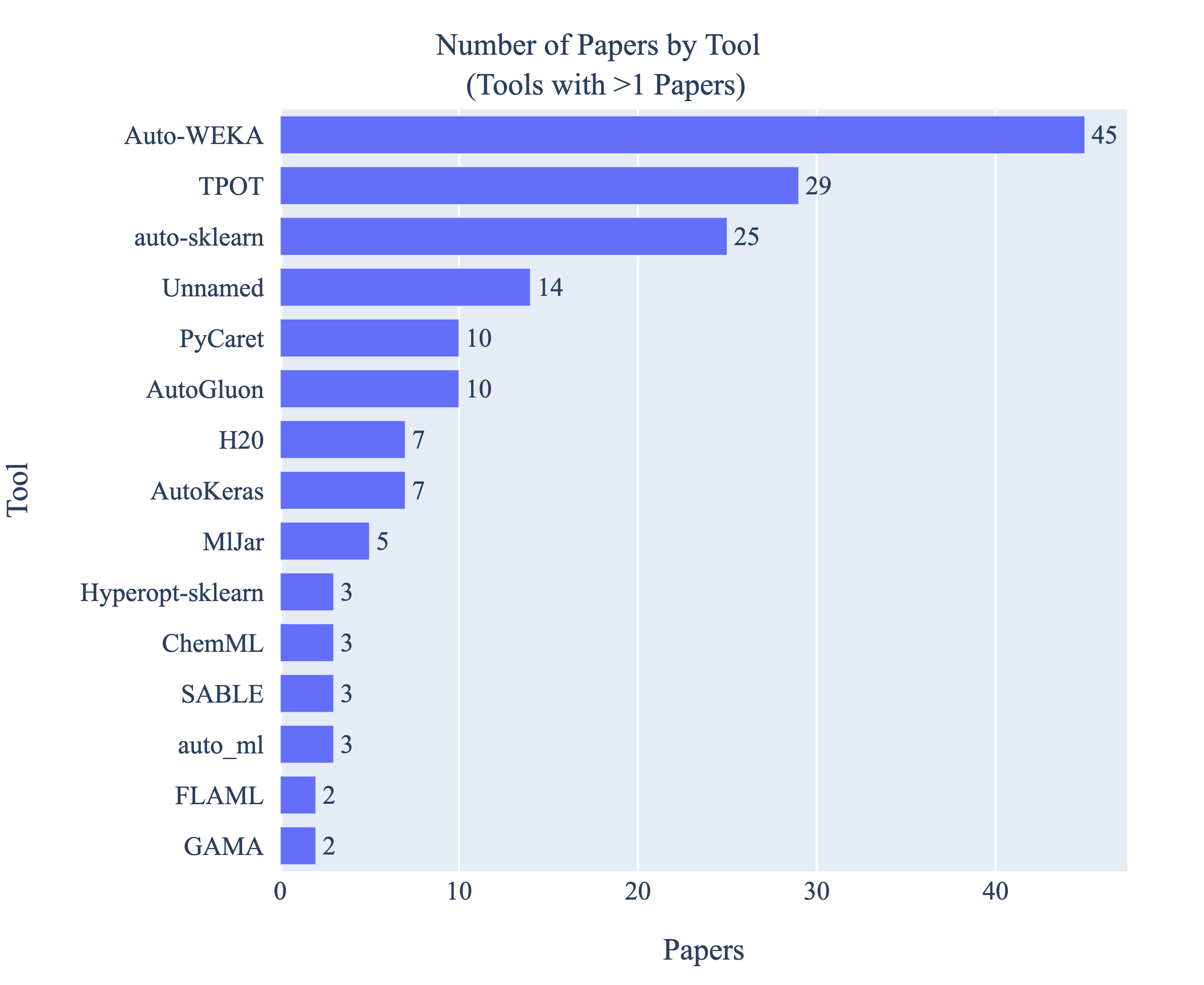}
  \caption{The AutoML tools associated with at least two application papers.}
  \label{Fig:papers_by_tool_top}
\end{figure}

\begin{figure}[h]
  \centering
  \includegraphics[width=1\linewidth]{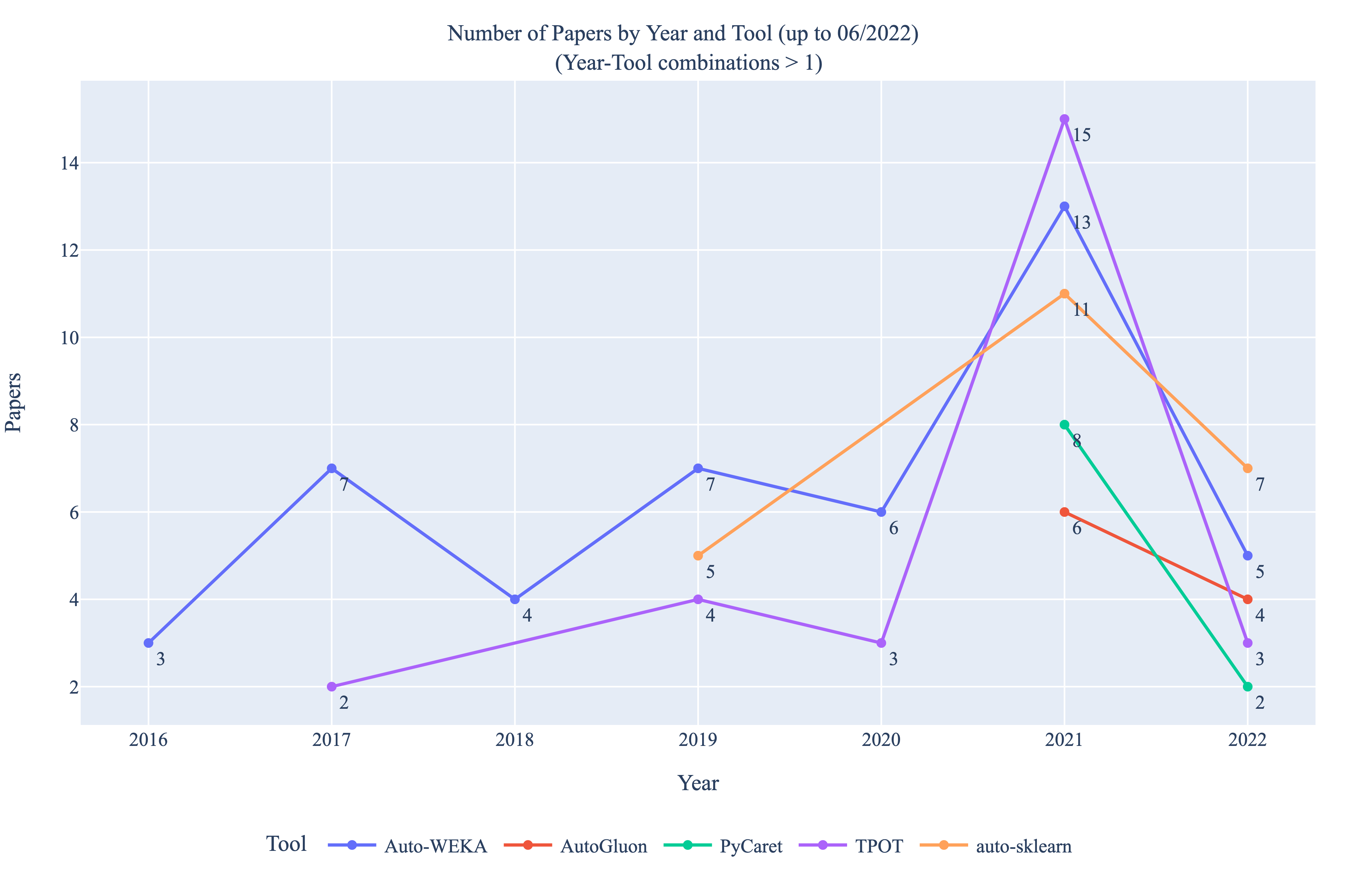}
  \caption{The yearly numbers of academic publications reporting on AutoML applications associated with popular AutoML tools.}
  \label{Fig:papers_by_year_tool}
\end{figure}

Now, while the surveyed applications have employed numerous AutoML frameworks, only a few have found repeat use.
These are listed in Fig.~\ref{Fig:papers_by_tool_top}.
Accordingly, the chart of yearly reporting can be broken down further to investigate the most popular tools, at least by publication volume.
It can immediately be seen from Fig.~\ref{Fig:papers_by_year_tool} that there is a `first mover' effect in play.
Despite the increasing amount and diversity of AutoML tools, many of which are likely to be more technically advanced and generally performant, Auto-WEKA and TPOT are still dominant in usage as of 2021.
Both open-source packages were very early entrants into the AutoML `marketplace'.
Accordingly, there may be a couple of factors involved.
Firstly, a system without competitors has the luxury of cultivating a solid reputation, assuming it meets a minimum threshold of functionality.
By the time there are competitors, the snowball effect is complete; prospective users are drawn to industry standards.
Secondly, whether or not a stakeholder buys into the reputation of an established system, it does serve as a consistent baseline for comparative performance, especially where ML applications might try a few AutoML frameworks.

Of course, there are many more reasons why stakeholders may have turned to some of the listed AutoML frameworks.
For instance, the rise of TPOT is likely supported by its relatively extensive coverage of an MLWF, e.g.~in terms of data preparation and feature generation/selection.
Other AutoML systems are also gradually developing a solid presence as of 2021, including auto-sklearn, AutoGluon, and PyCaret.
These tools benefit from strong communities, good documentation, and high levels of active development/maintenance.

Turning to broader trends, this review notes that many of the popular AutoML systems are considered generalist, e.g.~Auto-WEKA, auto-sklearn, TPOT, and AutoGluon.
Within the limits of their technical coverage, they are designed to be agnostic in terms of domain and dataset.
However, most application contexts are highly specialised.
Some examples are as follows:
\begin{itemize}
    \item Using AutoML to predict medical outcomes for diabetes patients~\cite{koko19}. 
    \item Applying AutoML to the domain of traffic forecasting~\cite{antr18, anma20}.
    \item Testing the ability of AutoML to anticipate crash severity in Colombia~\cite{anma21}.
\end{itemize}
Admittedly, the core ML processes remain the same for many of these specialised applications, e.g.~model selection for supervised learning.
Thus, with the technological evolution of AutoML, it is clear why there is a gathering proliferation of publications testing AutoML in different domains.
The technical hurdles of ML have been lowered, providing a prime opportunity for enterprising stakeholders to exploit associated techniques in numerous fields.

Nonetheless, applying generalist AutoML packages to varying domains is not trivial.
There is undoubtedly differentiation that needs to be accounted for, the majority involving the preparation of data and the translation of model outcomes.
The former, in particular, is an intense time sink for a typical data scientist.
Perhaps this will eventually be ameliorated by the continuing evolution of general AutoML frameworks~\cite{kemu20}, but, in the meantime, developers of some AutoML software have vaulted ahead by specialising in a particular domain.
Arguably, this might indeed be an appropriate way forward.
Fine-tuning AutoML technology for the realities and demands of a specific context could avoid any software bloat that generality requires.
However, ancillary features aside, specialised tools are no more revolutionary than generalist AutoML.
Further discussion on this topic is reserved for Section~\ref{Sec:specialisedtools}.

\begin{figure}[h]
  \centering
  \includegraphics[width=1\linewidth]{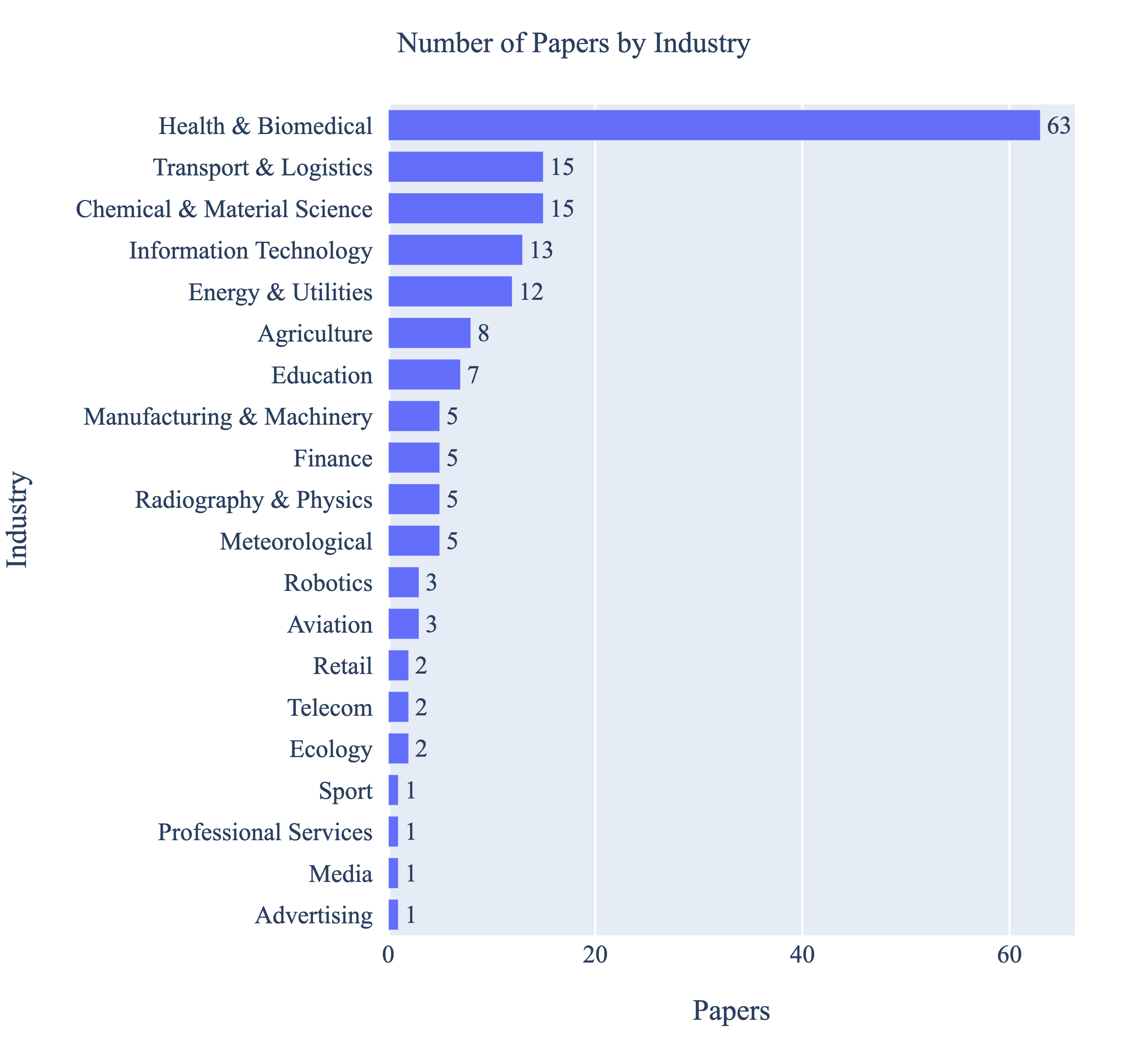}
  \caption{The number of academic publications reporting on AutoML applications associated with each major industry.}
  \label{Fig:papers_by_industry}
\end{figure}



Next, examining the major industries associated with AutoML applications renders the results in Fig.~\ref{Fig:papers_by_industry}.
They are essentially aggregations of the finer subfield breakdown, which was applied where possible, in Table~\ref{table:academic_applications_all_papers}.
Overwhelmingly, if 63 out of 169 academic publications are to be seen as a reflection of broader use, AutoML appears to be predominantly employed in the Health \& Biomedical sectors.
The second place is shared between Transport \& Logistics and Chemical \& Material Science, with 15 reports each.
Then comes Information Technology at 13 references, Energy \& Utilities at 12 references, and many other domains in the long tail of the distribution.

Here, it is tricky to identify why AutoML use has been concentrated in the way it has.
Chance likely plays some part, as there is little fundamental reason why the health sector was chosen as a backdrop for some of the earliest practical use of the technology.
Of course, once proof of principle is established, other interested stakeholders can better see how to use AutoML in such a context.
Perhaps the medical domain has simply had one of the most extended times since preliminary successes to cultivate a bandwagon effect. 
Nonetheless, in practice, some factors probably do weight the probability of a domain receiving attention from AutoML entrepreneurs.
For instance, there is the availability and quality of data, as well as the legal authorisation to use it and publish work around it.
It is also arguable that several of the best-represented sectors in a survey of \textit{academic} literature are themselves proximal to academia, i.e.~Health \& Biomedical, Chemical \& Material Science, and Information Technology.
The research, trials, investments and advancements in these fields are all better suited for scientific dissemination than ML problems in advertising or professional services, e.g.~law, accounting, consulting, etc.
Indeed, if a view of AutoML uptake is instead based on vendor reports, as discussed in Section~\ref{Sec:IndustrialApplications}, the top domains for applications are expected to be driven by commercial interests, e.g.~Finance, Retail, Marketing, and Insurance.
Simply put, perspective matters.

\begin{figure}[h]
  \centering
  \includegraphics[width=1\linewidth]{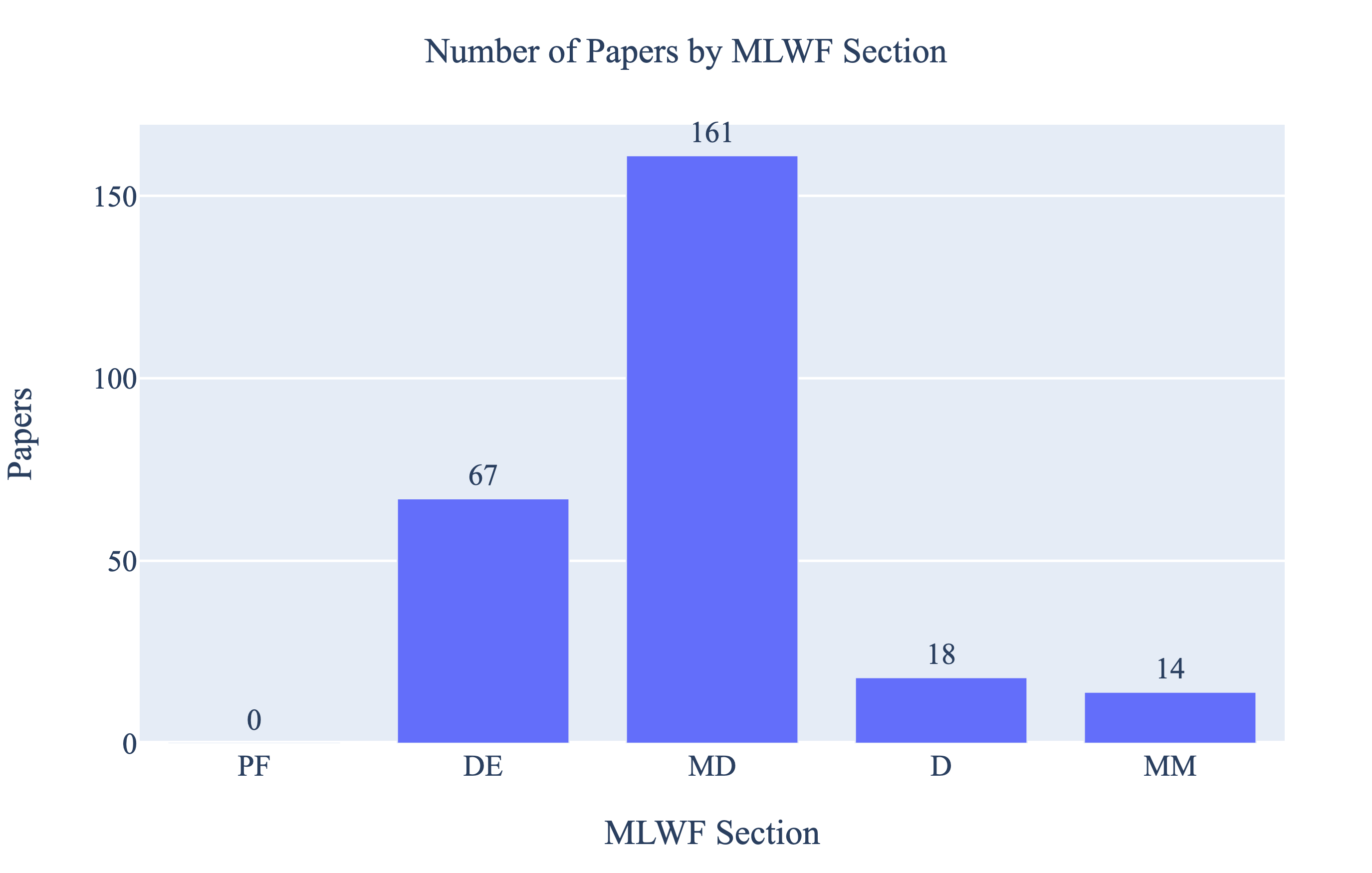}
  \caption{The number of academic publications reporting on AutoML applications associated with each stage of an MLWF. Abbreviations: PF for problem formulation \& context understanding, DE for data engineering, MD for model development, D for deployment, and MM for monitoring \& maintenance.}
  \label{Fig:papers_by_mlwf}
\end{figure}

Beyond domains, academic publications can also be categorised by the stages of an MLWF that are involved in their reported AutoML applications.
Accordingly, acknowledging that one article can be associated with multiple phases, the resulting distribution is depicted in Fig.~\ref{Fig:papers_by_mlwf}.
Two expected details immediately stand out.
Firstly, no academic report focusses on the preliminary stage of an MLWF; all applications begin with a well-defined and well-scoped problem.
Secondly, almost all applications involve model development, primarily HPO, as this is widely seen as the core offering of AutoML.

More informatively, 67 papers, around $40\%$ of the survey, discuss data engineering, i.e.~preprocessing and feature generation/selection.
Sometimes, this element appears as a direct by-product of an AutoML package, e.g.~TPOT.
In one particular case~\cite{bo21}, the data-engineering discussion is even the central focus, as the associated application involves assessing AutoFE.
Specifically, the report examines the comparative utility of features generated by human experts, Featuretools, and tsfresh.
In contrast, while data engineering receives a reasonable amount of attention, the last two stages of an MLWF are relatively ignored, with no more than about $10\%$ coverage.
This result may be because, as mentioned earlier, many applications reported in academia are seminal proof-of-principle works.
They are typically one-and-done projects, often used to harvest contextual insights rather than establish long-term productionised models.
That said, the automation of MLOps is also nascent, especially in terms of continuous learning, so time will tell whether the progression of these capabilities, as applied to real-world contexts, will receive commensurate academic interest.

\begin{figure}[h]
  \centering
  \includegraphics[width=1\linewidth]{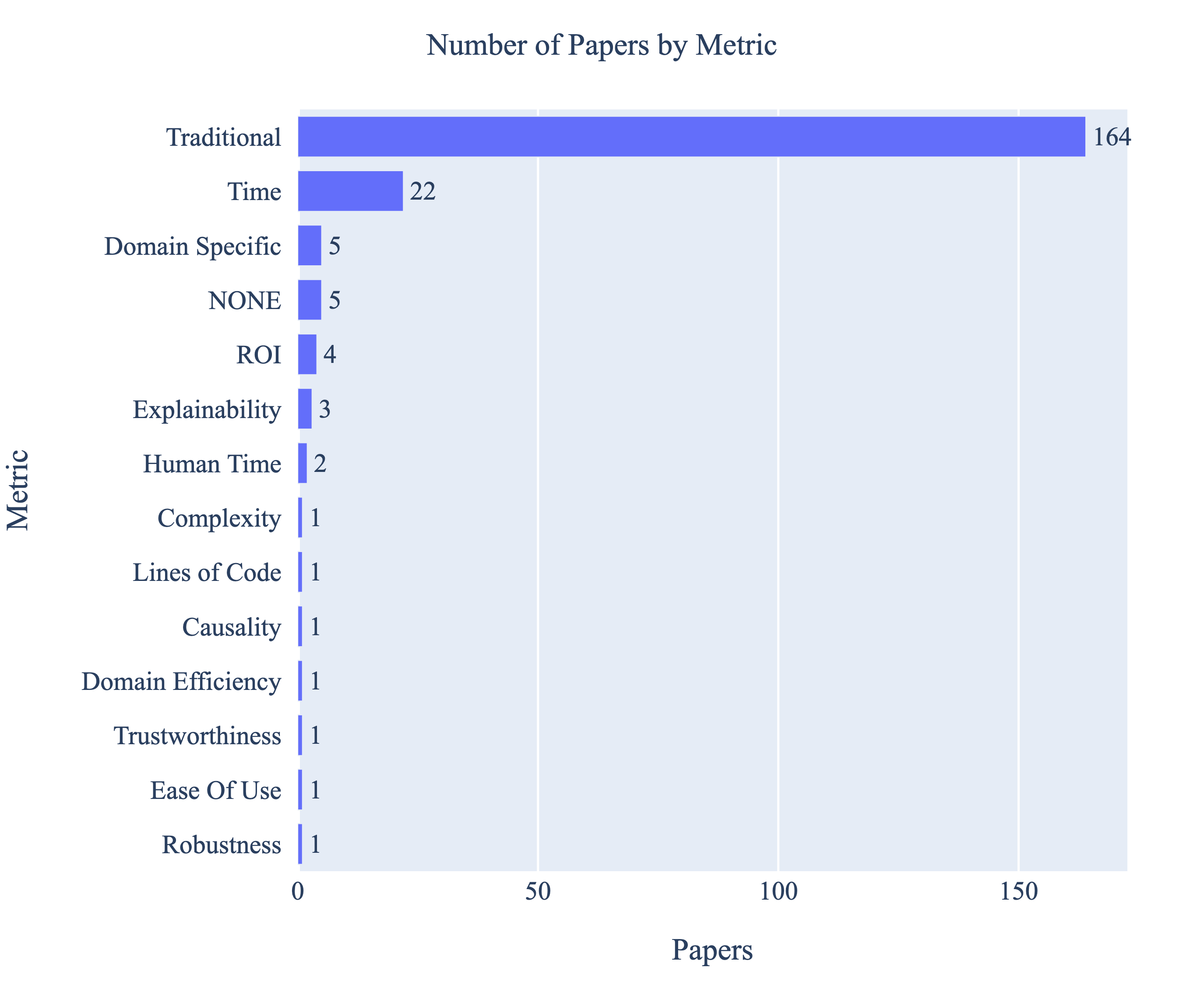}
  \caption{The number of academic publications reporting on AutoML applications associated with evaluation metrics of interest.}
  \label{Fig:papers_by_metric}
\end{figure}

At this point, it is worth recalling a central tenet to this monograph: performant ML, from the perspective of industry and broader society, relies on much more than model validity.
So, as part of the current application survey, academic publications have been dissected for the metrics they care about.
The resulting breakdown is shown in Fig.~\ref{Fig:papers_by_metric}, where the `traditional' term represents evaluations based on a confusion matrix, e.g.~accuracy, sensitivity, specificity, AUC, the $F_1$ score, etc.
Evidently, the academic lens is still firmly fixated on model error.
In fact, it is remarkable that computation `time', a metric for technical efficiency related to model training, is only considered by 22 papers, i.e.~$13\%$ of surveyed works.
This result suggests deficiencies in scientific thoroughness and rigour, even in the absence of loftier contemplations on the nature of ML performance.
In fairness, a few works do focus more on data preparation or deployment; many reports that do not name specific evaluation metrics fall in this category.
Nonetheless, the preoccupation with model validity in academia remains clear.

Now, even though only 23 reports, $13.6\%$ of the surveyed academic publications, consider non-traditional metrics beyond computation time, it is still instructive to see where the attention falls.
For instance, as a consequence of domain specialisation, several ML applications are judged by custom criteria related to context~\cite{chtr13, cobe17, chya20, shxi21, maar22}.
These projects include one that uses metrics associated with protein folding~\cite{cobe17} and another that evaluates advertising relevance~\cite{chya20}.
Then there are performance measures that lean towards ROI.
These are not always concretely quantified, such as the vague `business benefits' discussed by one publication~\cite{zhye20}.
Nonetheless, as exemplified by the expected benefit ratio (EBR) defined within another report~\cite{gezi22}, commercial metrics have a definite presence.
The only oddity is that this presence is so small, despite the volume of papers dedicated to exploring AutoML in applied industrial settings.
Again, perhaps discussions around business benefits are not considered of academic interest; any financial exploitation of uncovered contextual insights may occur subsequently to their reporting.

The remaining tail of the distribution in Fig.~\ref{Fig:papers_by_metric} gives tiny glimpses that suggest the criteria proposed in this monograph for performant ML are well founded.
Granted, they do also highlight the challenge of quantification, especially where subjectivity is involved.
For instance, the one application that explores ease of use does so by working with stakeholders to develop an associated score~\cite{flga21}.
Regardless, finding some form of representation for these facets of performant ML is still worthwhile.
Operational efficiency, for example, is captured by the notion of `human time', i.e.~how many hours it takes for a human to generate an ML solution with an AutoML tool.
Two of the surveyed academic publications record and tabulate this metric~\cite{payu19, maar22}.
One of the applications even goes further by assessing efficiencies based on lines of code~\cite{payu19}.

\begin{longtable}{|L{6cm}|L{7cm}|}
\caption{Possible justifications for using AutoML for an ML application.}
\label{table:justifications}
\\ \hline
\rowcolor[HTML]{B4C6E7} 
\multicolumn{1}{|c|}{\cellcolor[HTML]{B4C6E7}\textbf{Reason}} &
  \multicolumn{1}{c|}{\cellcolor[HTML]{B4C6E7}\textbf{Explanation}} \\ \hline\endhead
Computational   Cost                   & AutoML   reduces the computational cost of running/optimising ML   models.        \\ \hline
Data Prep                              & AutoML   can assist with data preparation, including preprocessing and   feature engineering. \\ \hline
Deployment   Considerations &
  AutoML   can assist with deployment and/or monitoring \& maintenance. \\ \hline
Domain (or   Existing Process) Utility & AutoML might be useful in a new domain or in comparison with an existing process.                   \\ \hline
Expertise   (Difficulty)               & AutoML   compensates for ML being difficult and requiring special   expertise.              \\ \hline
Explainability   of AutoML             & AutoML   can assist with enhancing the explainability of ML work.                       \\ \hline
Guesswork   (Human Error)              & AutoML   eliminates human error and guesswork involved in manual ML.         \\ \hline
Model   Simplicity                     & AutoML   creates simple models.                                                     \\ \hline
N - Justifying   ML                    & None. AutoML is used without justification, but the use of ML is justified.  \\ \hline
N - No   Justification                 & None. AutoML is used without justification.            \\ \hline
Open Source                            & A specific   AutoML tool is open-source.                                                   \\ \hline
Package is   SOTA (Popular)            & A specific   AutoML tool is the current state of the art (SOTA) or popular.         \\ \hline
Reference   Layperson (Democratisation) &
  AutoML   can be used by a layperson, helping to   democratise ML. \\ \hline
Reproducibility                        & AutoML can ensure ML work is reproducible.                                        \\ \hline
Running Many   Models                  & A specific   AutoML tool can run many models at once.                                        \\ \hline
Technical   Power (Accuracy) of AutoML &
  AutoML   provides superior technical power/performance, e.g.~in terms of model validity. \\ \hline
Time (Effort)                          & AutoML significantly reduces the time/effort a human must invest in ML.                           \\ \hline
\end{longtable}

\begin{figure}[h]
  \centering
  \includegraphics[width=1\linewidth]{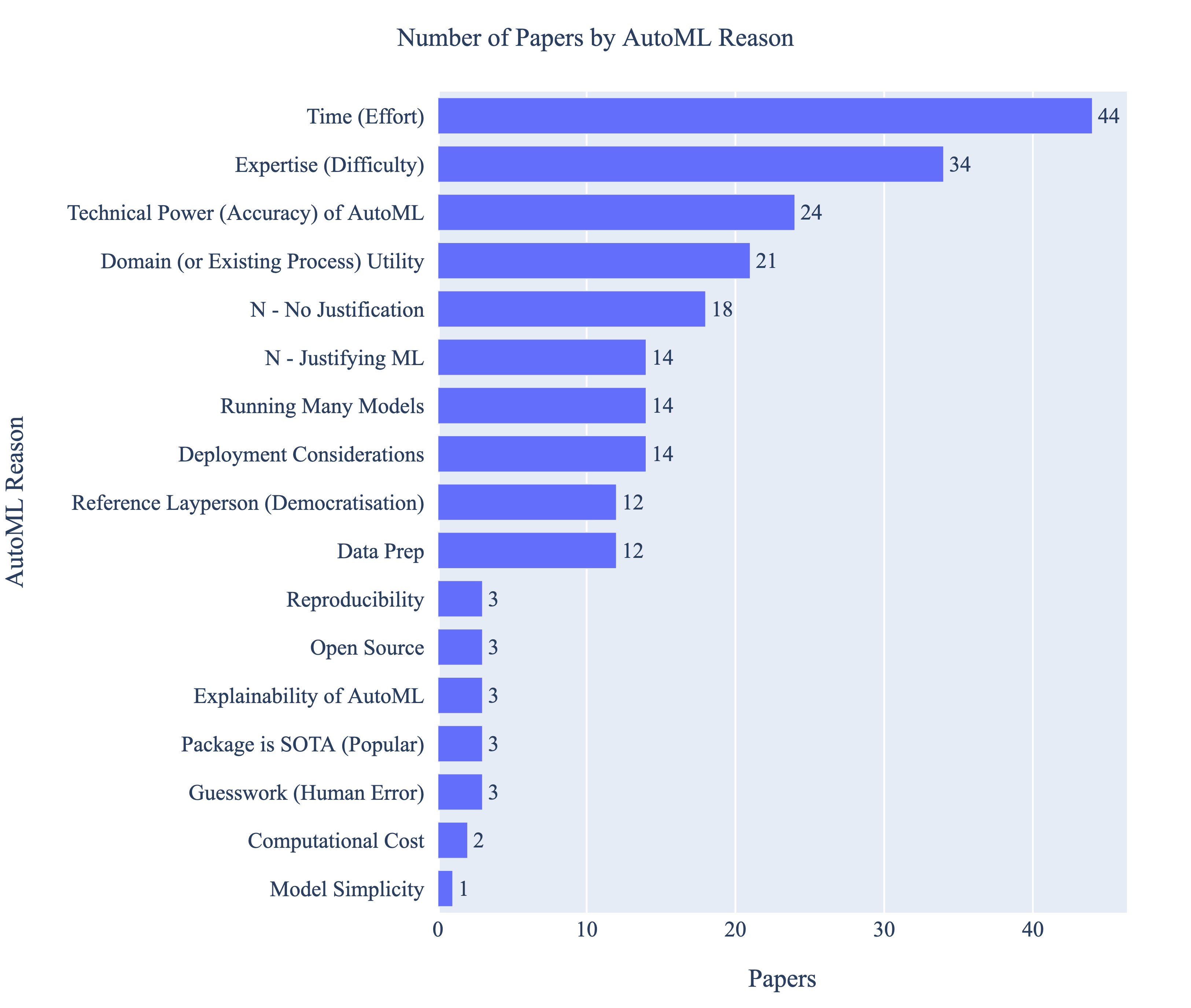}
  \caption{The number of academic publications reporting on AutoML applications associated with each justification for AutoML use.}
  \label{Fig:papers_by_reason}
\end{figure}

Admittedly, many of the exotic forms of evaluation metrics would perhaps be considered irrelevant if the primary motivation for using AutoML was enhanced predictive/prescriptive accuracy.
To better gauge how true this is, a list of possible reasons for why one might employ AutoML was curated, detailed in Table~\ref{table:justifications}.
The 169 academic reports were then dissected for any mention of the listed justifications, with the resulting counts of associations charted in Fig.~\ref{Fig:papers_by_reason}.
It is immediately apparent that, according to academia, the technical proficiency of AutoML is neither first nor second in the mind of an average stakeholder when considering its use.
Granted, the fact that it ranks third is still an interesting outcome, given that there is an ongoing debate about whether the outputs of AutoML, in terms of accuracy, significantly/consistently outperform ML models handcrafted by experts~\cite{lu19, habl20, zohu21}.
Nonetheless, nothing here supports the seeming obsession with model-validity metrics revealed by Fig.~\ref{Fig:papers_by_metric}.
One may thus contemplate the discrepancy: why are the benefits espoused about AutoML seemingly left unsubstantiated?
An optimist may infer that the community already holds these truths to be self-evident.
A cynic may question the sincerity of the assertions, perhaps seeing them exploited to elevate the importance of an otherwise rote ML publication.
Ultimately, the pragmatic truth may lie somewhere in the middle.
Of course, in fairness, every published AutoML application need not grapple with the challenge of quantifying the utility of the technology.
However, this monograph hopes to encourage a greater degree of concrete analysis to support/refute broader performance claims about AutoML.

Returning to the chart of justifications in Fig.~\ref{Fig:papers_by_reason}, the two most common assertions among academic publications posit that AutoML reduces the effort and expertise needed to employ ML.
Obviously, there is a caveat here.
Stakeholders who publish AutoML applications in academic journals are typically technically experienced individuals, and the tools involved often require code to be written.
Therefore, academically published judgements on the utility of AutoML may not be representative for lay users.
Indeed, one report talks explicitly in terms of ``the researcher''~\cite{camc21} and another promotes AutoML as a time-saving technology for data scientists~\cite{bero21}.
Nonetheless, various publications are more inclusive in how they perceive AutoML~\cite{vese21}, and 12 specifically espouse the democratisation angle.
A handful of the surveyed ML applications even discuss closely working with non-technical stakeholders~\cite{flga21, chli22}.
For instance, one acknowledges that selecting evaluation metrics is a technical process; it proposes personalising AutoML better by iteratively assisting users in making this choice~\cite{kuph20}.
Another example contemplates a similar stakeholder engagement process called `exploratory model analysis'~\cite{cahu19}.

Next up, the fourth most popular reason for AutoML use, following faith in its technical prowess, is that the technology may be able to crack unsolved problems in new domains and improve upon existing processes that have not previously involved ML.
Essentially, stakeholders that embrace this view are not primarily looking to push the limits of ML accuracy.
They instead seek to pursue interdisciplinary endeavours, forging fruitful connections between methodology and context that provide `good enough' preliminary outcomes.
Certainly, when the barrier-to-use for ML techniques and approaches is reduced, interested researchers in other fields are freed up to focus on domain insights and otherwise accelerate their paper-publishing workflow.

As for the remaining justifications, several skew more towards explicit functionality rather than high-level motivations.
For instance, one AutoML application is connected with the claim that automated processes can produce simpler ML models than a human~\cite{prra22}.
This argument is somewhat contentious and depends very much on both implementation and problem context.
What is much more uncontroversially appreciated is that AutoML software allows many possible models to be sampled, e.g.~Auto-WEKA~\cite{sisa16, cobe17, brst18, ngsa18, tile19, damu20, adka21a, deme21, qomo21} and PyCaret~\cite{hefi21, kopa21, lano21}.
In fact, for some AutoML applications, even the baseline offerings are seemingly insufficient for CASH; one project supplements Auto-WEKA with additional ML algorithms~\cite{alga17}.
Finally, it is worth noting that a moderate amount of papers do not justify the use of AutoML, with a large portion of these not even bothering to rationalise employing ML for a particular domain.
This result is interesting because it suggests AutoML may have already generated a reasonably authoritative reputation among prospective users.
Unfortunately, assessing whether this reputation is warranted would require much deeper post hoc analyses of ML application outcomes as perceived by their stakeholders.

To some extent, a more balanced appraisal of AutoML usage is at least possible by noting any challenges reported while employing the technology.
For instance, a couple of publications identify issues arising from small datasets~\cite{payu19, zhye20}.
Of course, these complications are less likely to be problematic within larger enterprise environments, where the difficulty lies in extracting value from big data, not accumulating it.
However, the more pertinent point here is that sparse data is a challenge for ML overall, not just AutoML.
Presumably, the takeaway is that automation is not a magic solution for limitations \textit{intrinsic} to an ML problem.
Elsewhere, a couple of ML applications dwell on the inability to customise certain tools for domain-specific metrics and nuances~\cite{ormo17, homa19}.
This matter seems to be a common concern, which is, again, why there has been a movement by some AutoML developers to cater for domain specialisation; see Section~\ref{Sec:specialisedtools}.

Occasionally, aligning closely with the interests of this review, some academic reports on AutoML applications have provided deeper commentary about potential obstacles to AutoML uptake.
For example, a few papers consider a lack of explainability as a possible hindrance to broader user engagement~\cite{megr17, zhge18}.
This recognition supplements current industry concerns and stakeholder requirements, legitimising the inclusion of the associated criteria for performant ML within Section~\ref{Sec:summarisedcriteria}.
However, one publication stands out in particular for its fine-grained focus on small-to-medium enterprise (SME) and its uptake of AutoML technology~\cite{badi20}.
The report cites a lack of expertise and funding to hire data-science talent -- this implicitly rejects AutoML being sufficiently democratised -- as obvious obstacles, but it also asserts that identifying business use cases for the technology may also be challenging.
Such a concern is a rare and thus notable reference to the initial stage of an MLWF: problem formulation and context understanding.
Until AutonoML becomes exceptionally advanced~\cite{khke21}, supporting the translation of business objectives into ML tasks will need to be addressed in other ways.

Ultimately, a survey of AutoML applications published within academic literature reveals that the utilisation phase of the technology has begun in earnest.
Many industries are proving fruitful for automated exploration, although, given this academic perspective, there is an obvious selection bias for observability that leans towards the research-heavy fields.
In any case, this surge has only occurred within the last few years, so, while the proliferation is noteworthy, AutoML technology is far from being exploited to the limits of its capability.
Partially, these constraints are governed by the software available at any particular time, which explains why few applications go far beyond the model-selection stage of an MLWF.
However, among potential users, there is also perhaps a limited understanding of what AutoML truly promises regarding its roles and benefits.
Indeed, this suggestion is most starkly supported by a disproportionate obsession with model-validity metrics that does not reflect what the application runners \textit{themselves} claim to be the advantages of AutoML.
Thus, the literature survey results end up serving as the perfect justification for bringing communal attention to this monograph and its more comprehensive definition of `performant ML'.

\subsection{Domain Specialisation}
\label{Sec:specialisedtools}

In the course of reviewing AutoML software and applications for Section~\ref{Sec:Tools} and Section~\ref{Sec:AcademicApplications}, respectively, it became apparent that several tools have been designed for highly specialised use cases.
This statement does not refer to the packages discussed in Section~\ref{Sec:ancillarytools}, which are specialised only in that they are `dedicated' to particular tasks and stages within an MLWF.
For instance, software that focusses on HPO in Section~\ref{Sec:purehpotools} is still general, developed to optimise -- ideally -- arbitrary functions, so long as they are both parameterisable and evaluable.
This section, instead, examines the specialisation of AutoML tools/applications for particular technical problems and industrial domains.


\begin{table}[h]
\caption{Specialised AutoML Tools. Notes the technical/industrial domain and whether the following mechanisms are included: Exp. for explainability, Viz. for visualisation, DP for data preparation, FG for feature generation, and FS for feature selection. Also details UI modes, HPO mechanisms, and library dependencies.}
\label{table:specialisedtools}
\resizebox{\textwidth}{!}{%
\begin{tabular}{|L{1.5cm}|L{1.9cm}|c|c|l|c|c|c|L{1.5cm}|L{2.5cm}|l}
\cline{1-11} \hline
\rowcolor[HTML]{B4C6E7} 
\multicolumn{1}{|c|}{\textbf{Name}} &
  \multicolumn{1}{c|}{\textbf{Domain}} &
  \multicolumn{1}{c|}{\textbf{Exp.}} &
  \multicolumn{1}{c|}{\textbf{Viz.}} &
  \multicolumn{1}{c|}{\textbf{UI}} &
  \multicolumn{1}{c|}{\textbf{DP}} &
  \multicolumn{1}{c|}{\textbf{FG}} &
  \multicolumn{1}{c|}{\textbf{FS}} &
  \multicolumn{1}{c|}{\textbf{HPO}} &
  \multicolumn{1}{c|}{\textbf{Wraps}} &
  \multicolumn{1}{c|}{\textbf{Ref.}} \\ \hline
Auto\_TS &
  Time   Series &
  {{N}} &
  {N} &
  {Code} &
  {Y} &
  {Y} &
  {N} &
 NONE &
  {[}Algorithms{]} Sklearn, FB Prophet, XGBoost, pmdarima &
  \multicolumn{1}{l|}{\cite{AutoTS}} \\ \hline
Luminaire &
  Outlier   (Time Series) &
  {{N}} &
  {Y} &
  {Code} &
  {Y} &
  {N} &
  {N} &
  Bayesian &
  {[}HPO{]}   Hyperopt &
  \multicolumn{1}{l|}{\cite{luminaire,chsh20}} \\ \hline
TODS &
  Outlier   (Time Series) &
  {{N}} &
  {N} &
  {Code} &
  {Y} &
  {Y} &
  {N} &
 NONE &
  {[}Algorithms{]} Sklearn, Tensorflow (TF), Keras, PyOD &
  \multicolumn{1}{l|}{\cite{tods, lazh20}} \\ \hline
AlphaPy &
  Finance,   Sport &
  {N} &
  {Y} &
  {Code/CLI} &
  {Y} &
  {N} &
  {Y} &
  Grid,   Random &
  {[}Algorithms{]} Sklearn, Keras, XGBoost, LightGBM, CatBoost &
  \multicolumn{1}{l|}{\cite{alphapy}} \\ \hline
Cardea &
  Medical   Documents &
  {{N}} &
  {N} &
  {Code} &
  {Y} &
  {Y} &
  {N} &
 NONE &
  {[}Other{]}  Compose, Featuretools &
  \multicolumn{1}{l|}{\cite{Cardea, alal20}} \\ \hline
ChemML &
  Chemistry &
  {{N}} &
  {N} &
  {Code/GUI} &
  {N} &
  {N} &
  {N} &
  Active   Learning, Genetic &
 NONE &
  \multicolumn{1}{l|}{\cite{chemml, havi19}} \\ \hline
EvalML &
  Fraud,   Lead Scoring &
  {{N}} &
  {Y} &
  {Code} &
  {Y} &
  {N} &
  {N} &
  Bayesian,   Grid, Random &
  {[}Algorithms{]} CatBoost, Sklearn, LightGBM, XGBoost, {[}HPO{]} skopt &
  \multicolumn{1}{l|}{\cite{evalml}} \\ \hline
\end{tabular}
}
\end{table}

Accordingly, the first lens through which to examine specialisation considers niche forms of data analysis.
For instance, there is the objective of outlier detection.
Certain AutoML tools, some discussed in Section~\ref{Sec:OpenDirtyData}, identify/manage obvious anomalies on the path to constructing better-fitting ML solutions, but this is not what is being referred to here.
Instead, we highlight that some ML applications and their models are entirely dedicated to predicting unusual patterns of behaviour within select data environments.
Academic examples of these applications have examined financial fraud~\cite{nghu11}, medical information about patients~\cite{haba13}, and issues with machine maintenance~\cite{kasa17}.
Of course, there are many techniques that one may leverage in such a use case, with the Python library PyOD~\cite{zhna19} implementing 33 algorithms from various sources at the time of this review.
However, the tool itself is not considered AutoML, as substantial manual effort is needed to employ its offerings.
In contrast, Luminaire~\cite{luminaire} and TODS~\cite{tods}, both dissected in Table~\ref{table:specialisedtools}, offer outlier-detection capabilities for time series data with significant automation, e.g.~in terms of feature extraction and running/optimising algorithms.
Both packages use scikit-learn, with the former depending on Hyperopt and the latter leveraging Keras, TF, and PyOD.

Incidentally, from a broader perspective, working exclusively with time-series data is itself a form of analytical specialisation.
The Auto\_ts package~\cite{AutoTS}, also in the table, is one example that does so, automating many related processes.
It supports users in generating ML solutions based on a limited but diverse pool of techniques, e.g.~a random forest supplied by scikit-learn, an autoregressive integrated moving average (ARIMA)~\cite{boje15}, or the Facebook Prophet~\cite{tale17}.

Continuing onwards, the second lens through which to examine specialisation considers developments for particular industrial domains.
Now, crucially, configuring for a specific problem \textit{setting} typically has less of an impact on core ML processes than configuring for a specific problem \textit{type}.
For instance, in the AutoDL case, planning for time-series forecasting or image recognition tasks may suggest limiting CASH, specifically NAS, to recurrent or convolutional neural networks.
In contrast, CASH designed for supervised ML on tabular data does not significantly change whether the data is financial, medical, and so on.
Admittedly, if a problem type is virtually intrinsic to a problem setting, e.g.~image recognition and the field of radiography, then the model-development phase of an MLWF can be configured appropriately in advance.
However, in most cases, specialisation for an industrial domain predominantly impacts the input-output edges of an ML application, i.e.~how data is prepared/engineered and how results are used/interpreted.
In other words, convenience features are developed for the transition boundary between domain insights and computational data.

One of the earliest examples of AutoML software specialised in such a way is PredicT-ML~\cite{lu16}, designed to perform ML on meaningful features that it extracts from big clinical data.
Unfortunately, it cannot be robustly assessed beyond its associated publication, lacking either a public repository or any extensive follow-up.
However, its existence does align well with the trend identified in Section~\ref{Sec:AcademicApplications}, i.e.~the heavy use of AutoML in the health \& biomedical domains.
Another more recent example also adhering to this trend is GenoML~\cite{male21}, designed explicitly for genomics, although it is too early to determine whether it will find significant purchase in academia.

Arguably exhibiting more uptake is ChemML~\cite{viha19}, an open-source tool that has been cited in several chemistry investigations and applications~\cite{hash19, pa20a}.
It can manipulate chemical data conveniently, e.g.~by encoding and visualising molecules, and supports automated pathways to generating ML models based on relevant features.
A similar package designed to work with RNA/DNA data is PyFeat~\cite{muah19}, which has likewise seen decent use~\cite{amra20, lvwa20, ahmu21, zhwa21}.
It first constructs features from RNA/DNA sequences according to domain-specific options that a user can select, then optionally runs ML classifiers on the associated feature-enriched dataset.
That said, the package is not particularly current in developer activity and only offers a CLI as an advanced UI, with which arguments are specified.

Here, the definition of AutoML already starts to blur, as the scope of ChemML and PyFeat is dictated more by the domain than by the practices of automating high-level ML operations.
They can still solidly be classified as AutoFE packages, at the very least, but the priorities involved further colour the discussion in Section~\ref{Sec:CommercialDiscussion} on the `AutoML agenda'.
An interesting question thus arises: in an age of closely integrated software, how rigid will the bounds on AutoML remain?
For instance, Cardea~\cite{Cardea} is an open-source tool designed to work with electronic health records.
It does not intrinsically provide AutoML capability but does so holistically by wrapping other packages such as Featuretools, Compose, and MLBlocks.
This software contrasts with another implementation in a similar space, i.e.~AutoPrognosis~\cite{AutoPrognosis, alsc18}, which, although it has not been tended to in a while, was explicitly developed to construct ML pipelines for clinical prognoses automatically.
The point here is that some packages broadly dedicated to data analytics for a specific domain have begun to integrate, either optionally or immediately, automated high-level ML processes.

The inverse, then, to a domain-specific package that includes AutoML functions is an AutoML package that includes domain-specific functions.
Both are valid manifestations of specialisation, even if each prioritises one characteristic over the other.
Accordingly, AlphaPy~\cite{alphapy} is an example of this inverse, offering generic AutoML functionality at its base and wrapping around popular ML libraries such as XGBoost, LightGBM, CatBoost, and scikit-learn.
What makes the software stand out is its supplementary `MarketFlow' and `SportFlow' pipelines, which are geared for financial analysis and sporting-event predictions, respectively.
Interestingly, the AlphaPy framework seemingly sticks to grid/random search for its optimisations, which suggests the industry-proximal project is either (1) unaware of theoretical advances in AutoML, (2) incapable of implementing them, or (3) dismissive of their utility.
Given finite developmental resources, the last option could be well justified if leveraging domain insights has more impact than employing SOTA techniques.
Elsewhere, EvalML~\cite{evalml} similarly provides general AutoML functions, this time including Bayesian HPO via wrapping up the Scikit-Optimize library~\cite{scikitoptimize}.
However, it too offers specially designed capabilities for industrial applications, i.e.~fraud detection and lead scoring.
These two use cases happen to be among the most recurrent according to vendor perspectives of AutoML; see Section~\ref{Sec:IndustrialApplications}.
So, for the many stakeholders solely interested in these particular domains, EvalML likely holds a unique appeal, allowing users to avoid the brunt of system configuration and data formulation.

\begin{figure}[h]
  \centering
  \includegraphics[width=1\linewidth]{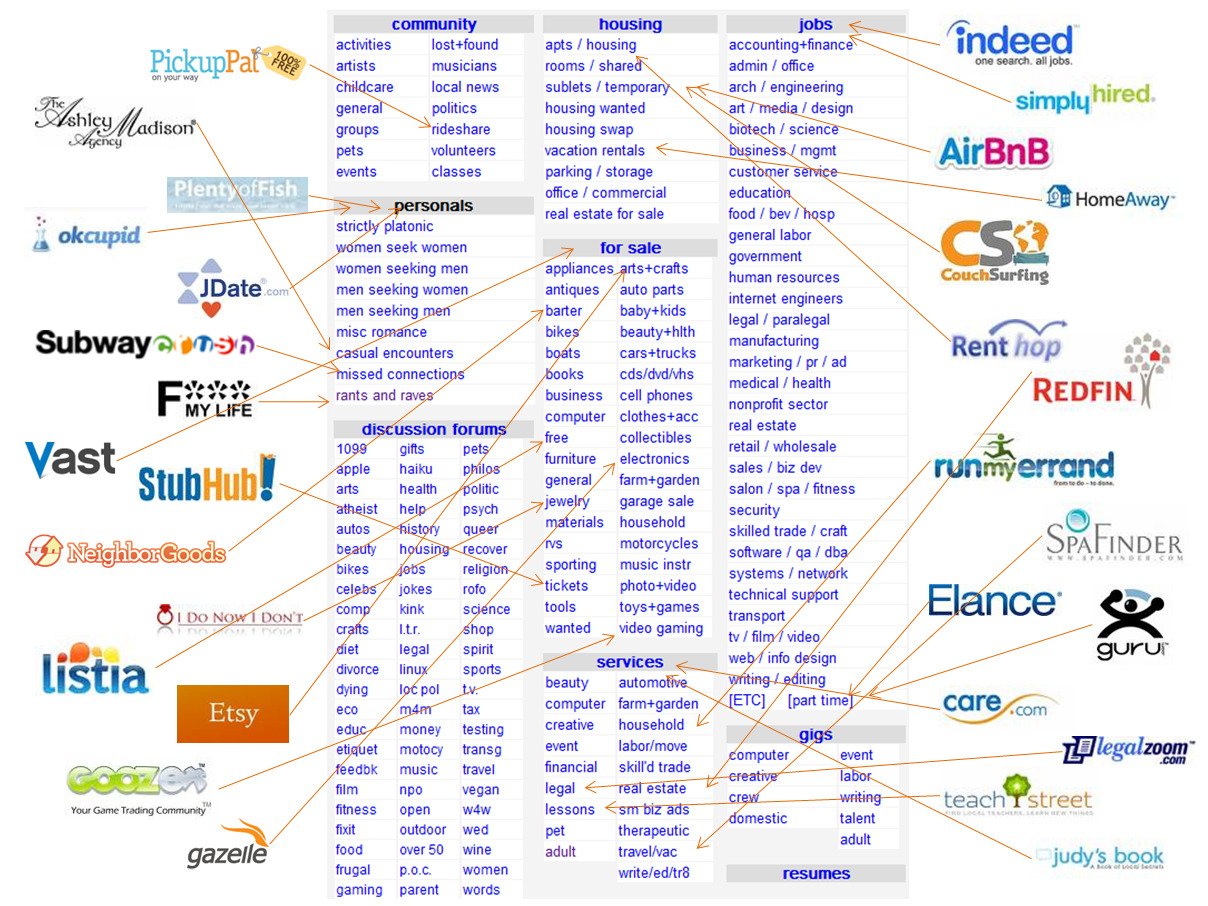}
  \caption{A well-known diagram~\cite{pa10} linking startups to popular niches on Craigslist, which acts as an analogy to how AutoML developers may similarly pursue vertical growth strategies.}
  \label{Fig:craigliststartups}
\end{figure}

Ultimately, domain specialisation is still a nascent trend, even given the technological evolution of AutoML and its overall recency.
After all, there is no point in investing substantial resources into fine-tuning software for a particular purpose/industry until the associated demand profile is evident.
Accordingly, this phenomenon is closely tied with, if not reactive to, the body of AutoML applications that arises.
Nonetheless, considering the speculation in Section~\ref{Sec:CommercialDiscussion} that, fragmented agendas aside, the general AutoML marketplace might eventually become an oligopoly, domain specialisation is one possible pathway to accommodate ongoing competition and differentiation.
Indeed, analogised by Fig.~\ref{Fig:craigliststartups} to how many startups grow vertically from existing niches, the future of AutoML technology may heavily involve prospective developers catering to specific tasks and industries.
Such a process would undoubtedly contribute to the overall aims of the AutoML endeavour.
For instance, consider a sports analyst lacking an understanding of general ML, let alone the skills to run an application without falling into common traps, e.g.~data leakage~\cite{sach20}.
While generic AutoML software might suffice for advancing their analytical ambitions, a closely assistive tool like AlphaPy could prove much less daunting, accelerating overall industrial uptake of the technology.
Essentially, domain specialisation promises to further weaken the barriers to ML democratisation.
Of course, only time will tell whether enterprising developers have \textit{both} the domain expertise and academic skills to capably satisfy any emerging niche demand.

\subsection{Vendor Reports}
\label{Sec:IndustrialApplications}

While academic literature tends to be transparent and peer-reviewed, it biases observability around AutoML applications towards those with apparent scientific benefits.
However, a core tenet of the AutoML endeavour is that ML should ideally be made accessible wherever it can be found useful, including as part of far more mundane decision-making processes.
So, short of exhaustively surveying ML stakeholders, the best way to assess the broader uptake of AutoML technology is to see what vendors have to say about it.
Of course, all the usual caveats of taking commercial commentary at face value apply.
There is no reliable window into sales volume or industrial activity; case studies and portfolios of services may be aspirational rather than representative.
Nonetheless, this section collates promotional material from numerous vendors to gain at least a surface-level insight into the industries/problems the commercial AutoML sector targets.

\begin{figure}[h]
  \centering
  \includegraphics[width=1\linewidth]{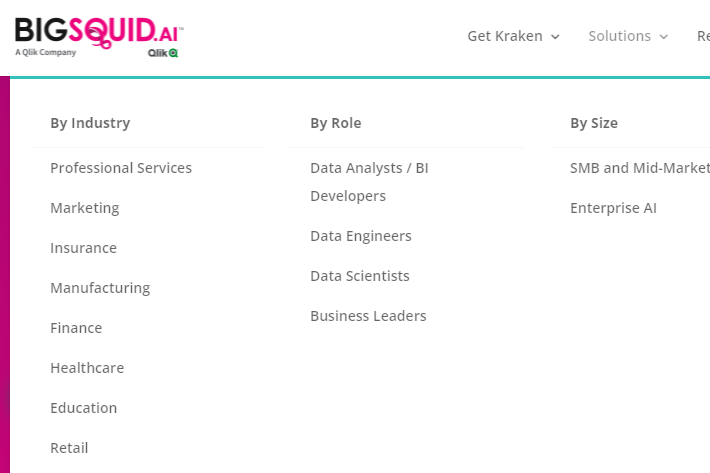}
  \caption{A snapshot of a commercial AutoML website, which lists industries targeted by its services. The vendor associated with this example is BigSquid.}
  \label{Fig:bigsquidindustries}
\end{figure}

Now, many AutoML vendors tend to use websites to list the domains in which they prioritise the targeting of their services.
One example is shown in Fig.~\ref{Fig:bigsquidindustries}.
Such summaries are often the result of two calculations: where a particular implementation is likely to have the most significant impact and, simply put, where the money is.
Granted, while commercial income is a decent proxy for demand, other nuances can drive commercial agendas, e.g.~maintaining a few well-paying clients in one industry may be more attractive than cheaply servicing numerous end-users in another.
Regardless, setting aside contemplations of business strategy, we review 23 vendor websites and categorise the domains they target.
Listed terms are occasionally split, combined or rephrased to sensibly fit a common framework.
For instance, `Banking \& Insurance' targeted by Einblick translates to one tick for two categories, i.e.~`Finance' and `Insurance'.

\begin{figure}[h]
  \centering
  \includegraphics[width=1\linewidth]{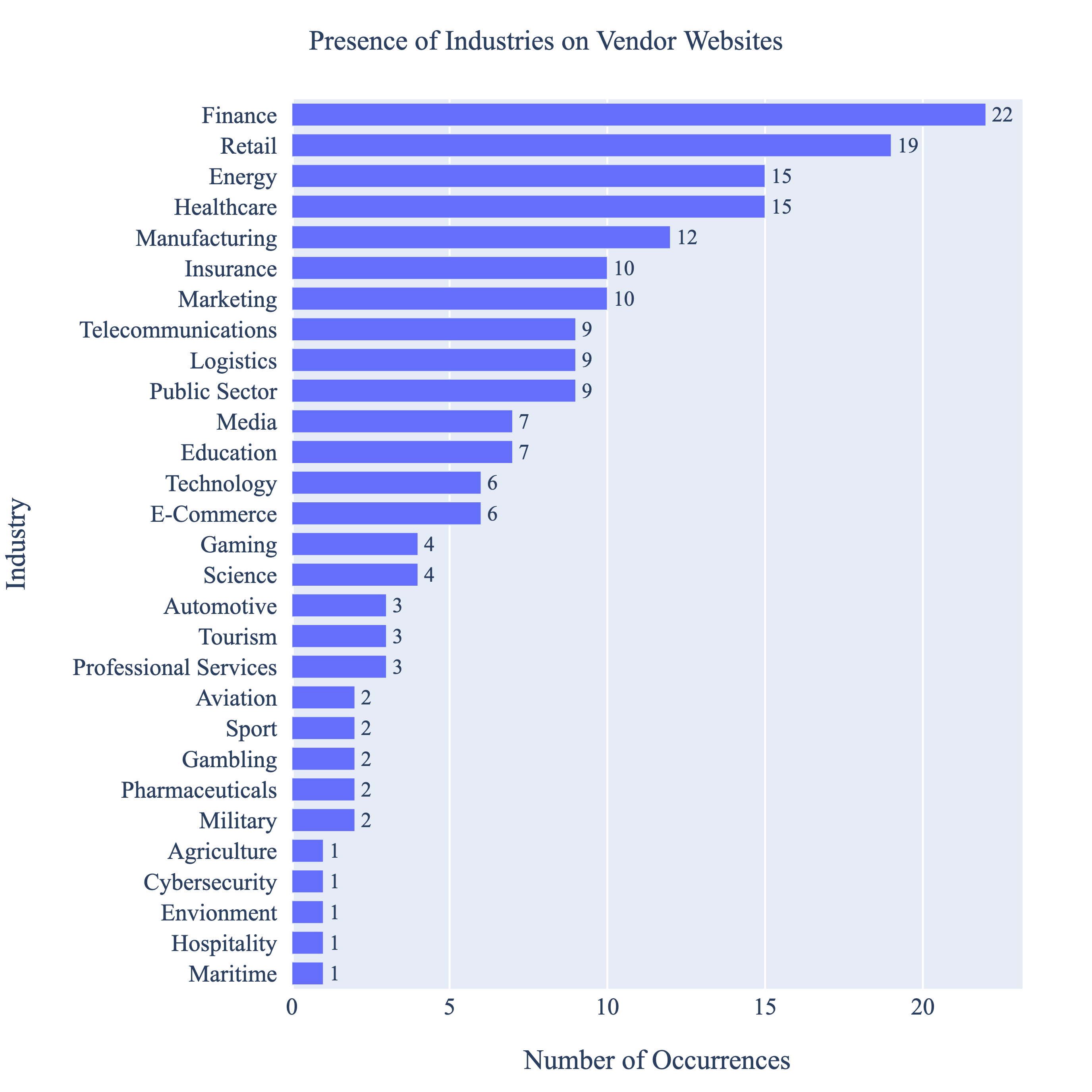}
  \caption{Major industries suitable for ML. Each one is associated with the number of commercial AutoML websites that target the industry. This survey covers 23 vendors.}
  \label{Fig:vendorindustries}
\end{figure}

The resulting distribution of the commercial focus on industries is depicted in Fig.~\ref{Fig:vendorindustries}.
A long tail highlights the general utility of AutoML, but some domains receive higher levels of attention than others.
There are numerous reasons why this may be the case, including a prevalence of big data and client funds.
It is also possible that the types of ML problems arising in these targeted domains conform to well-defined archetypes, thus requiring less effort in the first MLWF phase, i.e.~problem formulation and context understanding.
Accordingly, the following sectors are well represented: finance, retail, marketing, and insurance.
The ML model outcomes in these industries also tend to be proxies for profit, such as the volume of sales, so the value of AutoML can be quantified relatively quickly and objectively.
Healthcare and energy also receive serious attention, as these are typically considered social goods and thus are often backed by government resources, either directly or via policies and less tangible benefits.
For the same reason, more than a third of surveyed vendors seek to offer AutoML services to the public sector.
In the meantime, privacy concerns and national defence considerations may stifle activity within some of the lower-ranked sectors, e.g.~military, aviation, maritime, and cybersecurity.
However, associated contracts may simply be less publicised.

The final analysis in this monograph involves looking into case studies promoted by AutoML vendors.
These reports are typically accessible via associated websites, e.g.~within sections named `blog' or `resources'.
Of course, due to the lack of peer review and often both depth and detail, only a glancing commentary can be made.
Moreover, not all vendor publications are relevant or useful to this survey.
For instance, partnership announcements are occasionally lumped in with the case studies.
Alternatively, reports may be written from the perspective of someone `experimenting' with the commercial AutoML product.
Even among the remainder, case studies that do not clearly detail a business problem serve no purpose for a rigorous review.

\begin{figure}[h]
  \centering
  \includegraphics[width=0.9\linewidth]{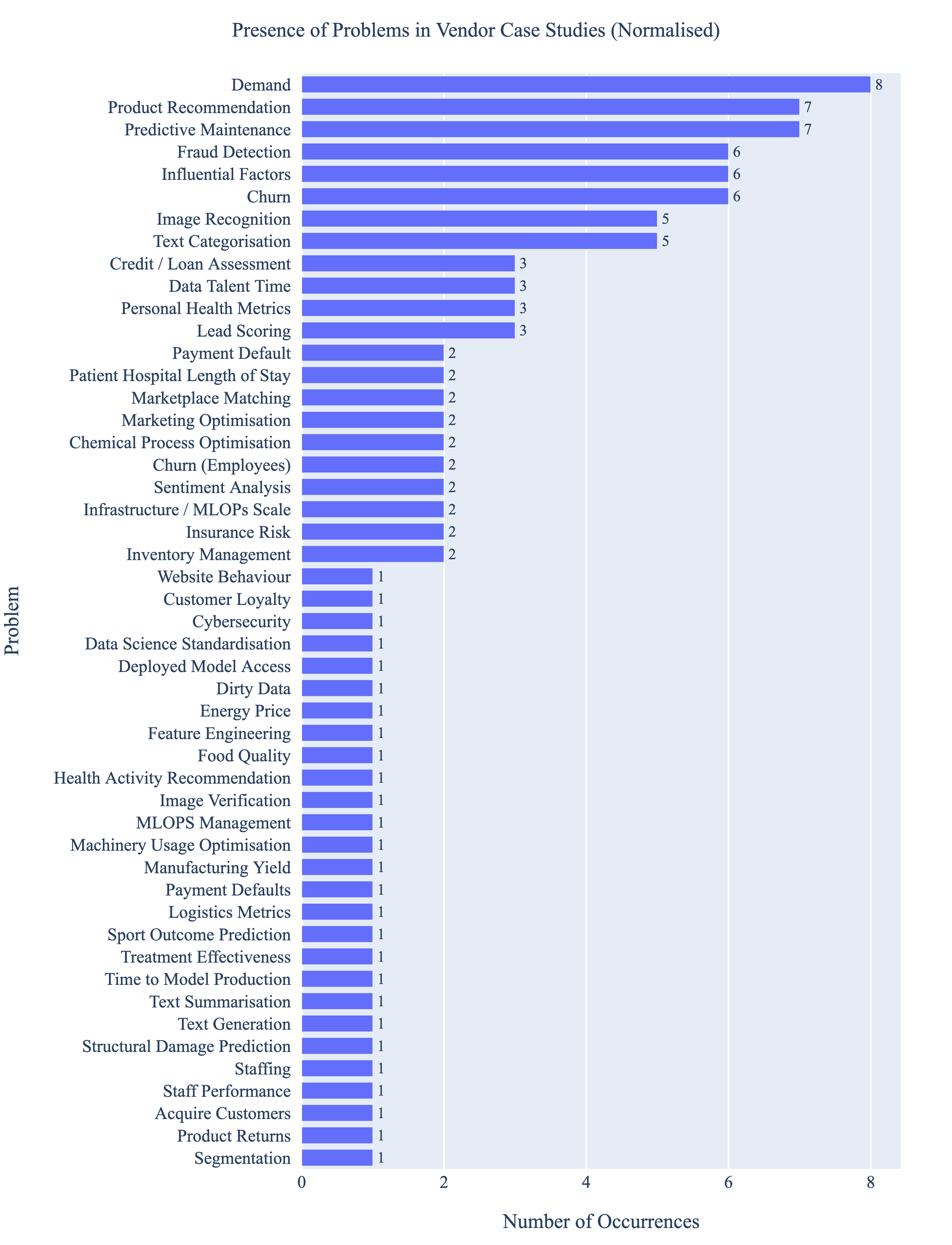}
  \caption{Problem types suitable for ML. Each one is associated with the number of commercial AutoML websites that dedicate at least one case study to the problem. This survey covers 20 vendors.}
  \label{Fig:vendorcontentproblemsnorm}
\end{figure}

Thus, the end result is a survey of 20 vendors and 130 documents of associated content.
While attempts were made to cover the same ground as in Section~\ref{Sec:AcademicApplications}, a lack of published information hampers an identical analysis.
This outcome is not unexpected.
For instance, vendors are unlikely to discuss the challenges of using AutoML while spruiking their products.
Ultimately, the only significant insights that can be extracted from this study relate to the problem types targeted by AutoML vendors, displayed in Fig.~\ref{Fig:vendorcontentproblemsnorm}.
As a side note, given that some vendors dedicate multiple reports to a single use case, the charted counts are normalised.
Accordingly, while product recommendation would otherwise top the rankings with 16 case studies, only seven vendors advertise this use case.

Unsurprisingly, the long tail on the chart indicates that AutoML is deemed useful for a diverse array of insight-gathering and decision-making problems.
Certainly, with supervised learning as the default form of ML, classification/regression tasks can be formulated within numerous industrial contexts.
Of course, as with the targeted industries, particular problems see the most \textit{apparent} use of AutoML.
These ML applications include demand forecasting for products/services, predictive maintenance, and product recommendation to new and existing customers, i.e.~cross-selling and up-selling.
Such problem types may be well-suited to AutoML, as they are often accompanied by plenty of labelled data and a clear target/objective.

However, it is notable that the term `influential factors' also ranks very high; associated ML tasks involve not just making predictions but also understanding what is \textit{driving} those predictions.
Admittedly, it is unclear why so many case studies focus on this problem type, i.e.~whether vendors are preparing this promotional material based on the features they have developed or on the demand they expect.
Nevertheless, the point here is that traditional technical performance is not the be-all and end-all for such ML tasks, and other factors, including explainability, necessarily influence successful outcomes.
In effect, commercial entrepreneurs are exploring just how versatile AutoML can be.
Thus, for those in the broader AutoML research community, it is a timely opportunity to consider what it means for the technology to be `performant' in all the use cases listed within Fig.~\ref{Fig:vendorcontentproblemsnorm}.
Such considerations and associated innovations may strongly influence future technological uptake and impact.

\section{Critical Discussion and Future Directions}
\label{Sec:discussion}

This comprehensive review of AutoML as a technology has comprised multiple analyses representing various angles of attack, all grappling with several fundamental questions.
Simply put, as of the early 2020s, what does AutoML technology look like?
How is it being offered?
How is it being used?
And is it any good?
In fact, this last question of `goodness' motivated substantial contemplation in Section~\ref{Sec:Motivation}, urging AutoML researchers to look beyond standard technical performance and consider the bigger picture, i.e.~the potential requirements of all possible stakeholders in an ML application.
Accordingly, equipped with a refined perspective, this monograph has made seminally broad assessments of supply, demand, and service quality.
Specifically, surveys of open-source/commercial software and academic/vendor reports have all aided in appraising industrial engagement with AutoML, both present and potential.
Nonetheless, with so much content, it is worth summarising/discussing the most prominent insights extracted from the analyses.

\textbf{Almost no one is touching automated problem formulation and context understanding.}
As a reminder, such an assessment concerns the first stage of an MLWF, i.e.~the conceptual framework introduced in both Section~\ref{Sec:mlwf} and Fig.~\ref{Fig:mlworkflowsubtasks} to encompass a diversity of high-level ML operations within one inclusive workflow.
Accordingly, it makes sense here to structure some of the overarching commentaries according to the main phases of an MLWF.
However, there is little to say about the initial stage, when business plans transform into a machine-interpretable problem.
Indeed, sophisticated mechanisms do not exist within the commercial sphere, despite several vendors offering templates and code repositories for common use cases.
The situation is not much different among open-source packages either, with only a couple trying NLP and adjacent methodologies to interpret user desires efficiently~\cite{libra, otto}.
In effect, virtually all AutoML software starts with the premise that stakeholders have already created a well-defined project plan, for which resourcing is approved and project controls have been set up.
This assumption is a poor one in reality, especially as organisations new to ML will likely struggle to translate their business objectives into well-scoped projects~\cite{badi20}. 
Admittedly, automating the formulation phase is a matter of HCI and may additionally require an AutoML paradigm shift~\cite{khke21}, e.g.~the design of reasoning mechanisms.
After all, business agendas are only rigidly limited by human creativity, so mapping an infinite space of real-world problems into concrete ML processes very much remains an open research question.

\textbf{Automated data engineering is reasonably well supported, but there is room for improvement.}
This second phase of an MLWF is the most closely linked to model development, given that data transformations can themselves be bundled into an ML pipeline or a deep neural network, so it makes logical sense that, historically, it was the first to be enveloped by the expanding scope of AutoML.
Unsurprisingly, at the current time, there are several tools dedicated solely to data engineering, as listed in Section~\ref{Sec:datafeaturetools}.
Likewise, the majority of surveyed `comprehensive' systems, whether open-source or commercial, provide assistance/automation with basic facets of cleaning or preparation.
However, the coverage is incomplete, and packages that, for instance, do not manipulate date-time stamps or tokenise text will struggle to construct the most information-rich version of data for ML modelling.
Similarly, Table~\ref{Table:framework_E9_E13_tool} and Table~\ref{Table:framework_E9_E13} show that AutoFE, for both feature generation and selection, has not yet been consistently integrated into the broader AutoML ecosystem.
This state of the technology means that users still require the technical expertise/support to manually clean, prepare and enhance data for many ML applications.
Such an outcome is not optimal, given that the data-engineering stage of an MLWF is a large time-sink.
Moreover, Fig.~\ref{Fig:papers_by_mlwf} confirms that there is a high demand for automated data engineering among academically published ML applications.
Thus, the idealised form of AutoML services, even for the core data-to-model conversion process, cannot be realised without further developmental progress in this phase.

\textbf{The focus on model development is most substantial, but notable gaps in coverage exist.}
First of all, it is no surprise that all comprehensive AutoML software heavily automates the construction of ML models, given that CASH is the capability that initially brought the field to prominence.
Virtually all surveyed academic reports of AutoML applications involve some form of model selection.
Thus, as detailed by Table~\ref{Table:framework_CC1_CC9_tool} and Table~\ref{Table:framework_CC1_CC9}, AutoML tools, whether open-source or commercial, almost universally support standard forms of supervised ML.
Time-series forecasting is also decently well represented.
However, the provision of more exotic services is scattered.
Stakeholders will need to do their research when shopping for AutoML software if they have problems related to unsupervised learning, computer vision, and NLP.
Furthermore, particularly on the commercial side, Section~\ref{Sec:CommercialDiscussion} notes that transparency is an issue; just because a vendor claims to support an ML use case does not mean that the details of this support cover all possible stakeholder requirements.
Accordingly, the quality of model outcomes can be tough to guarantee under the broad framework for `performant ML' introduced in Section~\ref{Sec:summarisedcriteria}.

\textbf{Thus far, there has been little effort while developing AutoML technologies to ensure trust in ML.}
Indeed, this assertion is among the most predominant concerns when discussing possible deficiencies within model outcomes.
Many factors feed into whether a human is willing to trust the validity/utility of ML, and two are particularly pertinent here: explainability and fairness.
Now, Table~\ref{Table:framework_EX4_EX7_tool} and Table~\ref{Table:framework_EX3_EX7} indicate that existing AutoML software does generally assist interpretability on a global level, e.g.~identifying feature importance.
However, more sophisticated treatments are rare in commercial products and, according to this survey, are entirely absent in major open-source offerings.
This situation means stakeholders wishing to understand ML predictions/prescriptions for individual queries, e.g.~distinct people, may be forced to supplement a comprehensive system with the dedicated tools in Section~\ref{Sec:biasfairnesstools}.
As for scenario analysis, which could shed further light on problem contexts and constructed ML models by supporting counterfactual thinking, very few AutoML systems incorporate such a mechanism.
Of course, explainability can verify whether ML worked free of technical error, but it is also a prerequisite for identifying whether an ML solution is fair.
Bias can crop up via human decisions, non-representative data or algorithmic assumptions, so it is a significant deficiency that, among comprehensive AutoML systems, its automated detection and management are rare and practically nonexistent, respectively.
Consumers, businesses and general society are becoming increasingly aware of the socioethical impacts that real-world ML applications can have, and regulatory policies and legislation are steadily coming online.
Thus, while the topic is understandably complex~\cite{khke21}, it does not suit the ethos of AutoML to leave the daunting task of selecting/interpreting bias-and-fairness metrics to a non-expert.

\textbf{Publicly available AutoML software is hardly `cutting edge', but that may be okay; it is still unclear what contributes most to an optimal ML solution anyway.}
In Section~\ref{Sec:CommercialCompleteness}, a comparative analysis of HPO-mechanism availability within open-source systems and commercial products was accompanied by a discussion asserting that the technical depth of the latter seems relatively shallow.
Granted, this assessment is made holistically and is also possibly skewed by the heavy presence of active academics within the open-source community.
Nonetheless, it does seem like, for the model-development phase, commercial software is more likely to employ rudimentary grid/random search than any bandit-based refinement of Bayesian optimisation.
As for more sophisticated concepts explored within academia, their presence is felt sparsely across \textit{all} offerings, i.e.~meta-learning, ensemble techniques, ML pipeline optimisation, and so on.
Now, as mentioned in Section~\ref{Sec:specialisedtools}, the absence of theoretical advances in a package can be due to (1) ignorance, (2) an inability to implement, or (3) intentional rejection.
In this last case, it is possible that such an attitude could be justified.
Maybe technical depth is not actually required for an AutoML system to support performant ML.
Many modern SOTA approaches in data science burn up increasing amounts of resources to squeeze out decreasing increments of improvement, often all focussed on accuracy metrics that, as this review has argued, may not be critically important.
Even if they were, perhaps ML applications would be better served by a data-centric approach rather than fixating on fine-tuning models for a static dataset.
Intelligently controlling data flow, not just through AutoFE but also by the selective presentation of samples, may potentially impact model outcomes more than CASH.
Whatever the case, many commercial AutoML providers are presently surviving without needing to employ SOTA techniques.
Perhaps operational efficiencies gained across an entire MLWF make up for any model-development shortcomings.
Alternatively, AutoML may be so innovative among its target industries that any rudimentary approaches still provide immense value.
Only time will tell if developers start following academic research more closely once the easy pickings dry up.

\textbf{The limited provision of automated deployment, monitoring and maintenance is a predominantly commercial affair.}
After all, as a sweeping generalisation, academia appears to be preoccupied with the static accuracy of ML models.
This assertion is supported by Fig.~\ref{Fig:papers_by_metric}, which shows the metrics that overwhelmingly matter when publishing AutoML applications via academic journals.
Essentially, all scientific interest seems to end with the model-development phase of an MLWF.
One may argue that what follows tends to be dismissed as a mundane and messy engineering problem.
This perspective may also explain why free AutoML services that are not directly linked to academia likewise do not bother assisting with the deployment stage.
In contrast, vendors, whose clientele can include particularly non-technical stakeholders, are contractually obligated to provide end-to-end support.
Outside of insight-gathering applications, an ML model object suits no business organisation if it cannot be put into production to supply end-users with predictions/prescriptions.
Accordingly, most surveyed commercial systems have found it necessary to venture into the space of MLOps.
That said, Table~\ref{Table:framework_DM2_DM4} questions the contemporary sophistication of these services, e.g.~in terms of testing and updating deployments.
The situation is similar for the final phase of an MLWF: monitoring and maintenance.
Many surveyed vendors do actually have basic mechanisms in place to alert clients about degrading model performance, triggering retraining if desired.
However, Table~\ref{Table:framework_DM5_DM11} reveals a dearth of systems that actively monitor data for concept drift, let alone proactively update an ML model.
Of course, it is presently unclear what the best policy for triggering adaptation is~\cite{zlbu15, zlbi12}, and vendors are also likely to prefer humans in the loop to limit liability for poor model outcomes.
Nonetheless, there is a long way to go before AutoML services can be claimed to support continuous learning, a prerequisite for next-generation AutonoML frameworks~\cite{kemu20, khke21}.

\begin{table}[h]
\caption{Summary of how the surveyed open-source and commercial AutoML systems scored for major criteria. Includes (1) total scores incorporating all the included criteria and (2) average scores for the open-source and commercial software tools. Each score is relative to the maximum achievable and is contained within the 0 to 1 range, with 0 meaning that none of the assessed aspects have been met and 1 that all of them are perfectly fulfilled. Colour coding: the darker green colour means better performance. Abbreviations: E - efficiency, DD - dirty data, CC - completeness \& currency, EX - explainability, EU - ease of use, DM - deployment and management effort, and G - governance and security.}
\label{table:aggregate_comparison_table}
\resizebox{\textwidth}{!}{%
{\footnotesize
\begin{tabular}{|l|c|c|c|c|c|
>{\columncolor[HTML]{FCFCFF}}c |
>{\columncolor[HTML]{FCFCFF}}c |c|}
\hline
\cellcolor[HTML]{B4C6E7}Name &
  \cellcolor[HTML]{B4C6E7}E Total &
  \cellcolor[HTML]{B4C6E7}DD Total &
  \cellcolor[HTML]{B4C6E7}CC Total &
  \cellcolor[HTML]{B4C6E7}EX Total &
  \cellcolor[HTML]{B4C6E7}EU Total &
  \cellcolor[HTML]{B4C6E7}DM Total &
  \cellcolor[HTML]{B4C6E7}G Total &
  \cellcolor[HTML]{B4C6E7}Overall   Proportion 
  \\ \hline
Auto ViML &
  \cellcolor[HTML]{B7E0C4}0.452 &
  \cellcolor[HTML]{B0DDBD}0.500 &
  \cellcolor[HTML]{BDE3C8}0.417 &
  \cellcolor[HTML]{DAEFE2}0.222 &
  \cellcolor[HTML]{D1EBDA}0.286 &
  0.000 &
  0.000 &
  \cellcolor[HTML]{C9E8D3}0.333 \\ \hline
Auto-PyTorch &
  \cellcolor[HTML]{CBE8D5}0.323 &
  \cellcolor[HTML]{B0DDBD}0.500 &
  \cellcolor[HTML]{B6E0C3}0.458 &
  \cellcolor[HTML]{FCFCFF}0.000 &
  \cellcolor[HTML]{D1EBDA}0.286 &
  0.000 &
  0.000 &
  \cellcolor[HTML]{D1EBDA}0.281 \\ \hline
auto-sklearn &
  \cellcolor[HTML]{DAEEE2}0.226 &
  \cellcolor[HTML]{9DD6AD}0.625 &
  \cellcolor[HTML]{96D3A7}0.667 &
  \cellcolor[HTML]{FCFCFF}0.000 &
  \cellcolor[HTML]{BBE2C7}0.429 &
  0.000 &
  0.000 &
  \cellcolor[HTML]{CBE8D5}0.323 \\ \hline
AutoGluon &
  \cellcolor[HTML]{D0EAD9}0.290 &
  \cellcolor[HTML]{C3E5CE}0.375 &
  \cellcolor[HTML]{96D3A7}0.667 &
  \cellcolor[HTML]{DAEFE2}0.222 &
  \cellcolor[HTML]{BBE2C7}0.429 &
  0.000 &
  0.000 &
  \cellcolor[HTML]{C8E7D2}0.344 \\ \hline
AutoKeras &
  \cellcolor[HTML]{DFF0E6}0.194 &
  \cellcolor[HTML]{C3E5CE}0.375 &
  \cellcolor[HTML]{90D1A2}0.708 &
  \cellcolor[HTML]{FCFCFF}0.000 &
  \cellcolor[HTML]{BBE2C7}0.429 &
  0.000 &
  0.000 &
  \cellcolor[HTML]{CEEAD8}0.302 \\ \hline
AutoML Alex &
  \cellcolor[HTML]{D5ECDD}0.258 &
  \cellcolor[HTML]{8ACE9C}0.750 &
  \cellcolor[HTML]{AADBB8}0.542 &
  \cellcolor[HTML]{FCFCFF}0.000 &
  \cellcolor[HTML]{BBE2C7}0.429 &
  0.000 &
  0.000 &
  \cellcolor[HTML]{CDE9D6}0.313 \\ \hline
carefree-learn &
  \cellcolor[HTML]{D0EAD9}0.290 &
  \cellcolor[HTML]{B0DDBD}0.500 &
  \cellcolor[HTML]{AADBB8}0.542 &
  \cellcolor[HTML]{FCFCFF}0.000 &
  \cellcolor[HTML]{BBE2C7}0.429 &
  0.000 &
  0.000 &
  \cellcolor[HTML]{CEEAD8}0.302 \\ \hline
FLAML &
  \cellcolor[HTML]{D5ECDD}0.258 &
  \cellcolor[HTML]{C3E5CE}0.375 &
  \cellcolor[HTML]{A3D8B2}0.583 &
  \cellcolor[HTML]{FCFCFF}0.000 &
  \cellcolor[HTML]{D1EBDA}0.286 &
  0.000 &
  0.000 &
  \cellcolor[HTML]{D1EBDA}0.281 \\ \hline
GAMA &
  \cellcolor[HTML]{E4F2EA}0.161 &
  \cellcolor[HTML]{C3E5CE}0.375 &
  \cellcolor[HTML]{BDE3C8}0.417 &
  \cellcolor[HTML]{DAEFE2}0.222 &
  \cellcolor[HTML]{63BE7B}1.000 &
  0.000 &
  0.000 &
  \cellcolor[HTML]{D1EBDA}0.281 \\ \hline
HyperGBM &
  \cellcolor[HTML]{C6E6D1}0.355 &
  \cellcolor[HTML]{B0DDBD}0.500 &
  \cellcolor[HTML]{B6E0C3}0.458 &
  \cellcolor[HTML]{FCFCFF}0.000 &
  \cellcolor[HTML]{79C78E}0.857 &
  0.000 &
  0.000 &
  \cellcolor[HTML]{C9E8D3}0.333 \\ \hline
Hyperopt-sklearn &
  \cellcolor[HTML]{C1E4CC}0.387 &
  \cellcolor[HTML]{C3E5CE}0.375 &
  \cellcolor[HTML]{BDE3C8}0.417 &
  \cellcolor[HTML]{FCFCFF}0.000 &
  \cellcolor[HTML]{D1EBDA}0.286 &
  0.000 &
  0.000 &
  \cellcolor[HTML]{D1EBDA}0.281 \\ \hline
lgel &
  \cellcolor[HTML]{DAEEE2}0.226 &
  \cellcolor[HTML]{9DD6AD}0.625 &
  \cellcolor[HTML]{A3D8B2}0.583 &
  \cellcolor[HTML]{FCFCFF}0.000 &
  \cellcolor[HTML]{79C78E}0.857 &
  0.000 &
  0.000 &
  \cellcolor[HTML]{C9E8D3}0.333 \\ \hline
Lightwood &
  \cellcolor[HTML]{DFF0E6}0.194 &
  \cellcolor[HTML]{C3E5CE}0.375 &
  \cellcolor[HTML]{BDE3C8}0.417 &
  \cellcolor[HTML]{DAEFE2}0.222 &
  \cellcolor[HTML]{D1EBDA}0.286 &
  0.000 &
  0.000 &
  \cellcolor[HTML]{D8EEE0}0.240 \\ \hline
Ludwig &
  \cellcolor[HTML]{DAEEE2}0.226 &
  \cellcolor[HTML]{B0DDBD}0.500 &
  \cellcolor[HTML]{A3D8B2}0.583 &
  \cellcolor[HTML]{DAEFE2}0.222 &
  \cellcolor[HTML]{A5D9B4}0.571 &
  0.000 &
  0.000 &
  \cellcolor[HTML]{CBE8D5}0.323 \\ \hline
Mljar &
  \cellcolor[HTML]{D5ECDD}0.258 &
  \cellcolor[HTML]{B0DDBD}0.500 &
  \cellcolor[HTML]{AADBB8}0.542 &
  \cellcolor[HTML]{DAEFE2}0.222 &
  \cellcolor[HTML]{BBE2C7}0.429 &
  0.000 &
  0.000 &
  \cellcolor[HTML]{CDE9D6}0.313 \\ \hline
mlr3automl &
  \cellcolor[HTML]{D5ECDD}0.258 &
  \cellcolor[HTML]{B0DDBD}0.500 &
  \cellcolor[HTML]{BDE3C8}0.417 &
  \cellcolor[HTML]{FCFCFF}0.000 &
  \cellcolor[HTML]{D1EBDA}0.286 &
  0.000 &
  0.000 &
  \cellcolor[HTML]{D6EDDE}0.250 \\ \hline
OBOE &
  \cellcolor[HTML]{D0EAD9}0.290 &
  \cellcolor[HTML]{8ACE9C}0.750 &
  \cellcolor[HTML]{C3E5CE}0.375 &
  \cellcolor[HTML]{FCFCFF}0.000 &
  \cellcolor[HTML]{D1EBDA}0.286 &
  0.000 &
  0.000 &
  \cellcolor[HTML]{D3ECDC}0.271 \\ \hline
PyCaret &
  \cellcolor[HTML]{D5ECDD}0.258 &
  \cellcolor[HTML]{77C68C}0.875 &
  \cellcolor[HTML]{96D3A7}0.667 &
  \cellcolor[HTML]{DAEFE2}0.222 &
  \cellcolor[HTML]{BBE2C7}0.429 &
  0.000 &
  0.000 &
  \cellcolor[HTML]{C3E5CE}0.375 \\ \hline
TPOT &
  \cellcolor[HTML]{D0EAD9}0.290 &
  \cellcolor[HTML]{9DD6AD}0.625 &
  \cellcolor[HTML]{8ACE9C}0.750 &
  \cellcolor[HTML]{FCFCFF}0.000 &
  \cellcolor[HTML]{A5D9B4}0.571 &
  0.000 &
  0.000 &
  \cellcolor[HTML]{C3E5CE}0.375 \\ \hline \hline
\cellcolor[HTML]{B4C6E7}Average   score - open-source systems &
  \cellcolor[HTML]{D3ECDB}0.273 &
  \cellcolor[HTML]{ACDCBA}0.526 &
  \cellcolor[HTML]{AADBB9}0.537 &
  \cellcolor[HTML]{F0F7F5}0.082 &
  \cellcolor[HTML]{B5E0C2}0.466 &
  0.000 &
  0.000 &
  \cellcolor[HTML]{CDE9D7}0.308 \\ \hline \hline
\cellcolor[HTML]{B4C6E7}Name &
  \cellcolor[HTML]{B4C6E7}E Total &
  \cellcolor[HTML]{B4C6E7}DD Total &
  \cellcolor[HTML]{B4C6E7}CC Total &
  \cellcolor[HTML]{B4C6E7}EX Total &
  \cellcolor[HTML]{B4C6E7}EU Total &
  \cellcolor[HTML]{B4C6E7}DM Total &
  \cellcolor[HTML]{B4C6E7}G Total &
  \cellcolor[HTML]{B4C6E7}Overall Proportion \\ \hline
Alteryx &
  \cellcolor[HTML]{CEE9D7}0.306 &
  \cellcolor[HTML]{77C68C}0.875 &
  \cellcolor[HTML]{C9E8D3}0.333 &
  \cellcolor[HTML]{FCFCFF}0.000 &
  \cellcolor[HTML]{8FD0A1}0.714 &
  \cellcolor[HTML]{BBE2C7}0.429 &
  \cellcolor[HTML]{96D3A7}0.667 &
  \cellcolor[HTML]{C1E4CC}0.391 \\ \hline
Auger &
  \cellcolor[HTML]{B7E0C4}0.452 &
  \cellcolor[HTML]{8ACE9C}0.750 &
  \cellcolor[HTML]{C3E5CE}0.375 &
  \cellcolor[HTML]{FCFCFF}0.000 &
  \cellcolor[HTML]{8FD0A1}0.714 &
  \cellcolor[HTML]{C6E6D0}0.357 &
  0.000 &
  \cellcolor[HTML]{BEE3CA}0.406 \\ \hline
B2Metric &
  \cellcolor[HTML]{E4F2EA}0.161 &
  \cellcolor[HTML]{B0DDBD}0.500 &
  \cellcolor[HTML]{B6E0C3}0.458 &
  \cellcolor[HTML]{FCFCFF}0.000 &
  \cellcolor[HTML]{D1EBDA}0.286 &
  \cellcolor[HTML]{BBE2C7}0.429 &
  \cellcolor[HTML]{96D3A7}0.667 &
  \cellcolor[HTML]{CDE9D6}0.313 \\ \hline
Big Squid &
  \cellcolor[HTML]{CBE8D5}0.323 &
  \cellcolor[HTML]{9DD6AD}0.625 &
  \cellcolor[HTML]{C9E8D3}0.333 &
  \cellcolor[HTML]{A7DAB6}0.556 &
  \cellcolor[HTML]{A5D9B4}0.571 &
  \cellcolor[HTML]{E7F4ED}0.143 &
  0.000 &
  \cellcolor[HTML]{C6E7D1}0.354 \\ \hline
BigML &
  \cellcolor[HTML]{DAEEE2}0.226 &
  \cellcolor[HTML]{B0DDBD}0.500 &
  \cellcolor[HTML]{A3D8B2}0.583 &
  \cellcolor[HTML]{DAEFE2}0.222 &
  \cellcolor[HTML]{79C78E}0.857 &
  \cellcolor[HTML]{DCEFE3}0.214 &
  \cellcolor[HTML]{63BE7B}1.000 &
  \cellcolor[HTML]{BEE3CA}0.406 \\ \hline
cnvrg.io &
  \cellcolor[HTML]{CBE8D5}0.323 &
  \cellcolor[HTML]{B0DDBD}0.500 &
  \cellcolor[HTML]{C3E5CE}0.375 &
  \cellcolor[HTML]{EBF6F1}0.111 &
  \cellcolor[HTML]{79C78E}0.857 &
  \cellcolor[HTML]{8FD0A1}0.714 &
  \cellcolor[HTML]{C9E8D3}0.333 &
  \cellcolor[HTML]{BBE2C7}0.427 \\ \hline
Compellon &
  \cellcolor[HTML]{E4F2EA}0.161 &
  \cellcolor[HTML]{C3E5CE}0.375 &
  \cellcolor[HTML]{F0F7F4}0.083 &
  \cellcolor[HTML]{DAEFE2}0.222 &
  \cellcolor[HTML]{D1EBDA}0.286 &
  0.000 &
  0.000 &
  \cellcolor[HTML]{E6F3EC}0.146 \\ \hline
D2iQ &
  \cellcolor[HTML]{E4F2EA}0.161 &
  \cellcolor[HTML]{D6EDDE}0.250 &
  \cellcolor[HTML]{DDF0E4}0.208 &
  \cellcolor[HTML]{EBF6F1}0.111 &
  \cellcolor[HTML]{8FD0A1}0.714 &
  \cellcolor[HTML]{C6E6D0}0.357 &
  \cellcolor[HTML]{96D3A7}0.667 &
  \cellcolor[HTML]{D5ECDD}0.260 \\ \hline
Databricks &
  \cellcolor[HTML]{B7E0C4}0.452 &
  \cellcolor[HTML]{E9F5EF}0.125 &
  \cellcolor[HTML]{B6E0C3}0.458 &
  \cellcolor[HTML]{DAEFE2}0.222 &
  \cellcolor[HTML]{A5D9B4}0.571 &
  \cellcolor[HTML]{DCEFE3}0.214 &
  \cellcolor[HTML]{63BE7B}1.000 &
  \cellcolor[HTML]{C0E4CB}0.396 \\ \hline
Dataiku &
  \cellcolor[HTML]{84CB97}0.790 &
  \cellcolor[HTML]{D6EDDE}0.250 &
  \cellcolor[HTML]{83CB97}0.792 &
  \cellcolor[HTML]{85CC99}0.778 &
  \cellcolor[HTML]{79C78E}0.857 &
  \cellcolor[HTML]{9AD5AB}0.643 &
  \cellcolor[HTML]{63BE7B}1.000 &
  \cellcolor[HTML]{8CCF9F}0.734 \\ \hline
DataRobot &
  \cellcolor[HTML]{7CC891}0.839 &
  \cellcolor[HTML]{D6EDDE}0.250 &
  \cellcolor[HTML]{90D1A2}0.708 &
  \cellcolor[HTML]{96D3A7}0.667 &
  \cellcolor[HTML]{8FD0A1}0.714 &
  \cellcolor[HTML]{84CC98}0.786 &
  \cellcolor[HTML]{63BE7B}1.000 &
  \cellcolor[HTML]{8DCF9F}0.729 \\ \hline
Deep   Cognition &
  \cellcolor[HTML]{DAEEE2}0.226 &
  \cellcolor[HTML]{E9F5EF}0.125 &
  \cellcolor[HTML]{D6EDDE}0.250 &
  \cellcolor[HTML]{DAEFE2}0.222 &
  \cellcolor[HTML]{8FD0A1}0.714 &
  \cellcolor[HTML]{E7F4ED}0.143 &
  0.000 &
  \cellcolor[HTML]{D8EEE0}0.240 \\ \hline
Einblick &
  \cellcolor[HTML]{DAEEE2}0.226 &
  \cellcolor[HTML]{E9F5EF}0.125 &
  \cellcolor[HTML]{B0DDBD}0.500 &
  \cellcolor[HTML]{DAEFE2}0.222 &
  \cellcolor[HTML]{A5D9B4}0.571 &
  0.000 &
  \cellcolor[HTML]{96D3A7}0.667 &
  \cellcolor[HTML]{D0EAD9}0.292 \\ \hline
Google &
  \cellcolor[HTML]{C1E4CC}0.387 &
  \cellcolor[HTML]{D6EDDE}0.250 &
  \cellcolor[HTML]{AADBB8}0.542 &
  \cellcolor[HTML]{96D3A7}0.667 &
  \cellcolor[HTML]{79C78E}0.857 &
  \cellcolor[HTML]{C6E6D0}0.357 &
  \cellcolor[HTML]{63BE7B}1.000 &
  \cellcolor[HTML]{B2DEBF}0.490 \\ \hline
H2O &
  \cellcolor[HTML]{9FD6AF}0.613 &
  \cellcolor[HTML]{D6EDDE}0.250 &
  \cellcolor[HTML]{7DC991}0.833 &
  \cellcolor[HTML]{A7DAB6}0.556 &
  \cellcolor[HTML]{79C78E}0.857 &
  \cellcolor[HTML]{9AD5AB}0.643 &
  0.000 &
  \cellcolor[HTML]{9BD5AC}0.635 \\ \hline
IBM &
  \cellcolor[HTML]{C6E6D1}0.355 &
  \cellcolor[HTML]{E9F5EF}0.125 &
  \cellcolor[HTML]{C9E8D3}0.333 &
  \cellcolor[HTML]{85CC99}0.778 &
  \cellcolor[HTML]{8FD0A1}0.714 &
  \cellcolor[HTML]{A5D9B4}0.571 &
  \cellcolor[HTML]{63BE7B}1.000 &
  \cellcolor[HTML]{B8E1C4}0.448 \\ \hline
KNIME &
  \cellcolor[HTML]{DAEEE2}0.226 &
  \cellcolor[HTML]{E9F5EF}0.125 &
  \cellcolor[HTML]{B6E0C3}0.458 &
  \cellcolor[HTML]{FCFCFF}0.000 &
  \cellcolor[HTML]{79C78E}0.857 &
  \cellcolor[HTML]{BBE2C7}0.429 &
  \cellcolor[HTML]{63BE7B}1.000 &
  \cellcolor[HTML]{C6E7D1}0.354 \\ \hline
Microsoft &
  \cellcolor[HTML]{9FD6AF}0.613 &
  \cellcolor[HTML]{D6EDDE}0.250 &
  \cellcolor[HTML]{9DD6AD}0.625 &
  \cellcolor[HTML]{A7DAB6}0.556 &
  \cellcolor[HTML]{79C78E}0.857 &
  \cellcolor[HTML]{9AD5AB}0.643 &
  \cellcolor[HTML]{63BE7B}1.000 &
  \cellcolor[HTML]{9ED6AE}0.615 \\ \hline
MyDataModels &
  \cellcolor[HTML]{DFF0E6}0.194 &
  \cellcolor[HTML]{D6EDDE}0.250 &
  \cellcolor[HTML]{D6EDDE}0.250 &
  \cellcolor[HTML]{C9E8D3}0.333 &
  \cellcolor[HTML]{BBE2C7}0.429 &
  0.000 &
  0.000 &
  \cellcolor[HTML]{DDF0E4}0.208 \\ \hline
Number Theory &
  \cellcolor[HTML]{D0EAD9}0.290 &
  \cellcolor[HTML]{E9F5EF}0.125 &
  \cellcolor[HTML]{AADBB8}0.542 &
  \cellcolor[HTML]{DAEFE2}0.222 &
  \cellcolor[HTML]{D1EBDA}0.286 &
  \cellcolor[HTML]{A5D9B4}0.571 &
  0.000 &
  \cellcolor[HTML]{C5E6CF}0.365 \\ \hline
RapidMiner &
  \cellcolor[HTML]{CBE8D5}0.323 &
  \cellcolor[HTML]{E9F5EF}0.125 &
  \cellcolor[HTML]{A6DAB5}0.563 &
  \cellcolor[HTML]{B8E1C5}0.444 &
  \cellcolor[HTML]{79C78E}0.857 &
  \cellcolor[HTML]{B0DDBD}0.500 &
  \cellcolor[HTML]{63BE7B}1.000 &
  \cellcolor[HTML]{B6E0C2}0.464 \\ \hline
SageMaker &
  \cellcolor[HTML]{C1E4CC}0.387 &
  \cellcolor[HTML]{E9F5EF}0.125 &
  \cellcolor[HTML]{AADBB8}0.542 &
  \cellcolor[HTML]{85CC99}0.778 &
  \cellcolor[HTML]{79C78E}0.857 &
  \cellcolor[HTML]{A5D9B4}0.571 &
  \cellcolor[HTML]{63BE7B}1.000 &
  \cellcolor[HTML]{ADDCBB}0.521 \\ \hline
SAS &
  \cellcolor[HTML]{CBE8D5}0.323 &
  \cellcolor[HTML]{E9F5EF}0.125 &
  \cellcolor[HTML]{B0DDBD}0.500 &
  \cellcolor[HTML]{B8E1C5}0.444 &
  \cellcolor[HTML]{8FD0A1}0.714 &
  \cellcolor[HTML]{B0DDBD}0.500 &
  \cellcolor[HTML]{63BE7B}1.000 &
  \cellcolor[HTML]{BAE1C6}0.438 \\ \hline
Spell &
  \cellcolor[HTML]{DFF0E6}0.194 &
  \cellcolor[HTML]{FCFCFF}0.000 &
  \cellcolor[HTML]{BDE3C8}0.417 &
  \cellcolor[HTML]{EBF6F1}0.111 &
  \cellcolor[HTML]{8FD0A1}0.714 &
  \cellcolor[HTML]{C6E6D0}0.357 &
  \cellcolor[HTML]{96D3A7}0.667 &
  \cellcolor[HTML]{CEEAD8}0.302 \\ \hline
TIMi &
  \cellcolor[HTML]{F8FAFB}0.032 &
  \cellcolor[HTML]{FCFCFF}0.000 &
  \cellcolor[HTML]{E3F2E9}0.167 &
  \cellcolor[HTML]{DAEFE2}0.222 &
  \cellcolor[HTML]{E7F4ED}0.143 &
  0.000 &
  0.000 &
  \cellcolor[HTML]{F0F7F4}0.083 \\ \hline \hline
\cellcolor[HTML]{B4C6E7}Average   score - commercial systems &
  \cellcolor[HTML]{C8E7D2}0.343 &
  \cellcolor[HTML]{D2EBDB}0.280 &
  \cellcolor[HTML]{B8E1C4}0.449 &
  \cellcolor[HTML]{C9E8D3}0.338 &
  \cellcolor[HTML]{97D3A8}0.663 &
  \cellcolor[HTML]{C2E5CD}0.383 &
  \cellcolor[HTML]{A3D8B2}0.587 &
  \cellcolor[HTML]{C0E4CB}0.397 \\ \hline
\end{tabular}
}
}
\end{table}

\textbf{Making a business out of AutoML drives different priorities compared with providing it for free.}
To support this point and underscore several preceding insights, Table~\ref{table:aggregate_comparison_table} highlights just how differently this monograph rated open-source and commercial AutoML systems for major criteria.
This comparison includes considerations of the deployment and management effort (DM), plus governance and security (G), though it is noted that all free providers effectively ignore these facets and score zero, bringing down all their overall scores.
Nevertheless, taking into account all the criteria listed in the table, it is worth noting that five packages each obtained over half of the available points, namely Dataiku (0.734), DataRobot (0.729), H2O (0.635), Microsoft (0.615), and SageMaker (0.521).
They are all commercial.
So, after first accounting for the skewing effect of these five, we find there is little overall difference between open-source and closed-source tools in terms of operational/technical efficiencies (E).
In fact, when it comes to matters of technical depth, e.g.~cleaning/preparing dirty data (DD) and providing coverage of ML approaches that is complete and current (CC), free AutoML services are generally ahead.
One can even hypothesise that, if benchmarking technical performance was included in this survey, open-source systems would likely have the greater range of options to pursue SOTA ML results.
However, when it comes to explainability (EX) and ease of use (EU), critical requirements for democratisation, vendors are solidly ahead, scoring 0.338 versus 0.082 and 0.663 versus 0.466, respectively.
Admittedly, the quantification in this monograph is open to debate, and we do not claim there are well-defined thresholds for service quality or business value.
Nonetheless, it is clear that when the stakes are high, and the survival of AutoML software is closely coupled with stakeholder engagement, certain aspects attain an importance they did not previously have.

\textbf{There is no singular roadmap for the future of AutoML technology.}
Admittedly, this particular finding may seem at odds with previous suggestions that imply some logical order to advancing the field.
However, as discussed in Section~\ref{Sec:CommercialDiscussion}, there are many actors with varying origins and diverse agendas telling this story.
So, while global trends and directions may be emergently evident, the localised threads of development are much more unpredictable and chaotic; there may yet be surprises in store for what `AutoML Tech' comes to mean.
For instance, concerning the latter stages of an MLWF, it is clear that different parties are converging on assisted/automated MLOps from two different directions.
Some are pushing forwards from core AutoML practices such as CASH, while others are entering the space by expanding the scope of generic DevOps platforms.
It is not even clear which entities will outcompete each other.
The latter tend to have well-established infrastructures to lean on, while, as recently mentioned, it is not apparent whether any advanced model-selection methodologies offered by the former constitute a genuine advantage.
Thus, from an industrial consumer perspective, will AutoML platforms offer MLOps?
Or will MLOps platforms offer AutoML?
Then there are higher-level questions.
It is clear that an integrated comprehensive system, both end-to-end and generally applicable, is theoretically ideal in terms of broad utility~\cite{kemu20}.
Yet AutoML tools dedicated solely to individual MLWF phases/processes, as in Section~\ref{Sec:ancillarytools}, and technical/industrial domains, as in Section~\ref{Sec:specialisedtools}, benefit from being lightweight and focussed.
Maybe it is more practical for sufficiently technical stakeholders to stitch together a patchwork of AutoML services as desired, assuming commercial tools are transparent enough to allow this, leaving comprehensive systems to those who need/want to relinquish finer control.
Again, this review cannot pass judgement here, as it is unclear how much a dedicated developmental focus compensates for restricted utility and the technical challenges of subsequent integration.

\textbf{Ultimately, further studies based on a broader vision of `performant ML' are required to honestly assess whether AutoML technology is living up to its full potential.}
That is not to say that the endeavour of automating high-level ML operations has not already made a splash in its last decade of mainstream dissemination.
Numerous enterprising developers have jumped on the bandwagon, offering a diverse supply of AutoML products.
Demand is likewise surging if the yearly numbers of published AutoML applications charted in Fig.~\ref{Fig:papers_by_year} are anything to go by.
Nonetheless, this means that it is even \textit{more} crucial in this phase of initial contact between technology and industry that the best first impression is achieved and maintained.
Already, this review has identified numerous gaps in the services offered across both the open-source sphere and commercial marketplace.
Admittedly, the impact of these deficiencies will obviously differ.
For instance, not being able to cover certain types of ML problem merely limits the prospective clientele for associated tools.
On the other hand, obsessing with technical metrics at the cost of appreciating the holistic stakeholder experience may frustrate engaged users and damage future uptake.
Developers should thus carefully consider the broader criteria introduced in Section~\ref{Sec:summarisedcriteria} when assessing their implementations.
Of course, further deliberation and debate are invited so such factors can be quantified as objectively as possible; a scientific consensus can only make ensuing benchmarks for AutoML services much more authoritative.
Finally, while this monograph has primarily attempted to assess the current state of AutoML technology via literature surveys, its limitations are clear.
For instance, vendor-promoted case studies are a poor substitute for academic reports, often exhibiting bias, spin, and a lack of verifiable details.
Instead, directly approaching AutoML users is likely a better litmus test for demand, although such surveys should also consider a diversity of potential stakeholders beyond just data scientists~\cite{wawe19, drwe20}.
Fundamentally, the AutoML community cannot optimise the translation of theory to technology if it cannot effectively observe and identify what industry likes and what industry wants.
Simply put, there is more work to be done.

\section{Conclusions}
\label{Sec:Conclusion}

This monograph has reviewed the technological emergence and industrial uptake of AutoML as they appear in the early 2020s.
It is the first to comprehensively assess how this translation of academic theory to mainstream practice has fared, distinguishing itself from its forerunners~\cite{kemu20, doke21} and other overviews in the literature that focus instead on the fundamental concepts behind AutoML.
Unsurprisingly, such an endeavour grapples with many questions.
Who are the people that are likely to engage with or be impacted by AutoML, both presently and potentially?
What do they want from AutoML?
How is the technology currently being supplied?
What is it capable of?
And what is it being used for?
Commentary on these topics has been heavily informed by surveys of documented codebases, promotional material, and application reports.

Before undertaking any analyses, it was first necessary to motivate and define a lens through which the state of AutoML technology could be appraised.
The following is a summary of this preamble:
\begin{itemize}
	\item Section~\ref{Sec:mlwf} formalised the nature of an ML application, i.e.~a collection of activities aiming to create/exploit a data-driven solution to a problem of descriptive/predictive/prescriptive analytics via ML techniques.
	Specifically, the section introduced the notion of an MLWF, designating a systematic way to organise tasks common to an ML application within one encompassing but segmented workflow representation.
	Such a conceptual framework makes it possible to assess which of the following operational phases are targeted by AutoML technology: problem formulation \& context understanding, data engineering, model development, deployment, and monitoring \& maintenance.
	\item Section~\ref{Sec:keystakeholders} introduced the notion of an ML stakeholder, i.e.~a person or party with direct or indirect interests in the processes and outcomes of an ML application.
	The section then elaborated on the likely desires and requirements each type of stakeholder brings to the table when engaging with ML.
	Such considerations allow assessments of AutoML to extend beyond the data-scientist experience and, in particular, a fixation on technical metrics for ML model validity.
	\item Section~\ref{Sec:summarisedcriteria} proposed an extensive assessment framework to determine how well an AutoML service supports `performant ML'.
	The associated criteria are based on a holistic view of what matters to stakeholders and are grouped into the following categories: efficiency, dirty data, completeness \& currency, explainability, ease of use, deployment \& management effort, and governance.
	\item Section~\ref{Sec:automlrole} finally condensed the assessment framework based on stakeholder requirements to generate insight and commentary about what industry presently sees as the role of AutoML.
	Overall, there appear to be three primary goals for the technology as it relates to data science practices: enhancement, democratisation, and standardisation.
\end{itemize}

Having established a comprehensive reference frame for what makes AutoML `good' within a real-world setting, this monograph then proceeded to present and dissect several surveys of existing tools and application reports.
The following summarises the analyses specific to AutoML software:
\begin{itemize}
	\item Section~\ref{Sec:ancillarytools} began appraising supply by examining AutoML tools that are termed `dedicated'; these do not intrinsically feature the core processes of generating/managing ML model objects.
	Instead, the packages are meant to be used with other software, focussing only on specific segments and responsibilities within an MLWF, such as (1) data and feature engineering, (2) bias, fairness and explainability, and (3) HPO.
	Overall, these `detachable' tools can provide flexibility and control to a technical user, but integration challenges may hinder broader uptake.
	\item Section~\ref{Sec:coretools} considered open-source AutoML software termed `comprehensive'; these implementations take charge of training ML models at a low level while also automating higher-level operations.
	Essentially, this survey found that the field of free options available to interested stakeholders is broad, with nuanced discussions arising for each of the aforementioned criteria.
	However, there is a general sense that these systems, often tied to academia, are frequently inspired to innovate at technical depth, e.g.~with sophisticated model selection.
	The trade-off is that almost all have no interest in the productionisation stages of an MLWF, and it is also rare to see HCI options for non-technical stakeholders receive serious developmental attention.
	\item Section~\ref{Sec:commercialtools} continued assessing the supply of comprehensive AutoML systems, although now focussing on the commercial sphere.
	Again, many products ready for organisational use were identified, even though the capabilities and features their developers invest in vary.
	Unsurprisingly, transparency was an obstacle to this analysis, but the details suggest that commercial AutoML goes broad, not deep, contrasting with open-source offerings.
	For instance, vendors seeking non-technical clients are motivated to focus on accessible UIs and support some form of MLOps, rather than implement a SOTA CASH mechanism.
	However, the survey results also indicated that no one particular agenda dominates the space of AutoML technology; developers are scattered in their origins, approaches, and priorities.
\end{itemize}
The following summarises the analyses specific to AutoML applications:
\begin{itemize}
	\item Section~\ref{Sec:AcademicApplications} began appraising demand by examining academically published AutoML projects run for real-world problems.
	The rate of these applications being published was found to be accelerating, and, although usage appears to be dominated by the health \& biomedical sectors, the spectrum of industries involved affirms the broad utility of AutoML.
	However, the academic context of these publications obviously distorts what could be surmised about the technology.
	In particular, there was a jarring discrepancy between the contemplative ways authors promoted AutoML and the routine ways it was assessed to be performant; judgements predominantly fell back on technical metrics for model validity.
	\item Section~\ref{Sec:specialisedtools} briefly acknowledged a rising trend, i.e.~the increasing number of AutoML tools being specialised for particular use cases, such as niche forms of data analysis or associations with specific industries.
	This phenomenon is arguably driven by expectations of demand, as expending effort on specialised functionality has little to gain if it is not used.
	For now, the trend is too nascent to earn significant commentary, but it reaffirms a finding in this monograph that there are many ways for AutoML technology to evolve `in the wild'.
	\item Section~\ref{Sec:IndustrialApplications} finally capped off analysing demand by inspecting AutoML vendor websites and associated case studies.
	Although such promotional material should be reasonably well representative of what customers want from their commercial AutoML services, the usefulness of the reports was obviously limited by a lack of peer review and detail.
	Nonetheless, vendors were found to be targeting a selection of industries somewhat dissimilar to those linked with academically published AutoML applications, which again highlights the general utility of the technology.
	Moreover, the diverse array of problem types that AutoML is servicing further suggests that high-quality outcomes depend on more than just technical accuracy metrics.
\end{itemize}

Ultimately, the value in this review is ideally found in the broader perspective it promotes for evaluating a \textit{technology} as opposed to a science.
By this stage, preceding monographs have already argued strongly that the theoretical potential of AutoML~\cite{kemu20} and AutoDL~\cite{doke21} is immense.
An enterprising researcher may envisage logical pathways -- not necessarily easy ones -- towards inspirational ambitions, including AutonoML systems capable of continuous learning, dynamic self-assembly, and general applicability.
However, AutoML implementations ready for societal use are clearly nowhere near such lofty heights.
This review even finds suggestions that publicly available products lag behind contemporary SOTA techniques, implying developer ignorance, an inability to implement, or deliberate rejection.
Indeed, if fundamental AutoML advances are being actively dismissed by developers, despite researchers finding them promising, then this disconnect is a troublesome one and should be understood; failure to do so will only cripple the translation and impact of AutoML.
The fact is, for better or worse, technology is what showcases a science to investors and heavily influences, if not determines, how many resources are cycled back for further research and development.
Accordingly, the AutoML community cannot neglect to examine how industry is engaging with the technology, and this requires a solid grasp of what `performance' means in the general case.
Real-world ML applications involving a diversity of stakeholders are subject to many more complicated needs and pressures than a sanitised benchmark challenge tackled by a group of expert ML practitioners.

Granted, to the credit of AutoML developers, this review has found many promising implementations of AutoML software, whether open-source or commercial, end-to-end or dedicated, generic or specialised.
Nonetheless, at the same time, numerous gaps have been identified in the services provided.
Perhaps these do not matter, and clients will still appreciate any added value that ML can extract from their business contexts, provided the platforms they use look professional.
Alternatively, complacency may lead to flagging public interest in AutoML, and the technology could end up being perceived as a gimmick.
All that is clear is that the lessons learned and decisions made during this initial contact between AutoML technology and industry will set the tone for future societal engagement.

\bibliographystyle{ACM-Reference-Format}
\bibliography{automl_academic_refs_preprint, automl_academic_refs, automl_githubs, automl_non_academic_report, automl_non_academic_web, automl_vendors
}

\newpage
\begin{appendices}
\section{Faded Hyperparameter Optimisation Tools}
\label{Sec:fadedpurehpo}
\begin{table}[H]
\begin{tabular}{|l|l|l|}
\hline
\rowcolor[HTML]{B4C6E7} 
\multicolumn{1}{|c|}{\cellcolor[HTML]{B4C6E7}\textbf{Name}} &
  \multicolumn{1}{c|}{\cellcolor[HTML]{B4C6E7}\textbf{GitHub}} &
  \multicolumn{1}{c|}{\cellcolor[HTML]{B4C6E7}\textbf{Ref.}} \\ \hline
Adatune       & {\color[HTML]{0563C1} { https://github.com/awslabs/adatune}}          & \cite{adatune} \\ \hline
Darts         & {\color[HTML]{0563C1} { https://github.com/quark0/darts}}             & \cite{darts} \\ \hline
DeepArchitect & {\color[HTML]{0563C1} { https://github.com/negrinho/deep\_architect}} &  \cite{deeparchitect} \\ \hline
FAR-HO        & {\color[HTML]{0563C1} { https://github.com/lucfra/FAR-HO}}           & \cite{farho} \\ \hline
GPyOpt        & {\color[HTML]{0563C1} { https://github.com/SheffieldML/GPyOpt}}       & \cite{gpyopt} \\ \hline
Optunity      & {\color[HTML]{0563C1} { https://github.com/claesenm/optunity}}        & \cite{optunity} \\ \hline
Osprey        & {\color[HTML]{0563C1} { https://github.com/msmbuilder/osprey}}        & \cite{osprey} \\ \hline
pyGPGO        & {\color[HTML]{0563C1} { https://github.com/josejimenezluna/pyGPGO}}   & \cite{pyGPGO} \\ \hline
RoBO          & {\color[HTML]{0563C1} { https://github.com/automl/RoBO}}              &  \cite{RoBO} \\ \hline
sklearn-deap  & {\color[HTML]{0563C1} { https://github.com/rsteca/sklearn-deap}}      & \cite{sklearndeap} \\ \hline
Spearmint     & {\color[HTML]{0563C1} { https://github.com/HIPS/Spearmint}}           & \cite{Spearmint} \\ \hline
\end{tabular}
\end{table}

\section{Faded AutoML Systems}
\label{Sec:fadedcoreautoml}
\begin{table}[H]
\begin{tabular}{|l|l|l|l|}
\hline
\rowcolor[HTML]{B4C6E7} 
\multicolumn{1}{|c|}{\cellcolor[HTML]{B4C6E7}\textbf{Name}} &
  \multicolumn{1}{c|}{\cellcolor[HTML]{B4C6E7}\textbf{GitHub}} &
  \multicolumn{1}{c|}{\cellcolor[HTML]{B4C6E7}\textbf{Company}} &
  \multicolumn{1}{c|}{\cellcolor[HTML]{B4C6E7}\textbf{Ref.}} \\ \hline
Adanet      & {\color[HTML]{0563C1} { https://github.com/tensorflow/adanet}}     & Google & \cite{adanet} \\ \hline
Advisor     & {\color[HTML]{0563C1} { https://github.com/tobegit3hub/advisor}}   & Personal &  \cite{advisor}  \\ \hline
Aethos      & {\color[HTML]{0563C1} { https://github.com/Ashton-Sidhu/aethos}}   & Personal &  \cite{aethos}  \\ \hline
Amla        & {\color[HTML]{0563C1} { https://github.com/CiscoAI/amla}}          & Cisco    &  \cite{amla}  \\ \hline
ATM         & {\color[HTML]{0563C1} { https://github.com/HDI-Project/ATM}}       & Research &  \cite{atm}  \\ \hline
auto\_ml    & {\color[HTML]{0563C1} { https://github.com/ClimbsRocks/auto\_ml}}  & Personal &  \cite{automl}  \\ \hline
Auto-Weka   & {\color[HTML]{0563C1} { https://github.com/automl/pyautoweka}}     & Research &  \cite{pyautoweka}  \\ \hline
AutoXGBoost & {\color[HTML]{0563C1} { https://github.com/ja-thomas/autoxgboost}} & Personal &  \cite{autoxgboost}  \\ \hline
DeepMining                            & {\color[HTML]{0563C1} { https://github.com/sds-dubois/DeepMining}}    & Research &  \cite{an17, Deepmining}  \\ \hline
FedNas                            & {\color[HTML]{0563C1} { https://github.com/chaoyanghe/FedNAS}}    & Research &  \cite{fednas, hean20}  \\ \hline
HpBandSter  & {\color[HTML]{0563C1} { https://github.com/automl/HpBandSter}}     & Research &  \cite{HpBandSter}  \\ \hline
MetaQNN                            & {\color[HTML]{0563C1} { https://github.com/bowenbaker/metaqnn}}    & Research &  \cite{metaqnn, bagu17}  \\ \hline
MLBox       & {\color[HTML]{0563C1} { https://github.com/AxeldeRomblay/MLBox}}   & Personal &   \cite{MLBox} \\ \hline
Recipe      & {\color[HTML]{0563C1} { https://github.com/laic-ufmg/Recipe}}      & Research &  \cite{Recipe}  \\ \hline
tuneRanger                            & {\color[HTML]{0563C1} { https://github.com/PhilippPro/tuneRanger}}    & Research &  \cite{tuneRanger}  \\ \hline
Xcessiv                            & {\color[HTML]{0563C1} { https://github.com/reiinakano/xcessiv}}    & Research &  \cite{xcessiv}  \\ \hline
\end{tabular}
\end{table}

\section{Insufficiently Detailed Commercial Systems}
\label{Sec:opaquevendor}
\begin{table}[H]
\begin{tabular}{|l|l|l|}
\hline
\rowcolor[HTML]{B4C6E7} 
\multicolumn{1}{|c|}{\cellcolor[HTML]{B4C6E7}\textbf{Name}} &
  \multicolumn{1}{c|}{\cellcolor[HTML]{B4C6E7}\textbf{Website}} &
  \multicolumn{1}{c|}{\cellcolor[HTML]{B4C6E7}\textbf{Ref.}} \\ \hline
Aible     & {\color[HTML]{0563C1} { https://www.aible.com/}}           & \cite{Aible} \\ \hline
Algolytics         & {\color[HTML]{0563C1} { https://algolytics.com/products/abm/}}             & \cite{Algolytics} \\ \hline
DMway       & {\color[HTML]{0563C1} { http://dmway.com/}}          & \cite{DMway} \\ \hline
dotData        & {\color[HTML]{0563C1} { https://dotdata.com/}}       & \cite{dotData} \\ \hline
Kortical        & {\color[HTML]{0563C1} { https://kortical.com/}}        & \cite{Kortical} \\ \hline
neuralstudio.ai        & {\color[HTML]{0563C1} { https://neuralstudio.ai/}}            & \cite{NeuralStudioai} \\ \hline
OptiScorer & {\color[HTML]{0563C1} { https://optiscorer.com/}} &  \cite{OptiScorer} \\ \hline
Pecan        & {\color[HTML]{0563C1} { https://www.pecan.ai/}}       & \cite{Pecan} \\ \hline 
Prevision.io      & {\color[HTML]{0563C1} { https://prevision.io/}}        & \cite{Prevision} \\ \hline
SparkCognition        & {\color[HTML]{0563C1} { https://www.sparkcognition.com/products/darwin/}}   & \cite{SparkCognition} \\ \hline
TAZI          & {\color[HTML]{0563C1} { https://www.tazi.ai/}}              &  \cite{Tazi} \\ \hline
Xpanse  & {\color[HTML]{0563C1} { https://xpanse.ai/}}      & \cite{XpanseAI} \\ \hline
\end{tabular}
\end{table}

\end{appendices}

\end{document}